UNIVERSITY OF CALGARY

# Multi-Sensor Integration for Indoor 3D Reconstruction

by

Jacky C.K. Chow

A THESIS

SUBMITTED TO THE FACULTY OF GRADUATE STUDIES

IN PARTIAL FULFILMENT OF THE REQUIREMENTS FOR THE

DEGREE OF DOCTOR OF PHILOSOPHY

DEPARTMENT OF GEOMATICS ENGINEERING

CALGARY, ALBERTA

APRIL, 2014

## Abstract

Outdoor maps and navigation information delivered by modern services and technologies like Google Maps and Garmin navigators have revolutionized the lifestyle of many people. Motivated by the desire for similar navigation systems for indoor usage from consumers, advertisers, emergency rescuers/responders, etc., many indoor environments such as shopping malls, museums, casinos, airports, transit stations, offices, and schools need to be mapped. Typically, the environment is first reconstructed by capturing many point clouds from various stations and defining their spatial relationships. Currently, there is a lack of an accurate, rigorous, and speedy method for relating point clouds in indoor, urban, satellite-denied environments. This thesis presents a novel and automatic way for fusing calibrated point clouds obtained using a terrestrial laser scanner and the Microsoft Kinect by integrating them with a low-cost inertial measurement unit. The developed system, titled the Scannect, is the first joint-static-kinematic indoor 3D mapper.

Manmade instruments are susceptible to systematic distortions. These uncompensated errors can cause inconsistencies between the map and reality; for example a scale factor error can lead firefighters to the wrong rooms during rescue missions. For terrestrial laser scanners, marker-based user self-calibration has shown effectiveness, but has yet to gain popularity because it can be cumbersome to affix hundreds of targets inside a large room. Previous attempts to expedite this process involved removing the dependency on artificial signalized targets. However, the commonalities and differences of markerless approaches with respect to the marker-based method were not established. This research demonstrated with simulations and real data that using planar features can yield similar calibration results as using markers and much of the knowledge about marker-based calibration is transferable to the plane-based calibration method.

For the Microsoft Kinect, there is a limited amount of research dedicated to calibrating the system. Most methods cannot handle the different onboard optical sensors simultaneously. Therefore, a novel and accurate total system calibration algorithm based on the bundle adjustment that can account for all inter-sensor correlations was developed. It is the first and only algorithm that functionally models the misalignments between the infrared camera and projector in addition to providing full variance-covariance information for all calibration parameters.

Scan-matching and Kalman filtering are often used together by robots to fuse data from different sensors and perform Simultaneous Localisation and Mapping. However, most research that utilized the Kinect adopted the loosely-coupled paradigm and ignored the time synchronization error between the depth and colour information. A new measurement model that facilitates a tightly-coupled Kalman filter with an arbitrary number of Kinects was proposed. The depth and colour information were treated independently while being aided by inertial measurements, which yielded advantages over existing approaches. For the depth data, a new tightly-coupled iterative closest point algorithm that minimizes the reprojection errors of a triangulation-based 3D camera within an implicit iterative extended Kalman filter framework was developed. When texture variation becomes available, the RGB images automatically update the state vector by using a novel tightly-coupled 5-point visual odometry algorithm without state augmentation.

## Acknowledgements

Thank you Dr. Derek Lichti and Dr. William Teskey for being such considerate and trusting supervisors. I would not have made it through these years at the university without your support and the freedom given to me for finding my way in life. I hope my occasional recklessness and spontaneity didn't give you too many grey hairs. Thank you Derek for teaching me everything I know about photogrammetry and how to live the life of an academic. The knowledge you have given me made it possible for me to reach new heights and do things that would be unimaginable a few years ago. The skills I have acquired under your supervision are valuable to me for the rest of my career and I owe this to you. Thank you Dr. Teskey for offering me the research opportunity while I was still an undergraduate student. That first summer experience inspired me to pursue research ever since. Thank you for being like a father to me all these years and giving me guidance on various things in life and sharing all your stories with me.

The time and effort my supervisory committee has put into reviewing this thesis is greatly appreciated. Thank you Dr. Ayman Habib, Dr. Costas Armenakis, Dr. Mohamed Elhabiby, and Dr. Simon Park for your feedback and assistance in improving this dissertation. I would like to express my sincere gratitude to Dr. Habib and Mohamed for being my educators from the time I was an undergraduate student and continuing to educate me to this day.

Thank you Brad, Courtenay, Gail, Garth, Julia, June, Kathy, Kirk, Kristal, Lu-Anne, Marcia, and Monica for assisting me and all the other students over the years. Our department could not have functioned without you. Thank you for presenting me with the "Winner of the Once in a Lifetime KING Among Geomatics Nerds Award", it the most prestigious recognition any graduate student can receive and I will cherish the plaque dearly. I apologize for all the last minute requests/submissions I have made over the years.

I am grateful for the wonderful professors in our department that have made a positive impact to my life and/or have given me knowledge to complete this dissertation and have a career in geomatics. Thank you Dr. Alex Braun, Dr. Andrew Hunter, Dr. Danielle Marceau, Dr. Jeong Woo Kim, Dr. Kyle O'Keefe, Dr. Mark Petovello, Dr. Michael Barry, Dr. Michael Collins, Dr. Michael Sideris, Dr. Quazi Hassan, Dr. Steve Liang, and Dr. Susan Skone.

Financial support from the Natural Sciences and Engineering Research Council (NSERC) of Canada, Alberta Innovates Technology Futures (AITF) Innovates Centre of Research Excellence (iCORE), Killam Trust, International Society for Optics and Photonics (SPIE), American Society of Photogrammetry and Remote Sensing (ASPRS), Terramatic Technologies Inc., SarPoint Engineering Ltd., Micro Engineering Tech Inc., the University of Calgary, Government of Alberta, and other sources are greatly appreciated.

I am honoured to have a great family, thank you for all the positive influences from my parents, grandparents, Uncle Alan, Uncle Chris, Uncle Kim, Auntie Kitty, Nicholas, Ryan, and Auntie Sue. The undeniable fact that my parents and grandparents spoiled me is truly a blessing. Thank you grandma for always cooking delicious comfort food for me. Thank you grandpa that you are always there to put things into perspective for me and to teach me useful skills in life.

Thank you Aivaras, Denise, and Fiona for welcoming me to your team in Belgium wholeheartedly. I learned a lot about computer vision from my friends Chris, Geert, Stephane, Stijn, and Yannick. Special thanks to Stephane for introducing me to Speculoospasta, which is still my favourite.

Thank you Amy, Gregory, Laurence, Stephanie, Xin for offering me this eye-opening experience in Paris. Thank you Stephane and both Guillaumes for educating me about French cuisine and acting like my personal trip advisors during my journey in France. Thank you David and Volker for offering me a place to stay when I was still desperately learning to adapt to the new culture. Thank you Pascal for looking out for me and teaching me the real meaning of crêpe.

Thank you Bart, Caecilia, and Sylvain for making my stay in Geneva so pleasant and exciting. I really enjoyed my exchange and am appreciative of everything you have taught me. I will always remember the Raclette and wine.

Thank you Henk for recruiting me and giving me flexibility to explore my own interests and complete part of my thesis in the Netherlands. I appreciate everything I have learned from the research group (Claudio, Giovanni, Jeroen, Laurens, Matteo, Padraig, and Raymond).

Thank you to my colleagues in the Imaging Metrology Group (Adam, Hervé, Hooman, Jeremy, Justin, Kristian, Mehdi, Mostafa, Reza, Sherif, Sonam, Tanvir, Ting, and Xiaojuan) and PEDS Group (Axel, Daniel, and JP) for all the wonderful and constructive discussions. In particular, salute to Axel for teaching me surveying and adjustments, bringing me on some of the most difficult hikes I have done, being a tolerating manager, being enthusiastic about beer everywhere we go, and for showing me that the graduate student lifestyle can be fascinating and fulfilling.

My time as a graduate student would not have been so enjoyable if I wasn't surrounded by amazing friends. Thank you Alexandra, Ali, Andrés, Anna, Anup, Bernhard, Claudius, CJ, Erin-Jennifer, Ehsan, Erwan, Eunju, Ivan, Jin, Ki In, Landon, Mike, Mohannad Monica, Navdeep, Sinem, Tasnuva, Oday, Ossama, Rohana, Vidya, Wesley, and Wouter for giving me a reason to be on campus. Thank you Erin-Jennifer and Ivan for sharing your overseas experiences with me, which led me to follow your footsteps. I will always remember the times we've travelled together, especially when we all slept on the floor of a gymnasium in Croatia. Ivan, thank you for getting me involved with all the extracurricular activities, I would have missed a lot of fun without you. Thank you Eunju and Mohannad for acting like my older siblings and creating an intervention when you saw fit. The computer used for processing majority of the data in this thesis was assembled by Mohannad, therefore your geekiness is gratefully acknowledged. Thank you Wes for sharing your business perspectives with me, you definitely inspired me to be an entrepreneur. Thank you Erwan for making the best turkey with stuffing, introducing me to Gougères, and helping me plan my travels to Europe. Thank you Mike for your constant inappropriate jokes and uplifting personality. Thank you Andrés for helping me organize many events for the students (especially the barbeque), teaching me how to rock climb, and reminding me to never stop being a kid.

I am grateful for the wonderful friendships I have made over the years that brought me here, thank you Adrian, Andrew, Anthony, Christine, Damian, Dorothy, Elisha, Elle, Ericson, Erik,

Gary, Hai, Jason, Jonathan, Kelson, Kirk, Kris, Melissa, Peter, Rachel, Shiga, Sinisa, and Wilson.

Thank you K'dee for the endless hours you've spent helping me proofread this dissertation.

Thank you for the wonderful family of geomatics students that made my university experience rewarding and amusing:

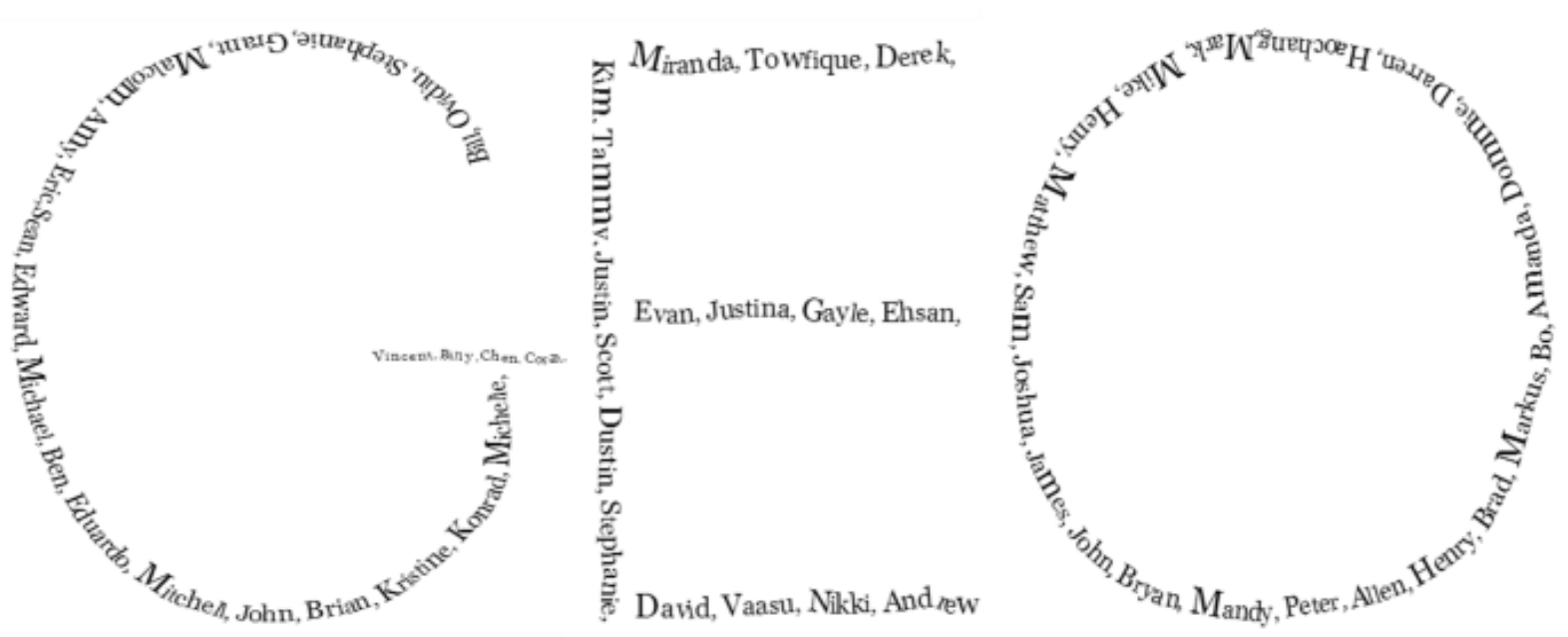

*To my courageous, wise, and loving mother*

*Thank you for your patience and time in teaching and taking care of me*

*Thank you for giving up everything for me*

*Thank you for all your prayers*

*Your unconditional love and encouragements made me who I am*

*P.S. Sorry for my only child syndrome*

**Table of Contents**

## List of Tables

## List of Figures

## List of Abbreviations

| *Abbreviation* | *Definition* |
|---|---|
| AM-CW | Amplitude Modulated Continuous Wave |
| AP | Additional Parameter |
| ARW | Angular Random Walk |
| AUPT | Attitude UPdaTe |
| CPU | Central Processing Unit |
| CUPT | Coordinate UPdaTe |
| CSHOT | Color Signature of Histograms of OrienTations |
| EKF | Extended Kalman Filter |
| EOP | Exterior Orientation Parameter |
| FAST | Features from Accelerated Segment Test |
| FLANN | Fast Library for Approximate Nearest Neighbors |
| FOV | Field Of View |
| GNSS | Global Navigation Satellite System |
| GPGPU | General-Purpose computing on Graphics Processing Units |
| GPU | Graphics Processing Unit |
| HFOV | Horizontal Field Of View |
| ICP | Iterative Closest Point |
| IR | InfraRed |
| IEKF | Iterative Extended Kalman Filter |
| IMU | Inertial Measurement Unit |
| INS | Inertial Navigation System |
| IOP | Interior Orientation Parameter |
| ISS | Intrinsic Shape Signature |
| KF | Kalman Filter |
| KLT | Kanade–Lucas–Tomasi |
| LiDAR | Light Detection And Ranging |
| MEMS | MicroElectroMechanical Systems |
| PCL | Point Cloud Library |
| PTAM | Parallel Tracking And Mapping |
| RANSAC | RANdom SAmple Consensus |
| RGB | Red Green Blue |
| RGB-D | Red Green Blue and Depth |
| RMSE | Root Mean Squared Error |
| ROP | Relative Orientation Parameter |
| SBA | Sparse Bundle Adjustment |
| SFM | Structure From Motion |
| SHOTColor | Signature of Histograms of OrienTations Color |
| SIFT | Scale-Invariant Feature Transform |
| SLAM | Simultaneous Localisation and Mapping |
| SURF | Speeded Up Robust Feature |
| TLS | Terrestrial Laser Scanning |
| TOF | Time Of Flight |
| TORO | Tree-based netwORk Optimizer |
| TSDF | Taylor Surface Distance Function |
| VFOV | Vertical Field Of View |
| VO | Visual Odometry |
| VSLAM | Visual Simultaneous Localisation and Mapping |
| VTK | Visualization ToolKit |
| ZUPT | Zero velocity UPdaTe |

# Chapter One: Introduction

The natural habitat of human beings is indoors because it provides comfort and shelter from harsh weather as well as predators. Based on the World Health Organization, as of 2010, urbanization has resulted in more than half of the world's population residing in urban areas [1]. Although maps have been a crucial tool in human history for describing locations and navigation since 6400 B.C., most indoor urban environments are still mapless [2]. Most state-of-the-art maps are defined in 3D rather than being a 2D projection of 3D space because it conveys more information. Such detailed 3D spatial digitization of the surroundings has many advantages and applications if it can be accomplished in haste and without compromising accuracy.

## 1.1 Motivation and Problem Statement

There is currently a huge demand for efficient and affordable kinematic mapping systems for indoor infrastructure from architectural, engineering and construction contractors (AEC); mechanical, electrical and plumbing (MEP); mining; and law enforcement, public safety, and first responders. An indoor mobile mapping system allows large areas such as plant and factory facilities; high-rise offices, residential, and government buildings; airports and train stations; music halls, theatres, and auditoriums; covered pedestrian concourses; and underground mines and tunnels to be reconstructed efficiently [3] [4] [5]. An accurate 3D model of indoor urban environments can be a valuable asset to cultural heritage documentation, virtual environments, city planning, urban design, military applications, and fire & police planning [6] [7] [4] [8] [9] [10]. For instance, every year in the United States hundreds of highly qualified firefighters suffer injuries or fatalities because they were lost while escaping from a burning building [11]. A navigation device equipped with indoor 3D maps could help them find escape routes and assist the captain with planning more effective missions. Similarly, this technology can be used to help customers/visitors manage their trip to Ikea, museums, and malls by providing directions to nearby washrooms, food vendors, specific products, paintings, etc. [12].

One of the most frequently encountered tasks in mapping is registration. Because most geospatial imaging sensors have limited range and field of view (FOV), smaller local maps need to be merged or registered together to achieve more complete coverage. This can be achieved either directly via external sensors that provide the absolute position and orientation to the imaging sensor (typically a laser scanner), or indirectly through surveyed control points. It is an expensive task to establish control points, especially indoors where there are many obstructions and doorways. If range and incidence angle constraints are considered, to scan an indoor environment with sufficient overlap can require more static scan stations than anticipated [13]. For outdoor mapping, a well-established direct georeferencing solution is to integrate Global Navigation Satellite System (GNSS) and Inertial Navigation System (INS) with either a laser scanner or photogrammetric system. However, for indoor applications where satellite signals are obstructed, an effective substitute for GNSS is still under extensive research. When using a Microelectromechanical Systems (MEMS) Inertial Measurement Unit (IMU) instead of a navigation-grade IMU, it is even more critical for the drift in the navigation states to be controlled. Other researchers have attempted to use various sensors to aid the IMU, such as Wi-Fi [14], magnetometers [15], and ultra-wide band [16]. To address this problem, in this project, low-cost off-the-shelf RGB-D cameras were combined with a MEMS IMU through a tightly-coupled Kalman filter (KF). For close-range mapping applications, such as the measurement of a statue, the range camera and IMU system alone will suffice, as the range camera will be used for both mapping and assisting the IMU solution through resection. If a larger area such as a cathedral or the Olympic Oval is to be mapped, a sensor with longer range, such as a laser scanner needs to be appended to the system. Therefore, the problem statement is whether low-cost autonomous 2D/3D vision sensors and IMU can expedite the workflow of static terrestrial laser scanners while ensuring sufficient accuracy for surveying applications. The desired outcome is a low-cost mobile mapping system suitable for generating 3D point clouds for geomatics applications with reduced time and labour in both the acquisition and processing stages. As stated in [3], mobile mapping and navigation involves a number of issues such as the *system design*, *choice of sensors*, and *methods for localisation and mapping*; this thesis will address these three aspects of mobile mapping in addition to *sensor calibration*.

## 1.2 Proposed Solution: Scannect

In this thesis, a new system for performing indoor 3D mapping and localisation entitled the Scannect is presented. It is one of the first systems that combined the accuracy of static laser scanning with the efficiency of kinematic mapping. A FARO Focus$^{3D}$ S terrestrial laser scanner, RGB-D cameras from Microsoft, and an Xsens MEMS IMU were assembled together on a wheel-based platform. The onboard FARO Focus$^{3D}$ S 3D terrestrial laser scanner operating in stop-and-go mode captures data at longer ranges over a wide angular FOV. Occasionally performing a full 360$^{o}$ scan in static mode with sufficient scene overlap can provide valuable coordinate updates (CUPT) and attitude updates (AUPT), in addition to assisting loop-closures. When a static scan is completed, the Scannect enters mobile mode where the Microsoft Kinect systems continuously estimate the exterior orientation parameters (EOPs) using the RGB images and point clouds independently. This information about the change in position and orientation will update the navigation states and suppress the drift in the IMU-only solution. At the same time the IMU provides relative position and orientation information to the 3D range cameras, which were used for acquiring 3D images of the scene and expanding the map coverage. Between vision updates (which has a much lower frequency than an IMU) or during visual outages (e.g. lack of feature/texture or out-of-range) the IMU keeps an accurate prediction of the navigation states for a short time frame. When the quality of the state estimates degrades significantly, zero velocity updates (ZUPTs) can be performed and a full $360^o$ laser scan can be used to reposition the system by relating itself to previously reconstructed areas. Outputs from the sensors were all fused using a Kalman filter to generate a 3D point cloud of the scene.

In the proposed KF, the vision data are assumed to be calibrated (i.e. free of unmodelled distortions). In preparation for integration, systematic errors in the optical sensors were analyzed and methods for reducing these errors were studied. For terrestrial laser scanning (TLS) instruments, the self-calibration approach, which models the systematic errors in distance and angles much like a total station was adopted and had demonstrated wide success on different scanners. However, the unavoidable tedious chore of setting up hundreds of signalized targets is still a limitation of this approach. The concept of substituting signalized targets with planar features for quicker calibrations was explored in this research. It was shown using synthetic and real data that the markerless plane-based self-calibration method can indeed replace self-calibration based on signalized targets. However, special considerations need to be made when designing the network and they are outlined in this thesis. Regarding the self-calibration of the Kinect, most existing methods are only devoted to aligning the colour data with the depth data. Merely a handful of researches were dedicated to the RGB and depth co-registration and error modelling of the depth data simultaneously. Even then, the non-parallelism between the camera-projector pair was often ignored, even though they have been found to be statistically significant. To alleviate these limitations a new self-calibration algorithm for the Kinect was developed, tested, and analyzed for minimizing errors of all the onboard vision sensors. This method was shown to be capable of reducing systematic errors effectively and yielded point clouds with better accuracy and precision. Besides error modelling of each instrument, the translational and rotational parameters relating all the sensors were also estimated numerically using least squares adjustment. After calibration, the individual sensor data were ready to be fused.

### 1.3 Research Objectives

The main research objectives of this thesis are:

- To improve the quality and/or efficiency of current self-calibration methods for modern static 3D terrestrial laser scanners by:
    - Enhancing the measurement precision and accuracy of low-cost TLS instruments by identifying and modelling systematic errors using the user self-calibration approach,
    - Studying the observability of additional parameters in the multi-station self-calibration of hybrid and panoramic-type TLS instruments as well as their dependency on the network design,
    - Comparing the newer and more efficient plane-based self-calibration of TLS instruments to the better studied point-based approach and identify their dissimilarities.

- To devise an accurate, rigorous, and user-friendly method for compensating the systematic distortions in the Microsoft Kinect system that can overcome the shortcomings of existing calibration methods by:
    - Improving the co-registration between the point cloud and colour image,
    - Improving both the accuracy and precision of the depth map.
- To design an accurate indoor mapping system that can reduce the field surveying time by introducing kinematic mapping concepts by:
    - Selecting and fusing a suitable group of sensors for indoor 3D documentation applications; evaluating the suitability of using a low-cost IMU, digital cameras, and 3D cameras to keep track of the laser scanner's pose to aid its registration process,
    - Develop a new Iterative Closest Point (ICP) algorithm based on the measurement principle of the Kinect for more effective scan-matching and EOP estimation,
    - Designing a tightly-coupled extended Kalman filter (EKF) for integrating the inertial data, texture data, and 3D shape data for mapping and navigation purposes.

## 1.4 New Research Contributions

The most notable scientific contributions of this thesis are:

- A new and effective means for recovering the horizontal collimation axis error, one of the most critical systematic errors in hybrid-type TLS instruments through improved network geometry (i.e. tilted scans and slanted planes).

- Demonstration that the point-based and plane-based TLS self-calibration methods can deliver comparable results with proper network design measures.
- A novel Kinect optical system self-calibration method based on the photogrammetric bundle adjustment that simultaneously estimates the EOPs, interior orientation parameters (IOPs), additional parameters (APs), object space coordinates, feature parameters, and relative orientation parameters (ROPs) of the Kinect.
- The first hybrid mobile mapper design that operates in both stop-and-go and kinematic modes to harvest the efficiency and continuous trajectory of a full kinematic mapping solution while maintaining a degree of accuracy and integrity from stop-and-go scanning. The balance between the two approaches can be adapted to the field situation without changing the software.
- A novel point-to-plane ICP implementation using the implicit iterative extended Kalman filter (IEKF) to minimize the reprojection errors of a stereo-vision system while incorporating IMU data and texture information (after accounting for temporal synchronization errors). Its main benefit is that it is solvable even when a single textureless wall and/or less than the minimum number of matches were found, which can be commonly encountered in indoor urban environments. Furthermore, while travelling down flat homogenous hallways, it does not “anchor” the translation along the principal axis of the corridor, allowing the system to slide past these areas.
- The first multi-Kinect system integrated with a MEMS-based IMU to assist a 3D TLS instrument by filling occlusions efficiently and continuously updating the system’s

EOPs. This provides the scanner with a good initial alignment necessary to ensure good convergence when using ICP methods.

- A new 5-point visual odometry algorithm in an implicit IEKF framework to overcome the growing state vector problem in typical EKF monocular vision navigation. This tightly-coupled approach only provides updates to the EOPs by implicitly solving the object space coordinates while minimizing the reprojection errors. This textural information can aid EOP estimation over textured regions even if topography information is limited and can detect periods when the camera is stationary.

## 1.5 Thesis Outline

This thesis is organized into six chapters. The first chapter covers the motivations and objectives behind this research, problem statement being addressed, proposed solution, and main contributions. The remaining chapters are arranged as documented below, with chapters 3, 4, and 5 presenting articles that have been previously published or submitted to internationally-recognized refereed journals.

**Chapter 2** provides the reader with relevant background information about the terrestrial laser scanner, Microsoft Kinect, and IMU that composes the Scannect.

**Chapter 3** investigates the point-based and plane-based TLS self-calibration methods. The ideas of including tilted scans and/or planes to improve calibration quality in the network adjustment were explored.  Recommendations for better recovery of APs in the self-calibration of different scanner architectures were made.

**Chapter 4** analyzes the systematic errors found in the Microsoft Kinect and presents a novel bundle adjustment approach for compensating the non-random geometric distortions in the colour and depth data simultaneously.

**Chapter 5** describes the *system design*, *choice of sensors*, and *methods for localisation and mapping* of the Scannect.  The mathematical models for tightly-coupling the Kinects with an IMU using a Kalman filter is presented.  Mapping results using the Scannect were reported and the quality was assessed.

**Chapter 6** presents concluding remarks and recommendations for future investigations.

### 1.6 Copyright

Chapters 3, 4, and 5 in addition to Appendix A and B are publications used for compiling this manuscript-based thesis.  Their contents are slightly edited and reformatted for consistency. References for these publications are declared below.

*Chapter 3:*

**Chow, J.**, Lichti D., Glennie, C., and Hartzell, P. (2013).  Improvements to and comparison of static terrestrial LiDAR self-calibration methods.  Sensors 13(6), 7224 - 7249.

*Chapter 4:*

**Chow, J.** and Lichti D. (2013). Photogrammetric bundle adjustment with self-calibration of the PrimeSense 3D camera technology: Microsoft Kinect. IEEE Access 1(1), 465-474.

*Chapter 5:*

**Chow, J.**, Lichti, D., Hol, J., Bellusci, G., and Luinge, H. (2014). IMU and multiple RGB-D camera fusion for assisting indoor stop-and-go 3D terrestrial laser scanning. Robotics: Robot Vision. 35 pages. Submitted: February 19, 2014.

*Appendix A:*

**Chow, J.**, Ang, K., Lichti, D., and Teskey, W. (2012). Performance analysis of a low-cost triangulation-based 3D camera: Microsoft Kinect system. International Society of Photogrammetry and Remote Sensing Archives XXXIX-B5, 175-180.

*Appendix B:*

**Chow, J.**, Lichti, D., and Teskey, W. (2012). Accuracy assessment of the FARO Focus$^{3D}$ and Leica HDS6100 panoramic-type terrestrial laser scanner through point-based and plane-based user self-calibration. *FIG Working Week 2012: Knowing to manage the territory, protect the environment, evaluate the cultural heritage.* Rome, Italy. May 6-10.

# Chapter Two: Background

Terrestrial laser scanners and 3D range cameras are current state-of-the-art active imaging modalities. These optical systems are capable of measuring detailed dense geometric information of the object space efficiently. However, these sensors cannot penetrate objects and occlusions can exist in the data because of their perspective imaging geometry. Therefore, multiple 3D point clouds are usually captured from different positions and orientations to generate a complete model of the object space. To achieve this goal, the individual 3D point clouds must be registered or transformed into a common coordinate system. Traditionally this can be accomplished using signalized targets; however this is a very time-consuming procedure and is sometimes impossible (e.g. due to accessibility). Alternatively, feature-based registration techniques (Gielsdorf et al., 2004 [17]; Rabbani et al., 2007 [18]) and the ICP algorithms (Besl & McKay, 1992 [19]; Chen & Medioni, 1992 [20]) have been developed to register point clouds together based on geometric features in the scene. Such methods can produce a good fit between adjacent point clouds if sufficient features exist in the scene and there is a significant degree of overlap. However, because a large degree of overlap is required, scans need to be captured closer together which results in more scan time and hinders the productivity in the field. Methods for using digital images to reduce the amount of overlap between point clouds have been suggested (Al-Manasir & Fraser, 2006 [21]; Renaudin et al., 2011 [22]), however texture/features/topography are scene-dependent and may not always be available. For example, cloud-based registration can be a challenge when moving through featureless hallways or offices with many doorways. To overcome these challenging situations, additional sensors may be used to provide the navigation solution needed for mapping.

Outdoor mobile LiDAR mapping systems began with airborne platforms, and later transitioned to include terrestrial systems as well. They use one or more GNSS receivers coupled with an expensive INS to obtain the trajectory information necessary for the imaging sensor to map the scene (Talaya et al., 2004 [23]; Asai et al., 2005 [24]). This is known as direct georeferencing and commercial systems based on this principle are widely available: for instance Riegl VMX-450 and Topcon IP-S2 (Figures 1a and 1b). Terrestrial systems are not restricted to automobile-mounted systems, for example Google has outdoor systems on snowmobiles and tricycles (Figures 1c and 1d). Traditionally the EOPs are estimated independent of the mapping process; however this is changing as more researchers are seeking improvements in the mapping quality through image/point cloud matching [25]. In general, these systems can perform rather poorly when the GNSS signals are lost for long periods of time; therefore they are not the most suitable for indoor applications. When indoors, a clear replacement for GNSS is still being researched and common navaids include Wi-Fi, IMU, Bluetooth, ultra-wide band, visible light communication, pseudolites, cameras, ultrasound, odometer, cell phone signals, magnetometers, and barometers [26] [16] [27] [28]. A selection of these sensor(s) is/are required to assist the IMU and control the rapid drift in the navigation states over time, in particular for low-cost MEMS IMUs. More information about each of the three sensors used in this project (i.e. the FARO Focus$^{3D}$ S terrestrial laser scanner, Microsoft Kinect 3D camera, and Xsens MTi MEMS IMU) is presented in the following sections.

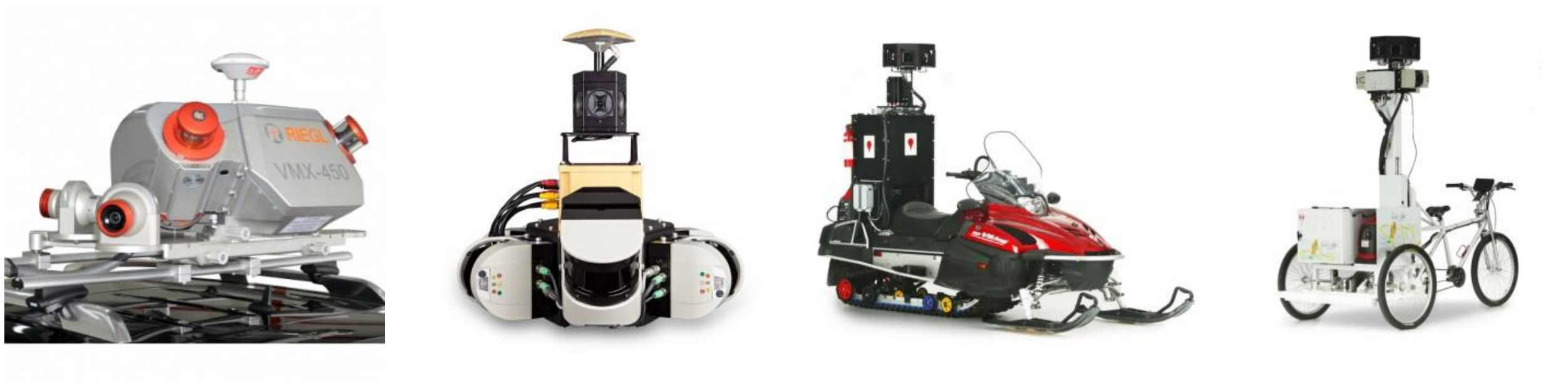

(a) (b) (c) (d)

Figure 1: Modern mobile terrestrial mapping systems – a) Riegl VMX-450; b) Topcon IP-S2 Compact +; c) Google Street View Snowmobile; d) Google Street View Trike [29] [30] [31]

### 2.1 Terrestrial Laser Scanners

In this research, a terrestrial laser scanning instrument operates in stop-and-go fashion on a wheel-based moving platform. TLS instruments are a type of accurate 3D imaging sensor that can output 3D point clouds describing the geometry in the scene. Although the point cloud is commonly expressed in the Cartesian coordinate system, most TLS instruments actually observe the distance to the object space at uniform increments of arcs in two orthogonal directions. This observation principle is comparable to a reflectorless total station. Laser scanners based on this principle will be the main focus in the remaining discussion as they are commonly found in surveying and mapping.

In the last decade, TLS instruments have gained significant recognition in the areas of surveying, forensic crime scene documentation, mapping, deformation monitoring, biomedical engineering, and many more. Recent advances in the terrestrial laser scanner designs have made this optical imaging modality more user-friendly, practical, and popular. For example, on-board touch screen user control interfaces have made data acquisition by non-experts easy to accomplish. Some attractive features of modern TLS instruments include:

- *Wide angular field of view (FOV):* TLS scanners like the Z+F Imager® 5010c and FARO Photon 120/20 offer a horizontal FOV of 360$^{o}$ and a vertical FOV of 320$^{o}$. Only a small cone underneath the tripod is not observable.
- *Large unambiguous range interval:* Close-range scanners such as the Z+F Imager® 5010 can measure objects that are from 0.3 m to 187.3 m away from the sensor. Long-range scanners such as Riegl VZ-6000 can measure objects up to 6 km away.
- *High data acquisition rate:* Currently, the fastest phase-based scanners (Z+F Imager® 5010 and Leica HDS6200) can measure up to 1 016 727 points per second. With recent breakthroughs, even pulse-based scanners (Leica ScanStation P20 and Trimble TX8) have reached acquisition rates up to 1 000 000 points per second.
- *Active sensor:* Unlike traditional photogrammetric systems, laser scanners can operate day or night and are capable of making dense observations on homogeneous surfaces without signalized targets or locating conjugate points in overlapping images.
- *Accurate geometric information:* For close-range measurements, the 3D position of every point can be determined with millimeter-level accuracy.
- *Radiometric information:* Besides geometric information, many scanners can also deliver intensity information for every point in the point cloud based on the object’s reflectivity.
- *Additional information:* The new generation of TLS instruments (e.g. Riegl VZ-1000) can deliver full return waveform information so that multiple echoes can be analyzed for each point.

- *Additional sensors:* Scanners can have a dual-axis compensator, barometer, magnetometer, GPS, and a high-resolution digital camera for enhancing the point cloud with spectral information.
- *Operable conditions:* Scanners like the Leica HDS8810 and MAPTEK I-Site 8810 can operate in temperatures between -40$^{o}$C and +50$^{o}$C.
- *Affordable cost:* The latest FARO Focus$^{3D}$ S is for sale at approximately CAD$30 000 including software and accessories.
- *Compact and light weight*: The FARO Focus$^{3D}$ S and Trimble TX5 weighs only 5 kg and has dimensions of 240 mm by 200 mm by 100 mm.

Modern TLS instruments demonstrate incomparable capabilities that are superior to other sensors on the market and have opened the door for many new applications. Unlike total stations that typically only measure to the centroid of signalized targets, TLS instruments can measure the entire scene automatically, with high density and efficiency. There are no correspondence issues as in traditional photogrammetry because 3D information is gathered from a single scan station. However, there are still some noticeable limitations of modern TLS instruments and they are listed below:

- *Scan time delay:* The operating principle of TLS instruments requires a complete profile to be captured in the vertical direction before the scanner head turns in the horizontal direction. This restricts TLS instruments to applications where only static objects are measured.

- *Less portable and more expensive than off-the-shelf camera systems:* Off-the-shelf amateur digital single-lens reflex (DSLR) cameras and 3D cameras such as the Microsoft Kinect system can be purchased for less than CAD$400, which is comparably more affordable than even the FARO Focus$^{3D}$.
- *Systematic errors:* After the manufacturer's precise laboratory calibration, systematic errors can still be observed in the point cloud of many TLS instruments. For precise applications such as deformation monitoring, these systematic deviations need to be eliminated, either through calibration or by adopting special observation procedures.

### *2.1.1 Classification of Terrestrial Laser Scanners*

Unlike DSLR cameras where the hardware design used by different manufacturers is more or less comparable, the architecture of modern TLS instruments can differ significantly depending on the manufacturer. Regardless of the assembly of the individual components, the type of deflection mirror(s) used, or the type of motors installed, TLS instruments can be broadly categorized based on their distance measurement principle and field of view [32]. It is important to distinguish the different types of laser scanners at this point, before the error sources and user self-calibration procedure is explained.

#### 2.1.1.1 Distance Measurement Principle

In general, the distance from the sensor to a point in the object space can be measured based on one of the three principles: pulse-based, phase-based, and triangulation-based (Figure 2). The most accurate distance measurement principle is triangulation. Distances can be measured accurate to a hundredth of a millimeter (e.g. Minolta Vivid 9i). But this comes at the expense of a shorter measurement range, which is typically less than 10 m. Due to the limited measurement range, triangulation-based laser scanners are not usually used for mapping and are therefore not the focus of this project. This thesis will concentrate on the time-of-flight systems (pulse-based and phase-based type laser scanners) as they are the most commonly utilized scanners in geomatics engineering for mapping purposes.

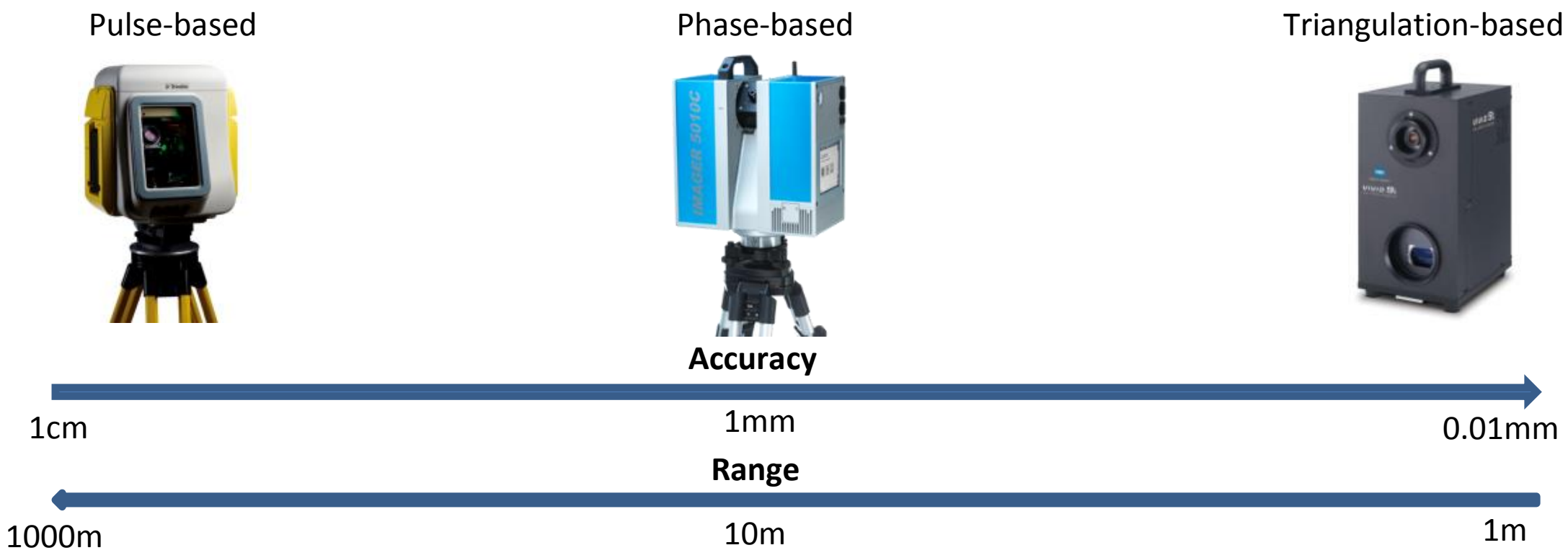

Figure 2: Classification of terrestrial laser scanners [33] [34] [35]

2.1.1.1.1 Pulse-based Terrestrial Laser Scanners

As shown in Figure 2, pulse-based scanners usually have centimeter-level distance precision or better and can measure distances that are kilometres away. To accomplish this, a laser pulse is emitted from the scanner at uniform increments of horizontal and vertical angles. When the laser pulse exits the sensor, it triggers the scanner's internal clock to begin measuring the time. The laser pulse (belonging to either the visible or near-infrared spectrum of electromagnetic radiation) travels through space, is reflected by the object, and received by the sensor, which then stops the scanner's clock. By knowing the speed of light and the two-way travel time, the distance to the object can be derived based on Equation 1.

$$\rho = \frac{c\Delta t}{2} \tag{1}$$

where $\rho$ is the range from the scanner to the object
$c$ is the speed of light
$\Delta t$ is the two-way travel time of the light pulse

To ensure that there is no ambiguity in the distance measurements, a pulse is typically not emitted until the previous pulse has been received. This causes a limitation on the pulse repetition rate. The distance measurement precision is mainly a function of timing and the signal-to-noise ratio of the received laser pulse. Hence, to achieve high range precision, an accurate clock is fundamental.

2.1.1.1.2 Phase-based Terrestrial Laser Scanners

Amplitude modulated continuous wave (AM-CW) light is emitted and received by phase-based scanners. The phase-shift between the emitted and received signal is used to calculate the two-way travel time of the laser. The range in this case can be calculated based on Equation 2.

$$\rho = U\left(\frac{\psi}{2\pi} + n\right) \tag{2}$$

where $U$ is half the modulation wavelength known as unit length
$\psi$ is the phase difference
$n$ is the integer number of full wavelengths

Distances can be measured with greater precision than the pulse-based method because an accurate clock is not mandatory for this system. Also, because light is continuously emitted and received, distances can typically be measured at a much higher frequency. The range precision is proportional to the modulation frequency, while the maximum unambiguous range is inversely proportional to the modulation frequency. Thus, if the laser is modulated with high frequency, the range precision will be higher but will be limited to a shorter distance measurement interval. In general, phase-based TLS instruments can be more accurate than pulse-based instruments, especially for shorter distances.

#### 2.1.1.2 Scanner Architecture and Field of View

TLS instruments can also be broadly categorized based on the scanner's field of view. In ascending order from the smallest to the largest angular field of view, modern static terrestrial laser scanners can be classified as having a camera-type, hybrid-type, or panoramic-type architecture. Each of the architectures will be explained in greater detail in the following subsections.

##### 2.1.1.2.1 Camera-Type Terrestrial Laser Scanners

This type of scanner is becoming obsolete as the horizontal and vertical FOV is usually only $40^o$ (Figure 3). An example of the camera-type scanner is the Optech ILRIS-3D. The deflection of the laser beam is usually achieved using two orthogonal galvanometers. Since this type of scanner architecture is uncommon in modern systems, it is not explained further.

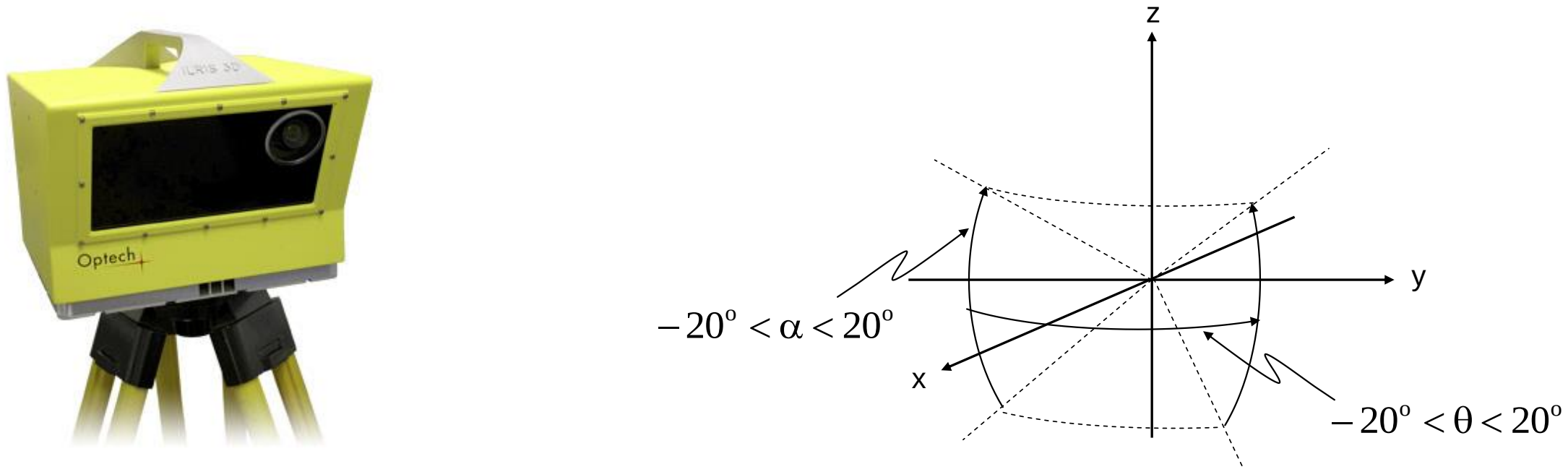

Figure 3: FOV of camera-type laser scanners [36] [37]

2.1.1.2.2 Hybrid-Type Terrestrial Laser Scanners

In most hybrid-type scanners, the laser beam is deflected in the vertical direction using an oscillating mirror or multiple-facet rotating mirror, while the horizontal FOV is maximized through a rotating scanner head (Figure 4). When capturing a full spherical scan, the scanner head rotates a full 360$^o$ and the oscillating mirror will direct the laser beam below the horizontal plane by an angle $\alpha_{min}$ and above the horizontal plane by $\alpha_{max}$. The maximum elevation angle $\alpha_{max}$ does not exceed 90$^o$ and the minimum elevation angle $\alpha_{min}$ is usually less than nadir (-90$^o$) because of the tripod legs.

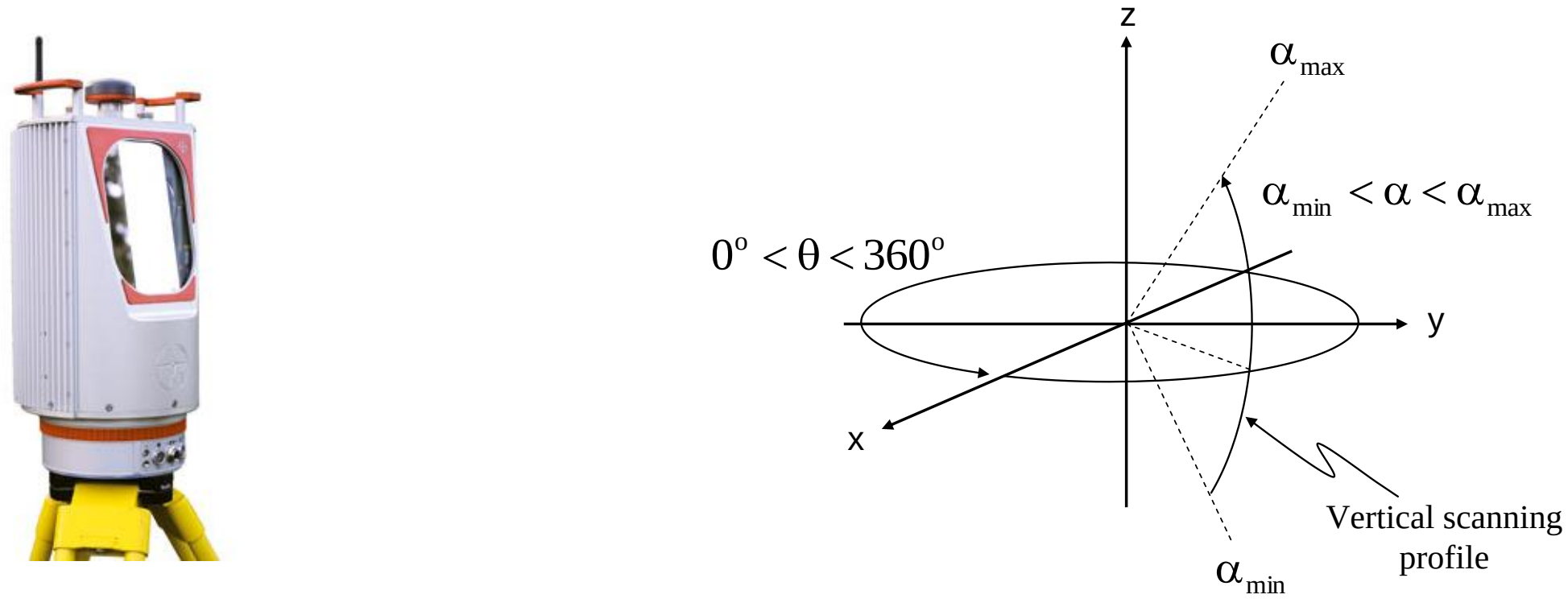

Figure 4: FOV of hybrid-type laser scanners [37] [38]

2.1.1.2.3 Panoramic-Type Terrestrial Laser Scanners

An increasing number of laser scanners are adopting this type of scanner architecture as they usually offer the largest angular FOV and fastest scan time. Similar to a hybrid-type scanner, the horizontal FOV is $360^o$ because the scanner is capable of rotating around in a full circle. Although the scanner head can rotate $360^o$, panoramic-type scanners only need to rotate the scanner head $180^o$ to capture a full spherical scan. The rotating mirror can deflect the laser beam from below the horizontal plane in front of the scanner ($\alpha_o$), up past the vertical axis, and down below the horizontal plane behind the scanner ($270^o$- $\alpha_o$). A graphical representation of the scanner's FOV is shown in Figure 5. Since the mirror is rotating with constant angular velocity and deflects the laser beam in front and behind the scanner, the scan time is greatly reduced and the angular FOV is only restricted by the tripod. This category of laser scanners is the most suitable for indoor mobile scanning because of its large coverage and high frequency.

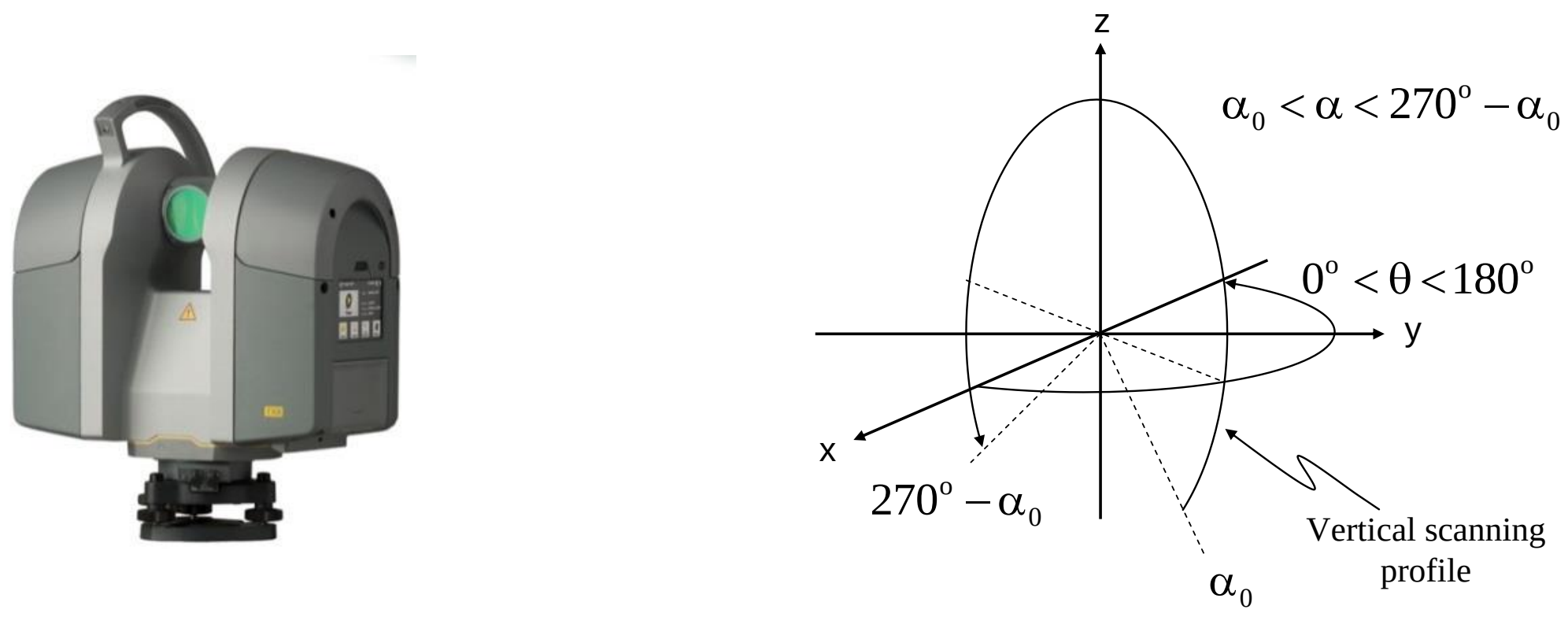

Figure 5: FOV of panoramic-type laser scanners [37] [39]

### *2.1.2 Systematic Errors in Terrestrial Laser Scanners*

There are many companies that manufacture survey-grade TLS instruments, including Leica, Trimble, Riegl, Z+F, FARO, Optech, and Maptek. The scanners produced by these companies are all different, and prior to shipping these high-quality instruments, they were all carefully calibrated by the manufacturers. Every manufacturer has their own calibration facility and procedure tuned specifically for their instrument, therefore calibration of scanners is not as standardized as digital cameras. Unfortunately, the raw data and empirical registration results from most modern TLS instruments indicate that unmodelled systematic errors can still exist in the point cloud. Numerous researchers from around the world have independently identified systematic trends in the laser scanner's residuals that deteriorate the range and angular measurement precision and accuracy of the laser scanner [40] [41] [42] [43] [44]. These systematic errors originate from two main sources: 1) systematic errors in the individual components; and 2) axes misalignments [45]. The most crucial assumption for TLS instruments is that the three fundamental axes (trunnion axis, vertical axis, and collimation axis) are mutually orthogonal and intersect at a common point in space (Figure 6). Therefore, any deviation from this ideal assembly will result in systematic errors in the point cloud. Furthermore, these systematic errors may be unstable and change over time or during transportation. To ensure optimum performance of the TLS instrument, it is important to remove these systematic distortions from the point cloud. Instead of sending the scanner back to the manufacturer for their laboratory calibration, more efficient quality assurance techniques have been explored. One suitable solution that can improve the raw observation precision and accuracy of every point in the point cloud is the user self-calibration approach. This technique allows systematic errors of each component and inter-components in hybrid/panoramic-type pulse/phase-based scanners to be modelled without disassembling the instrument [46]. The calibration does not require any specialized tools other than some form of targets and can be performed as frequently as the user desires with ease. A more detailed explanation of the self-calibration method and the error models is given in Chapter 3.

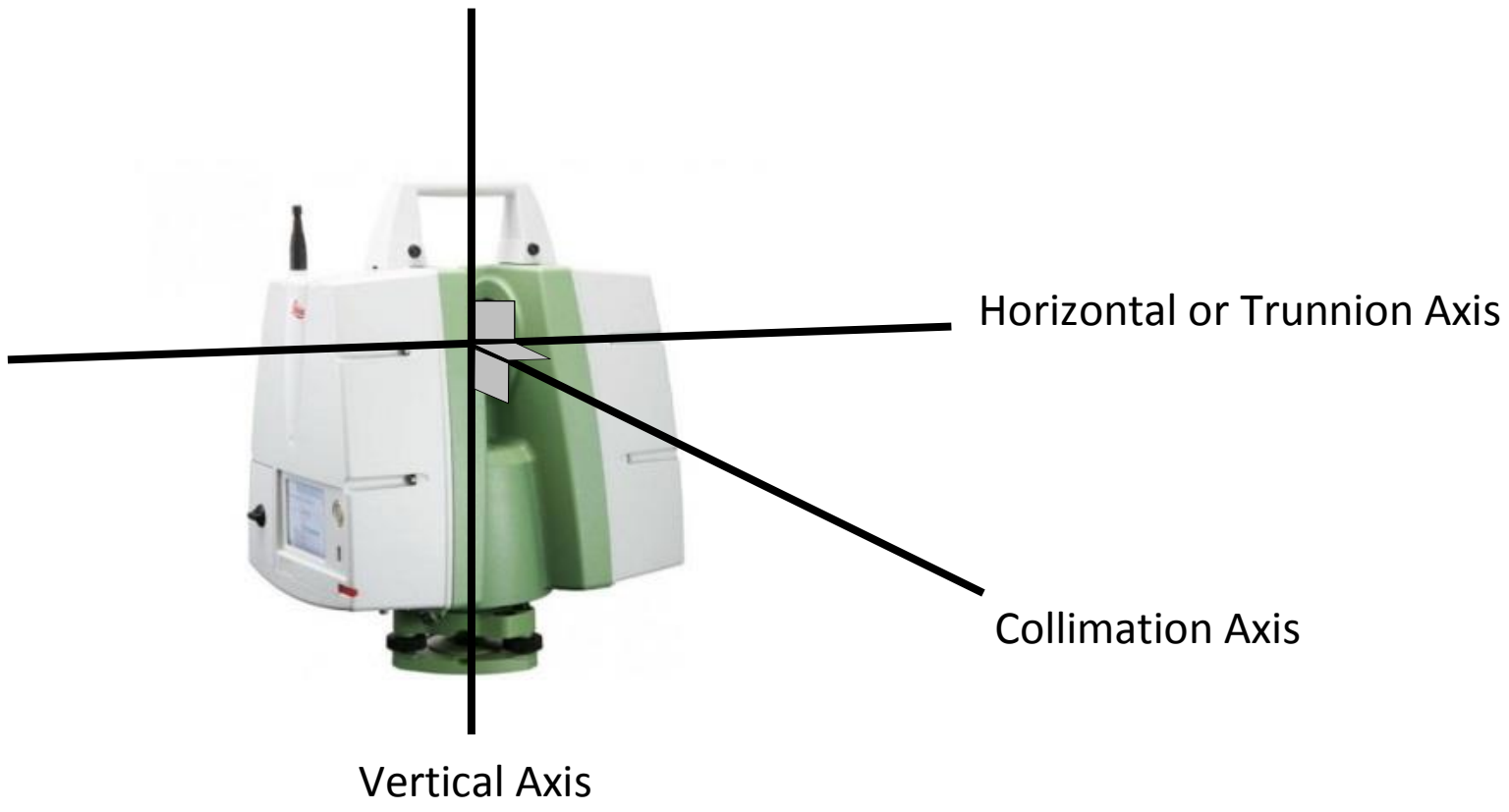

Figure 6: Fundamental axes of terrestrial laser scanners [47]

### *2.1.3 FARO Focus$^{3D}$ S 120*

Motivated by its cost (~CAD$30 000), the FARO Focus$^{3D}$ S 120 (Figure 7) was chosen to be an integral part of the Scannect. This survey-grade 3D terrestrial laser scanner is a phase-based panoramic-type scanner equipped with a built-in tilt sensor, barometer, and magnetic compass. The manufacturer's specifications for the Focus$^{3D}$ S are given in Table 1.

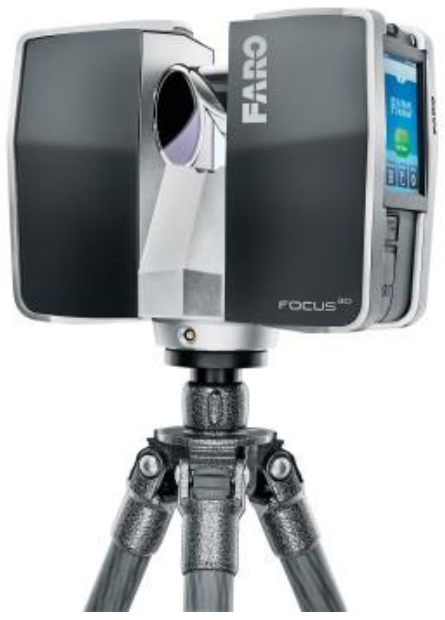

Figure 7: FARO Focus$^{3D}$ S 120 terrestrial laser scanner [48]

Table 1: Technical specifications of the FARO Focus$^{3D}$ S 120 [48]

| | |
|---|---|
| **Architecture** | Panoramic |
| **HFOV/VFOV** | 360°/305° |
| **Range measurement principle** | Phase-based |
| **Maximum Scan rate** | 976 000 Hz |
| **Unambiguity interval** | 153.49 m |
| **Spot size** | 3.8 mm + 0.16 mrad |
| **Range precision @ 25 m (90% albedo)** | 0.95 mm (0.50 mm) |
| **Range accuracy @ 25 m (90% albedo)** | ± 2 mm |
| **Weight** | 5 kg |
| **Size** | 240 x 200 x 100 mm |
| **Operating temperature** | 5°C – 40°C |
| **Levelling sensor** | Dual-axis tilt sensor |
| **Heading sensor** | Electronic compass |
| **Height sensor** | Barometer |
| **RGB** | Built-in camera |

## 2.2 3D Range Cameras

3D range cameras are a relatively new active imaging modality that is gaining popularity in robotics, automotive, gaming, surveillance, and medical applications. Companies such as Mesa, PMDTechnologies, Canesta, Microsoft, Panasonic Electric Works, Fotonic, and SoftKinetic are all manufacturing time-of-flight (TOF) 3D range cameras (Figure 8). These cameras illuminate the scene using near-infrared light and have a special CCD/CMOS array that can demodulate the reflected light at every pixel. By measuring the phase difference between the emitted and received amplitude-modulated light at every pixel location, the 3D geometric information of the scene can be captured instantaneously in a single exposure. There is another category of 3D range cameras that is based on the triangulation principle instead; examples from this category include the Microsoft Kinect, Asus Xtion Pro, and Fotonic P70. Some of the advantages and disadvantages of 3D range cameras over traditional photogrammetric systems and TLS instruments are listed below.

Advantages:

- *3D Data capture at video frame rates:* 3D geometric information can be captured at up to 100 Hz without any scan time delay.
- *Active sensor:* There is no correspondence issue as the camera illuminates the scene and acquires dense 3D geometric information in a single exposure.

Disadvantages:

- *Short maximum distance:* The maximum unambiguous range (typically under 10 m) of a phase-based 3D range camera is limited by the modulation frequency. Maximum range of a triangulation-based 3D range camera is restricted by the baseline separation between the emitter and receiver.
- *Small FOV:* The largest angular FOV for 3D cameras currently on the market is 90°.
- *Low measurement accuracy:* Even in close-range applications, only centimetre-level accuracy should be expected for each pixel due to systematic errors and a low signal-to-noise ratio.

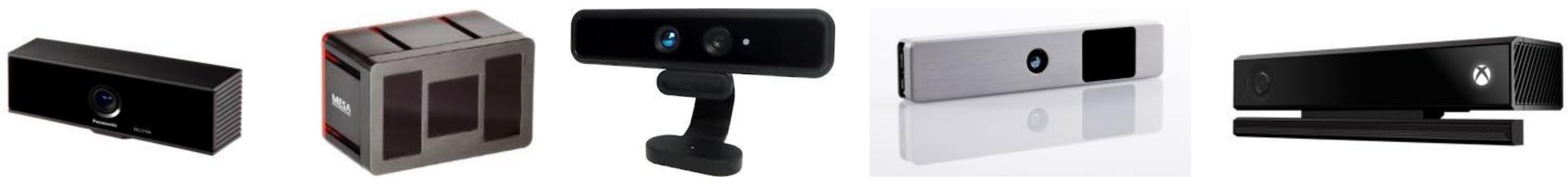

(a) (b) (c) (d) (e)

Figure 8: State-of-the-art 3D TOF Range Cameras – a) Panasonic D-IMager; b) Mesa Image SwissRanger SR4500; c) SoftKinectic DepthSense® 325; d) PMD CamBoard pico; e) Microsoft Kinect 2 [49] [50] [51] [52] [53]

Although 3D range cameras will not be replacing TLS systems at the current stage of technology, they can certainly complement each other. The most obvious drawbacks of 3D range cameras in surveying and mapping applications are their centimeter-level measurement accuracy and shorter unambiguous range measurement interval. However, their ability to acquire 3D information of the scene instantaneously at video frame rate makes them more suitable for measuring dynamic scenes. In addition, they can either serve as a deadreckoning navigation sensor or assist an IMU by supplying position, orientation, and velocity updates. As mentioned, one of the objectives of this thesis is to design a low-cost system; therefore, the Microsoft Kinect was chosen for its economic benefits over other 3D cameras (which are approximately 3 to 50 times more expensive). Hence, the remainder of this section will focus on the Kinect.

### *2.2.1 Systematic Errors in Cameras*

Manufacturing imperfections can result in radiometric and geometric systematic distortions. For 3D reconstruction and navigation, the geometric accuracy is perhaps more crucial. In both computer vision and photogrammetry, cameras are typically assumed to obey the pin-hole camera model. However, in order to focus the light, compound lenses are often installed, and they are rarely crafted and assembled flawlessly. The most common systematic errors found in modern day digital cameras due to the aforementioned manufacturing limitations are radial lens distortion and decentring lens distortion (their corrections are part of the APs). The important IOPs required for describing metric cameras include the principal point offsets and principal distance. The former can give the position in the image plane where it intersects with the optical axis and the latter describes the separation between the image plane and the perspective centre. All these parameters together are sufficient to describe the behaviour of bundles of light rays as they enter the camera at the moment of exposure. For expensive survey cameras, special calibration tools and procedures may be used by the manufacturer to calibrate the camera. However, most low-cost amateur cameras are found off-the-shelf, with their main audience being consumers seeking aesthetically-pleasing artistic photos. Therefore, the users are often responsible for their own camera calibration if they are to be used for mapping purposes. Among the various methods, the bundle adjustment with self-calibration approach is one of the most widely adopted. With slight modifications, it is also applicable to modern 3D cameras. When calibrating multi-camera systems like the Kinect, the ROPs, IOPs, as well as the APs need to be considered. For more information about the mathematical equations for performing bundle adjustment with self-calibration using the Kinect, please refer to Chapter 4.

### *2.2.2 Triangulation-based 3D Cameras: Microsoft Kinect*

In this project, the Microsoft Kinect system designed by PrimeSense (Figure 9), which is a triangulation-based 3D range camera, was utilized. Besides using it to map the scene, it was used alongside with the IMU to provide position and orientation information to the scanner. The 3D geometric data streamed by the Kinect system can be used to perform resection and estimate the change in position and orientation of the camera at each epoch using the iterative closest point algorithm. The 2D colour image can assist the resection and potentially reduce, if not eliminate, the need for large degrees of overlap in the point cloud [21][22]. The Kinect system is composed of a multi-array microphone, an RGB video camera, and a near-infrared projector coupled with a near-infrared camera (which together act as a depth sensor). The raw vision outputs of the Kinect include the infrared (IR) image, depth image, and RGB image (Figure 10). The depth sensor determines the depth at every pixel based on photogrammetric intersection like coded light (a.k.a. structured light) systems. The accuracy of this approach relies heavily on the accuracy with which the lever-arm (relative 3D translation) and boresight (relative 3D rotation) parameters are known. The depth map can be converted to a point cloud when the IOPs are known. If the lever-arm and boresight between the IR camera and RGB camera are also known, the point cloud can be texturized (Figure 11). The specifications of the Kinect's imaging sensors, which are the emphasis of this project, are given in Table 2.

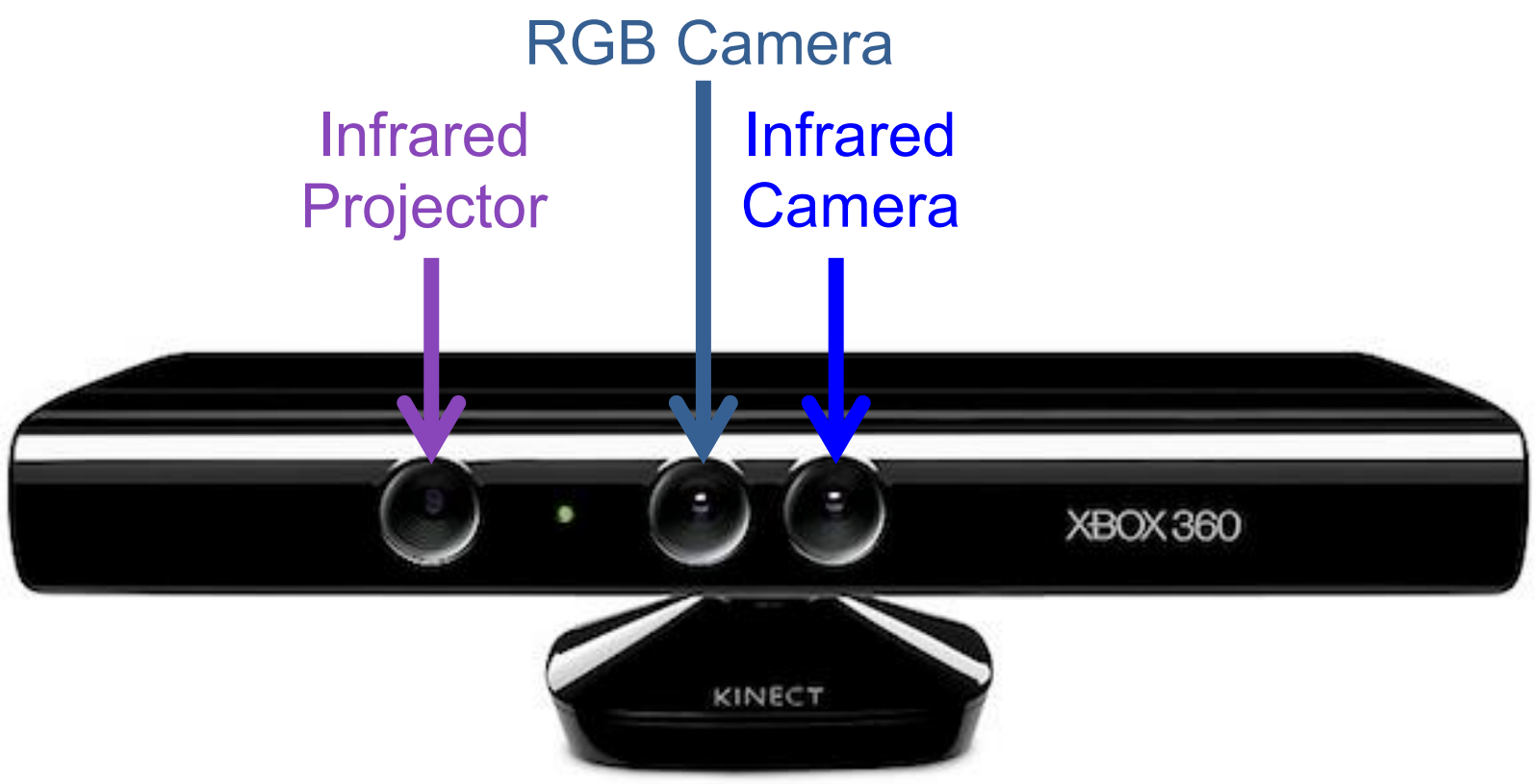

Figure 9: The Microsoft Kinect system [54]

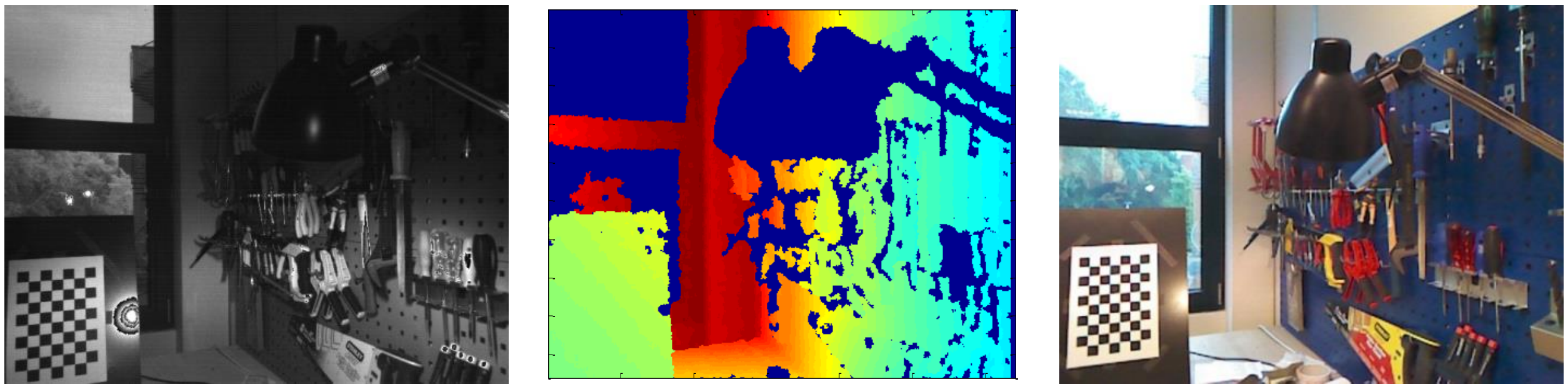

Figure 10: Raw output images from the Kinect - (left) IR image, (middle) depth image, (right) RGB image

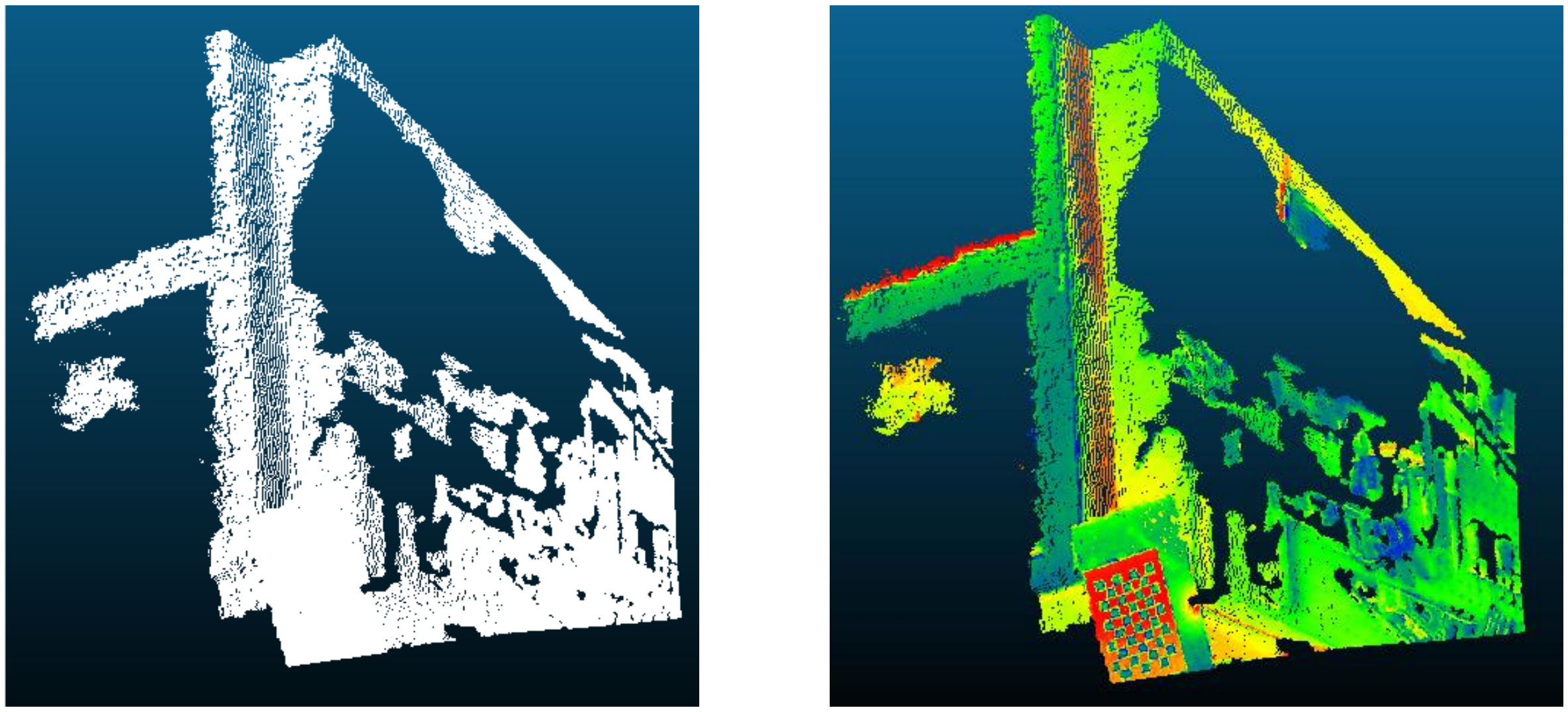

Figure 11: (Left) 3D point cloud captured by the Kinect using IR camera and projector. (Right) 3D point cloud texturized using the RGB camera

Table 2: Technical specifications of the Microsoft Kinect cameras [55]

| | |
|---|---|
| **HFOV/VFOV** | $57^o/43^o$ |
| **Depth sensor range** | 1.2 m – 3.5 m |
| **Depth sensor array dimensions** | 640 x 480 pixels |
| **RGB sensor array dimensions** | 640 x 480 pixels |
| **Depth resolution** | 11-bit depth |
| **RGB radiometric resolution** | 8-bit 3 channels |
| **Approximate baseline between IR projector and IR camera** | 7.5 cm |
| **Approximate baseline between IR camera and RGB camera** | 2.5 cm |
| **Maximum Frame rate** | 30 Hz |

The suitability of the Kinect for geomatics applications was tested and reported in the publication provided in Appendix A. Based on the results, the Kinect system was chosen over TOF 3D range cameras for several reasons:

1. *Low cost:* It is an economical, off-the-shelf, consumer-grade 3D range camera that costs approximately CAD$150. Even though it was originally designed for gaming, it has demonstrated high geometric accuracy in some preliminary testing.
2. *Free of internal scattering errors:* It does not experience any scene-dependent internal scattering, which is a problem found in many TOF 3D range cameras. Most TOF 3D range cameras experience scene-dependent range biases where the light returning from foreground objects reflect internally inside the camera, resulting in biased range measurements to the background objects [56].
3. *Colour information:* Most TOF 3D cameras, unlike the Kinect, do not come with a built-in calibrated colour camera for increased semantic information.
4. *Independent of radiometric properties and incidence angle:* Empirical testing has shown that the Kinect's depth measurement accuracy is less dependent on the colour of the object and the incidence angle when compared to TOF 3D cameras.

As demonstrated in [57] the Kinect system is suitable for indoor mapping applications. But before the Kinect system can be used to map a scene for geomatics purposes or used as a navigation aid, it needs to be calibrated. Systematic errors have been identified in the Microsoft Kinect system and various calibration methods can be found in the literature [58] [59] [60]. These calibration models have their shortcomings: [58] did not estimate the lens distortions in the depth image; [59] did not account for the projector's lens distortions and co-registration error of the RGB and depth images; and [60] did not calibrate for the misalignments between the projector and infrared camera. More information about the new Kinect photogrammetric self-calibration method developed in this project will be discussed in Chapter 4. The proposed approach can calibrate the APs, IOPs, lever-arms, and boresight angles between the projector, infrared camera, and RGB camera simultaneously (along with estimating all their variance-covariance information).

### 2.3 Inertial Measurement Units: Xsens MTi IMU

Inertial measurement units (IMUs) are a type of accurate deadreckoning system capable of delivering the navigation states (position, orientation, and velocity) at high frequencies (e.g. up to 2 kHz) in all environments, even underground. IMUs do not require direct line-of-sight to any reference beacons/markers and can deliver a reliable navigation solution regardless of the scene being imaged. The sensor itself consists of a triad of accelerometers and a triad of gyroscopes. The accelerometers measure the acceleration of the unit in three orthogonal directions and the gyroscopes measure the rotation rate in three orthogonal directions. After the accelerometer measurements are transformed into the local level frame and the effect of local gravity is removed, the first integration gives the change in velocity while the second integration gives the change in position. The change in orientation comes from a single integration of the gyroscope readings after the rotation rate of the earth is removed. For more details about the mathematical formulation please refer to Chapter 5.

There are many different classes of IMUs and they can be broadly categorized from descending order of accuracy as navigation-grade, tactical-grade, and MEMS. The accuracy and cost comparisons between these three grades of IMUs are presented in [61]. While navigation-grade IMUs can cost more than CAD$90 000, their gyro turn-on bias is negligible and their ARW is usually less than $10^{o}/h/\sqrt{Hz}$. Although MEMS IMUs are the least accurate out of these three

grades (e.g. their gyro turn-on bias can be thousands of degrees per hour with an ARW greater than $200^{o}/h/\sqrt{Hz}$), they are the most portable and lowest in cost. As a result of the advancements in the fabrication of silicon micro-machining techniques and new IMU-processing algorithms for strapdown systems, the popularity of MEMS IMUs has grown rapidly over the years. In this project, the MTi MEMS IMU developed by Xsens was used for supplying the navigation states. Some specifications of the Xsens MTi are given in Table 3.

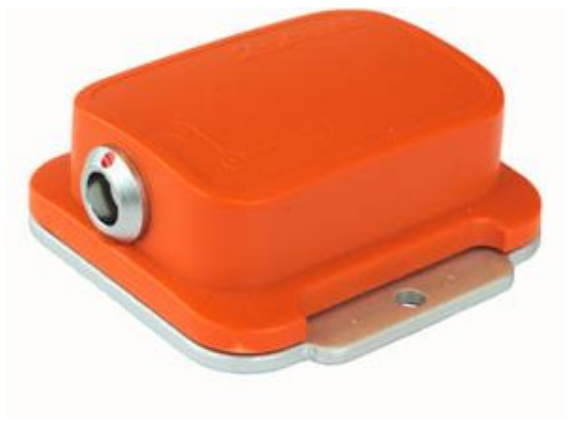

Figure 12: Xsens MTi MEMS IMU [62]

Table 3: Technical specifications of the Xsens MTi [62]

| | |
|---|---|
| **Accelerometer noise** | $0.002\ m/s^2/\sqrt{Hz}$ |
| **Accelerometer bias instability** | $0.02\ m/s^2$ |
| **Gyroscope noise** | $0.05\ ^{o}/s/\sqrt{Hz}$ |
| **Gyroscope bias instability** | $1\ ^{o}/s$ |
| **Maximum update rate** | 512 Hz |
| **Weight** | 50 g |
| **Dimensions** | 58 x 58 x 22 mm |

Due to the nature of numerical integration in IMU-processing, errors in the navigation states can accumulate quite rapidly over time. With low-cost MEMS IMUs, the accumulation of errors is even more prominent than higher quality IMUs. To control the error growth in the navigation states, techniques such as ZUPTS, CUPTS, and AUPTS can be adopted. Also, navigation states coming from other sensors, such as LiDAR, 2D cameras, and 3D cameras can be used to update the IMU solution and help reduce the drift. The IMU solution can be further improved by doing backward smoothing following forward filtering.

### *2.3.1 IMU Self-Calibration*

Like any other manmade instrument, IMUs are inevitably prone to systematic errors. The most noticeable systematic errors in the triad of accelerometers and gyroscopes are the bias, scale factor error, and axes non-orthogonality. In addition, the triad of accelerometers may not be parallel with the triad of gyroscopes. Different methodologies have been adopted to calibrate the sensor; for example the accelerometer bias, scale factor, and non-orthogonality error can all be estimated using the modified six-position static acceleration test explained in [63]. The mathematical models for performing the self-calibration of accelerometers and gyroscopes are given in Equations 3 and 4.

$$\begin{aligned} l_{gx} &= b_{gx} + (1+s_{gx})g_x \\ l_{gy} &= b_{gy} + (1+s_{gy})\left(-g_x \sin\theta_{gyz} + g_y \cos\theta_{gyz}\right) \\ l_{gz} &= b_{gz} + (1+s_{gz})\left(g_x \sin\theta_{gyz} - g_y \sin\theta_{gzx}\cos\theta_{gyz} + g_z \cos\theta_{gzx}\cos\theta_{gzy}\right) \end{aligned} \tag{3}$$

where $l_{gi}$ is the reading of the accelerometer triad's $i$ axis ($i = x, y$, or $z$)
$b_{gi}$ is the bias of the accelerometer triad's $i$ axis
$s_{gi}$ is the scale factor error of the accelerometer triad's $i$ axis
$\theta_{gij}$ is the non-orthogonality between the $i$ and $j$ axes of the accelerometer triad ($j = x, y$, or $z$, and $i \neq j$)
$g$ is the local gravity

$$
\begin{aligned}
\omega_x &= \frac{l_{\omega x} - b_{\omega x}}{1 + s_{\omega x}} \\
\omega_y &= \tan\theta_{\omega zy}\left(\frac{l_{\omega x} - b_{\omega x}}{1 + s_{\omega x}}\right) + \left(\frac{1}{\cos\theta_{\omega yz}}\right)\left(\frac{l_{\omega x} - b_{\omega x}}{1 + s_{\omega x}}\right) \\
\omega_z &= \left(\tan\theta_{\omega zx}\tan\theta_{\omega zy} - \frac{\tan\theta_{\omega zy}}{\cos\theta_{\omega zx}}\right)\left(\frac{l_{\omega x} - b_{\omega x}}{1 + s_{\omega x}}\right) + \left(\frac{\tan\theta_{\omega zx}}{\cos\theta_{\omega zy}}\right)\left(\frac{l_{\omega y} - b_{\omega y}}{1 + s_{\omega y}}\right) \\
&\quad + \left(\frac{1}{\cos\theta_{\omega zx}\cos\theta_{\omega zy}}\right)\left(\frac{l_{\omega y} - b_{\omega y}}{1 + s_{\omega y}}\right)
\end{aligned} \tag{4}
$$

where $l_{\omega i}$ is the reading of the gyroscope triad's $i$ axis ($i = x$, $y$, or $z$)
$b_{\omega i}$ is the bias of the gyroscope triad's $i$ axis
$s_{\omega i}$ is the scale factor error of the gyroscope triad's $i$ axis
$\theta_{\omega ij}$ is the non-orthogonality between the $i$ and $j$ axes of the gyroscope triad ($j = x$, $y$, or $z$, and $i \neq j$)

This method does not require any specialized tools, the IMU just needs to be in different orientations and capture a series of static data. The systematic errors can be estimated by enforcing the condition that, at any static orientation/position, the magnitude of acceleration and the rotation rate sensed by all three axes should be equal to the acceleration of local gravity and the rotation rate of the Earth, respectively. However, this method is only suitable for estimating the systematic errors inherent in the accelerometer, not the gyroscope (except for the gyro bias). The systematic errors for the gyroscope can usually be recovered by using a precision rate table. This is usually better than static calibration because unlike the gravity signal, the rotation rate of the Earth is a weak signal and can be difficult to sense if the gyroscopes have a low signal-to-noise ratio (e.g. MEMS IMU). Therefore, by rotating the IMU at precisely known rates in different directions, the systematic errors in the gyroscope can be easily calibrated.

## 2.4 Simultaneous Localisation and Mapping (SLAM)

In the area of robotics and artificial intelligence, Simultaneous Localisation and Mapping (SLAM) is already a well-studied concept and has reached a state of considerable maturity [64]. As the name SLAM suggests, the theory is to map an unknown environment and estimate the navigation states simultaneously by observing landmarks, as these processes are highly correlated [65]. Successful probabilistic SLAM solutions can allow robots to autonomously explore deep waters, mines, unstable structures, and other dangerous scenarios instead humans. Typically, sensors such as an odometer are equipped for localisation, but mechanical errors in the wheel system and/or slippage is a common problem [66]. Visual-based systems are popular navigation aids for compensating these mechanical errors and the basic form utilizes artificial landmarks [67] [68]. The simplest case, without additional infrastructures uses a single camera and matches interest points detected by Scale-Invariant Feature Transform (SIFT) or Speeded Up Robust Feature (SURF) using RANdom SAmple Consensus (RANSAC). For indoor environments, stereo-vision systems have been applied and features such as points, lines, and planes can be used to aid the navigation solution [69]. If a high density point cloud is available from laser scanning, ICP can be used with SLAM to avoid the need of feature extraction [70] [71] [72]. Recently, the Microsoft Kinect has been used in RGB-D SLAM and showed promising results [73]. However, the success of visual-based systems depends on the scene being imaged. Visual-based systems are vulnerable to scenes that contain high dynamics; for example, having people walking around can make extracting and matching features between successive frames difficult [74]. The landmarks being tracked in adjacent image frames can disappear [57]. During these visual system outages, other sensors such as an IMU (which does not require clear lines-of-sight) can be used to provide accurate navigation states [75]. The estimation of the navigation states are usually achieved using a Kalman filter [76] [77], although other estimation techniques such as the particle filter have gained noticeable attention in recent years [78]. More information about modern SLAM solutions using RGB-D cameras can be found in Chapter 5.

# Chapter Three: Systematic Error Modelling for the FARO Focus$^{3D}$ S 120 Terrestrial Laser Scanner via the Self-Calibration Approach

With the emergence of low-cost survey-grade 3D TLS instruments, the potential of these sensors for accurate 3D reconstruction has drawn more attention from an increasing number of communities. However, seldom does the user take special precaution in calibrating their own scanner; despite the fact that many users have reported geometric accuracy deterioration in their acquired data. The presented article below studies the effectiveness of using different primitives (i.e. points and planes) for a multi-station self-calibration. This paper argued that plane-based calibration was less time-consuming (more cost-effective) and demonstrated through simulation and empirical data that with a strong network configuration it can compensate for the systematic errors similar to the point-based approach.

Using this method, the FARO Focus$^{3D}$ S was calibrated and the results are given in Appendix B. The Focus$^{3D}$ S tested showed higher measurement noise than the more expensive Leica HDS6100 even after calibration. The trunnion axis error and horizontal collimation axis error were significantly more prominent, but were successfully modelled using the self-calibration approach; improvement in angular measurement precision up to 60% was found. Post-calibration, the Focus$^{3D}$ S was able to deliver similar results (sub-millimetre differences) as the HDS6100 at close range (i.e. less than 10 m). For the price tag and specifications listed in Table 1, the results of the Focus$^{3D}$ S showed promise for applications in geomatics.

## 3.1 Article: Improvements to and Comparison of Static Terrestrial LiDAR Self-Calibration Methods

Jacky C.K. Chow[1,*], Derek D. Lichti[1], Craig Glennie[2] and Preston Hartzell[2]

1 Department of Geomatics Engineering, The University of Calgary, 2500 University Drive N.W., Calgary, AB T2N 1N4, Canada; E-Mail: ddlichti@ucalgary.ca

2 Department of Civil and Environmental Engineering, The University of Houston, N107 Engineering Building 1, Houston, TX 77204-4003, USA; E-Mails: clglennie@uh.edu (C.G.); pjhartzell@uh.edu (P.H.)

* Author to whom correspondence should be addressed; E-Mail: jckchow@ucalgary.ca; Tel.: +1-403-210-9741; Fax: +1-403-284-1980.

Abstract: Terrestrial laser scanners are sophisticated instruments that operate much like high-speed total stations. It has previously been shown that unmodelled systematic errors can exist in modern terrestrial laser scanners that deteriorate their geometric measurement precision and accuracy. Typically, signalised targets are used in point-based self-calibrations to identify and model the systematic errors. Although this method has proven its effectiveness, a large quantity of signalised targets is required and is therefore labour-intensive and limits its practicality. In recent years, feature-based self-calibration of aerial, mobile terrestrial, and static terrestrial laser scanning systems has been demonstrated. In this paper, the commonalities and differences between point-based and plane-based self-calibration (in terms of model identification and parameter correlation) are explored. The results of this research indicate that much of the knowledge from point-based self-calibration can be directly transferred to plane-based calibration and that the two calibration approaches are nearly equivalent. New network configurations, such as the inclusion of tilted scans, were also studied and prove to be an effective means for strengthening the self-calibration solution, and improved recoverability of the horizontal collimation axis error for hybrid scanners, which has always posed a challenge in the past.

## 1. Introduction

Three-dimensional point clouds from terrestrial laser scanning (TLS) instruments are a valuable asset for a variety of real-world problems. Beside their traditional use in areas such as digital terrain modelling, geological exploration, surveying, archeological and architectural documentation, they have found applications in forensics and crime scene investigations, deformation monitoring of dams, buildings and landslides, and more recently for scanning movie sets. For many applications of TLS, it is assumed that the point cloud is error-free. Although this is far from reality, numerous efforts have attempted to produce more reliable and accurate point clouds. New and improved filters and measurement algorithms have limited a great deal of outliers and noise in the point cloud. Typical TLS projects involve grouping common points to model a surface or object to reduce the effect of random noise. Modern TLS instruments behave much like total stations; however their systematic errors cannot be eliminated by taking direct and reverse measurements. Instead, users usually have to rely on the manufacturer's black box instrument calibration. The manufacturer's calibration has been shown to be an effective method for improving the scanner's observation precision, but it has also been demonstrated that point-based user self-calibration can further reduce the standard deviation of the scanners' observations [1].

Some noticeable advantages of the TLS point-based self-calibration method include:

- Manufacturer-independent total system error modelling.
- Frequent and rapid calibration at the user's convenience for quality assurance.
- A common basis for comparing the measurement precision of different scanners.

In the past, point-based self-calibration has proven to be successful at identifying and modelling systematic errors in a variety of TLS instruments [2–5]. It can also be executed efficiently once the point clouds are captured; for example the point targets can be extracted and matched automatically (e.g., Leica Cyclone) and the selection of the relevant error model parameters can be made with minimal user interaction [6]. However, its largest drawbacks are the time and cost required to set up a large quantity of signalised targets, and the poor recovery of the horizontal collimation axis error in hybrid-type scanners. Self-printed paper targets could substitute for the expensive laser scanner targets [7], but it might still be a challenge to install targets in large rooms [8] despite the fact that they are beneficial to the calibration [9].

An alternative is to replace signalised targets with well-defined geometric features that are commonly encountered by the user. In [10] the authors presented mathematical models for simultaneous registration and modelling of objects that are frequently encountered at industrial sites. These mathematically well-defined surfaces include planes, cylinders, tori, and spheres. This paper concentrates on the utilization of planes for performing TLS self-calibration. The main motivations of this paper are: (1) to verify whether or not point-based and plane-based TLS self-calibration produce equivalent results; and (2) to propose new network configuration measures (e.g., the inclusion of tilted scans) to enhance the quality of TLS calibration. Special attention is placed on the precise estimation of the collimation axis error for hybrid-type scanners that has proven difficult to resolve in the past [3,8,9,11].

This article is organized as follows: in Section 2, previous research on TLS error modelling is reviewed. The functional and stochastic models used for carrying out the point-based and plane-based error modelling are explained in Section 3. Section 4 shows simulated calibrations using the point-based and plane-based methods. Cases where only levelled scans are used and when levelled and tilted scans are used in conjunction are analyzed in terms of the standard deviation and correlation of the systematic errors. Finally, Section 5 shows point-based and plane-based calibration results from TLS instruments built by Riegl, Z+F, Leica, and Trimble.

## 2. Previous Work

Unmodelled systematic distortions have been identified in TLS instruments from different manufacturers having unique internal components and architectures [2,12–15]. Previous calibration attempts have modelled the scanner based on the panoramic camera model [16], theodolites [17] and total stations [18,19]. The two former approaches are restricted to modelling systematic errors in the angular measurements and only the last method can simultaneously account for systematic errors in distances and directions. This desirable trait makes the third option the preferred approach.

Distance measurements in survey grade TLS instruments are achieved by measuring the two-way travel time of laser pulse(s) or the phase difference of amplitude modulated continuous light waves. Many of the errors in the Electronic Distance Measurement Instruments (EDMI) are well known and have been shown to be applicable to laser scanners [20]. To determine the 3D position of a single point, the laser needs to be redirected in two orthogonal directions. This can be achieved by deflecting the laser beam using a mirror device (e.g., oscillating, rotating, or polygonal mirror), physically rotating the scanner head, or using multiple rangefinders with fixed angular separations. The methodology used for sweeping the laser over the scene has a strong influence on the resulting field of view (FOV) of the TLS instrument. For example, scanners with oscillating mirrors (e.g., Trimble GX and Leica Scanstation2) usually have a much smaller vertical FOV than scanners using rotating mirrors (e.g., FARO Focus$^{3D}$, Z+F Imager 5010, Leica P20, and Trimble TX8).

According to [21], TLS instruments can be broadly categorized into hybrid-type and panoramic-type based on their FOV. The calibration of panoramic-type scanners is better understood because it allows data to be acquired in two layers (*i.e.*, in front of and behind the unit). This is comparable to total stations observing in both the face-left and face-right orientations; the end result is parameter de-correlation and better model identification. In [18] the systematic errors in a panoramic-type scanner, Faro 880, were studied in great detail and an extensive 17 parameter error model with both physical and empirical terms that is independent of the scanner architecture was presented. This method solves for the exterior orientation parameters (EOPs), object space target coordinates, and additional parameters (APs) simultaneously in a free station network adjustment, rather than solving for the systematic errors one at a time [17].

Further analyses in [8] identified and explored the correlation of APs with other parameters in self-calibration. It has been shown that problems with inflated correlation between parameters are largely mitigated by having independent measurements of the EOPs (e.g., levelling the scanner using inclinometer measurements). Levelling information is valuable — even if its quality is low — for the recovery of the vertical circle index error. Parameter de-correlation is most important for calibrating hybrid-type scanners, which suffer from stronger dependencies than panoramic scanners due to their data acquisition pattern (*i.e.*, data can only be captured in one face). This has been independently confirmed by [19] who studied the self-calibration of many different hybrid-type TLS systems. Instead of defining the datum using inner constraints, which is known to have the negative effect of giving higher parameter correlation [22], all parameters are treated as observations in a unified least-squares adjustment in [19]. Although the inclusion of independent observations of the EOPs can help de-correlate many of the parameters, it also can magnify the complexity of the calibration procedure and diminishes its ease of use. For example, to recover the rangefinder offset accurately, the position of the scanner needs to be measured with accuracy better than one millimetre. In practice, this is difficult to achieve unless stable pillars with known coordinates are used. Nonetheless, [9,19] have clearly explained the limitations of the current methodologies to solve for the horizontal collimation axis error, which is perfectly correlated with the tertiary rotation angle for hybrid-type scanners. High parameter correlation can pose a threat to data integrity if the scanner is not calibrated *in-situ* [23]. In [11], the correlation between the collimation axis error and the heading angle of the scanner was reduced through the introduction of a new mathematical model for the AP as well as approximating the relative heading angle between scans. The correlation was successfully reduced but the standard deviation of the recovered systematic error was still higher than expected (*i.e.*, in the order of arc minutes). To date, a solution to this issue has not yet been tested thoroughly and reported.

Another approach to laser scanner calibration focuses on minimizing the distance between measured points and a well-defined mathematical surface. In general, any geometric features that can be used for registration can be extended for self-calibration [10]. However, most calibration routines have focused on the use of planes, likely because of their simple mathematical representation and abundance in urban settings. This concept was first presented by [24] for calibrating their in-house laser scanner. In [25] it was studied further as an extension of their point-based calibration model for panoramic-type scanners. Through simulation they investigated different scanner configurations for performing plane-based calibration and suggested that a long baseline is helpful for recovering the collimation axis error when only two scans are acquired. [26] showed through simulation that the impact of the four fundamental systematic errors (*i.e.*, rangefinder offset, vertical circle index error, trunnion axis error, and horizontal collimation axis error) on the measurement residuals is very similar between the point-based and plane-based calibration. They also indicated some subtle differences between the two methods; in particular, the correlation between APs and EOPs needs to be further studied.

The Velodyne HDL-64E S2 scanner was calibrated using planar-features and this improved the geometric quality of the point cloud by a factor of three [27]. The inclusion of tilted scans in the adjustment was tested in [27,28] for reducing the correlation between horizontal offset and horizontal angular offset as well as vertical offset and vertical angular offset in the Velodyne calibration. [11] also tried tilting the Trimble GS200 by a small amount (<10°) and reported minor improvements in recovery of the horizontal collimation axis error. The use of planar features for calibrating laser scanners can also be found in airborne laser scanning. The authors of [29] presented the plane-based calibration results of a mobile scanning system mounted on a helicopter and have indicated the benefits of having tilted planes with various orientations for recovering the boresight angles and rangefinder offset. In [30] planes were used to solve for the internal calibration parameters and boresight angles of a mobile terrestrial scanner (Velodyne). Other geometric features such as cylinders and catenaries have also been used for calibrating static terrestrial laser scanners [31] and mobile terrestrial laser scanners [32], respectively.

## 3. Mathematical Model

The 3D Cartesian coordinate of every point in a TLS point cloud is determined from a distance observation and two angular observations. The conversion between the spherical coordinate system, in which the observations are made, and the Cartesian coordinate system, in which the point cloud is typically defined, is given by Equation (1). Every point is uniquely determined in TLS, and therefore the same point needs to be observed more than once to achieve any redundancy. Having redundancy in the solution is crucial for quality assurance purposes when registering scans and calibrating the scanner. This is rather difficult to achieve in TLS because the points are irregularly spaced and in general there are no point-to-point correspondences between scans. The conventional approach is to measure multiple points on the surface of a signalised target and calculate the centre of the target. Corresponding target centroids are used to relate scans captured from different positions and orientations, usually via a 3D rigid body transformation (Equation (2)). For the point-based registration model, one can substitute Equation (1) into Equation (2) and estimate the EOPs, which ensure best fit between the targets in object space. In other words, the 3D volume of uncertainty at every object space target position is minimized.

Alternatively, geometric features can be used to solve the correspondence problem. Features such as planes, spheres, tori, cylinders, pyramids, and catenaries can be used for registration. If planes are used, the orthogonal distance between every point observed on the plane is minimized (Equation (3)). The datum in both point-based and plane-based self-calibrations is defined by enforcing inner constraints on the object space primitives. For the plane-based calibration, the imposition of inner constraints on the plane parameters has been shown to be the preferred datum definition because it mitigates correlation problems [33]. In registration, if a highly redundant strong network of well-distributed targets is used, APs can be appended to the observations to model systematic errors in the scanner. APs can be categorised into physical terms (the source of error is known) and empirical terms (the source of error is unknown). The known physical APs in the range, horizontal direction, and vertical angle measurements are given by Equations (4–6), respectively. In both point-based and plane-based self-calibration, the sum of the weighted squared residuals is minimized while simultaneously solving for the EOPs, APs, and object space target coordinates or feature parameters. Details about the least squares estimation method can be found in [34]:

$$\rho_{ij} = \sqrt{x_{ij}^2 + y_{ij}^2 + z_{ij}^2} + \Delta\rho$$
$$\theta_{ij} = \tan^{-1}\left(\frac{y_{ij}}{x_{ij}}\right) + \Delta\theta \qquad (1)$$
$$\alpha_{ij} = \tan^{-1}\left(\frac{z_{ij}}{\sqrt{x_{ij}^2 + y_{ij}^2}}\right) + \Delta\alpha$$

where $\rho_{ij}$, $\theta_{ij}$, and $\alpha_{ij}$ are the slope distance, horizontal direction, and vertical angle of point i in scanner space j
$x_{ij}$, $y_{ij}$, and $z_{ij}$ are the Cartesian coordinates of point i in scanner space j
$\Delta\rho$, $\Delta\theta$, and $\Delta\alpha$ are the additional parameters for the scanner

$$\begin{bmatrix} x_{ij} \\ y_{ij} \\ z_{ij} \end{bmatrix} = R_3(\kappa_j)R_2(\phi_j)R_1(\omega_j)\left(\begin{bmatrix} X_i \\ Y_i \\ Z_i \end{bmatrix} - \begin{bmatrix} X_{oj} \\ Y_{oj} \\ Z_{oj} \end{bmatrix}\right) \qquad (2)$$

where $X_i$, $Y_i$, and $Z_i$ are the Cartesian coordinates of point i in object space
$X_{oi}$, $Y_{oi}$, and $Z_{oi}$ are the origin of scanner space j in object space
$\omega_j$, $\phi_j$, and $\kappa_j$ are the orientation of scanner space j relative to object space

$$(a_k \quad b_k \quad c_k)\left\{R_3(\kappa_j)R_2(\phi_j)R_1(\omega_j)\begin{bmatrix} (\rho_{ij} - \Delta\rho)\cos(\alpha_{ij} - \Delta\alpha)\cos(\theta_{ij} - \Delta\theta) \\ (\rho_{ij} - \Delta\rho)\cos(\alpha_{ij} - \Delta\alpha)\sin(\theta_{ij} - \Delta\theta) \\ (\rho_{ij} - \Delta\rho)\sin(\alpha_{ij} - \Delta\alpha) \end{bmatrix} + \begin{bmatrix} X_{oj} \\ Y_{oj} \\ Z_{oj} \end{bmatrix}\right\} - d_k = 0 \qquad (3)$$

where $a_k$, $b_k$, $c_k$, and $d_k$ are the plane parameters defining the normal axis and orthogonal distance to the plane

$$\Delta\rho = A_0 + A_1\rho_{ij} + A_2 \sin(\alpha_{ij}) + A_3 \sin\left(\frac{4\pi}{U}\rho_{ij}\right) + A_4 \cos\left(\frac{4\pi}{U}\rho_{ij}\right) + ET_\rho \quad (4)$$

where $A_0$ is the rangefinder offset
$A_1$ is the range scale factor error
$A_2$ is the laser axis vertical offset
$A_3$ and $A_4$ are the cyclic errors
$U$ is the unit length
$ET_\rho$ is the empirical range errors

$$\begin{aligned}\Delta\theta = {} & B_1\theta + B_2 \sin(\theta) + B_3 \cos(\theta) + B_4 \sin(2\theta) + B_5 \cos(2\theta) + B_6 \sec(\alpha)^* \\ & + B_7 \tan(\alpha) + B_8\rho^{-1} + B_9 \sin(\alpha) + B_{10} \cos(\alpha) + ET_\theta\end{aligned} \quad (5)$$

* Note: for hybrid-type scanners, the reduced collimation model is used instead, $B_6[\sec(\alpha) - 1]$

where $B_1$ is the horizontal direction scale factor error
$B_2$ and $B_3$ are the horizontal circle eccentricity
$B_4$ and $B_5$ are the non-orthogonality of horizontal encoder and vertical axis
$B_6$ is the horizontal collimation axis error
$B_7$ is the trunnion axis error
$B_8$ is the horizontal eccentricity of collimation axis
$B_9$ and $B_{10}$ are the trunnion axis wobble
$ET_\theta$ is the empirical horizontal direction errors

$$\begin{aligned}\Delta\alpha = {} & C_0 + C_1\alpha + C_2 \sin(\alpha) + C_3 \cos(\alpha) + C_4 \sin(2\alpha) + C_5 \cos(2\alpha) + C_6\rho^{-1} \\ & + C_7 \sin(\theta) + C_8 \cos(\theta) + ET_\alpha\end{aligned} \quad (6)$$

where $C_0$ is the vertical circle index error
$C_1$ is the scale factor error
$C_2$ and $C_3$ are the vertical circle eccentricity
$C_4$ and $C_5$ are the non-orthogonality of vertical encoder and trunnion axis
$C_6$ is the vertical eccentricity of collimation axis
$C_7$ and $C_8$ is the vertical axis wobble
$ET_\alpha$ is the empirical elevation angle errors

As documented in [8,9,11,19], weighted constraints can be added to the self-calibration to decouple some of the parameters, namely high correlations between the APs and EOPs. They can be easily integrated as linear pseudo-observation equations as shown in Equation (7). For the stochastic model, the three observations are assumed to be independent of each other. Correlations between observations can exist, especially if the observations are within close proximity of each other, which may be the case with newer scanners acquiring even denser point clouds than before. But this assumption of independent observations can simplify the adjustment model and allow the use of mathematical techniques such as the summation of normals [29]. Moreover, for point-based self-calibration this is largely mitigated by the spatial separation of the placement of signalised targets. For plane-based calibration this is more of a concern, but can also be reduced by downsampling the point cloud. The adopted stochastic model for the range and angular observations in this paper is shown in Equation (8). Both angular observations' standard deviations are assumed to be constant, while the range observation's standard deviation changes as a function of the incidence angle. This dependency is mainly due to the enlargement of the laser beam footprint at high incidence angle. Therefore a lower weight is assigned to range observations captured from an oblique angle [35]:

$$\begin{aligned} \omega_j &= \omega_{obs} \pm \sigma_\omega \\ \phi_j &= \phi_{obs} \pm \sigma_\phi \\ \kappa_j &= \kappa_{obs} \pm \sigma_\kappa \end{aligned} \tag{7}$$

$$\begin{gathered} E\left\{\varepsilon_\rho^2\right\} = \sigma_\rho^2 \sec^2(\beta) \\ \mathrm{E}\left\{\varepsilon_\theta^2\right\} = \sigma_\theta^2 \\ \mathrm{E}\left\{\varepsilon_\alpha^2\right\} = \sigma_\alpha^2 \\ \mathrm{E}\left\{\varepsilon_{\theta\alpha}^2\right\} = \mathrm{E}\left\{\varepsilon_{\rho\alpha}^2\right\} = \mathrm{E}\left\{\varepsilon_{\rho\theta}^2\right\} = 0 \end{gathered} \tag{8}$$

where $\beta$ is the incidence angle of the laser on the surface being measured

Variance component estimation (VCE) has been applied to better characterize the relative weights between the three observation groups (*i.e.*, $\rho$, $\theta$, $\alpha$). To reduce the chance of blunders, Baarda's data snooping is performed after the adjustment to identify outliers, which are subsequently removed. These tools have been widely adopted in conventional survey network adjustments and often adopted in point-based TLS self-calibration as well. However, for the plane-based self-calibration only Baarda's data snooping was implemented. It was performed first in scanner space during plane extraction and then again during the plane-based self-calibration. VCE was not used in this case because, as shown in Section 3, many observations have a zero-valued (or near zero) residuals, especially the angular observations. This is a drawback of using planes, because only observations made in the direction orthogonal to the planes are constrained in the adjustment. When considering a small plane whose normal is directed towards the origin of the scanner space, both angular observations are more or less perpendicular to the normal vector of the plane and do not contribute to the estimation of the plane parameters. This drawback can be found in other geometric features too, such as observations along the principal direction of a cylinder. These large quantities of zero residuals can bias the estimated variance components, resulting in optimistic standard deviation estimates for the observations. Therefore, in this paper the observation weights estimated from the point-based self-calibration were used and held fixed in the plane-based self-calibration.

## 4. Simulated Data: Results and Discussion

In this section, the propagation of systematic errors into the residuals of the observations for scanners having both panoramic and hybrid type architectures will be studied. This is rather important because the presented self-calibration method models the systematic error in the scanner's observation space. In [18], the model identification problem was mainly addressed by studying the residual plots of the observations and visualising the systematic trends. However, it has been reported that some systematic errors which cannot be visually identified in the observation residuals can still be modelled correctly; an example is the vertical circle index error and the trunnion axis error in hybrid type scanners [11]. In general, if a systematic error can be visually identified in the observation residuals, it can be recovered with greater confidence.

For this study, the APs presented in Equations (4–6) are tested one at a time in point-based and plane-based self-calibration for hybrid and panoramic scanners, with some exceptional sinusoidal terms being tested in pairs. A 14 m by 11 m by 3 m rectangular room with 120 randomly distributed signalized targets was simulated. This was chosen as a representation of common realistic calibration setups found in literature [11,18]. It has been shown that larger rooms which cover the scanner's minimum and maximum unambiguous range are ideal, but difficult to come across, especially when modern pulse-based TLS instruments can measure up to 6 km (e.g., the Riegl VZ-6000). Since it is known that systematic errors can be recovered better when both the horizontal and vertical angular field of view is maximized, no FOV restriction has been placed on the simulated scan data. This is justifiable as technological developments have continuously increased the FOV of scanners in the past decade — for example, the vertical FOV of the Trimble scanner increased from 60° to 320° (GX to TX8). The geometric arrangement of the targets/planes and scanner setups are shown in Figure 1. Six scans were simulated at two unique positions in the room. At both positions there were three leveled scans offset by 60° about the tertiary axis. A total of 120 points visible from every position were evenly distributed on each of the six planes. For the plane-based calibration, 20 points on each plane is a very low point density, but is done to ensure the same observations as point-based calibration are used. The distance and angular measurement noise was assumed to be 0.5 mm (at 0° incidence) and 20″, respectively. In the case where tilted scans were tested, two of the levelled scans from one of the corners were set to be tilted by −45° and +45°. Since the focus of this simulation is on model identification, the minimum number of scans/targets required for calibration is not addressed in this paper. All simulated range APs are 10 mm in magnitude with the range scale factor error being exaggerated to 2. The angular systematic errors are set to a magnitude of 3’, except for $B_8$ and $C_6$ which were 10 mm.

**Figure 1.** Network configuration of the simulated room: (left) top view and (right) oblique view. The blue line indicates the scanner’s z-axis and magenta line indicates the scanner’s x-axis (note that three scans were simulated at each of the two positions).

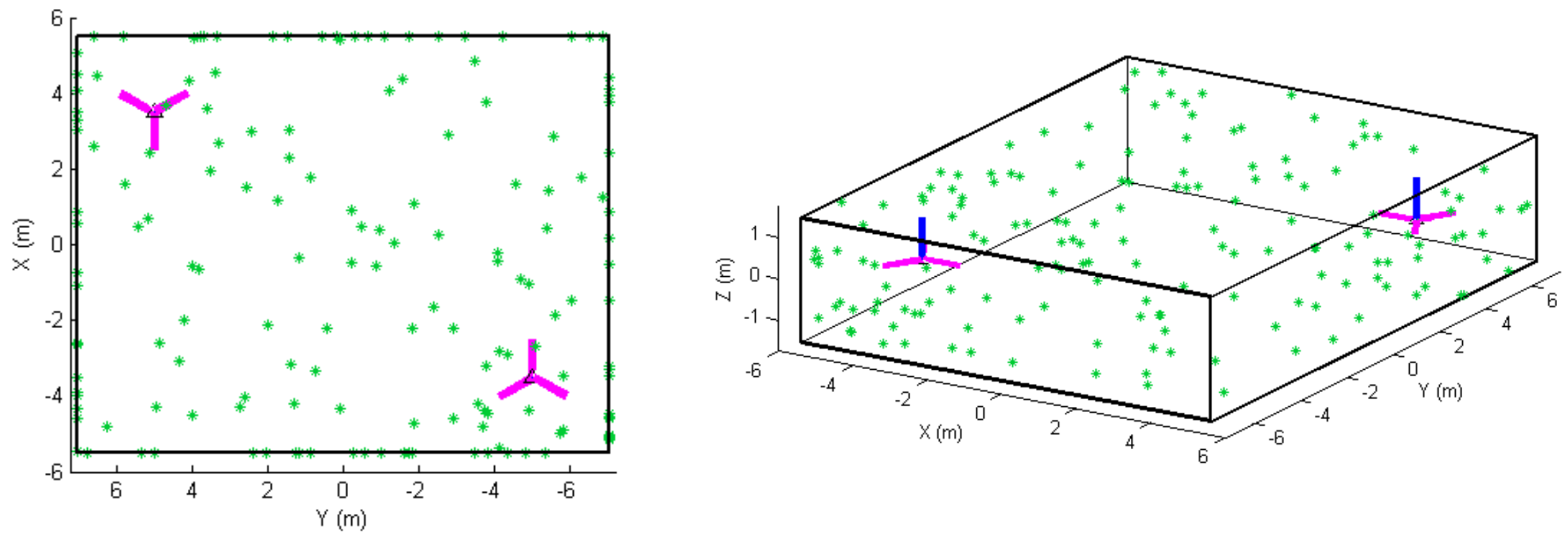

Figures 2–4 show the effect of the systematic error terms in distance, horizontal direction, and vertical angle observations respectively (as presented in Equations (4–6)) for hybrid scanners using point-based self-calibration. Similar plots showing the propagation of systematic errors into the observation residuals for panoramic type scanners are shown in Figures 5–7. Whenever a pattern in the residuals is visually identifiable and follows the mathematical model, a curve has been superimposed. Based on graphical analysis of these observation residual plots it is more difficult to identify the systematic error terms for scanners that can only observe in one face (*i.e.*, hybrid scanners); in Figure 4 only the trend of $C_7$ and $C_8$ can be visually identified while in Figure 7 trends for all APs can be identified. The majority of errors in the vertical direction cannot be identified by visual inspection of the plots in the case of hybrid scanners. Typically, error modelling of hybrid type scanners is more based on a trial-and-error approach with statistical tests and RMSE of residuals indicating parameter significance. The appearance of $C_0$ in the residuals for panoramic-type scanners is slightly different from the plot reported in [11]. This difference exists because its exact pattern is dependent on the network geometry. It is worth mentioning that the $C_3$ term was included in the error model, despite the fact that it was omitted in [18] due to its high correlation with other parameters. It will be shown later that this correlation issue can be mitigated with additional observations and/or tilting the scanner.

The residual plots from the same point clouds processed using the plane-based self-calibration are shown in Figures 8–10 for hybrid-type scanners, and Figures 11–13 for panoramic-type scanners. Points that are observed with a small incidence angle give rise to many zero-valued residuals in the angular measurements, which is an undesirable trait of plane-based self-calibration. These observations bias the statistics computed from the residuals and do not provide useful information for systematic error identification/modelling. Geometrically speaking, planar features can only constrain points in 1D, while point features are well-controlled in 3D. Regardless of the number of points in the point cloud, the average redundancy never exceeds 33% due to the nature of planes. The rangefinder offset is highly dependent on the incidence angle, and can only be measured from points observed with a small incidence angle [26]. When studying the internal reliability of each observation, it was clear that for the range observations the individual redundancy number decreases with increasing incidence angle and distance to target, while the opposite holds true for the angular observations. The $B_6$ term is completely hidden for hybrid-type scanners under both calibration routines. In this network configuration, some APs (e.g., $B_7$ and $B_9 + B_{10}$ for hybrid scanners, and $B_6$ and $B_9 + B_{10}$ for panoramic scanners) gave rise to similar trends in the residual plots and are difficult to distinguish.

**Figure 2.** Residual plots of systematic errors in the range observations ($\rho$) from point-based self-calibration for hybrid-type scanners: $A_0$ is the rangefinder offset, $A_1$ is the range scale factor error, $A_2$ is the laser axis vertical offset, and $A_3 + A_4$ are the cyclic errors.

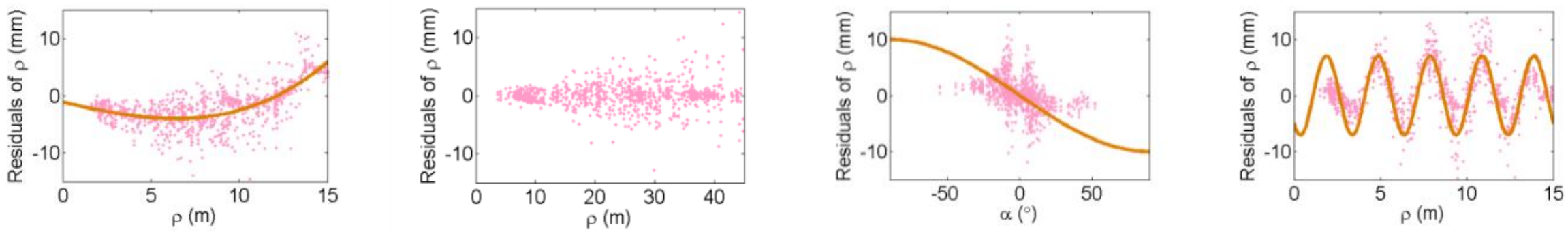

$A_0$ $A_1$ $A_2$ $A_3 + A_4$

**Figure 3.** Residual plots of systematic errors in the horizontal angle observations ($\theta$) from point-based self-calibration for hybrid-type scanners: $B_1$ is the horizontal direction scale factor error, $B_2$ + $B_3$ are the horizontal circle eccentricity, $B_4$ + $B_5$ are the non-orthogonality of horizontal encoder and vertical axis, $B_6$ is the horizontal collimation axis error, $B_7$ is the trunnion axis error, $B_8$ is the horizontal eccentricity of collimation axis, and $B_9$ + $B_{10}$ are the trunnion axis wobble.

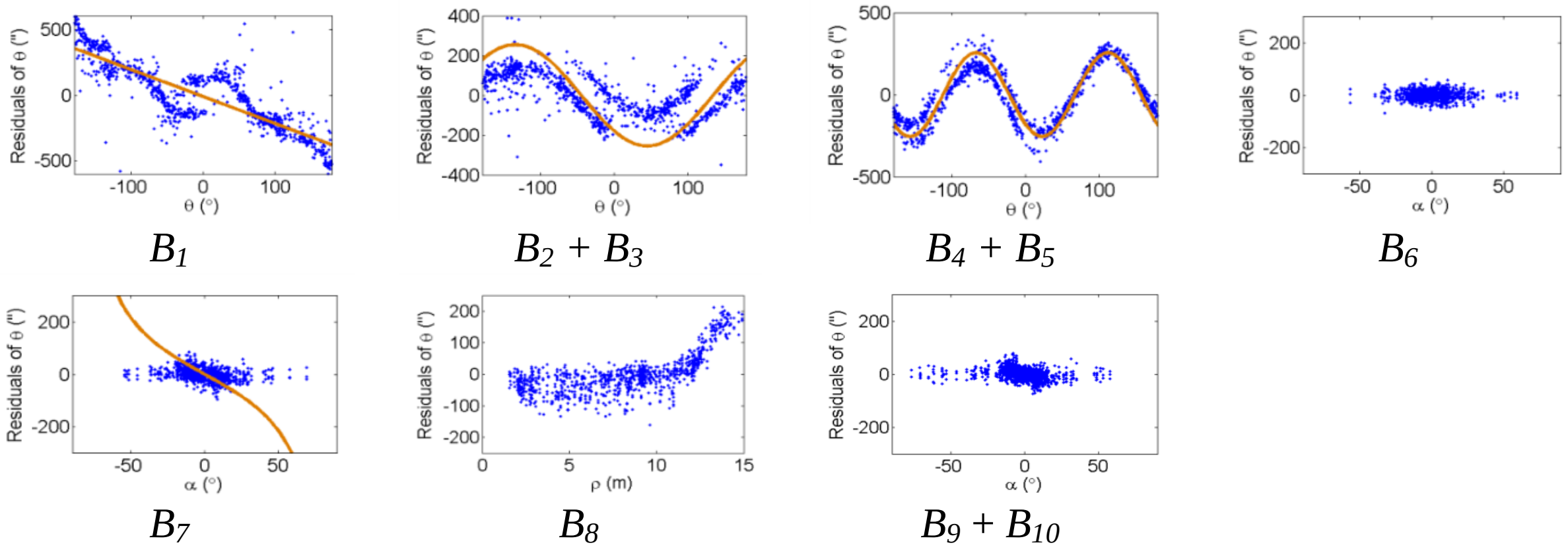

**Figure 4.** Residual plots of systematic errors in the vertical angle observations ($\alpha$) from point-based self-calibration for hybrid-type scanners: $C_0$ is the vertical circle index error, $C_1$ is the scale factor error, $C_2$ + $C_3$ are the vertical circle eccentricity, $C_4$ + $C_5$ are the non-orthogonality of vertical encoder and trunnion axis, $C_6$ is the vertical eccentricity of collimation axis, and $C_7$ + $C_8$ is the vertical axis wobble.

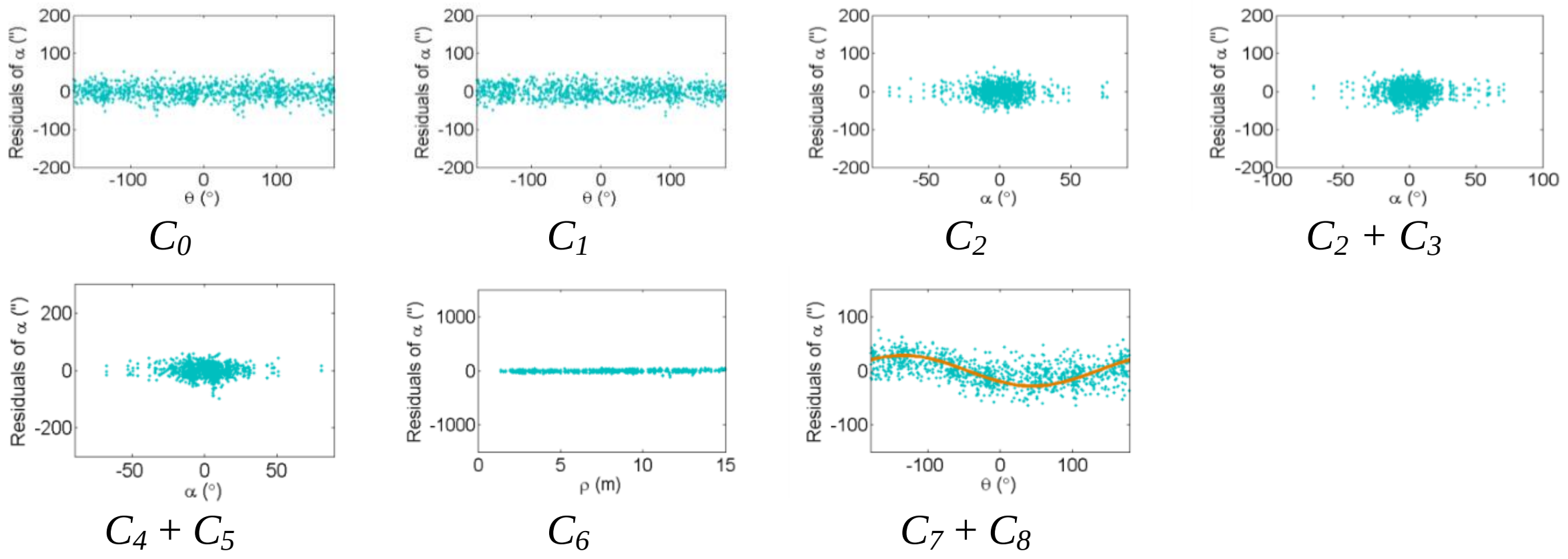

**Figure 5.** Residual plots of systematic errors in the range observations ($\rho$) from point-based self-calibration for panoramic-type scanners: $A_0$ is the rangefinder offset, $A_1$ is the range scale factor error, $A_2$ is the laser axis vertical offset, and $A_3 + A_4$ are the cyclic errors.

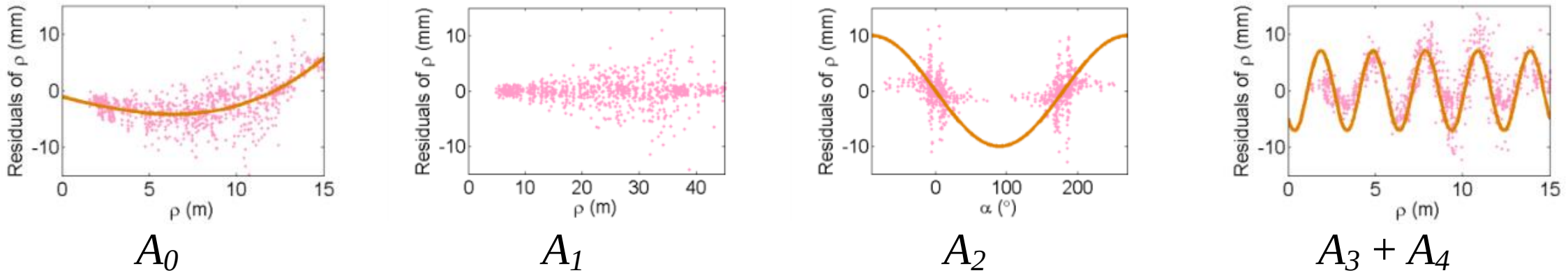

**Figure 6.** Residual plots of systematic errors in the horizontal angle observations ($\theta$) from point-based self-calibration for panoramic-type scanners: $B_1$ is the horizontal direction scale factor error, $B_2 + B_3$ are the horizontal circle eccentricity, $B_4 + B_5$ are the non-orthogonality of horizontal encoder and vertical axis, $B_6$ is the horizontal collimation axis error, $B_7$ is the trunnion axis error, $B_8$ is the horizontal eccentricity of collimation axis, and $B_9 + B_{10}$ are the trunnion axis wobble.

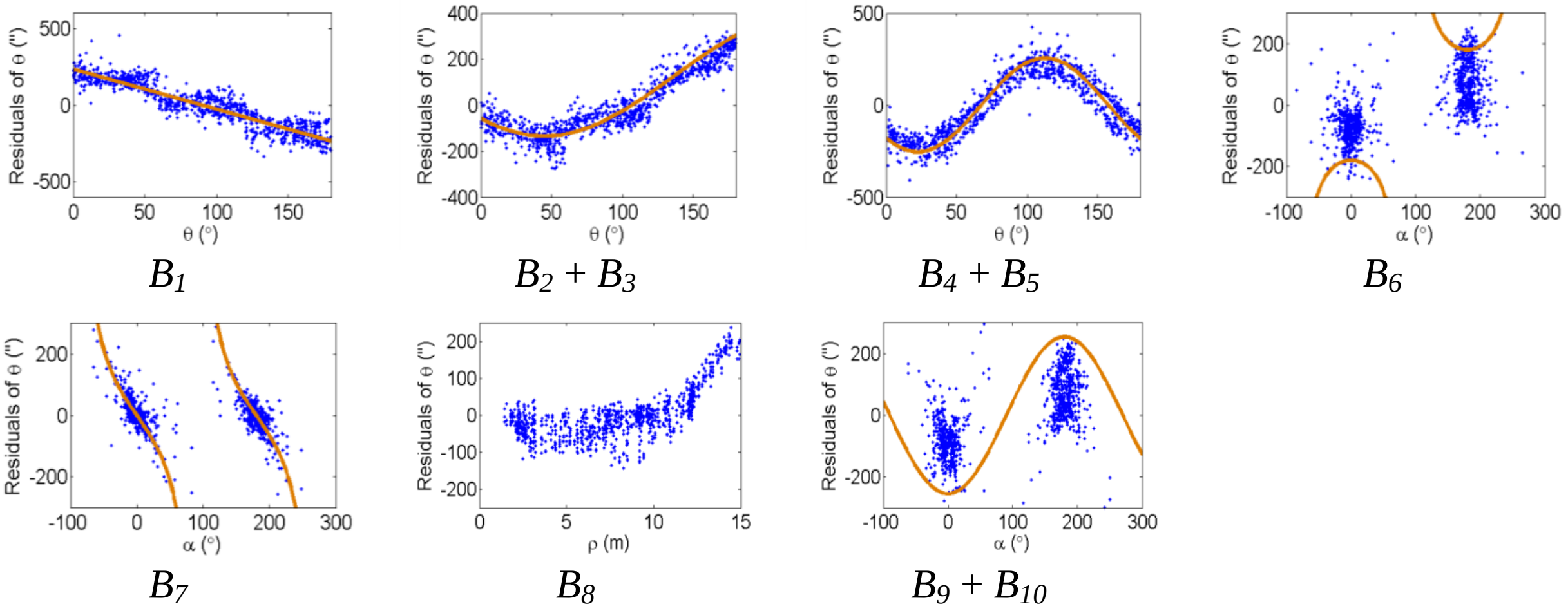

**Figure 7.** Residual plots of systematic errors in the vertical angle observations ($\alpha$) from point-based self-calibration for panoramic-type scanners: $C_0$ is the vertical circle index error, $C_1$ is the scale factor error, $C_2 + C_3$ are the vertical circle eccentricity, $C_4 + C_5$ are the non-orthogonality of vertical encoder and trunnion axis, $C_6$ is the vertical eccentricity of collimation axis, and $C_7 + C_8$ is the vertical axis wobble.

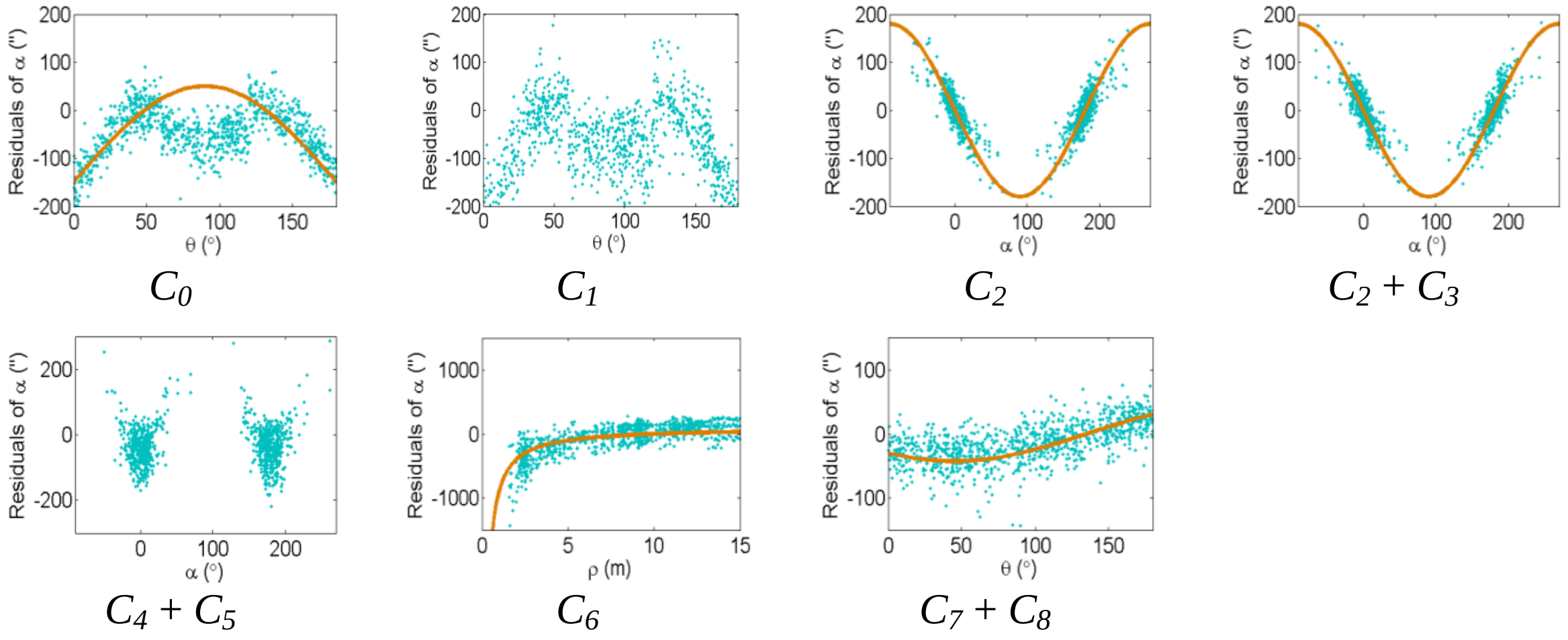

**Figure 8.** Residual plots of systematic errors in the range observations ($\rho$) from plane-based self-calibration for hybrid-type scanners: $A_0$ is the rangefinder offset, $A_1$ is the range scale factor error, $A_2$ is the laser axis vertical offset, and $A_3 + A_4$ are the cyclic errors.

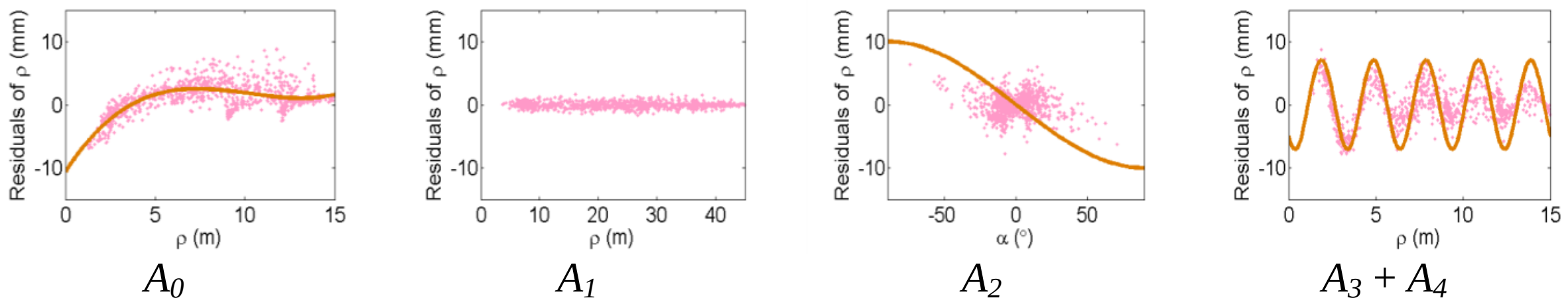

**Figure 9.** Residual plots of systematic errors in the horizontal angle observations ($\theta$) from plane-based self-calibration for hybrid-type scanners: $B_1$ is the horizontal direction scale factor error, $B_2 + B_3$ are the horizontal circle eccentricity, $B_4 + B_5$ are the non-orthogonality of horizontal encoder and vertical axis, $B_6$ is the horizontal collimation axis error, $B_7$ is the trunnion axis error, $B_8$ is the horizontal eccentricity of collimation axis, and $B_9 + B_{10}$ are the trunnion axis wobble.

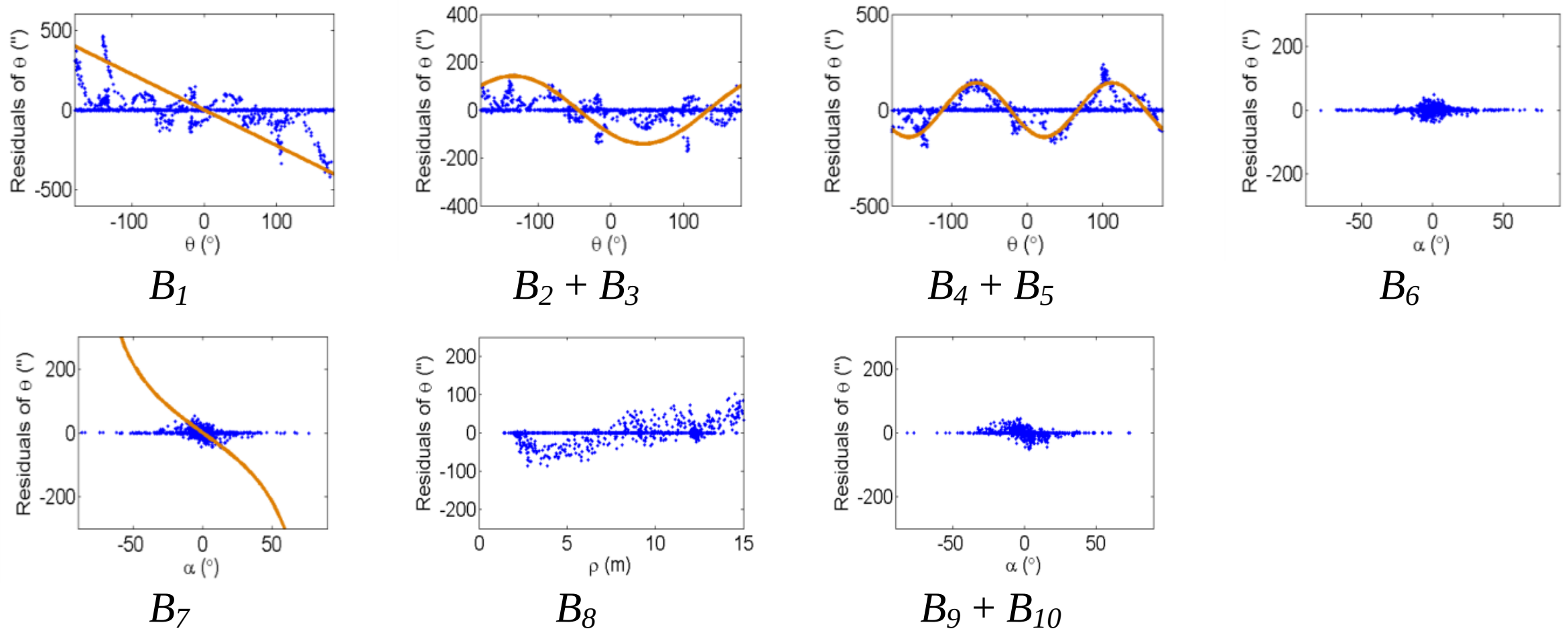

**Figure 10.** Residual plots of systematic errors in the vertical angle observations ($\alpha$) from plane-based self-calibration for hybrid-type scanners: $C_0$ is the vertical circle index error, $C_1$ is the scale factor error, $C_2 + C_3$ are the vertical circle eccentricity, $C_4 + C_5$ are the non-orthogonality of vertical encoder and trunnion axis, $C_6$ is the vertical eccentricity of collimation axis, and $C_7 + C_8$ is the vertical axis wobble.

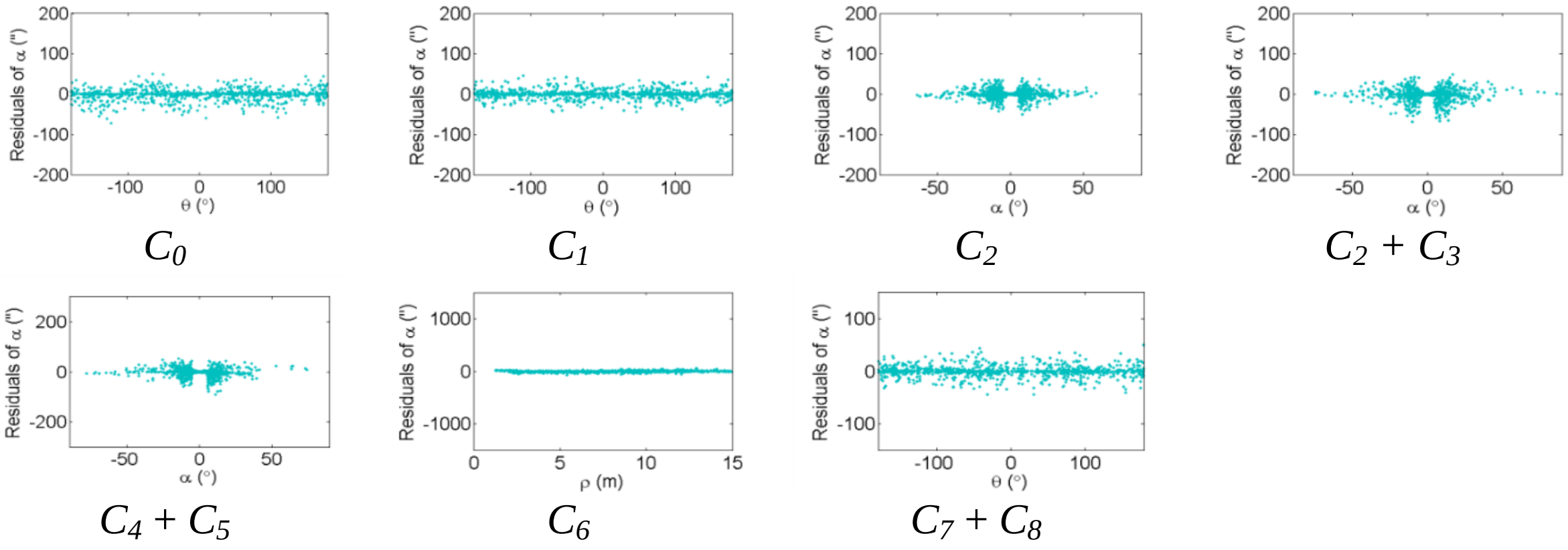

**Figure 11.** Residual plots of systematic errors in the range observations ($\rho$) from plane-based self-calibration for panoramic-type scanners: $A_0$ is the rangefinder offset, $A_1$ is the range scale factor error, $A_2$ is the laser axis vertical offset, and $A_3 + A_4$ are the cyclic errors.

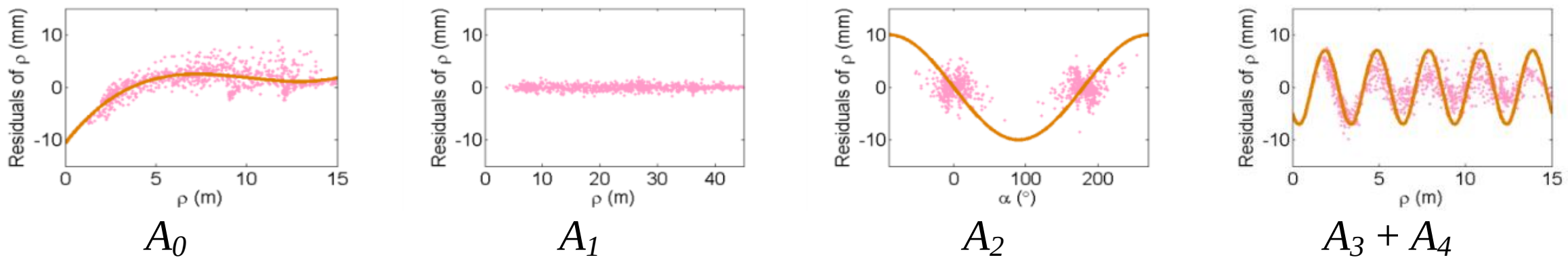

**Figure 12.** Residual plots of systematic errors in the horizontal angle observations ($\theta$) from plane-based self-calibration for panoramic-type scanners: $B_1$ is the horizontal direction scale factor error, $B_2 + B_3$ are the horizontal circle eccentricity, $B_4 + B_5$ are the non-orthogonality of horizontal encoder and vertical axis, $B_6$ is the horizontal collimation axis error, $B_7$ is the trunnion axis error, $B_8$ is the horizontal eccentricity of collimation axis, and $B_9 + B_{10}$ are the trunnion axis wobble.

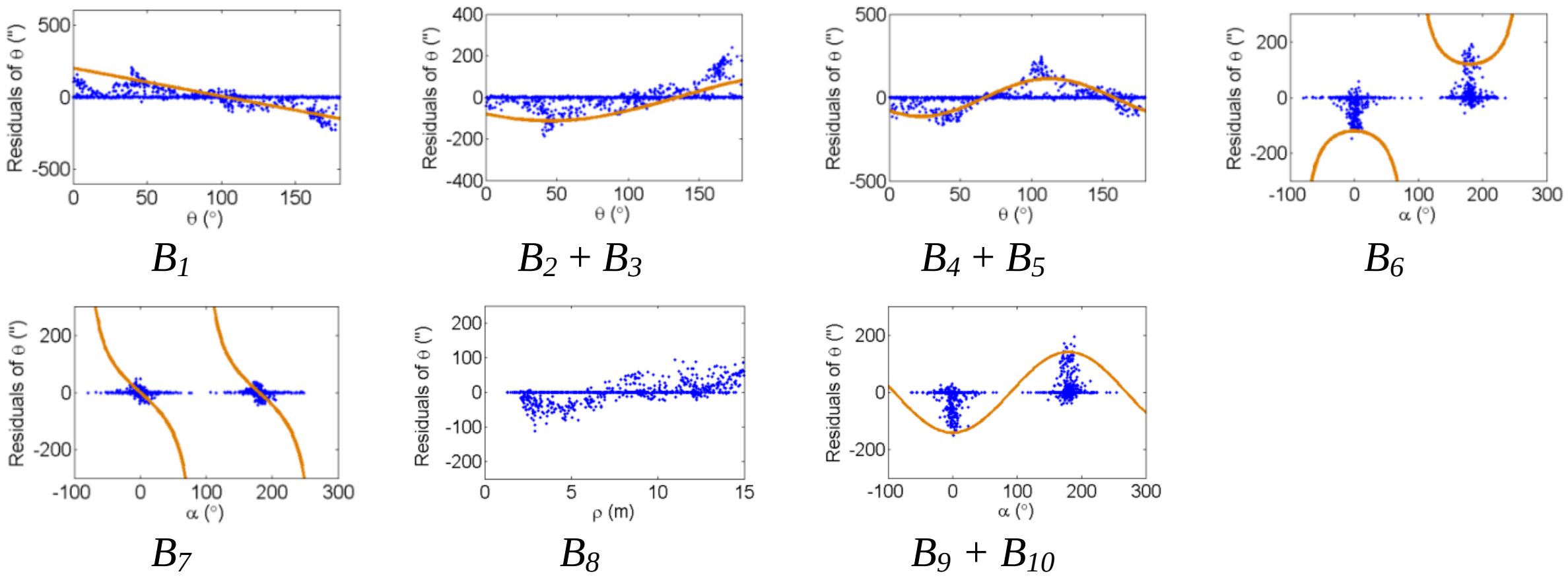

**Figure 13.** Residual plots of systematic errors in the vertical angle observations ($\alpha$) from plane-based self-calibration for panoramic-type scanners: $C_0$ is the vertical circle index error, $C_1$ is the scale factor error, $C_2$ + $C_3$ are the vertical circle eccentricity, $C_4$ + $C_5$ are the non-orthogonality of vertical encoder and trunnion axis, $C_6$ is the vertical eccentricity of collimation axis, and $C_7$ + $C_8$ is the vertical axis wobble.

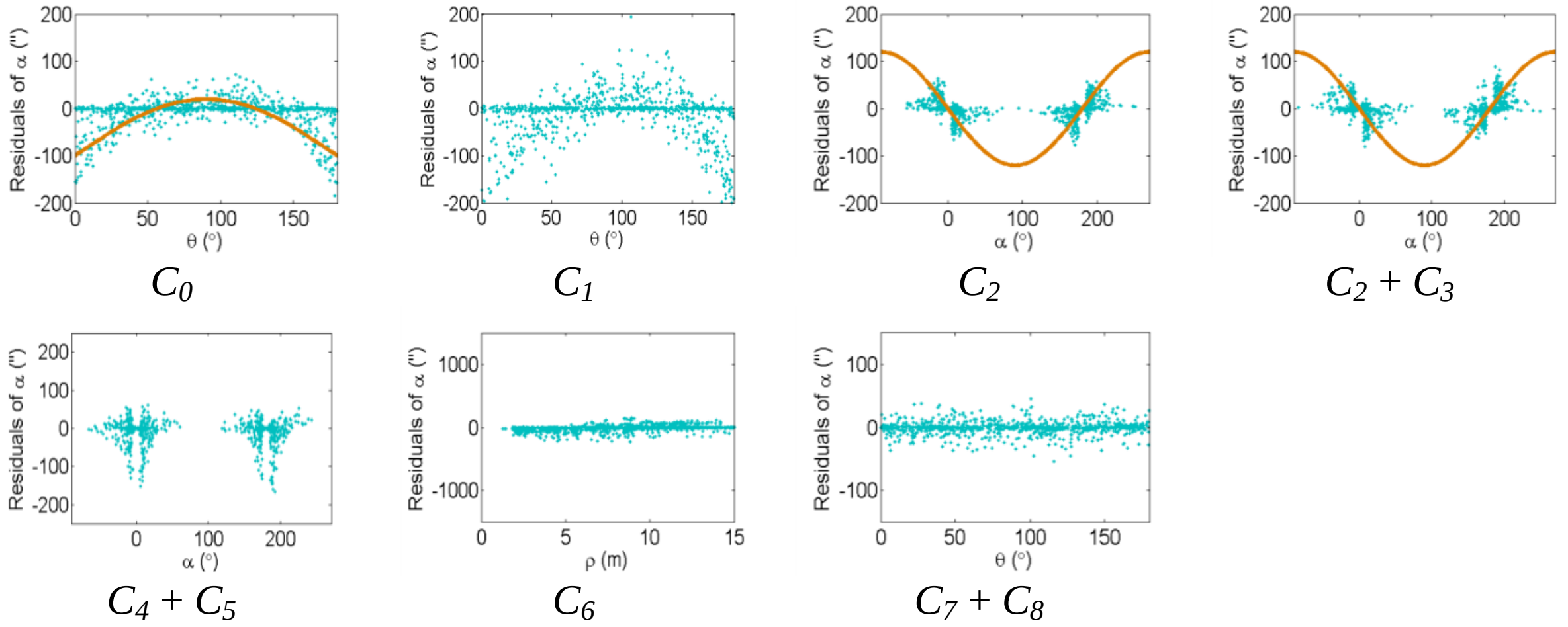

The corresponding residual plots between the point-based and plane-based self-calibrations for both scanner architectures are similar, suggesting that they can yield comparable results. However, there are some downfalls for the plane-based calibration — for example, the vertical eccentricity of collimation axis ($C_6$) and wobbling ($C_7$ and $C_8$) do not propagate into residual plots under the tested network geometry. As mentioned, visual identification of trends in the observation residuals can be an indication that the systematic error is decoupled from other observations and thus can be modelled in the observation space with greater confidence. Systematic error terms such as $B_8$ (horizontal eccentricity of collimation axis) in Figures 3 and 6 appear differently than the similar $C_6$ term (vertical eccentricity of collimation axis). Despite the similarities between the error models, yet different behavior in the observation residuals, both terms were recovered accurately. In general, systematic errors that were identifiable in the observation residual plots were accurately recovered using the calibration methods presented, including $C_0$ in hybrid-type scanners. Most systematic error terms tested to this point can be identified in the residual plots for panoramic type scanners, regardless of the primitives used for calibration. However, the same cannot be said for hybrid type scanners. In particular, there is poor recovery of vertical angular errors (Figures 4 and 10). The following section presents a feasible solution to these limitations of point-based calibration and in particular addresses the issue of solving for $B_6$ in hybrid-type scanners.

### *4.1. Tilted Scans*

The main advantage in self-calibration of panoramic scanners over hybrid scanners is observation of data in two layers, where vertical angle observations in front of and behind the scanner are in different quadrants. This implicitly decouples several systematic errors from other parameters because they have the same magnitude but opposite sign on opposing scanner faces. For hybrid-type scanners, a similar decoupling effect can be achieved by tilting the scanner with a roll angle of 90° as is done in photogrammetric self-calibrations [23]. To be consistent with the data captured in the real experiment, scans with a roll angle of −45° and +45° were simulated instead. Everything else in the simulation remains the same as before, with the replacement of two tilted scans at one of the scan stations. As shown in Figures 14 and 15, the desired correlation reduction is achieved, even for the horizontal collimation axis error in hybrid scanners.

**Figure 14.** Standard deviation and maximum correlation of the recovered APs for hybrid-type scanners in simulation.

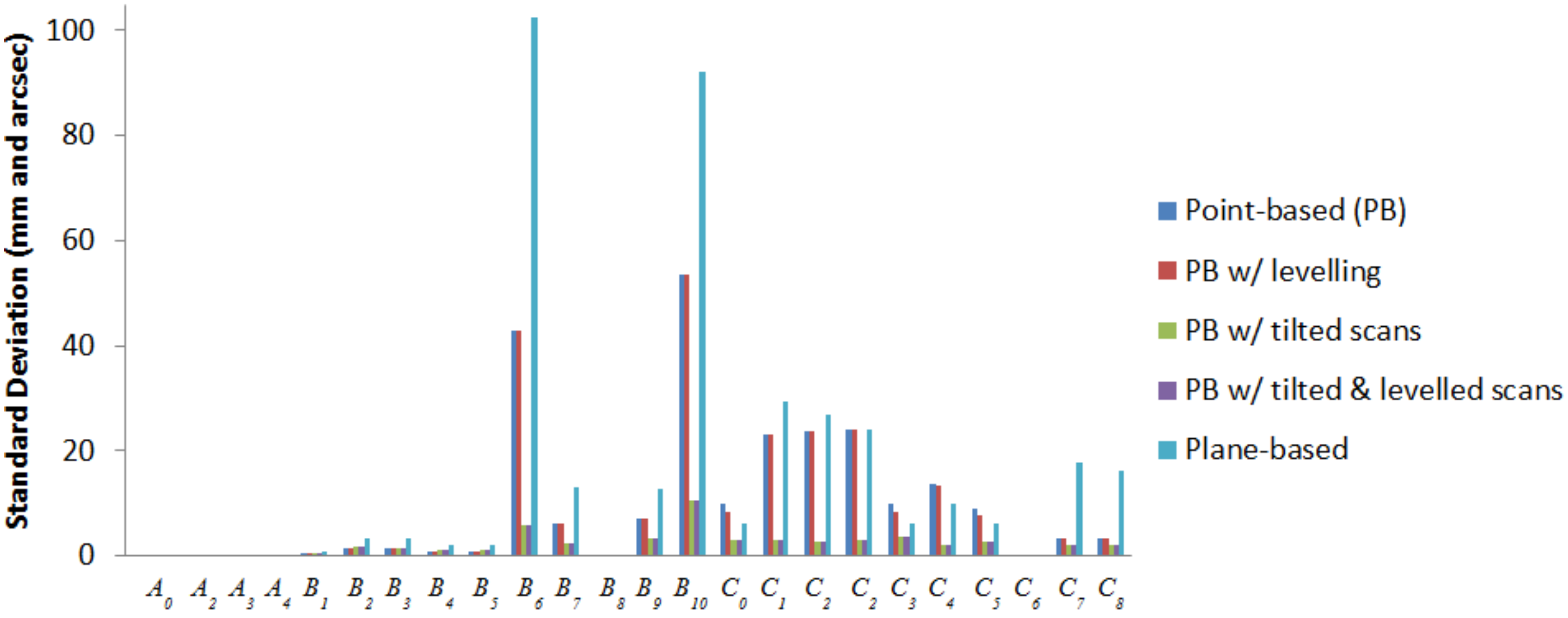

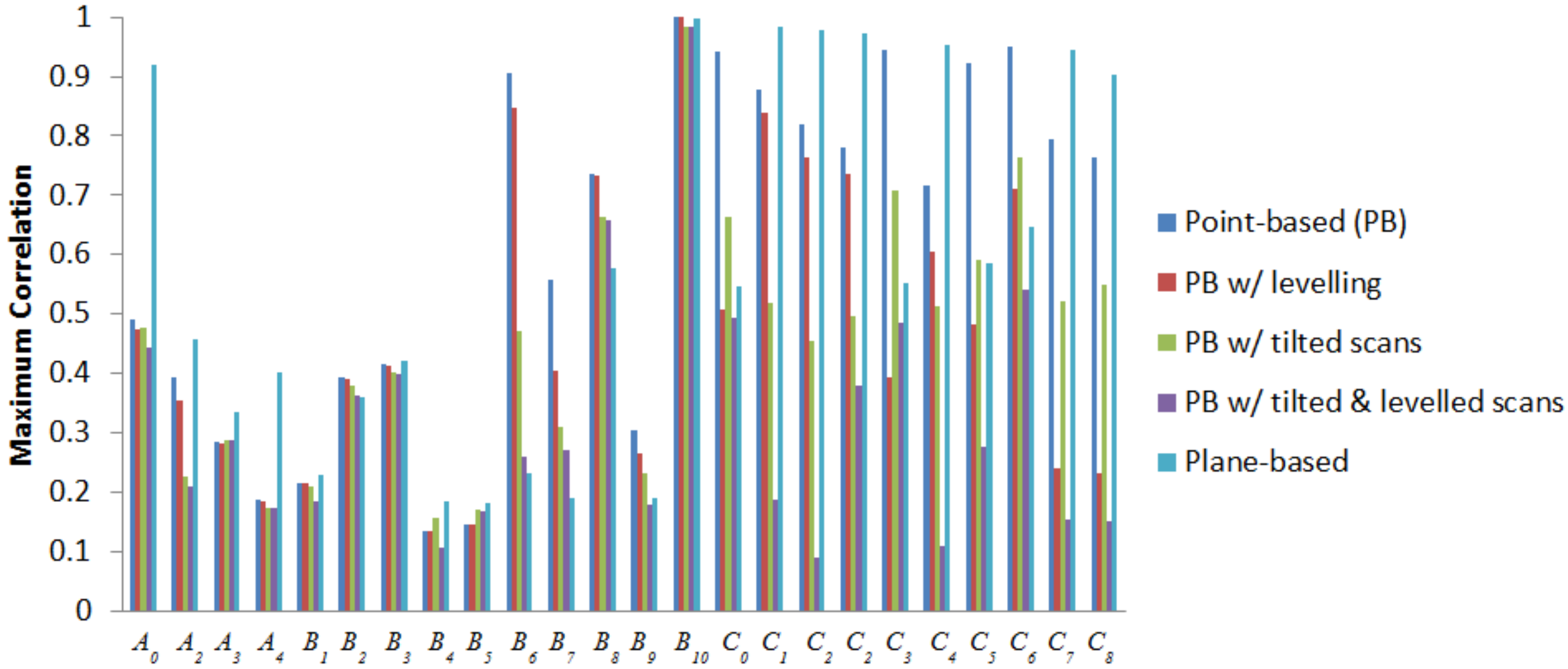

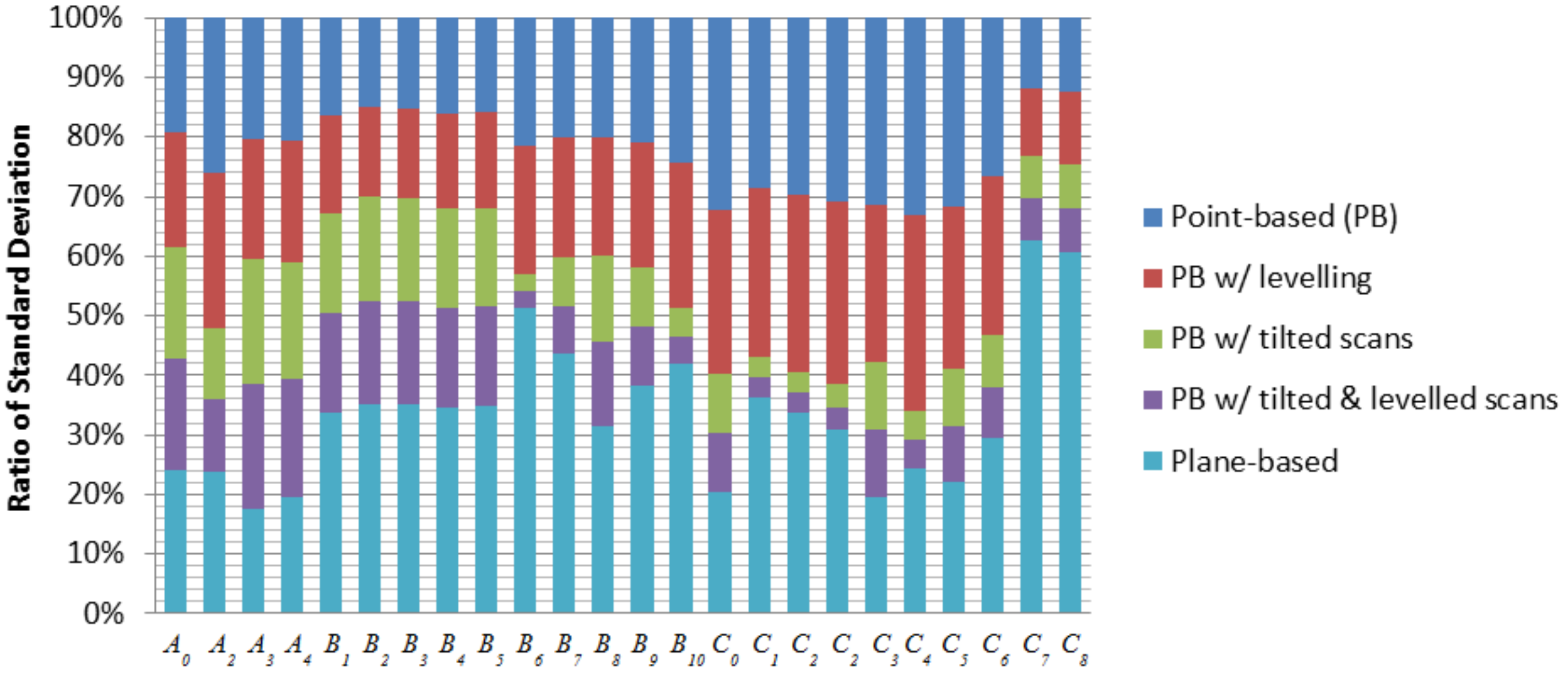

Ratio of Standard Deviation
100%
90%
80%
70%
60%
50%
40%
30%
20%
10%
0%
A0 A2 A3 A4 B1 B2 B3 B4 B5 B6 B7 B8 B9 B10 C0 C1 C2 C2 C3 C4 C5 C6 C7 C8
Point-based (PB)
PB w/ levelling
PB w/ tilted scans
PB w/ tilted & levelled scans
Plane-based

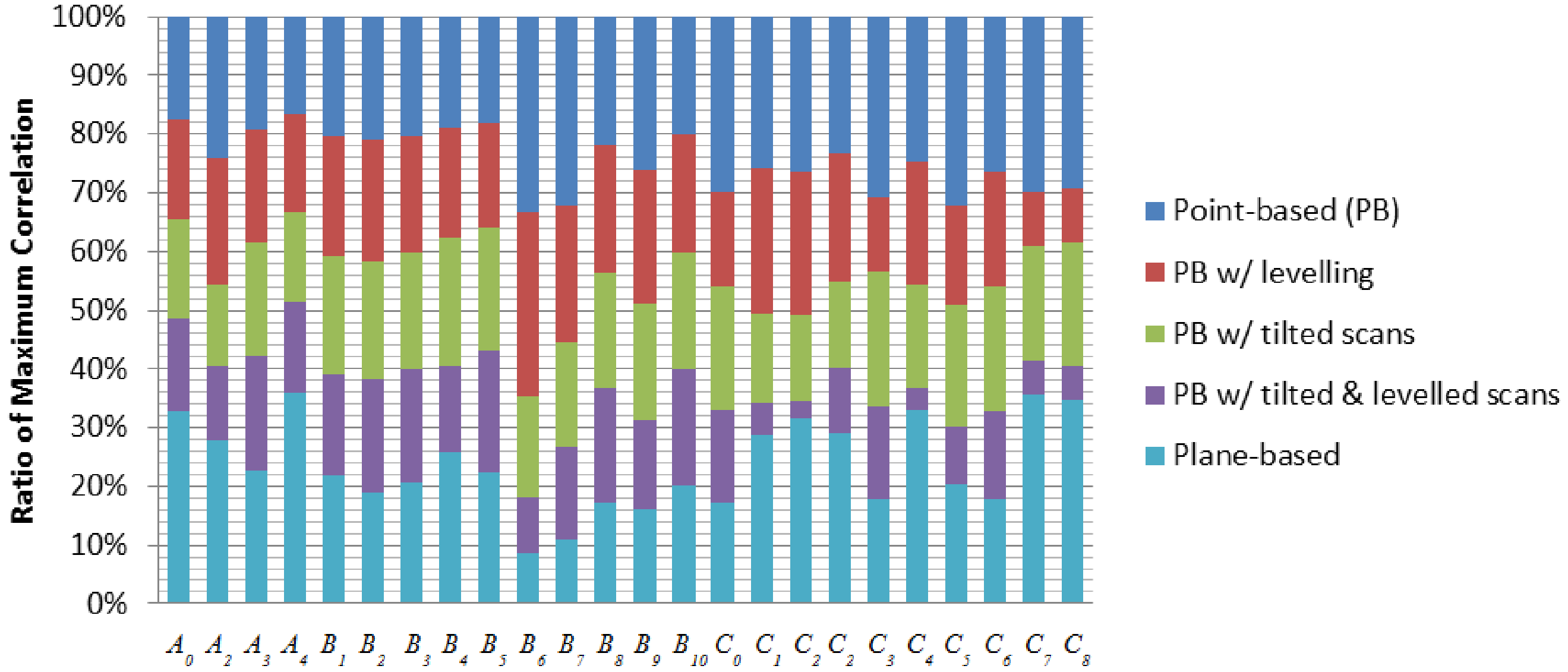

Ratio of Maximum Correlation
100%
90%
80%
70%
60%
50%
40%
30%
20%
10%
0%
A0 A2 A3 A4 B1 B2 B3 B4 B5 B6 B7 B8 B9 B10 C0 C1 C2 C2 C3 C4 C5 C6 C7 C8
Point-based (PB)
PB w/ levelling
PB w/ tilted scans
PB w/ tilted & levelled scans
Plane-based

**Figure 15.** Standard deviation and maximum correlation of the recovered APs for panoramic-type scanners in simulation.

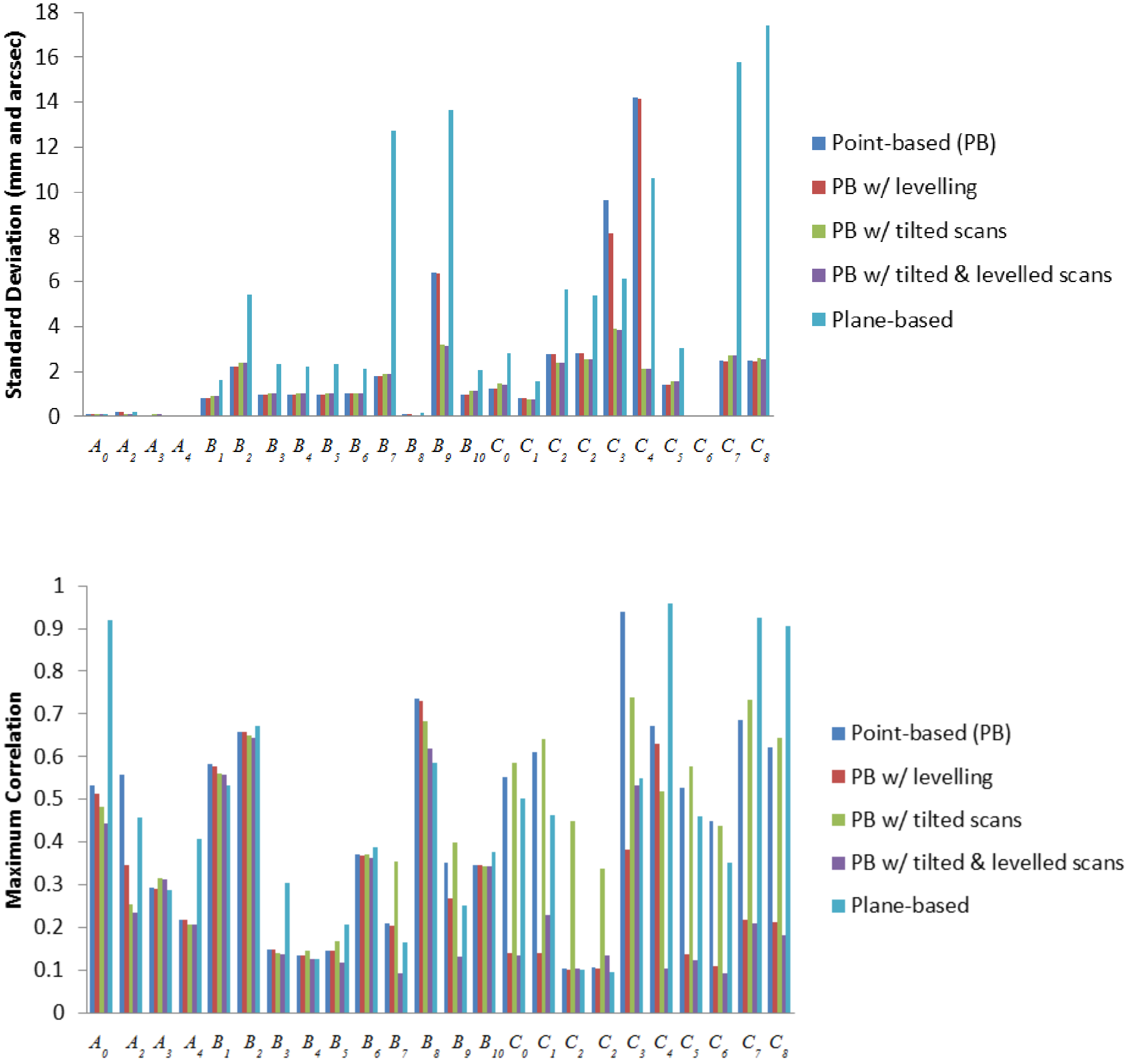

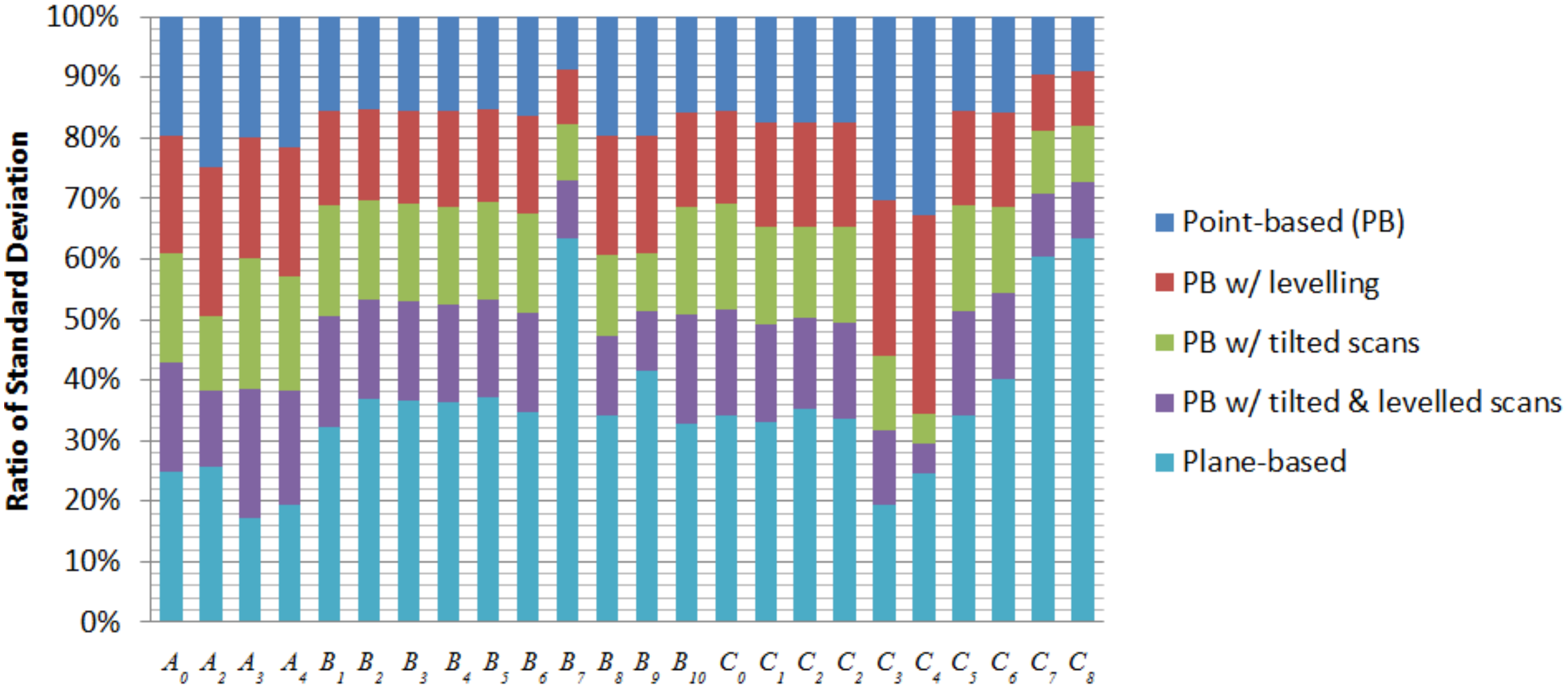

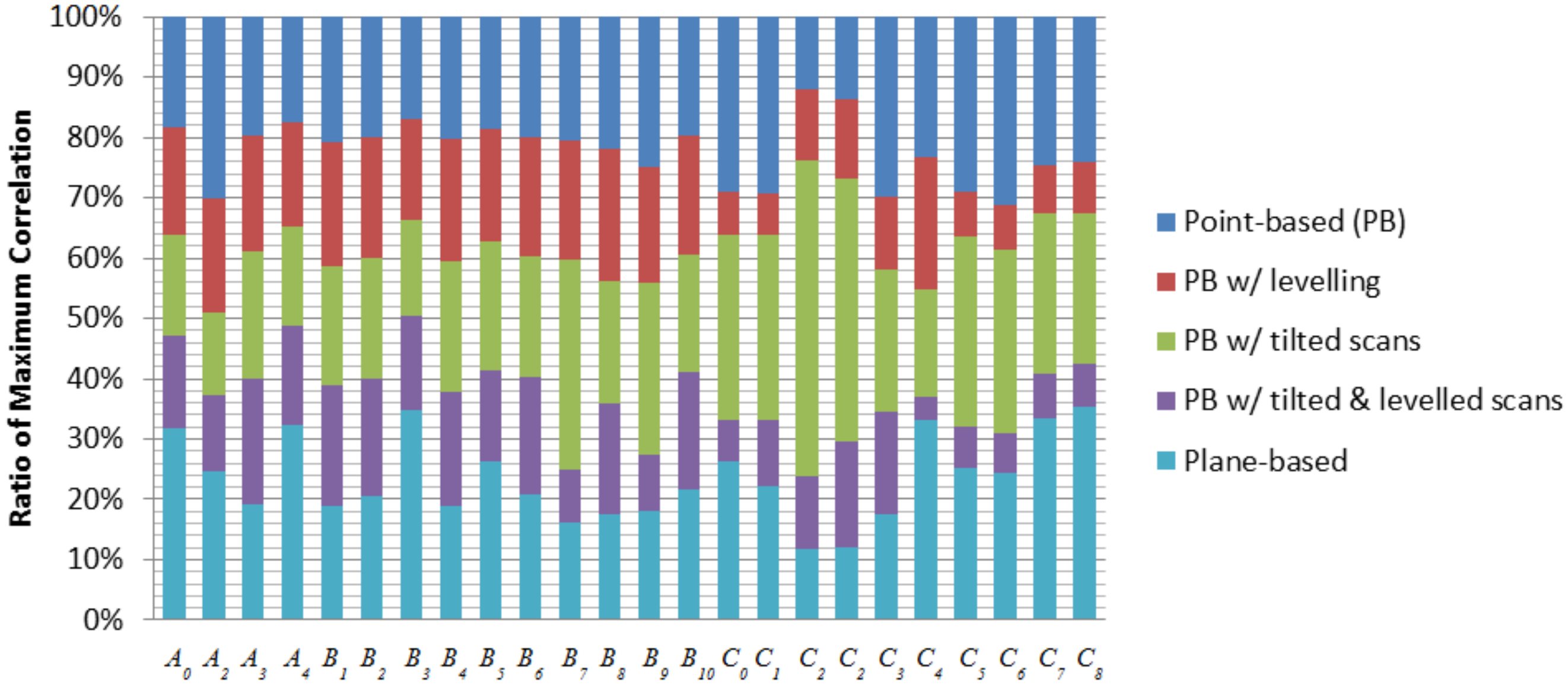

The standard deviations of the recovered systematic errors (given in millimetres and arc seconds) are reduced. It is worth mentioning that in [19] it was reported that beyond a fairly low threshold, additional scans or targets give only minimal improvement to the TLS self-calibration. Hence, the improvements being observed here are mainly a result of the geometric strength gained from tilting the scanner. Combining tilted scans with levelling constraints brings forth additional improvement to the correlations on all the vertical angular errors for hybrid scanners, with the exception of $C_3$. The standard deviation and correlation of $B_6$ and $B_{10}$ for hybrid scanners benefitted the most from tilting the scanner. The standard deviations dropped by 86% and 80% respectively; however, the maximum correlation for $B_{10}$ remained high. For panoramic scanners, their APs had better standard deviation and correlation than hybrid scans, and the only significant improvement to the standard deviations from tilting the scanner are found in $B_9$, $C_3$ and $C_4$ (50%, 60% and 85%, respectively). Most APs recovered in this simulation using planar features suffer from lower precision and higher correlation than the point-based calibration. However, it can be seen that most APs recoverable by point-based calibration are also recoverable by plane-based calibration.

## 5. Real Datasets: Results and Discussion

Point clouds captured by scanners with hybrid architecture (*i.e.*, a Riegl VZ-400 and a Leica HDS6100 operating in hybrid mode) and panoramic architecture (*i.e.*, a Z+F Imager 5003 and a HDS6100 operating in panoramic mode) are used for testing the proposed methodology. We first present the influence of tilted scans on the calibration of the VZ-400 and the Imager 5003. This will be followed by a comparison of the point-based calibration to plane-based calibration with real data, showing that they can yield the same results. For a fair comparison, the same point clouds were used in both point-based and plane-based self-calibrations with the only difference being the extracted primitives (*i.e.*, signalised targets or planes).

*5.1. Tilted Scans*

The Riegl VZ-400 and the Z+F Imager 5003 were calibrated in a 30 m by 33 m by 5 m room located at the University of Houston (Figure 16). A specialized tripod was used for acquiring scans on a −45° and +45° incline. Checkerboard-type paper targets were affixed to the ceiling, floor, wall, and the vertical support I-beams. The statistically significant systematic errors found in the calibration of the Riegl VZ-400 and the Z+F Imager 5003 with and without the tilted scans are shown in Tables 1 and 2, respectively.

**Figure 16.** TLS user self-calibration field setup with a Riegl VZ-400 and Z+F Imager 5003 at the University of Houston.

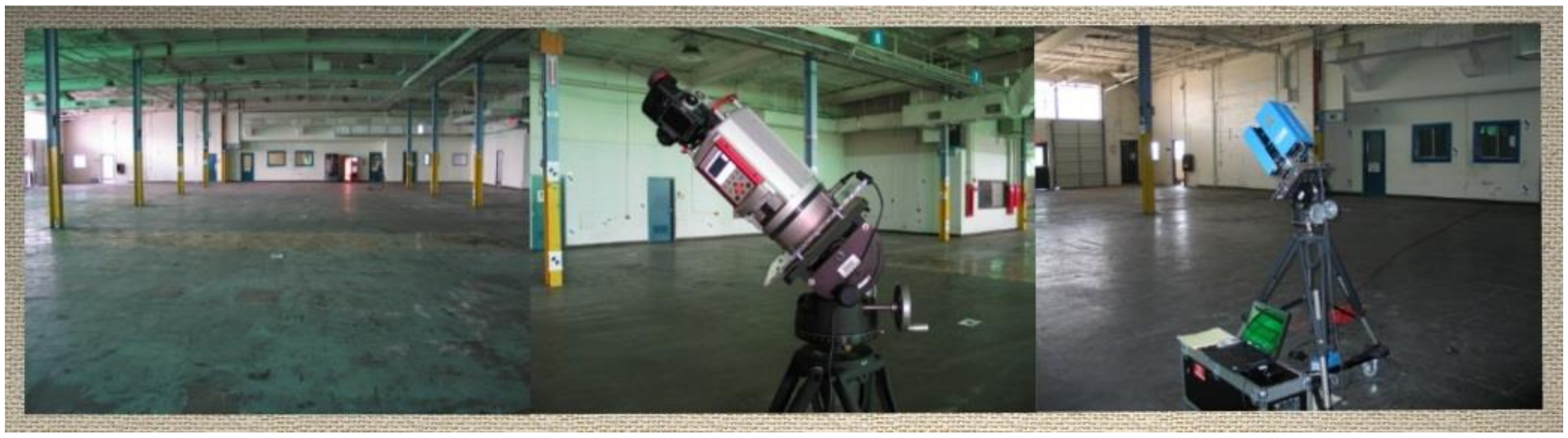

**Table 1.** Riegl VZ-400 point-based self-calibration.

| | Without Tilted Scans | | | With Tilted Scans | | |
|---|---|---|---|---|---|---|
| **# of Targets** | 61 | | | 61 | | |
| **# of Scans** | 4 (levelled) | | | 4 (levelled) + 2 (tilted) | | |
| **# of Obs.** | 456 | | | 639 | | |
| **# of Unk.** | 211 | | | 422 | | |
| $\rho_{min}, \rho_{max}$ **[m]** | 3.3, 30.6 | | | 3.3, 30.6 | | |
| $\alpha_{min}, \alpha_{max}$ **[°]** | −24, 35 | | | −40, 47 | | |
| | **Value** | **σ** | **Max Abs. Correlation (w/parameter)** | **Value** | **σ** | **Max Abs. Correlation (w/parameter)** |
| $A_0$ **[mm]** | 2.8 | 0.7 | 0.92 ($Y$) | 2.6 | 0.4 | 0.82 ($X$) |
| $C_0$ **[“]** | −16.1 | 20.6 | 1.00 ($Z_o$) | 9.6 | 3.9 | 0.88 ($Z_o$) |
| $B_6$ **[“]** | 3318.0 | 927.3 | 0.98 ($X$) | 109.5 | 24.1 | 0.62 ($\kappa$) |
| $B_7$ **[“]** | 472.0 | 131.4 | 0.92 ($Y$) | 23.1 | 5.3 | 0.29 ($Y_o$) |

**Table 2.** Z+F Imager 5003 point-based self-calibration.

| | Without Tilted Scans | | | With Tilted Scans | | |
|---|---|---|---|---|---|---|
| **# of Targets** | 65 | | | 67 | | |
| **# of Scans** | 4 (levelled) | | | 4 (levelled) + 2 (tilted) | | |
| **# of Obs.** | 390 | | | 480 | | |
| **# of Unk.** | 223 | | | 241 | | |
| $\rho_{min}, \rho_{max}$ **[m]** | 4.0, 30.6 | | | 3.3, 30.6 | | |
| $\alpha_{min}, \alpha_{max}$ **[°]** | −16.5, 199 | | | −49, 235 | | |
| | **Value** | **σ** | **Max Abs. Correlation (w/parameter)** | **Value** | **σ** | **Max Abs. Correlation (w/parameter)** |
| $A_0$ **[mm]** | 3.6 | 1.1 | 0.96 ($Y$) | −0.6 | 0.5 | 0.83 ($Z$) |
| $C_0$ **[“]** | 40.5 | 8.3 | 0.90 ($\omega$) | 36.9 | 5.3 | 0.77 ($\varphi$) |
| $B_6$ **[“]** | 3.5 | 3.6 | 0.49 ($X$) | −6.0 | 2.5 | 0.30 ($Y_o$) |
| $B_7$ **[“]** | 79.7 | 35.4 | 0.67 ($Y$) | −19.7 | 9.5 | 0.47 ($\kappa$) |

These APs were selected based on a combination of graphical analysis of the observation residual plots and statistical analysis. The calibration setup had targets covering distances from 3 m to 30 m, which is a desirable trait for range calibrations. However, due to the poor distribution of targets near the extrema of the vertical FOV and the scanner being only approximately levelled, the correlations between the APs and other parameters are rather high. The largest standard deviation in the above tables is for the horizontal collimation axis error ($B_6$) in the VZ-400. Although this error is statistically significant based on the t-test, the magnitude of the error appears to be unreasonably high. Even though the reduced horizontal collimation axis error model was used, it is expected to have no effect on the standard deviation as previously shown [11]. With the reduced model, the perfect correlation with the tertiary rotation angle has been removed, but it is still highly correlated with other parameters (e.g., up to 0.98 with object space target coordinates). By including two tilted scans, the maximum correlation of all the APs were reduced. Correlations with the $B_6$ and $B_7$ terms received the most benefit as they were reduced by 37% and 68%, respectively. The magnitude of the recovered horizontal angular errors was reduced drastically to a more reasonable level and the standard deviations for $B_6$ and $B_7$ were improved by 97% and 96%, respectively. This is lower than any standard deviation for $B_6$ found in literature for hybrid-type scanners.

The standard deviation of all the calibrated angular parameters is lower for the panoramic scanners than for the hybrid scanners, as expected. In this network configuration, where the vertical distribution of targets is not ideal and dual-axis compensation was not used for the scanners, tilted scans significantly improved the quality of the calibration. The improvements are even more pronounced than in the simulated results, where the targets were better spread out within the scanner's FOV.

*5.2. Comparison of Point-Based and Plane-Based TLS Self-Calibration*

Point-based and plane-based calibration results for the Leica HDS6100 in hybrid mode and in panoramic mode are presented in Tables 3 and 4, respectively. No tilted scans were included, but a large number of targets and planes with various orientations were used. Data were captured in a 14 m by 11 m by 3 m room at the University of Calgary (Figure 17). Special care was taken to ensure a large number of observations were made above the scanner and as close to the tripod legs as possible.

**Table 3.** Self-calibration of the Leica HDS6100 in hybrid mode.

| | **Point-Based Self-Calibration** | | | **Plane-Based Self-Calibration** | | |
|---|---|---|---|---|---|---|
| **# of Targets/Planes** | 285 | | | 118 | | |
| **# of Scans** | 6 | | | 6 | | |
| **# of Obs.** | 3405 | | | 98982 | | |
| **# of Unk.** | 895 | | | 512 | | |
| $\rho_{min}$, $\rho_{max}$ **[m]** | 1.1, 15.3 | | | 1.2, 15.2 | | |
| $\alpha_{min}$, $\alpha_{max}$ **[°]** | −60, 88 | | | −64, 58 | | |
| | **Value** | **σ** | **Max Abs. Correlation (w/parameter)** | **Value** | **σ** | **Max Abs. Correlation (w/parameter)** |
| $A_0$ **[mm]** | −0.6 | 0.05 | 0.62 ($X_o$) | −0.5 | 0.02 | 0.17 ($B_6$) |
| $C_0$ **[“]** | −87.3 | 3.2 | 0.94 ($Z_o$) | −77.5 | 1.5 | 0.11 ($\varphi$) |
| $B_6$ **[“]** | −24.2 | 5.8 | 0.87 ($Z_o$) | −27.6 | 12.8 | 0.48 ($B_7$) |
| $B_7$ **[“]** | 20.3 | 4.6 | 0.58 ($X$) | 40.0 | 2.4 | 0.48 ($B_6$) |

**Table 4.** Self-calibration of the Leica HDS6100 in panoramic mode.

| | Dataset 1 | | | | | |
|---|---|---|---|---|---|---|
| | Point-Based Self-Calibration | | | Plane-Based Self-Calibration | | |
| **# of Targets/Planes** | 285 | | | 118 | | |
| **# of Scans** | 6 | | | 6 | | |
| **# of Obs.** | 3039 | | | 98877 | | |
| **# of Unk.** | 895 | | | 512 | | |
| $\rho_{min}, \rho_{max}$ **[m]** | 1.1, 15.4 | | | 1.1, 15.2 | | |
| $\alpha_{min}, \alpha_{max}$ **[°]** | −59, 240 | | | −64, 244 | | |
| | **Value** | **σ** | **Max Abs. Correlation (w/parameter)** | **Value** | **σ** | **Max Abs. Correlation (w/parameter)** |
| $A_0$ **[mm]** | −0.8 | 0.05 | 0.59 ($Y_o$) | −0.5 | 0.02 | 0.12 ($B_6$) |
| $C_0$ **[“]** | 10.7 | 1.3 | 0.55 ($\omega$) | 3.8 | 0.5 | 0.09 ($B_6$) |
| $B_6$ **[“]** | −2.5 | 0.7 | 0.29 ($Y_o$) | 1.8 | 0.2 | 0.12 ($A_0$) |
| $B_7$ **[“]** | −48.6 | 1.9 | 0.31 ($Y$) | −57.0 | 1.8 | 0.06 ($B_6$) |

| | Dataset 2 (Levelled via dual-axis compensator) | | | | | |
|---|---|---|---|---|---|---|
| | Point-Based Self-Calibration | | | Plane-Based Self-Calibration | | |
| **# of Targets/Planes** | 384 | | | 223 | | |
| **# of Scans** | 5 | | | 5 | | |
| **# of Obs.** | 2878 | | | 126567 | | |
| **# of Unk.** | 1186 | | | 926 | | |
| $\rho_{min}, \rho_{max}$ **[m]** | 1.1, 15.5 | | | 1.1, 14.0 | | |
| $\alpha_{min}, \alpha_{max}$ **[°]** | −62, 239 | | | −64, 240 | | |
| | **Value** | **σ** | **Max Abs. Correlation (w/parameter)** | **Value** | **σ** | **Max Abs. Correlation (w/parameter)** |
| $A_0$ **[mm]** | −1.7 | 0.1 | 0.72 ($Y_o$) | −0.4 | 0.1 | 0.08 ($B_7$) |
| $C_0$ **[“]** | 10.7 | 2.0 | 0.16 ($\varphi$) | 8.1 | 1.5 | 0.08 ($B_7$) |
| $B_6$ **[“]** | −19.4 | 1.4 | 0.23 ($Y_o$) | −4.5 | 0.5 | 0.09 ($B_7$) |
| $B_7$ **[“]** | −84.8 | 2.1 | 0.22 ($Y$) | −69.4 | 4.0 | 0.09 ($B_6$) |

**Figure 17.** TLS user self-calibration at the University of Calgary.

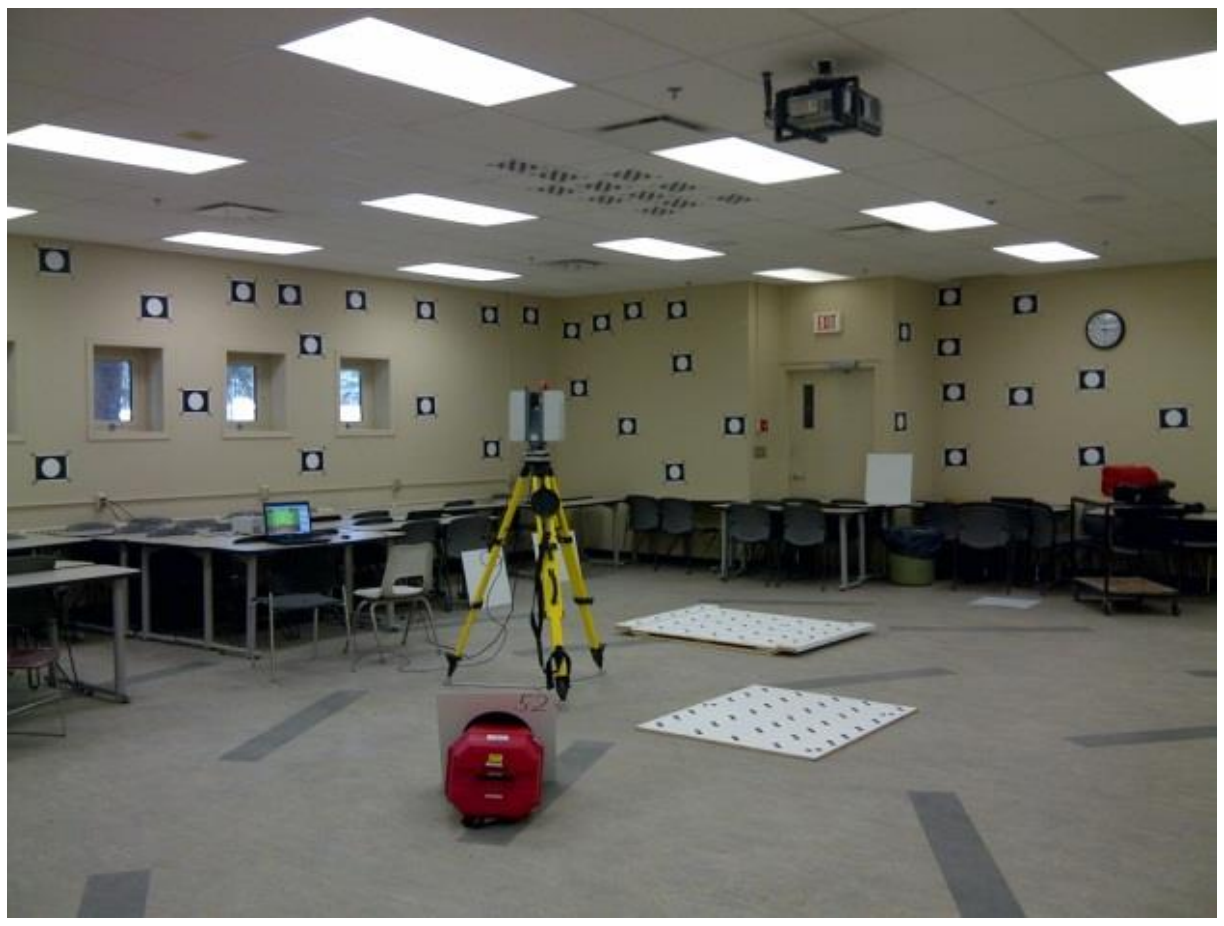

Evaluating the success of the point-based self-calibration is quite intuitive. For instance a check point analysis can be performed, or the RMSE of the residuals and/or the standard deviation of the observations estimated by VCE can be compared before and after error modeling [18]. On the other hand, evaluating the plane-based self-calibration is less intuitive. Planes are difficult to survey with accuracy at least an order of magnitude better than the TLS results, especially since some TLS instruments can measure ranges with sub-millimetre accuracy at close-range. As mentioned in Section 2, statistics computed from the residuals are biased because of the large amount of zero residuals. At close-range the improvements to the misclosure vector obtained from calibration are fairly small and are not an appropriate metric for evaluation of the error modeling. Therefore, in this paper the plane-based self-calibration is deemed effective when the recovered APs are comparable to the point-based self-calibration results. Since naively comparing the individual APs separately has been shown to be a flawed approach [36], comparing the APs as a group through simulation is used instead [37]. In all APs comparisons below, 1,000 simulations in a rectangular room having the same dimensions and same observational noise as the real data were tested. A total of 100 targets at different distances within the measurement volume were tested with 10 points chosen to be used as control.

In hybrid mode, the HDS6100 can observe up to the zenith, and by densifying the targets to cover the extrema of vertical angles, the horizontal collimation axis error can be recovered with a standard deviation of a few arc seconds. This is even lower than the value reported in Table 1, with one of the drawbacks being that the maximum correlation is higher than when tilted scans were added. It appears that if the hybrid scanner has a wide enough vertical field of view, then the four fundamental systematic errors can be recovered with high precision without any special constraints other than organizing the targets to have a good coverage at the maximum and minimum vertical angles. This finding has been confirmed through simulation. Using the same point clouds and calibrating the scanner using planar primitives, a slightly different set of APs were recovered. A difference in $A_0$ should not come as a surprise since the rangefinder offset is known to be a function of the scanner, target fitting algorithm, and properties of the target. Based on the APs compatibility test described earlier, the AP sets recovered using either primitive were found to be comparable for the HDS6100 (Tables 3 and 4). The results are similar regardless of whether the scanner was levelled using the built-in dual axis compensator and the levelling constraints were applied or not. In the plane-based self-calibration of the HDS6100, it is worth mentioning that planes with various orientations were used, and this led to parameter decoupling, as indicated by the low correlation of the APs. If the vertical FOV is restricted due to the scanner design, or targets could not be placed at the vertical extrema of the scanner, then tilted scans appears to be a viable alternative. Table 5 shows the plane-based calibration results of the Imager 5003 using the same data as shown in Table 2. With tilted scans the maximum correlations in the plane-based calibration are reduced for $B_6$ and $B_7$, at the expense of inflating the correlation for $A_0$ and $C_0$ when compared to only using levelled scans; however the precision of all the recovered systematic errors is improved. More importantly, when compared to the point-based calibration results, only the group of APs estimated with tilted scans is comparable.

**Table 5.** Z+F Imager 5003 plane-based self-calibration.

| | Without Tilted Scans | | | With Tilted Scans | | |
|---|---|---|---|---|---|---|
| **# of Targets/Planes** | 109 | | | 114 | | |
| **# of Scans** | 4 | | | 6 | | |
| **# of Obs.** | 50781 | | | 63507 | | |
| **# of Unk.** | 464 | | | 496 | | |
| $\rho_{min}, \rho_{max}$ **[m]** | 2.2, 24.1 | | | 2.2, 33.8 | | |
| $\alpha_{min}, \alpha_{max}$ **[°]** | −57, 236 | | | −65, 244 | | |
| | **Value** | **σ** | **Max Abs. Correlation (w/parameter)** | **Value** | **σ** | **Max Abs. Correlation (w/parameter)** |
| $A_0$ **[mm]** | 0.8 | 0.1 | 0.06 ($B_6$) | −1.0 | 0.1 | 0.15 ($B_6$) |
| $C_0$ **[“]** | 18.6 | 1.8 | 0.06 ($\omega$) | −0.3 | 1.4 | 0.33 ($B_7$) |
| $B_6$ **[“]** | 1.8 | 0.8 | 0.49 ($B_7$) | −3.3 | 0.6 | 0.11 ($B_7$) |
| $B_7$ **[“]** | 22.9 | 7.5 | 0.49 ($B_6$) | −19.1 | 2.0 | 0.33 ($C_0$) |

Not only does this show that the two methods can give similar results, some benefits of plane-based self-calibration are also realized. In general, when accounting for only the four fundamental systematic errors, the plane-based self-calibration can yield the same level or better standard deviations and correlations. This, along with the expedited workflow from data capture (*i.e.*, no need to prepare a room full of signalized targets) to processing (*i.e.*, automatic plane identification and matching), makes it the more desirable calibration method of the two.

# 6. Conclusions

The self-calibration of TLS instruments using point and planar primitives has been studied. Through simulated and real data, it has been demonstrated that having tilted scans in the network can improve the point-based self-calibration results, especially for hybrid scanners. For the first time, the horizontal collimation axis error in hybrid scanners was solved with reasonable correlation and standard deviation. If the hybrid scanner has a large field of view, the collimation axis error can even be recovered with low standard deviation by placing targets near zenith and the tripod legs. Tilted scans were also applied to the calibration of a panoramic scanner, and improvements in the quality of the recovered APs in both point-based and plane-based calibrations were observed.

Although plane-based calibration has shown potential to be a more efficient candidate for modelling systematic errors in laser scanners, some noticeable drawbacks such as poorer recoverability of vertical angular errors have been identified. In simulation, when given identical data, plane-based calibration resulted in higher standard deviations and correlations than point-based calibration. However, it was shown using real datasets that if some diversity exists in the plane orientations, then plane-based calibration can deliver a similar set of APs as the better studied point-based calibration. It can also achieve lower standard deviations and almost complete parameter de-coupling for the four fundamental systematic errors.

The results of this work can help improve the usability of TLS instruments by increasing the geometric accuracy of the data and reducing the need to send the scanner to the manufacturer for tuning. In applications that require 3D reconstruction and/or geometric modelling (e.g., infrastructure documentation) the fit between the observations and the model can be improved. In addition, the overall processing time may be decreased due to reduced necessity of point cloud filtering/smoothing [6]. For deformation monitoring it can even lead to higher detection sensitivity. Improved point cloud accuracy can also reduce errors during point cloud classification. Future research will focus on the attainable quality of the plane-based self-calibration routine under various network geometries and study ways of improving and independently validating the results from this method.

**Acknowledgments**

The authors would like to sincerely thank the Natural Sciences and Engineering Research Council of Canada, the Canada Foundation for Innovation, Alberta Innovates, Informatics Circle of Research Excellence, Killam Trusts, Micro Engineering Tech Inc., and SarPoint Engineering Ltd. for funding this research.

**References**

1. Chow, J.; Lichti, D.; Teskey, W. Accuracy Assessment of the Faro focus3D and leica HDS6100 Panoramic Type Terrestrial Laser Scanner through Point-based and Plane-based User Self-Calibration. In Proceedings of FIG Working Week **2012**: Knowing to Manage the Territory, Protect the Environment, Evaluate the Cultural Heritage, Rome, Italy, 6–10 May **2012**.

2. Lichti, D.; Stewart, M.; Tsakiri, M.; Snow, A. Calibration and Testing of a Terrestrial Laser Scanner. *The International Archives of Photogrammetry and Remote Sensing 33 (Part B5),* Amsterdam, Netherlands, 16–22 July **2000**; pp. 485-492.

3. Reshetyuk, Y. Calibration of terrestrial laser scanners callidus 1.1, Leica HDS 3000 and Leica HDS 2500. *Surv. Rev.* **2006**, *38*, 703–713.

4. Schulz, T. Calibration of a Terrestrial Laser Scanner for Engineering Geodesy. Ph.D Thesis, ETH Zurich, Zurich, Switerland, **2007**.

5. González-Aguilera, D.; Rodríguez-Gonzálvez, P.; Armesto, J.; Arias, P. Trimble GX200 and Riegl LMS-Z390i sensor self-calibration. *Opt. Express* **2011**, *19*, 2676–2693.

6. García-San-Miguel, D.; Lerma, J.L. Geometric calibration of a terrestrial laser scanner with
local additional parameters: An automatic strategy. *ISPRS J. Photogramm. Remote Sens.* **2013**, *79*, 122–136.

7. Lichti, D.; Brustle, S.; Franke, J. Self-Calibration and Analysis of the Surphaser 25HS 3D Scanner. In Proceedings of Strategic Integration of Surveying Services, FIG Working Week 2007; Hong Kong, China, 13–17 May **2007**; [On CD-ROM].

8. Reshetyuk, Y. A unified approach to self-calibration of terrestrial laser scanners. *ISPRS J. Photogramm. Remote Sens.* **2010**, *65*, 445–456.

9. Lichti, D. Terrestrial laser scanner self-calibration: Correlation sources and their mitigation. *ISPRS J. Photogramm. Remote Sens.* **2010**, *65*, 93–102.

10. Rabbani, T.; Dijkman, S.; van den Heuvel, F.; Vosselman, G. An integrated approach for modelling and global registration of point clouds. *ISPRS J. Photogramm. Remote Sens.* **2007**, *61*, 355–370.

11. Lichti, D.; Chow, J.; Lahamy, H. Parameter de-correlation and model-identification in hybrid-style terrestrial laser scanner self-calibration . *ISPRS J. Photogramm. Remote Sens.* **2011**, *66*, 317–326.

12. Lichti, D.; Franke, J. Self-Calibration of the iQsun 880 Laser Scanner. In Proceedings of the Optical 3-D Measurement Techniques VII, Vienna, Austria, 3–5 October **2005**; pp. 122–131.

13. Lichti, D.; Licht, M. Experiences with Terrestrial Laser Scanner Modelling and Accuracy Assessment. *The International Archives of Photogrammetry, Remote Sensing and Spatial Information Sciences 36 (Part 5),* Dresden, Germany, 25–27 September **2006**; pp. 155–160.

14. Dorninger, P.; Nothegger, C.; Pfeifer, N.; Molnár, G. On-the-job detection and correction of systematic cyclic distance measurement errors of terrestrial laser scanners. *J. Appl. Geod.* **2008**, *2*, 191–204.

15. Schneider, D. Calibration of a Riegl LMS-Z420i Based on a Multi-Station Adjustment and a Geometric Model with Additional Parameters. *International Archives of Photogrammetry, Remote Sensing and Spatial Information Sciences38 (Part 3/W8)*, Paris, France, 1–2 September **2009**; pp. 177–182.

16. Amiri Parian, J.; Gruen, A. Sensor modeling, self-calibration and accuracy testing of panoramic cameras and laser scanners. *ISPRS J. Photogramm. Remote Sens.* **2010**, *65*, 60–76.

17. Abmayr, T.; Dalton, G.; Hätrl, F.; Hines, D.; Liu, R.; Hirzinger, G.; Fröhlich, C. Standardization and Visualization of 2.5D Scanning Data and Color Information by Inverse Mapping. In Proceedings of Optical 3-D Measurement Techniques VII; Vienna, Austria, 3–5 October **2005**; pp. 164–173.

18. Lichti, D. Modelling, calibration and analysis of an AM-CW terrestrial laser scanner. *ISPRS J. Photogramm. Remote Sens.* **2007**, *61*, 307–324.

19. Reshetyuk, Y. Self-Calibration and Direct Georeferencing in Terrestrial Laser Scanning. Ph.D Thesis, Royal Institute of Technology, Stockholm, Sweden, **2009**.

20. Rüeger, J. *Electronic Distance Measurement: An Introduction*, 3rd ed.; Springer-Verlag: Heidelberg, Germany, **1990**.

21. Staiger, R. Terrestrial Laser Scanning — Technology, Systems and Applications. In Proceedings of the 2nd FIG Regional Conference; Marrakech, Morocco, 2–5 December **2003**.

22. Fraser, C. Optimization of precision in close-range photogrammetry. *Photogramm. Eng. Remote Sens.* **1982**, *48*, 561–570.

23. Luhmann, T.; Robson, S.; Kyle, S.; Harley, I. *Close Range Photogrammetry: Principles, Techniques and Applications*; Whittles Publishing: Caithness, UK, **2006**.

24. Gielsdorf, F.; Rietdorf, A.; Gründig, L. A Concept For the Calibration of Terrestrial Laser Scanners. In Proceedings of the FIG Working Week, Athens, Greece, May 22-27 **2004**; [On CD-ROM].

25. Bae, K.; Lichti, D. On-site self-calibration using planar features for terrestrial laser scanners. *The International Archives of the Photogrammery, Remote Sensing and Spatial Information Sciences 36 (Part 3/W52)*, Espoo, Finland, 12–14 September, **2007**; pp. 14–19.

26. Chow, J.; Lichti, D.; Glennie, C. Point-based *versus* plane-based self-calibration of static terrestrial laser scanners. *The International Archives of the Photogrammery, Remote Sensing and Spatial Information Sciences 38 (Part 5/W12)*, Calgary, Canada, 29–31 August, **2011**; pp. 121–126.

27. Glennie, C.; Lichti, D. Static calibration and analysis of the Velodyne HDL-64E S2 for high accuracy mobile scanning. *Remote Sens.* **2010**, *2*, 1610–1624.

28. Glennie, C.; Lichti, D. Temporal stability of the Velodyne HDL-64E S2 scanner for high accuracy scanning applications. *Remote Sens.* **2011**, *3*, 539–553.

29. Skaloud, J.; Lichti, D. Rigorous approach to bore-sight self-calibration in airborne laser scanning. *ISPRS J. Photogramm. Remote Sens.* **2006**, *61*, 47–59.

30. Glennie, C. Calibration and kinematic analysis of the Velodyne HDL-64E LiDAR sensor. *Photogramm. Eng. Remote Sens.* **2012**, *78*, 339–347.

31. Chan, T.; Lichti, D. 3D catenary curve fitting for geometric calibration. *The International Archives of the Photogrammery, Remote Sensing and Spatial Information Sciences 38 (Part 5/W12)*, Calgary, Canada, 29–31 August, **2011**; pp. 259–264.

32. Chan, T.; Lichti, D. Cylinder-based self-calibration of a panoramic terrestrial laser scanner. *The International Archives of the Photogrammery, Remote Sensing and Spatial Information Sciences 39 (Part B5)*, Melbourne, Australia, 25 August – 1 September **2012**; pp. 169–174.

33. Lichti, D.; Chow, J. Inner constraints for planar features. *Photogramm. Record 28(141).* **2013**, 74-85.

34. Förstner, W.; Wrobel, B. Mathematical Concepts in Photogrammetry. In *Manual of Photogrammetry*, 5th ed.; McGlone, J., Mikhail, E., Eds.; American Society of Photogrammetry and Remote Sensing: Bethesda, MD, USA, **2004**; pp 15–180.

35. Soudarissanane, S.; Lindenbergh, R.; Menenti, M.; Teunissen, P. Scanning geometry: Influencing factor on the quality of terrestrial laser scanning points. *ISPRS J. Photogramm. Remote Sens.* **2011**, *66*, 389–399.

36. Habib, A.; Morgan, M. Stability analysis and geometric calibration of off-the-shelf digital cameras. Photogramm. Eng. Remote Sens. **2005**, 71, 773–741.

37. Lichti, D. A method to test differences between additional parameter sets with a case study in terrestrial laser scanner self-calibration stability analysis. ISPRS J. Photogramm. Remote Sens. **2008**, 63, 169–180.

### 3.2 Contributions of Authors

The first author was responsible for composing this article; acquiring the datasets using the Leica scanner; writing the software for simulation, processing, and visualizing the results; and processing and analyzing the data. The second and third authors provided valuable advice and guidance throughout this project, including analyzing the results and suggesting experimental setups. The fourth author was responsible for acquiring the data using the Riegl and Z+F scanners.

# Chapter Four: Systematic Error Modelling for the Microsoft Kinect RGB-D Camera via the Self-Calibration Approach

Calibrating a gaming peripheral for engineering use is equally important, if not more critical, than calibrating the laser scanner. Unlike the survey-grade scanners considered in this thesis, the Kinect was not intended for photogrammetry or engineering, but merely for entertainment. This chapter explains the modified bundle adjustment method for modelling the IOPs, APs, and ROPs of the Kinect. The proposed method was developed based on the Kinect's behaviour reported in Appendix A and is designed to be practical for on-site calibration (to reduce the effect of system instability for these toys). For the preliminary testing of the Kinect's performance for photogrammetry, calibration results using only the depth and RGB images, as well as a multi-plane 3D calibration field, the reader can refer to Appendix A. The article in this chapter presents in detail a new self-calibration method for the Kinect using all retrievable image data (i.e. depth, RGB, and IR images). It is the first algorithm that explicitly models the boresight angles in the depth sensor and provides a measure of precision for all calibration parameters.

## 4.1 Article: Photogrammetric Bundle Adjustment with Self-Calibration of the PrimeSense 3D Camera Technology: Microsoft Kinect

Jacky C.K. Chow and Derek D. Lichti

Abstract—The Kinect system is arguably the most popular 3D camera technology currently on the market. Its application domain is vast and has been deployed in scenarios where accurate geometric measurements are needed. Regarding the PrimeSense technology, a limited amount of work has been devoted to calibrating the Kinect, especially the depth data. However, the Kinect is inevitably prone to distortions, as independently confirmed by numerous users. An effective method for improving the quality of the Kinect system is by modelling the sensor's systematic errors using bundle adjustment. In this paper a method for modelling the intrinsic and extrinsic parameters of the infrared and colour cameras, and more importantly the distortions in the depth image, is presented. Through an integrated marker- and feature-based self-calibration, two Kinects were calibrated. A novel approach for modelling the depth systematic errors as a function of lens distortion and relative orientation parameters is shown to be effective. The results show improvements in geometric accuracy up to 53% compared to uncalibrated point clouds captured using the popular software RGBDemo. Systematic depth discontinuities were also reduced and in the check-plane analysis the noise of the Kinect point cloud was reduced by 17%.

## 1 INTRODUCTION AND BACKGROUND

The Microsoft Kinect has unarguably made an impact in many scientific disciplines (e.g. computer vision, photogrammetry, and robotics) since its first release in November 2010. Although it began as a controller for the Xbox 360 video game console, it was one of the first low-cost and robust off-the-shelf 3D cameras on the market and thus its audience expanded quickly. One of the first uses adopted outside of gaming was in surgery, where a surgeon could scroll through medical images on a computer screen risk-free – simply with the wave of a hand – at Sunnybrook Hospital in Toronto (Canada) [1].

The Kinect is popular for both industrial and research applications. Commercial solutions based on the Kinect are readily available around the world. For example, iPiSoft offers a markerless optical motion capture system based on the Kinect; ReconstructMe, Manctl, and 4DDynamics turn the Kinect into a handheld scanner for 3D object reconstruction; Faceshift uses it to capture facial motions and expressions; and Fitnect uses it to build a virtual dressing room. In the area of research it has been tested as an aiding device for people who are visually impaired [2], for interactive teaching in classrooms [3], as a portable indoor mapper strapped to humans designed for first responders [4], and for motion capture of hands [5], to name a few.

The wide adoption of the Kinect system has resulted in more than 24 million units being sold as of February 2013. Most users of these units assume their Kinect is well-calibrated and is suitable for a wide-range of applications out-of-the-box. However, [6] tested multiple PrimeSense units and found inaccuracies of up to 1.5 cm. Although high precision of an individual unit is reported (suggesting that temporal averaging is not necessary), variations in accuracy of up to 2 cm between manufactured PrimeSense PS1080 devices were presented. Differences between precision and accuracy can suggest the existence of biases and many researchers have independently reported systematic distortions in the Kinect point clouds [7], [8], [9].

To reduce the effect of these systematic errors, numerous efforts have been made in the area of software calibration. One of the first Kinect calibrations was done by [10] in the popular software RGBDemo, where the intrinsic and extrinsic parameters of the infrared (IR) and RGB camera were calibrated based on the OpenCV calibration. Burrus [10] also proposed an algorithm for converting the disparity values to depth measurements; however no calibration routine for deriving these conversion parameters was suggested. This approach can estimate the alignment between the infrared image and the RGB image if libfreenect is used and can improve the alignment if OpenNI is used for data capture.

Similar calibrations for aligning the IR camera with the RGB camera using signalized targets can also be found in [8] where PhotoModeler and Australis were used, and in [11] where PhotoModeler was used. To account for the depth distortions in the point cloud, the disparity values were treated as observations and the baseline distance as well as the distance of the memorized reference pattern, were solved.

Khoshelham and Elberink [11] further added two normalization parameters to their depth calibration explained in [7] for accommodating the quantization of depth values. This approach assumed that there is a zero rotational offset between the infrared camera and projector, and that the depth calibration is independent of the lens distortions in a two-step independent procedure. However, as explained in [12], two-step independent 3D camera calibrations can have its shortcomings. In addition, the Kinect is based on triangulation and hence depth is a function of the image measurements.

Smisek et al. [13] modelled the depth errors in object space by solving for two coefficients of a linear mapping function that minimizes the residuals of a best fit plane. In this case, the depth correction is modelled as a function of distance, which has been shown to be inferior compared to expressing it in image space coordinates [14].

Draelos [15] used depth discontinuities to establish correspondences with the RGB image and used the corners of the planar board for registering the depth image with the RGB image. However, depth discontinuities are unstable and unsuitable for accurate pixel measurements. Furthermore, similar to [13], their depth calibration was independent of the image space.

Based on [16], all of the approaches described thus far do not deliver the optimal set of calibration parameters because 1) the calibrations of the cameras were not performed simultaneously; 2) the error modelling was not performed in image space; or 3) unstable points in the depth map were used. Herrera et al. [16] and [17] presented a method for aligning the RGB image with depth map using the point-on-plane constraint without any depth corrections. As an extension, [14] added a depth distortion model that is dependent on the image space rather than object space in their total system calibration and showed superior performance compared to [13].

Chow et al. [18] has also presented a method for calibrating the Kinect's depth image while simultaneously aligning it with the RGB image using the point-on-plane constraint, but have reported a poor precision in the estimated relative translation and rotation between the depth image and RGB image. To improve the precision they used three orthogonal planes to strengthen the depth and RGB co-registration. Even then, they indicated a lack of constraints to recover the lens distortions of the depth image because the IR images were not used.

Staranowicz and Mariottini [19] compared the approach of [14] and portions of the [17] calibration to a calibration method that uses spheres instead of planes. Their results agreed with [18] and indicated weak recovery of relative orientation parameters between the depth and RGB image when using the point-on-plane approach. However, their work with spherical objects only focused on aligning the depth and RGB images and no calibration model for correcting the depth map was proposed.

With mass production of the Kinect, primarily made for gaming, users attempting to employ this sensor for accuracy-demanding applications such as deformation monitoring and simultaneous localisation and mapping should consider calibrating the Kinect themselves. This paper is an extension of the work in [18] and now includes IR images in the bundle adjustment with self-calibration to improve the alignment between the IR and RGB cameras. It simultaneously calibrates all optical sensors in the Kinect system using a checkerboard pattern and depth measurements measured reliably on the surface of the plane. It is also the first calibration method that explicitly models the rotational offset between the depth image and projector of the Kinect. The depth error resulting from angular misalignments (in particular, from the yaw angle) between stereo pairs has been studied and its effect should not be understated [20]. Although all the data in this paper are captured with the Kinect, it is designed for any devices using the PrimeSense technology (e.g. Asus Xtion PRO, Xtion PRO LIVE and Fotonic P70).

This paper begins by giving a general overview of the Kinect hardware and software for operation in Section 2. Section 3 describes the calibration procedure undertaken in this paper.

Section 4 explains the calibration model developed and Section 5 shows the calibration results of two Kinects and discusses the quality of the calibration solution.

## 2 THE KINECT HARDWARE AND SOFTWARE

The initial Kinect hardware released on November 4, 2010 was designed with the intention of working with the Microsoft Xbox 360 only. The first Kinect for Windows version was not released until February 1, 2012. Compared to the Kinect for Xbox, the Kinect for Windows is not too different aside from offering a closer sensing distance (40 cm instead of 80 cm) and redesigned cabling for easier PC connection. However, in terms of the fundamental depth-sensing principle of the optical sensor at its core, the Kinect for Xbox and the Kinect for Windows are the same. They are also the same as the 3D triangulation-based cameras from Asus and Fotonic, as they are all based on the PrimeSense technology. Differences between these sensors stem from other design specifications; for example, Asus sells their 3D camera with or without a colour (RGB) camera built in.

In general, each one of these systems consists of three optical units: an RGB camera, an IR projector, and an IR camera. The projector emits light in the infrared spectrum and illuminates the scene with a speckle pattern generated from a set of diffraction gratings. Through a 9 by 9, 9 by 7, or 7 by 7 spatial multiplexing window, the pixel showing the highest correspondence among its 64 horizontal neighbours is selected in the infrared image as the corresponding point [21]. Further sub-pixel refinement then gives it a measurement accuracy of approximately $1/8^{th}$ of a pixel [22].

The optical axes of the projector and IR camera are nominally parallel and are separated by a baseline distance of approximately 7.5 cm. Through photogrammetric triangulation the depth can then be determined. The Kinect stores the disparity value of every pixel at a calibrated distance; therefore a difference between the measured disparity and reference disparity translates into a change in depth [23]. If colour information is desired (for instance in segmentation/classification applications), the RGB camera situated at approximately 2.5 cm from the IR camera can overlay 8-bit 3-channel red, green, blue information over the point cloud.

A standard Kinect has a vertical and horizontal field of view (FOV) of 43° and 57°, respectively. This can be extended by equipping a Kinect with the Nyko Zoom add-on. Although these additional lenses give the Kinect a wider FOV, they will likely increase the magnitude of lens distortion and will require a geometric calibration before they can be used for precise applications.

The raw RGB and IR images, as defined by the Aptina MT9M112 and Micron MT9M001 CMOS sensors respectively, are 1280 pixels by 1024 pixels [24]. Although most APIs allow access to higher resolution images (e.g. SXGA), it comes at the cost of a reduced frame rate. For depth acquisition at 30 Hz using a USB2.0 connection, VGA resolution is usually used due to bandwidth limitations.

For this paper, two Kinect for Xbox sensors were used. All images were captured using the standard VGA resolution to ensure that the IR images are calibrated at the same image resolution as the depth images. Among the various options for operating the Kinect, the Microsoft Kinect SDK was chosen instead of the open source library OpenKinect and the popular OpenNI. Most Kinect calibration work to date has been using OpenNI, as it is conveniently packaged into OpenCV and PCL, but the calibration model in this paper is software-independent and is applicable to Kinect data captured using any driver.

## 3 CALIBRATION PROCEDURE

Following a two hour warm-up period, depth images, IR images, and RGB images of a checkerboard target were acquired from various positions and orientations. At every exposure, 20 consecutive depth images were captured and averaged to reduce the random noise of the depth measurements and to fill in holes in the depth map. Although [6] and [18] suggested that improvements to range precision through temporal averaging is small (e.g. 1 mm improvement at 3 m distance), likely due to the low depth resolution, it can be done easily.

Since the Kinect cannot capture both the RGB and IR images at the same time, the RGB images and depth images are captured together first. Afterwards, the projector is covered and the IR image of the scene illuminated by an external light source is captured, which is an approach similar to [25]. This ensures good contrast in the IR image in the absence of disturbance from the projector. In the current Microsoft Kinect SDK 1.7 the projector can be switched off, making this step less cumbersome.

The observations in the adjustment can be categorized into three groups: image coordinates in the RGB images ($x^{RGB}$, $y^{RGB}$), image coordinates in the IR images ($x^{IR}$, $y^{IR}$), and image coordinates as seen by the projector ($x^{PRO}$, $y^{PRO}$), which were derived from the depth values retrieved from the SDK. The image coordinates from both cameras were obtained by measuring the corners of a checkerboard pattern using the MATLAB Camera Calibration Toolbox. The depth measurements were made by selecting a randomly distributed set of pixels in the depth image that belonged to the same plane as the checkerboard pattern. For pre-adjustment screening, a plane was fit to the point cloud derived from the depth images and points are removed using Baarda's data snooping.

# 4 MATHEMATICAL MODEL

The user self-calibration method presented in this paper is based on the pin-hole camera model given in Equation 1. To model departures from collinearity, Brown's model [26] for radial lens and decentring lens distortion is augmented with a model for in-plane distortions (i.e. affinity and shear), as shown in Equation 2. Equations 1 and 2 together form the standard photogrammetric bundle adjustment with self-calibration model and are the basis of our proposed calibration method [27].

$$x_{ij} = x_p - c\frac{m_{11}(X_i - X_{oj}) + m_{12}(Y_i - Y_{oj}) + m_{13}(Z_i - Z_{oj})}{m_{31}(X_i - X_{oj}) + m_{32}(Y_i - Y_{oj}) + m_{33}(Z_i - Z_{oj})} + \Delta x$$

$$y_{ij} = y_p - c\frac{m_{21}(X_i - X_{oj}) + m_{22}(Y_i - Y_{oj}) + m_{23}(Z_i - Z_{oj})}{m_{31}(X_i - X_{oj}) + m_{32}(Y_i - Y_{oj}) + m_{33}(Z_i - Z_{oj})} + \Delta y \quad (1)$$

where $x_{ij}$ and $y_{ij}$ are the image coordinates of point i in image j; $x_p$ and $y_p$ are the principal point offsets; c is the principal distance; $X_i$, $Y_i$ and $Z_i$ are the object space coordinates of point i; $X_{oj}$, $Y_{oj}$ and $Z_{oj}$ are the position of image j in object space; $m_{11} \ldots m_{33}$ are the elements of the rotation matrix defining the orientation of image j and expressed using the Cardan angle sequence ($\omega_j$, $\varphi_j$ and $\kappa_j$); $\Delta x$ and $\Delta y$ are the correction terms of additional calibration parameters.

$$\begin{aligned}\Delta x &= x'_{ij}\left(k_1 r_{ij}^2 + k_2 r_{ij}^4 + k_3 r_{ij}^6\right) + p_1\left(r_{ij}^2 + 2{x'_{ij}}^2\right) + 2p_2 x'_{ij}\, y'_{ij} + a_1 x'_{ij} + a_2 y'_{ij} \\ \Delta y &= y'_{ij}\left(k_1 r_{ij}^2 + k_2 r_{ij}^4 + k_3 r_{ij}^6\right) + p_2\left(r_{ij}^2 + 2{y'_{ij}}^2\right) + 2p_1 x'_{ij}\, y'_{ij}\end{aligned} \tag{2}$$

where $x'_{ij}$ and $y'_{ij}$ are the image coordinates of point i in image j after correcting for the principal point offset; $r_{ij}$ is the radial distance of point i in image j relative to the principal point; $k_1$, $k_2$ and $k_3$ are the radial lens distortion coefficients; $p_1$ and $p_2$ are the decentring lens distortion coefficients; $a_1$ and $a_2$ describe the in-plane affinity and shear distortions, respectively.

In this conventional form, the object space target coordinates {$X_i$, $Y_i$ and $Z_i$}; interior orientation parameters (IOPs) {$x_p$, $y_p$ and c}; additional parameters (APs) {$k_1$, $k_2$, $k_3$, $p_1$, $p_2$, $a_1$, $a_2$}; and exterior orientation parameters (EOPs) {$\omega_j$, $\varphi_j$, $\kappa_j$, $X_{oj}$, $Y_{oj}$, $Z_{oj}$} of both the IR and RGB cameras can already be calibrated simultaneously.

As shown in [28] and [29] the calibration of stereo cameras can be improved by constraining the six relative orientation parameters (ROPs) between the stereo pair to be the same at every exposure. Success of ROP constraints has also been demonstrated for systems with three cameras [30] and more [31]. In the least-squares adjustment, this constraint can be realized by adding equations/observations [32], [33] or be integrated directly into the collinearity equations [29], [34]. It was further explained in [34] and [35] that expressing this in the functional model rather than as a constraint equation can reduce the computational load and allow the rotations, translations, and their corresponding standard deviations to be estimated directly.

The optical sensors in the Kinect are rigidly mounted together on a metallic frame. Based on our literature review, there are no reasons to believe that the relative positions and orientations between the internal optical sensors change significantly when being handled with care over a short period of time [6]. Therefore a modified collinearity equation shown in Equation 3 is used for self-calibration instead, where the ROPs are introduced.

The four relevant right-handed coordinate systems (IR = infrared camera, RGB = colour camera, PRO = projector, and OBJ = object space) are illustrated in Figure 1. The notation adopted in this paper is as follows: a superscript of the frame alone means the quantity is expressed in that particular coordinate system (e.g. $[p_{ij}]^{RGB}$ is a vector observed in the RGB-frame); when both subscript and superscript of a coordinate system exist, it represents from *subscript-frame* to *superscript-frame* (e.g. $[R_j]_{OBJ}^{IR}$ is a matrix defining rotations from OBJ-frame to IR-frame).

$$p_{ij}^{RGB} - \frac{1}{\mu_{ij}^{RGB}} \Delta R_{IR}^{RGB} \left[R_j\right]_{OBJ}^{IR} \left(O_i^{OBJ} - \left[T_j\right]_{Obj}^{IR} - \left[R_j\right]_{IR}^{OBJ} b_{IR}^{RGB}\right) = 0 \tag{3}$$

where $p_{ij} = [x'_{ij} - \Delta x \quad y'_{ij} - \Delta y \quad c]^T$ is the corrected image coordinate vector for point i in image j ; $\mu_{ij}$ is the unique scale factor for point i in image j; $\Delta R$ is the relative rotation matrix defined by rotation angles about the primary ($\Delta\omega$), secondary ($\Delta\phi$) and tertiary axis ($\Delta\kappa$); $R_j$ is the rotation matrix of image j defined by rotations about the primary ($\omega_j$), secondary ($\phi_j$) and tertiary axis ($\kappa_j$); $O_i$ is the object space coordinates, $[X_i \quad Y_i \quad Z_i]^T$; $T_j$ is the translation vector of image j, $[X_{oj} \quad Y_{oj} \quad Z_{oj}]^T$; b is the relative translation vector, $[b_x \quad b_y \quad b_z]^T$.

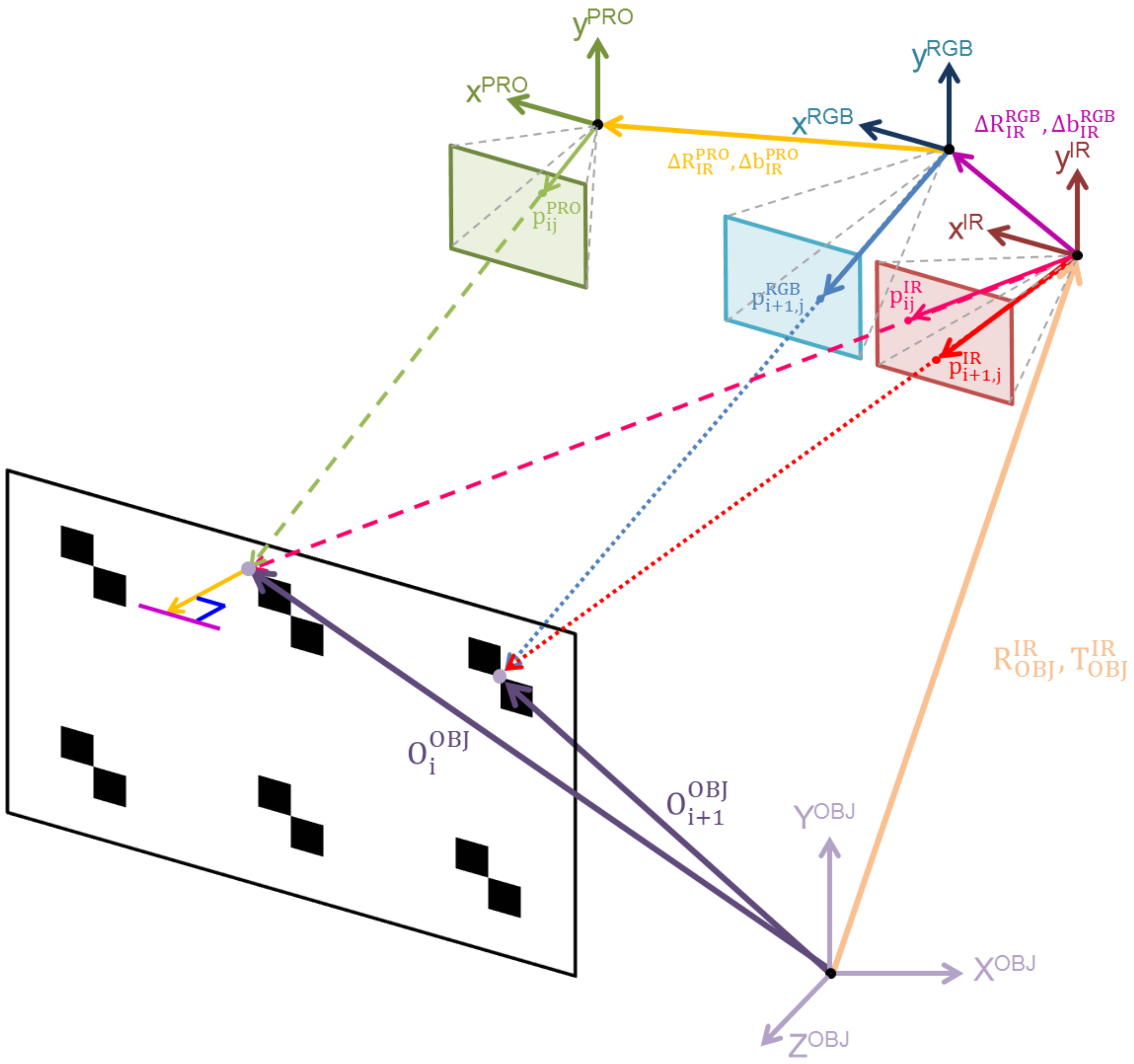

Fig. 1. Definition of the coordinate systems for the calibration.

The above model is sufficient to model the systematic errors in both the IR and RGB cameras; however, it does not characterize the depth measurements of the Kinect. The depth information is determined by triangulation from the IR camera and projector pair. Therefore, it is modelled as a function of the IOPs and APs of both sensors and the six ROPs defining the stereo pair.

As shown by [21] there is a null band in the depth images corresponding to the use of a 9 by 9 or 9 by 7 correlation window. Only after correcting for this offset can the IR images and depth images share the same EOPs, IOPs and APs. A similar correction was done in [11], [13] to align the depth map with the IR image. With knowledge about the intrinsics of the IR camera, the object space coordinates of every pixel in the depth map can be determined using Equation 4.

$$
\begin{aligned}
X_i^{IR} &= -\frac{Z_i^{IR}}{c^{IR}}\left(x_i^{IR} - x_p^{IR} - \Delta x^{IR}\right) \\
Y_i^{IR} &= -\frac{Z_i^{IR}}{c^{IR}}\left(y_i^{IR} - y_p^{IR} - \Delta y^{IR}\right)
\end{aligned}
\tag{4}
$$

By knowing the ROPs between the IR camera and projector, the object space coordinates can be back-projected into the image space of the projector. As in most camera-projector calibrations [36], [37] the projector can be treated like a camera; however in this case we do not know the structure of the projected pattern and thus, the IOPs and APs of the projector cannot be recovered reliably. To complicate the problem, the extrinsic parameters between the projector and IR camera are initially unknown. Hence the APs, IOPs, and ROPs of the projector need to be solved iteratively through forward and backward projections due to the non-linear nature of the collinearity equations.

With the image measurements from the RGB camera, the IR camera, and the projector, a bundle adjustment can be performed. To strengthen the calibration, the point-on-plane constraint has been included in the bundle adjustment, which minimizes the residuals orthogonal to the plane [38]. This is necessary for calibrating the depth map, as checkerboard patterns cannot be seen – only geometric features can be identified. Although this restricts the calibration field to a planar 2D target field rather than a 3D volume, a planar target field is portable and is practical for on-site calibration. The drawback of a 2D calibration field (i.e. projective compensation) can be mitigated by imaging a planar checkerboard pattern with converging geometry from various positions and perspectives.

The calibration model for the IR camera and RGB camera with relative translational and rotational constraints was shown in Equation 3. Likewise, the functional model for the depth-projector pair is given in Equation 5. The plane constraint is expressed by the scale factor term $\mu_{ij}$ which may be solved by substituting Equation 3 into Equation 6. This final calibration model minimizes the discrepancy between conjugate light rays while constraining them to lie on the best fit plane by solving for the EOPs, ROPs, IOPs, APs, and object space quantities simultaneously.

$$\begin{aligned} &\mu_{ij}^{PRO} R_{IR}^{OBJ} R_{PRO}^{IR} p_{ij}^{PRO} + \left[T_j\right]_{OBJ}^{IR} + R_{IR}^{OBJ} b_{Pro}^{IR} \\ &\quad - \left(\mu_{ij}^{IR} R_{IR}^{OBJ} p_{ij}^{IR} + \left[T_j\right]_{OBJ}^{IR}\right) = 0 \end{aligned} \tag{5}$$

$$[a_k \quad b_k \quad c_k] \bullet O_i^{OBJ} - d_k = 0 \tag{6}$$

where $a_k$, $b_k$, and $c_k$ are the direction cosines of the normal vector of the best-fit plane; $d_k$ is the orthogonal distance from the origin to the plane.

Unlike in the MATLAB Camera Calibration Toolbox [39] where the datum is defined by assuming all the object space coordinates are fixed, inner constraints are applied to the object space coordinates, the plane parameters and the EOPs [40]. This is done to improve the overall estimation precision and, most importantly, to prevent possible object space coordinate errors from propagating into the IOPs, APs, and/or ROPs which would result in a biased calibration. The collinearity and coplanarity equations are highly non-linear so the Gauss-Helmert least-square model has been chosen for minimizing the summation of the weighted residuals [41]. Baarda's data snooping with a 5% level of significance was used to minimize the possibility of outliers in the adjustment as the least-squares method is known to be highly sensitive to erroneous observations. Iterative variance component estimation has also been adopted for re-weighting the various observation groups (i.e. IR image, depth image, projector, and RGB image) as part of the adjustment [42].

The observations were assumed to be additive zero-mean Gaussian distributed errors uncorrelated with each other [22], and this has shown to be capable of describing the depth noise of the Kinect as a function of distance up to 10 m [18]. The effect of quantization of the disparity measurements is included in the stochastic model [43] and is given in Equation 7. The Kinect's disparities are normalized and quantized for streaming as 11-bit integers with the first bit indicating whether a depth measurement is valid or not, therefore further reducing their range to 1024 levels [11]. As suggested in [22], if the disparity values range between 2 and 88 pixels and are measured with a precision of 1/8 of a pixel ($\sigma_{Disparity}$), then by using the nominal focal length and baseline distance the effect of quantization step (q) on the depth reconstruction precision ($\sigma_Z$) as determined by Monte Carlo Simulation (1000 simulations per level) is given in Figure 2. This shows that the quantization has a relatively small effect on the depth reconstruction accuracy, which agrees with the findings reported in [11].

$$E\{\text{Total Disparity}^2\} = \sigma^2_{\text{Disparity}} + \frac{q^2}{12} \quad (5)$$

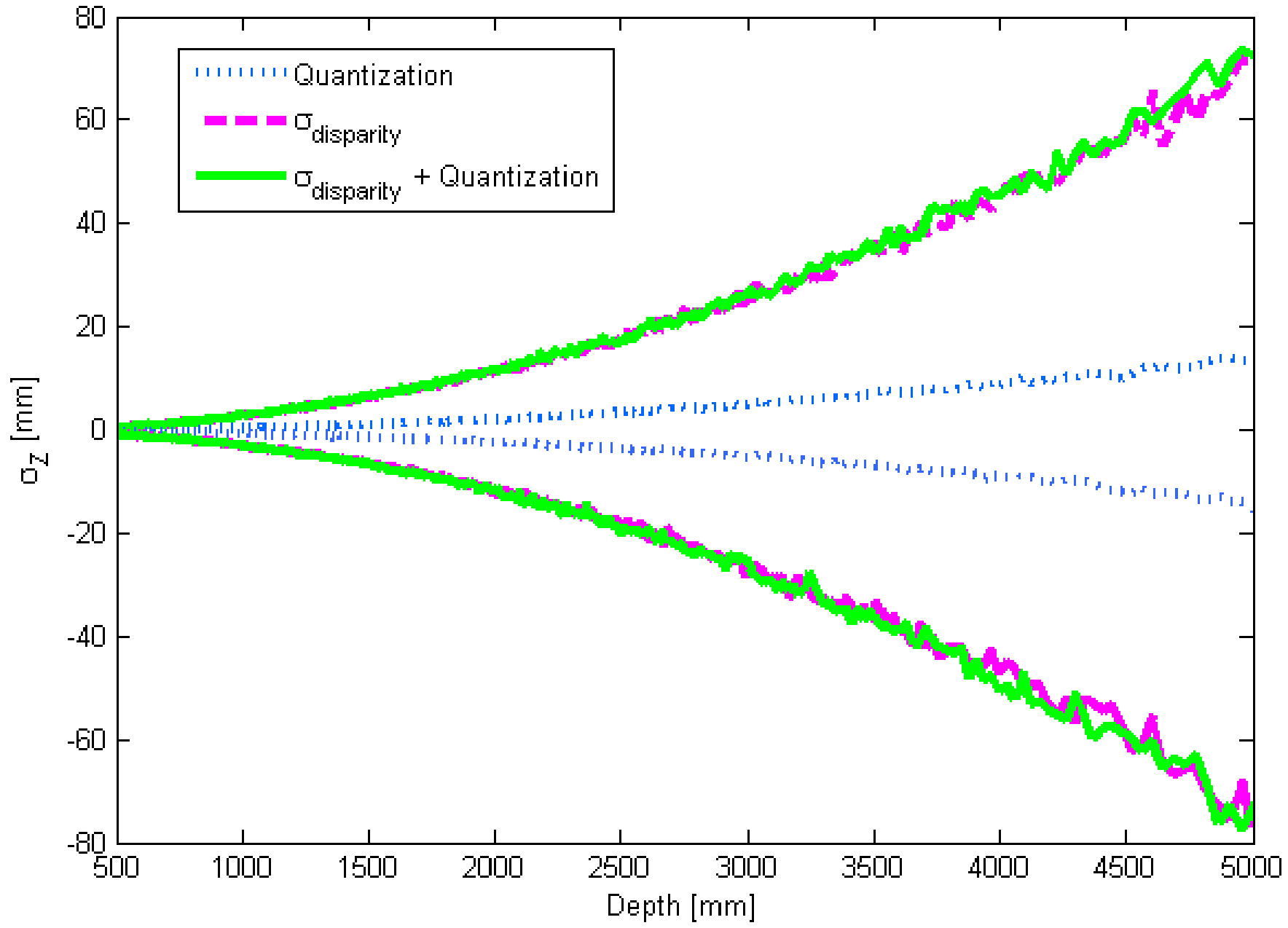

Fig. 2. Effect of the stochastic model on the accuracy of depth reconstruction. The dashed line (magenta) shows the uncertainty of the reconstructed depth if the disparity measures are continuous. The dotted line (blue) is the effect of quantization. The solid line (green) shows the result of quantized disparity values.

This calibration method follows the three requirements for a 3D camera calibration laid out by [16]. It is "accurate" because the object space target coordinates, plane parameters, EOPs, ROPs, IOPs and APs of all 3 optical sensors are estimated simultaneously in the same bundle adjustment; it is "practical" in that only a single planar checkerboard target is required; and it is "widely applicable" because depth measurements lying in the bounds of the plane are used instead of depth discontinuities.

Unlike [11], who assumed the calculated depth is independent of the lens distortion, we assume planimetric and depth coordinates are a function of the IR camera's APs. But as in [11] the proposed depth calibration is a function of the IR camera's principal distance and baseline distance. Parameters that account for misalignment between the IR camera's axes and the projector's axes are also included as the Cardan angle sequence.

Similar to [14], every depth pixel has a different calibration coefficient expressed in image coordinate units. However, the number of unknowns being solved is significantly lower in this case because they can be conveniently expressed by the APs. In addition, the IR images are used directly for the mutual de-correlation of parameters rather than using an external high-resolution camera since [14] has demonstrated that the improvement is small and external cameras can complicate the calibration procedure.

In summary, the unknown parameters in the adjustment are the principal point offset, principal distance, and lens distortion parameters of the IR and RGB cameras ($IOP^{IR}$, $AP^{IR}$, $IOP^{RGB}$, $AP^{RGB}$), the rotational and translational parameters of the projector relative to the IR camera ($ROP_{IR}^{PRO}$) and the RGB camera relative to the IR camera ($ROP_{IR}^{RGB}$), the IR camera orientation and position relative to the object space ($EOP_{OBJ}^{IR}$), the object space target coordinates, and the plane parameters.

Beginning with the initial approximations of zero principal point offset, APs and rotational offsets, a 2.9 mm principal distance and a 7.5 cm positional offset in the x-direction between the IR camera and projector, the depth observations ($D^{IR}$) were converted into image coordinate measurements of the projector ($x^{PRO}$, $y^{PRO}$) by back-projection (Equation 3). The linearized least-squares optimization is then carried out using the math model presented in this section. After bundle adjustment, with the current best estimate of the unknown parameters and $D^{IR}$, the point cloud is back-projected into the projector again and with the updated $x^{PRO}$ and $y^{PRO}$ the bundle adjustment is repeated until convergence.

# 5 RESULTS AND DISCUSSION

## 5.1 Self-Calibration Results

To evaluate the proposed method, two Kinect for Xbox sensors were calibrated following the above procedure. One planar target with 24 signalized targets was observed by one Kinect (hereafter "Kinect1") from 11 different poses and 48 planar signalized targets were imaged by another Kinect (hereafter "Kinect2") from 10 different stations (Figure 3). The number of observations in the bundle adjustment is 1972 for Kinect1 and 3152 for Kinect2, with an average redundancy number of 0.93 and 0.94 respectively, yielding a well-controlled network in both cases.

The estimated IOPs and the statistically significant APs for the IR camera and RGB camera are displayed in Table 1. The estimated relative translation and rotation between the three optical sensors are given in Table 2. The initial estimates of 7.5 cm and 2.5 cm for the baseline distances between the IR camera and projector and the IR camera and RGB camera are close to the recovered values. However, detectable rotational offsets between the IR camera and projector were found in one of the Kinects. Compared to the previous work in [18] the standard deviations of $ROP_{IR}^{RGB}$ are significantly improved (by up to 46x for some parameters) with the inclusion of the IR images.

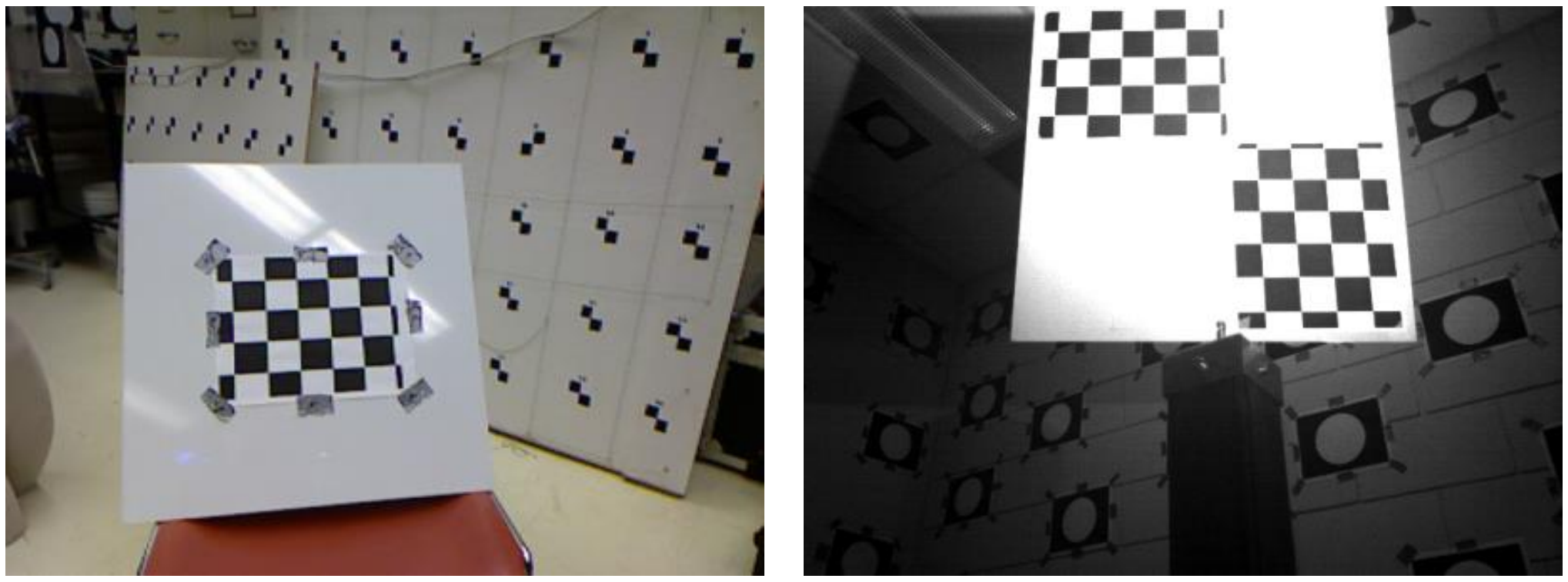

Fig. 3. (Left) RGB image of the calibration field used for calibrating Kinect1. (Right) IR image of the calibration field used for calibrating Kinect2.

TABLE 1

INTRINSIC PARAMETERS RECOVERED FROM CALIBRATION

| | Kinect1 | | Kinect2 | |
|---|---|---|---|---|
| | IR camera | RGB camera | IR camera | RGB camera |
| xp [µm ± µm] | -30.2 ± 2.7 | 52.1 ± 2.1 | -100 ± 1.1 | 8.5 ± 2.1 |
| yp [µm ± µm] | -0.4 ± 2.5 | -175.9 ± 2.3 | -56.8 ± 0.1 | -162 ± 1.9 |
| c [mm ± mm] | 6.045 ± 0.003 | 2.896 ± 0.002 | 6.136 ± 0.001 | 3.017 ± 0.003 |
| $k_1$ [$1/mm^2$ ± $1/mm^2$] | -3.9e-3 ± 1.8e-4 | 2.3e-2 ± 1.5e-3 | -2.5e-3 ± 8.5e-5 | 3.1e-2 ± 1.6e-3 |
| $k_2$ [$1/mm^4$ ± $1/mm^4$] | 3.8e-4 ± 3.3e-5 | -1.0e-2 ± 1.1e-3 | 3.6e-4 ± 1.8e-5 | -1.2e-2 ± 1.3e-3 |
| $k_3$ [$1/mm^6$ ± $1/mm^6$] | -1.2e-5 ± 1.8e-6 | 1.5e-3 ± 2.1e-4 | -1.5e-5 ± 1.1e-6 | 1.6e-3 ± 2.9e-4 |

TABLE 2

RELATIVE ORIENTATION PARAMETERS RECOVERED FROM CALIBRATION

| | Kinect1 | | Kinect2 | |
|---|---|---|---|---|
| | IR-PRO | IR-RGB | IR-PRO | IR-RGB |
| $\Delta\omega$ [arcsec ± arcsec] | 585 ± 81 | 431 ± 149 | 42 ± 23 | -2690 ± 132 |
| $\Delta\phi$ [arcsec ± arcsec] | 844 ± 75 | -1970 ± 111 | -12 ± 28 | -975 ± 121 |
| $\Delta\kappa$ [arcsec ± arcsec] | 152 ± 13 | -2185 ± 22.7 | 17 ± 5 | -1858 ± 24 |
| $b_x$ [mm ± mm] | 76.5 ± 0.1 | 25.6 ± 0.1 | 74.1 ± 0.1 | 27.0 ± 0.3 |
| $b_y$ [mm ± mm] | -0.1 ± 0.1 | 0.7 ± 0.1 | 0 ± 0.1 | 2.2 ± 0.2 |
| $b_z$ [mm ± mm] | -0.9 ± 0.3 | 0.7 ± 0.3 | 0.4 ± 0.2 | 15.4 ± 0.8 |

The RMSE of the misclosure of conjugate light rays lying on the best fit plane before and after modelling for the IOPs, APs, and ROPs of the depth image and projector pair is given in Table 3. The standard deviation of the plane parameter $d_k$ of the best fit plane as estimated simultaneously by all optical sensors is provided as well, as an indication of the quality of the plane used for this assessment; note that even when the systematic errors of the depth map are untreated, the plane parameters are still well-estimated because of the stereo-pair formed by the calibrated IR and RGB images.  Prior to calibration the precisions of the two Kinects differ, but after calibration they are more comparable.  An RMSE of up to 6.3 mm was observed before calibration, but after modeling for the systematic errors in the IR camera and projector the RMSE were all less than 0.2 mm.

A plot showing the residuals of the depth map before and after calibration is provided in Figure 4. Before calibration, reprojection errors up to 52 µm (5 pixels) can be perceived but they were reduced to the sub-pixel level after calibration.

Based on the Aptina and Micon sensor's specifications, the pixel size of the IR and RGB cameras is 10.4 µm and 5.6 µm, with a nominal focal length of 6 mm and 2.9 mm, respectively. With a lack of specifications for the projector, it is initially assumed to have the same specifications as the IR camera. Using variance component estimation, the depth image and projector were measuring with a standard deviation of 1/8 of a pixel for Kinect1 and 1/13 of a pixel for Kinect2 after calibration (as seen in Table 4). This is more precise than measurements of the checkerboard pattern made by either camera using the Camera Calibration Toolbox, which delivers approximately 1/4 of a pixel standard deviation for the IR cameras, and 1/3 and 1/2 of a pixel measurement precision for the RGB cameras of Kinect1 and Kinect2 respectively.

TABLE 3

QUALITY OF PLANE-FIT ESTIMATED IN THE BUNDLE ADJUSTMENT WITH SELF-CALIBRATION

| | Kinect1 (mm) | | | Kinect2 (mm) | | |
|---|---|---|---|---|---|---|
| | Before | After | % Improv. | Before | After | % Improv. |
| $RMSE_X$ | 5.3 | 0.1 | 98% | 2.9 | 0.1 | 97% |
| $RMSE_Y$ | 4.0 | 0.1 | 98% | 6.3 | 0.2 | 97% |
| $RMSE_Z$ | 4.2 | 0.1 | 98% | 3.9 | 0.1 | 97% |
| Std. Dev. of $d_k$ | 1.1 | 0.9 | 18% | 0.6 | 0.5 | 17% |

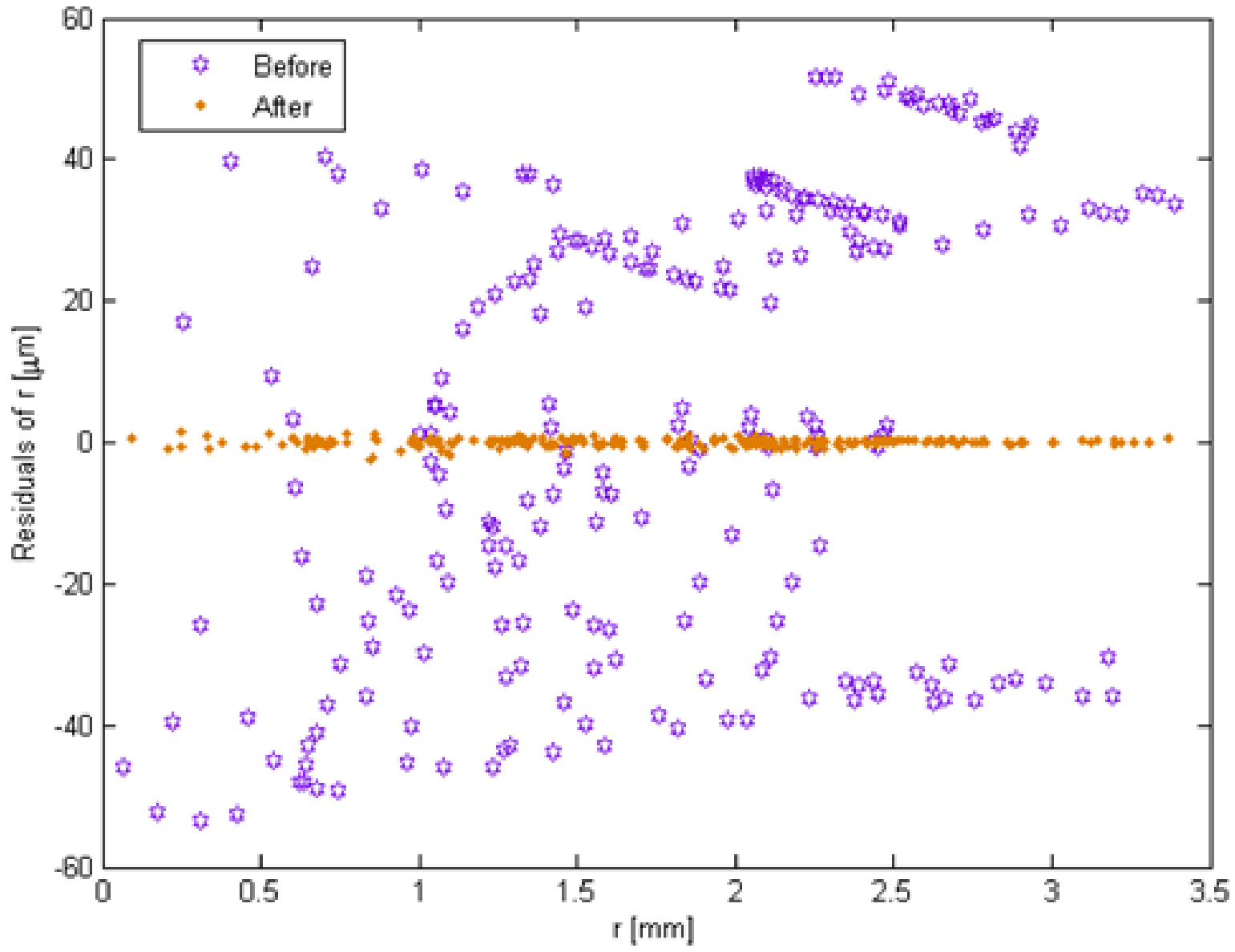

Fig. 4. Residuals of the depth map as a function of the radial distance from the principal point before and after systematic error modelling.

| TABLE 4<br>ESTIMATED STANDARD DEVIATION OF THE OBSERVATION RESIDUALS | | | | | | |
|---|---|---|---|---|---|---|
| | Kinect1 (µm) | | | Kinect2 (µm) | | |
| | Before | After | % Improv. | Before | After | % Improv. |
| IR | 2.1 | 2.8 | 0 | 2.5 | 2.6 | 0% |
| Depth | 28.2 | 1.3 | 95% | 24.8 | 0.8 | 97% |
| Projector | 29.5 | 1.2 | 96% | 26.0 | 0.8 | 97% |
| RGB | 1.6 | 1.7 | 0% | 2.9 | 2.8 | 0% |

With the current calibration model the overall parameter correlation is low: the maximum correlation of the IOPs/APs with the EOPs is 0.26, with the target coordinates is 0.25, and with the plane parameters is 0.14. Some noticeably significant correlations are highlighted in Table 5 (excluding the correlation between the APs, which are already known to exist). Most of the correlation patterns are common between both calibrations, except for the correlations between $x_p^{RGB}$-$\Delta\phi^{IR}$ and $c^{RGB}$-$b_y^{RGB}$ which may have higher dependency on the imaging geometry. As the only significant correlations are between the APs and ROPs, these calibration parameters should be transferable to other datasets captured by the same Kinect. To confirm this, additional data were acquired using Kinect1 to assess the accuracy of the calibration.

TABLE 5
SIGNFICANT PARAMETER CORRELATIONS IN THE BUNDLE ADJUSTMENT

| | Kinect1 | Kinect2 |
|---|---|---|
| $x_p^{IR}$-$\Delta\phi^{IR}$ | 0.97 | 0.82 |
| $y_p^{IR}$-$\Delta\omega^{IR}$ | 0.97 | 0.94 |
| $c^{IR}$-$\Delta Z^{IR}$ | 0.74 | 0.84 |
| $x_p^{IR}$-$\Delta X^{IR}$ | 0.59 | 0.61 |
| $x_p^{RGB}$-$\Delta\phi^{IR}$ | 0.53 | 0.24 |
| $x_p^{RGB}$-$\Delta\phi^{RGB}$ | 0.76 | 0.88 |
| $y_p^{RGB}$-$\Delta\omega^{RGB}$ | 0.85 | 0.94 |
| $c^{RGB}$-$b_y^{RGB}$ | 0.06 | 0.47 |
| $c^{RGB}$-$b_z^{RGB}$ | 0.47 | 0.75 |

**5.2 External Quality Assessment**

To quantify the external accuracy of the Kinect and the benefit of the proposed calibration, a target board located at 1.5-1.8 m away with 20 signalized targets was imaged using an in-house program based on the Microsoft Kinect SDK and with RGBDemo. Spatial distances between the targets were known from surveying using the FARO Focus3D terrestrial laser scanner with a standard deviation of 0.7 mm. By comparing the 10 independent spatial distances measured by the Kinect to those made by the Focus3D, the RMSE was 7.8 mm using RGBDemo and 3.7 mm using the calibrated Kinect results; showing a 53% improvement to the accuracy. This accuracy check assesses the quality of all the imaging sensors and not just the IR camera-projector pair alone.

To isolate the assessment of the IR camera-projector quality, the roughness of another scene consisting of a flat target board was computed. The surface roughness was calculated as the normalized smallest eigenvalue in a 30 mm radius neighbourhood. From Figure 5 one can observe that a few systematic abrupt changes in the depth appearing as vertical streaks (highlighted in orange) have been eliminated post calibration. The reason for these stripe artifacts that are parallel to the y-axis of the image coordinate system is unknown as details about the PrimeSense algorithm is still a trade secret. Nonetheless, this artifact was also identified in [8], but was not handled by their calibration scheme. In addition, similar to [14], the presented calibration approach achieves a certain degree of depth smoothing even though lowpass filters were not applied to any of the datasets.

The probability densities of the plane deviations pre- ($\sigma$ = 3.6 mm) and post-calibration ($\sigma$ = 3.0 mm) for these data are shown in Figure 6. The noise in depth still follows a Gaussian distribution before calibration, but with a larger standard deviation. In another scene, 20 planes with a 10 cm diameter were extracted. These planes vary in both orientation and position relative to the Kinect, which is important as the residuals from plane fitting are dependent on these parameters. Based on the check plane analysis shown in Figure 7 there is an overall improvement of 17% to the RMSE of the planes estimated using least-squares after calibration.

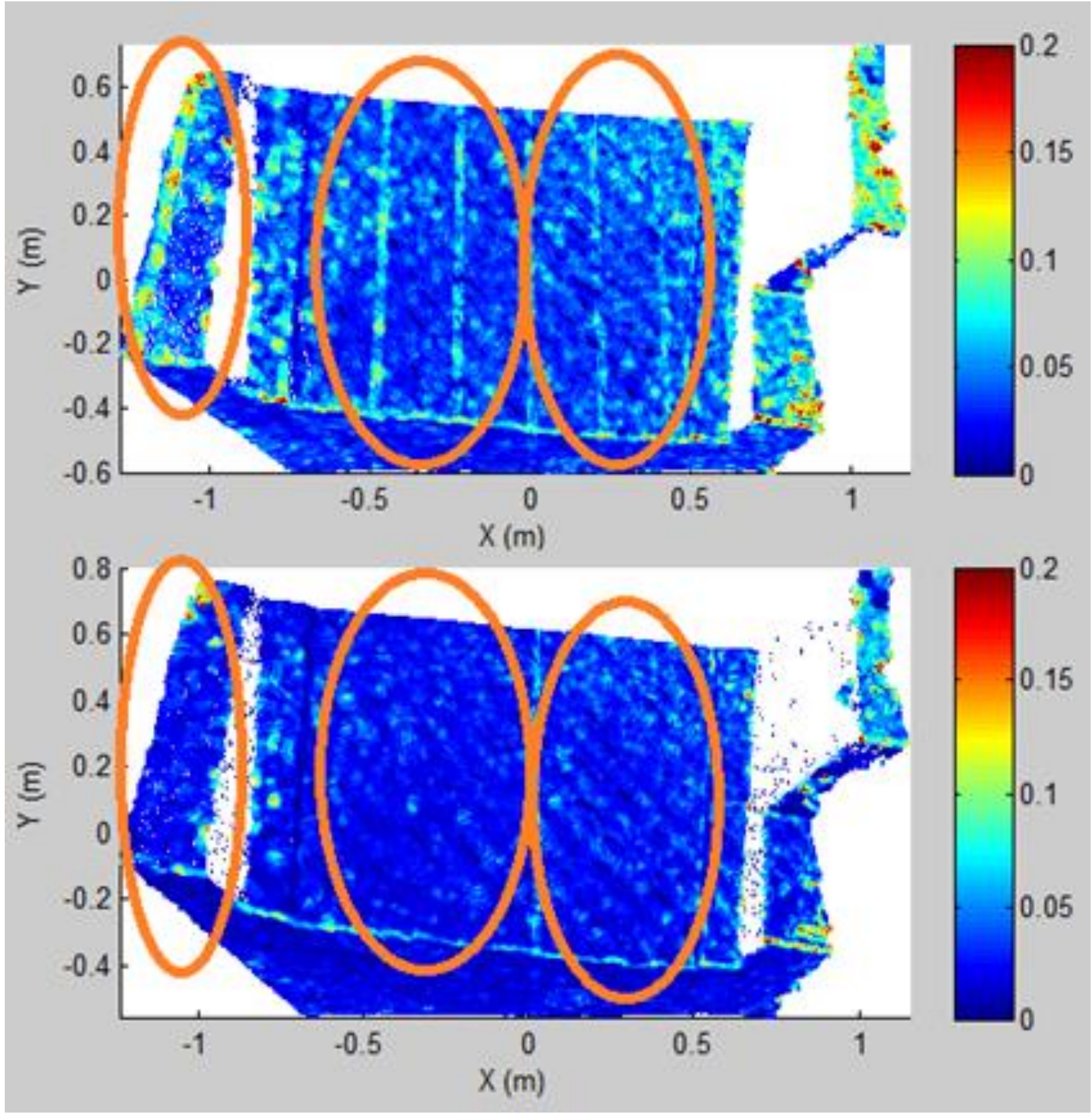

Fig. 5. (Top) Roughness of point cloud *before* calibration. (Bottom) Roughness of point cloud *after* calibration.

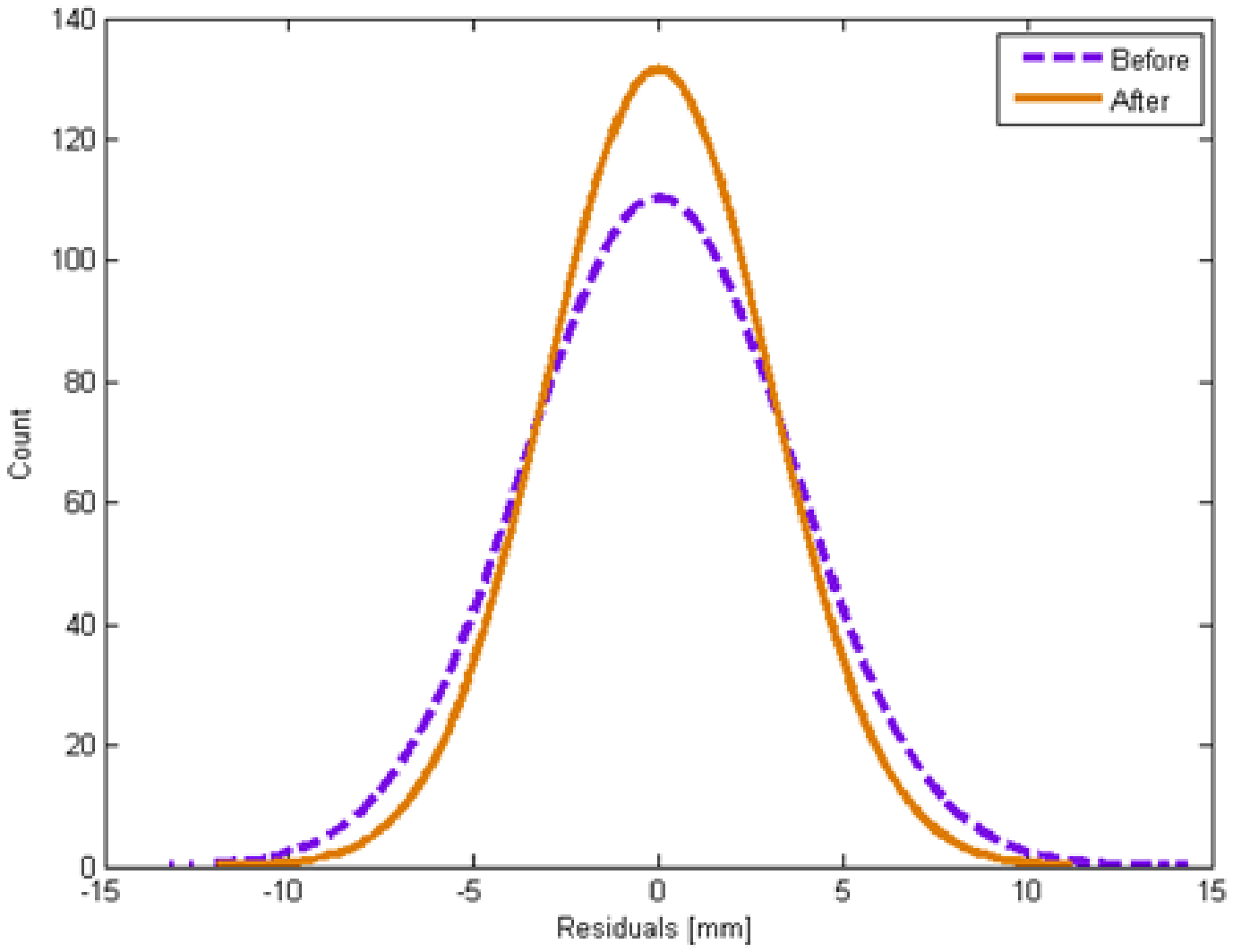

Fig. 6. Residuals of the plane-fitting before and after calibration.

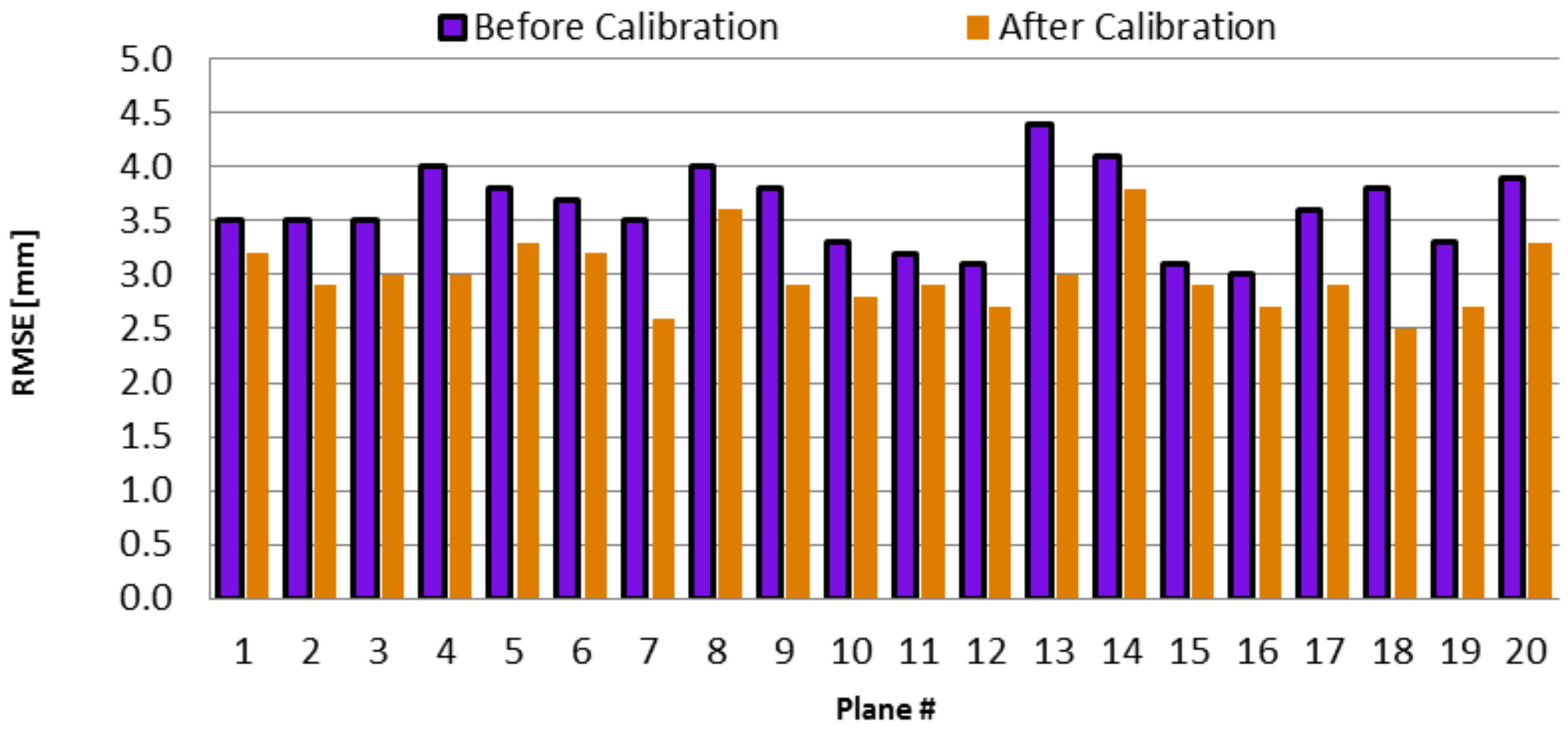

Fig. 7. RMSE of check planes before and after applying the calibration parameters.

## 6 CONCLUSION

A self-calibration method suitable for the Microsoft Kinect was presented and tested. The method solves for the relative translations and rotations between the IR camera, projector, and RGB camera. At the same time, it solves for the intrinsic parameters of both cameras, extrinisic parameters of the IR camera, object space target coordinates, and plane parameters. Geometric constraints have been included in the bundle adjustment to ensure points lie on the best fit plane and the optical sensors are all mounted rigidly on the same platform. The depth calibration is expressed as a function of three rotations, three translations, interior orientation parameters and the lens distortion of the IR camera. The effect of quantized disparity values on depth reconstruction is modelled stochastically in the bundle adjustment despite its small effect on reconstruction accuracy.

In the experimental results, significant rotational offsets up to 0.2 degs between the IR camera and projector have been recovered in the bundle adjustment. Through the inclusion of IR images and various geometric constraints, no significant correlations can be identified between the system calibration parameters and the scene dependent parameters. In the quality control stage, both the precision and accuracy of the Kinect were improved by 17% and 53%, respectively, following the presented calibration method. Furthermore, through qualitative assessment, some visually identifiable systematic artifacts in the Kinect point cloud have been removed.

Future work will study both the short-term and long-term stability of the calibration parameters for these low-cost gaming sensors. Additional features such as lines will be added to the bundle adjustment to improve the precision of the recovered calibration parameters. Recovery of the distortions in the projector could not be performed reliably using the proposed method and will be investigated.

## ACKNOWLEDGMENT

Support for this research was provided by the Natural Sciences and Engineering Research Council of Canada (NSERC), the Canada Foundation for Innovation (CFI), Alberta Innovates, and Killam Trust.

## REFERENCES

[1] P. Loriggio, Toronto doctors try Microsoft Kinect in OR, Toronto: Globe and Mail, 2011.

[2] M. Zöllner, S. Huber, H.-C. Jetter and H. Reiterer, "NAVI - a proof of concept of a mobile navigational aid for visually impaired based on the Microsoft Kinect," in 13th IFIP TC 13 International Conference, Lisbon, Portugal. September 5-9, 2011.

[3] M. Johnson and K. Hawick, "Teaching computational science and simulations using interactive depth-of-field technologies," in Proc. Int. Conf on Frontiers in Education: Computer Science and Computer Engineering, Las Vegas, USA, 2012.

[4] M. Fallon, H. Johannsson, J. Brookshire, S. Teller and J. Leonard, "Sensor fusion for flexible human-portable building-scale mapping," in IEEE/RSJ International Conference on Intelligent Robots and Systems, Algarve, Portugal. October 7-12, 2012.

[5] I. Oikonomidis, N. Kyriazis and A. Argyros, "Efficient model-based 3d tracking of hand articulations using kinect," In BMVC 2011. BMVA, 2011.

[6] J. Boehm, "Natural user interface sensors for human body measurement," International Archive of Photogrammetry and Remote Sensing Vol. XXXIX-B3, pp. 531-536, 2012.

[7] K. Khoshelham, "Accuracy analysis of kinect depth data," in Int. Arch. Photogramm. Remote Sens. Spatial Inf. Sci., XXXVIII-5/W12, pp. 133-138, 2011.

[8] F. Menna, F. Remondino, R. Battisti and E. Nocerino, "Geo-metric investigation of a gaming active device," in Proc. SPIE 8085, Videometrics, Range Imaging, and Applications XI, 80850G.

[9] C. Toth, B. Molnar, A. Zaydak and D. Grejner-Brzezinska, "Calibrating the MS kinect sensor," in ASPRS 2012 Annual Conference, Sacramento, USA, March 19-23, 2012.

[10] N. Burrus, "RGBDemo," 2012. [Online]. Available: http://labs.manctl.com/rgbdemo/. [Accessed 30 March 2012].

[11] K. Khoshelham and S. Oude Elberink, "Accuracy and resolution of kinect depth data for indoor mapping applications," Sensors 12, pp. 1437-1454, 2012.

[12] D. Lichti and C. Kim, "A comparison of three geometric self-calibration methods for range cameras," Remote Sensing 3(5), pp. 1014-1028, 2011.

[13] J. Smisek, M. Jancosek and T. Pajdla, "3D with kinect," in Consumer Depth Cameras for Computer Vision, Barcelona, Spain, November 12, 2011.

[14] D. Herrera, J. Kannala and J. Heikkilä, "Joint depth and color camera calibration with distortion correction," IEEE Transactions on Pattern Analysis and Machine Intelligence 34(10), pp. 2058-2064, 2012.

[15] M. Draelos, "The Kinect up close: modifications for short-range depth imaging," MSc Thesis, North Carolina State University, 2012.

[16] C. Herrera, J. Kannala and J. Heikkilä, "Accurate and practical calibration of a depth and color camera pair," in 14th International Conference on Computer Analysis of Images and Patterns, Seville, Spain, 2011.

[17] C. Zhang and Z. Zhang, "Calibration between depth and color sensors for commodity depth cameras," in: International Workshop on Hot Topics in 3D, in conjunction with ICME, 2011.

[18] J. Chow, K. Ang, D. Lichti and W. Teskey, "Performance analysis of a low-cost triangulation-based 3D camera: Microsoft Kinect system," Int. Arch. Photogramm. Remote Sens. Spatial Inf. Sci., XXXIX-B5, pp. 175-180, 2012.

[19] A. Staranowicz and G. Mariottini, "A comparative study of calibration methods for Kinect-style cameras," in Proceedings of the 5th International Conference on PErvasive Technologies Related to Assistive Environments. June 6-8, Heraklion, Crete, Greece., 2012.

[20] W. Zhao and N. Nandhakumar, "Effects of camera alignment errors on stereoscopic depth estimates," Pattern Recognition 29(12), pp. 2115-2126, 1996.

[21] K. Konolige and P. Mihelich, "Kinect Calibration: Technical," 27 December 2012. [Online]. Available: http://www.ros.org/wiki/kinect_calibration/technical. [Accessed 17 February 2013].

[22] C. Dal Mutto, P. Zanuttigh and G. Cortelazzo, Time-of-Flight Cameras and Microsoft Kinect, Springer, 2013.

[23] B. Freedman, A. Shpunt, M. Machline and Y. Arieli.United States of America Patent US2010/0018123 A1, 2010.

[24] OpenKinect, "Hardware Info," 24 Feburary 2011. [Online]. Available: http://openkinect.org/wiki/Hardware_info. [Accessed 19 February 2013].

[25] J. Smisek and T. Pajdla, "3D camera calibration," MSc Thesis, Czech Technical University of Prague, 2011.

[26] D. Brown, "Close-range camera calibration," Photogrammetric Engineering 37(8), pp. 855-866, 1971.

[27] C. Fraser, "Digital camera self-calibration," ISPRS Journal of Photogrammetry and Remote Sensing 52(4), pp. 149-159, 1997.

[28] G. He, K. Novak and W. Feng, "Stereo camera system calibration with relative orientation constraints," Proceedings of SPIE Vol. 1820 - Videometrics, pp. 2-8, 1992.

[29] B. King, "Bundle Adjustment of Constrained Stereopairs - Mathematical Models," Geomatics Research Australasia, No. 63, pp. 67-92, 1995.

[30] A. Tommaselli, M. Galo, J. Marcato, R. Ruy and R. Lopes, "Registration and fusion of multiple images acquired with medium format cameras," International Archives of Photo-grammetry and Remote Sensing, Vol. XXXVIII - Part 1, 6 pages, 2010.

[31] N. El-Sheimy, "A mobile multi-sensor system for GIS applications in urban centers," International Archives of Photogrammetry and Remote Sensing, Vol. XXXI, pp. 95-100, 1992.

[32] J. Lerma, S. Navarro, M. Cabrelles and A. Seguí, "Camera calibration with baseline distance constraints," The Photo-grammetric Record 25(130), pp. 140-158, 2010.

[33] A. Tommaselli, M. Moraes, J. Marcato, C. Caldeira, R. Lopes and M. Galo, "Using relative orientation constraints to produce virtual images from oblique frames," International Archives of Photogrammetry and Remote Sensing Vol. XXXIX-B1, pp. 61-66, 2012.

[34] A. Habib, A. Kersting, K. Bang and J. Rau, "A novel single-step procedure for the calibration of the mounting parameters of a multi-camera terrestrial mobile mapping system," Archives of Photogrammetry, Cartography and Remote Sensing Vol. 22, pp. 173-185, 2011.

[35] A. Kersting, "Quality Assurance of Multi-Sensor Systems," PhD Thesis. Department of Geomatics Engineering, University of Calgary. UCGE Report #20346., 2011.

[36] M. Trobina, "Error model of a coded-light range sensor," Technical Report, Communication Technology Laboratory Image Science Group, ETH-Zentrum, Zurich, 1995.

[37] G. Falcao, N. Hurtos and J. Massich, "Plane-based calibration of a projector-camera system," Tech. rep., VIBOT, 2008.

[38] M. Obaidat and K. Wong, "Geometric Calibration of CCD Camera Using Planar Object," Journal of Surveying Engineering 122(3), pp. 97-113, 1996.

[39] J. Bouguet, "Camera Calibration Toolbox for Matlab," 9 July 2010. [Online]. Available: http://www.vision.caltech.edu/bouguetj/calib_doc/index.html. [Accessed 30 March 2012].

[40] D. Lichti and J. Chow, "Inner constraints for planar features," Photogrammetric Record 28(141), pp. 74-85, 2013.

[41] W. Förstner and B. Wrobel, "Mathematical concepts in photogrammetry," in In Manual of Photogrammetry, 5th Ed., Bethesda, MD, USA, American Society of Photogrammetry and Remote Sensing, 2004, pp. 15-180.

[42] W. Caspary and J. Rüeger, Concepts of network and defor-mation analysis, Kensington, N.S.W. : School of Surveying, University of New South Wales, Monograph/School of Surveying, University of New South Wales, 183 pages, 1987.

[43] B. Widrow, I. Kollár and M. Liu, "Statistical theory of quantization," IEEE Transactions on Intrumentation and Measurement 45(2), pp. 353-361, 1996.

## 4.2 Contributions of Authors

The first author devised the presented mathematical models; wrote programs to implement the calibration models; captured and simulated the Kinect data; tested and analyzed the calibration method; and wrote the paper. The second author assisted in the paper writing and provided valuable guidance for this project.

# Chapter Five: Multi-Sensor Integration for 3D Mapping Using the Kalman Filter

So far, effective methods for reducing the effect of systematic errors in the TLS instrument, Kinect, and IMU have been presented. This chapter presents the final building blocks for the Scannect, namely the boresight and leverarm calibration, and sensor fusion using Kalman filtering (Figure 13). It summarizes the *system design*, *choice of sensors*, and *methods for localisation and mapping* used by the Scannect.

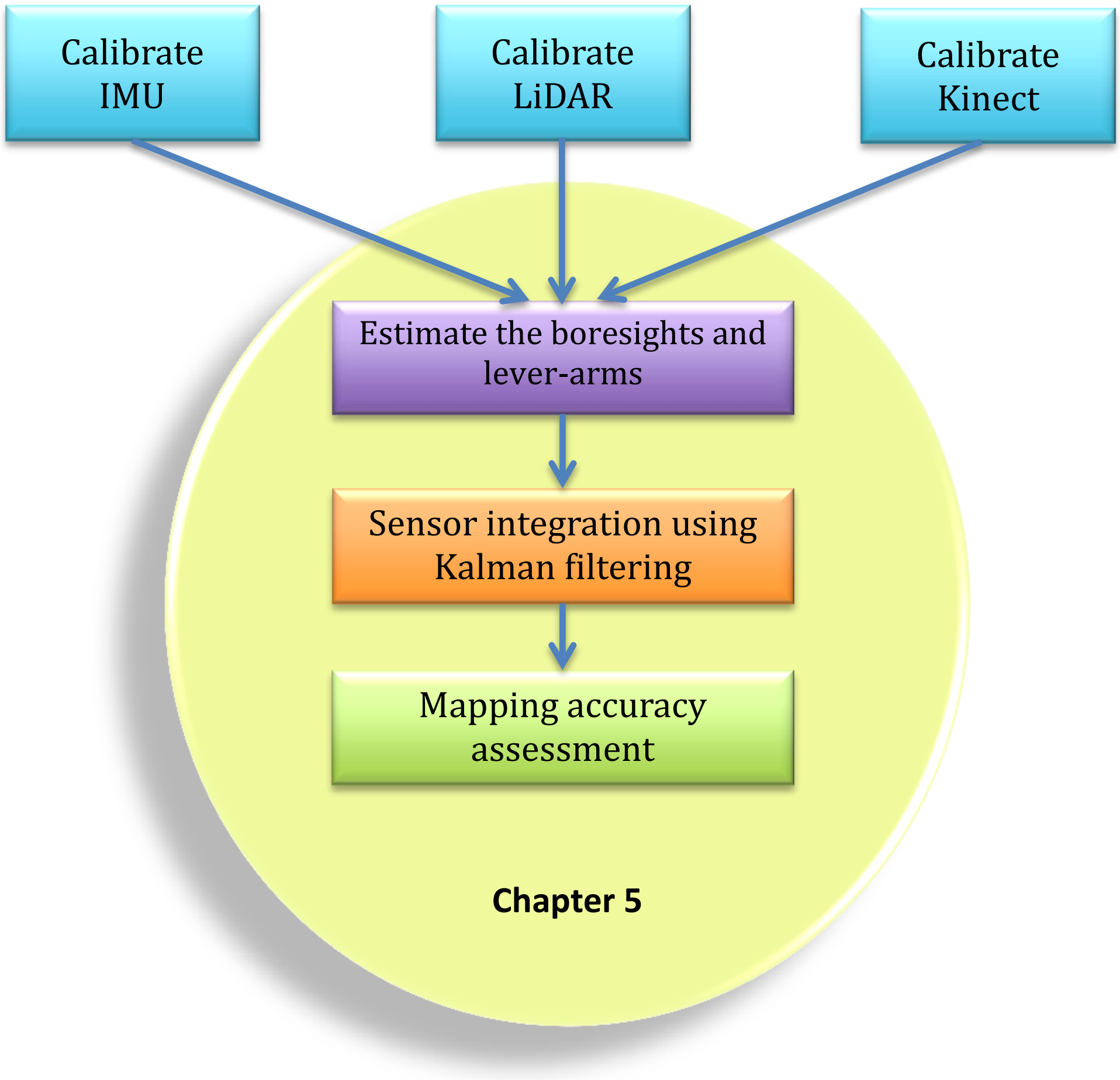

Figure 13: Overview of Chapter 5

## 5.1 Article: IMU and Multiple RGB-D Camera Fusion for Assisting Indoor Stop-and-Go 3D Terrestrial Laser Scanning

Jacky C.K. Chow[1,*], Derek D. Lichti[1], Jeroen D. Hol[2], Giovanni Bellusci[2] and Henk Luinge[2]

1 Department of Geomatics Engineering, Schulich School of Engineering, the University of Calgary, 2500 University Drive N.W. Calgary, AB T2N 1N4, Canada; E-Mails: ddlichti@ucalgary.ca

2 Xsens Technologies B.V., Pantheon 6a, 7521 PR, Enschede, The Netherlands; E-Mails: jeroen.hol@xsens.com (J.D.H.); giovanni.bellusci@xsens.com (G.B.); henk.luinge@xsens.com (H.L.)

* Author to whom correspondence should be addressed; E-Mail: jckchow@ucalgary.ca; Tel.: +1-403-889-1231; Fax: +1-403-284-1980.

Abstract: Autonomous Simultaneous Localisation and Mapping is an important topic in many engineering fields. Since stop-and-go systems are typically slow and full-kinematic systems may lack accuracy and integrity, this paper presents a novel hybrid "continuous stop-and-go" mobile mapping system called Scannect. A 3D terrestrial LiDAR system is integrated with a MEMS IMU and two Microsoft Kinect sensors to map indoor urban environments. The Kinects' depth maps were processed using a new point-to-plane ICP that minimizes the reprojection error of the infrared camera and projector pair in an implicit iterative extended Kalman filter (IEKF). A new formulation of the 5-point visual odometry method is tightly-coupled in the implicit IEKF without increasing the dimensions of the state space. The Scannect can map and navigate in areas with textureless walls and provides an effective means for mapping large areas with lots of occlusions. Mapping long corridors (total travel distance of 120 m) took approximately 30 minutes and achieved a Mean Radial Spherical Error of 17 cm before smoothing or global optimization.

## 1. Introduction

Visual Simultaneous Localisation and Mapping (V-SLAM) is an important topic for autonomous robot navigation. It is often branded as Structure from Motion (SFM) in computer vision and photogrammetric bundle adjustment in photogrammetry; two disciplines that also study the 3D reconstruction of an object/scene while inferring the 6 degrees-of-freedom movement of the optical system, albeit for different purposes. Traditionally the robotics community had been adopting computer vision techniques to address the SLAM problem; this research followed the photogrammetry methods. As part of the SLAM process, a 3D point cloud is generated, which is a valuable asset not just to robotics but also for cultural heritage documentation, infrastructure inspection, urban design, search and rescue operations, surveying, etc. The inferred pose can aid pedestrians/robots in navigating through unfamiliar indoor labyrinths and aid location-based services [1].

The term SLAM was first coined in 1995, but its concept and probability basis dates back to the 1980s [2]. The theoretical concept of SLAM is a matured subject [3] and over the years many

SLAM systems have been developed. These systems target specific environments (e.g. aerial [4], underwater [5], indoors [6], and outdoors [7]) because a superior system architecture and SLAM algorithm has yet to emerge. According to [8], assembling a mobile mapping and navigation system involves three main aspects: *System Design*, *Choice of Sensors*, and *Methods for Localisation and Mapping*. In this paper, an innovative indoor 3D mapping system with novel solutions to all three aspects, called the Scannect, is presented. The objective is to improve the efficiency of data acquisition (when using 3D terrestrial laser scanners) in large indoor environments with lots of occlusions through sensor fusion. The main contributions of this paper are:

- A novel design for a "continuous stop-and-go" indoor mapping system by fusing a low-cost 3D terrestrial laser scanner with two Microsoft Kinect sensors and a micro-electro-mechanical systems (MEMS) inertial measurement unit (IMU);
- A solution to the 5-point monocular visual odometry (VO) problem using a tightly-coupled implicit iterative extended Kalman filter (IEKF) without introducing additional states; and
- A new point-to-plane iterative closest point (ICP) algorithm suitable for triangulation-based 3D cameras solved in a tightly-coupled implicit IEKF framework.

Before describing the proposed system, recent scientific advancements in the three key topics are first highlighted. The advantages of the proposed indoor mapping system in the same three aspects relative to existing systems are then presented. The proposed mathematical model for separately updating the IEKF-SLAM using the RGB and depth information is explained. This is followed by results from real data captured with the Scannect.

*1.1. System Design*

For a lot of mobile robotic operations, efficient perception and pose estimation are critical. This motivated the design for most systems to function in full-kinematic mode, where the mapping operation is performed while the robot is moving [9]. More often than not, the robot's along-track movement extends the imaging sensor's coverage beyond its field of view [10]. However, because of imperfections in synchronization, pose estimation, control, and other sources, the resulting map accuracy and resolution can be rather poor. To quantify the mapping errors, a reference map is often captured in stop-and-go mode where the imaging sensor is only activated when the platform has halted its movements [11]. It is agreed for terrestrial applications that perception data captured while the platform is stationary can yield higher quality data because motion blur and any other robot motion induced errors are avoided [12]. Based on a recent survey done by [12] existing mobile mapping systems either solely operate in stop-and-go or in full-kinematic mode. The former approach is hindered by the slow data acquisition and the latter may lack accuracy. This survey further indicated a system that can operate in both modes has not yet been developed; specifically one that can capture higher density, stability and accuracy data when it is parked and lower quality but continuous data while it is moving.

*1.2. Choice of Sensors*

Indoor SLAM can be performed using many different sensors, for example it can be applied to Wi-Fi positioning [13]. However, due to the effectiveness and wide success of optical systems for mapping, vision-based SLAM is by far the most common scheme [14]. Popular sensors used for visual SLAM include monocular/stereo cameras [15] [16], 2D/3D Light Detection and Ranging (LiDAR) systems [11] [17], and 3D cameras [18], all of which have their own advantages and disadvantages.

Monocular cameras are passive sensors that can capture a 2D array of light intensity information instantaneously. Although they are sensitive to the ambient illumination and the information they encapsulate for SLAM is dependent on the scene's texture, being a bearing-only sensor it is largely independent of the object's reflectance properties and can observe details as far as their pixel resolution allows. Stereo cameras can further perceive depth from a single exposure station based on triangulation, a process similar to how human vision operates [19].

LiDAR systems on the other hand are active sensors that acquire 2D scan lines by measuring distance (based on the time-of-flight principle) and bearing information point-by-point at high speed. The third dimension is obtained either by the platform's trajectory (for 2D scanners) or by a rotating head/mirror (for 3D scanners). They can observe dense geometric information over homogeneous surfaces under any illumination conditions. However, their observable range is hindered by the amount of energy they can emit and receive after the laser signal has been reflected and absorbed along its path. In addition, since 2D scan lines are acquired sequentially, the time it takes to sweep the laser over the object of interest is typically long, making them slower than cameras [10].

These complementary characteristics of LiDAR and imagery have been detailed and harvested for airborne and terrestrial mapping by many [20] [21]. This fusion of LiDAR and imagery for more informative mapping is typically done mathematically in the software. In recent years, 3D cameras have emerged integrating the benefits of time-of-flight (ToF) measurements with the fast acquisition and gridded structure of digital images at the hardware level. Despite the fact that many commercial 3D cameras are based upon the ToF principle, currently the most widely used 3D camera on the market is the structured-light RGB-D camera known as the Microsoft Kinect. The per pixel metric depth data enhances the information content of the RGB image significantly and can make tasks like obstacle avoidance, object tracking and recognition more robust and accurate [22]. While it appears that RGB-D cameras may have eliminated the need for LiDAR in robotic applications, LiDAR has retained its place in the robotics community [23], and among others, due to the limited range, resolution, accuracy, and field of view of modern RGB-D cameras. Instead, RGB-D cameras might be more suitable for filling in gaps in the LiDAR point cloud at close-range as it can be more efficient for covering occluded areas [24]. In addition, even though the cost of a Kinect is lower than most 3D ToF cameras, [25], [26], and [27] have shown that the Kinect can produce more accurate results and are suitable for mapping; at close-range (less than 3.5 m) its accuracy can be similar to LiDAR and medium-resolution stereo cameras [25][26].

Existing indoor mapping systems based on the Kinect have so far been limited to a single system. In addition, they are often treated as substitutes for more expensive scanners, rather than being used as an aiding sensor to laser scanners. Furthermore, the integration of geometrically accurate LiDAR, semantic RGB, and efficient depth map for more informed mapping and pose estimation has not been thoroughly investigated, despite their known complementary characteristics.

### *1.3. Methods for Localisation and Mapping*

Indoor localisation-only solutions exist, for example pedestrian navigation devices [28]. Often IMUs are used in conjunction with a spatial resection based on bearings, distances, or a combination of both from primitives [29]. The primitives can be points, higher-dimensional 2D/3D geometric features (e.g. lines, planes, cylinders, and spheres), or more complex surfaces (e.g. a sofa and statue of a lion). The main assumption is that the positions of the primitives are known, either from a computer-aided design (CAD) model, surveying, or other means.

On the contrary, mapping-only solutions are also common, and are typically based on bearings, distances, or both measurements from one or more locations. In this scenario, the location of the sensor needs to be known, and often in combination with orientation information [10].

Localisation and mapping are highly correlated and are often described as the “chicken and egg” problem [30]. Naively performing mapping operations after localisation without any accounts for their correlation can lead to poorer accuracy, or even divergence in the solution [3]. Therefore, state-of-the-art methods usually solve the two problems simultaneously using optimizers such as least-squares, Kalman filters (KF), information filters, and particle filters [31]. The measurement models used by these optimizers depend on the data source and the features being matched.

For 2D camera images, popular matching methods include area-based matching (e.g. normalized cross-correlation), point-based matching (e.g. blob detectors like scale-invariant feature transform (SIFT) or corner detectors like Harris), and line-based matching. Keypoint matching is usually more accurate than area-based methods [32], and these points can be tracked over consecutive frames (e.g. by the Kanade–Lucas–Tomasi (KLT) feature tracker) for efficiency and reliability at the risk of experiencing tracking drift. Usually a pin-hole camera model is used to relate pixel observations to their homologous point in the world coordinate system using the collinearity condition [19]. For cameras the matching in general can be formulated as 3D-to-3D, 3D-to-2D, or 2D-to-2D matching, with the accuracy and complexity generally increasing in the order presented [32].

To match point clouds from 3D cameras or laser scanners, point-based, feature-based (e.g. lines, planes, and tori), and freeform matching (e.g. ICP) are possible. Since dense point clouds are usually available and the nature of objects/shapes in the scene can be unpredictable, freeform matching is one of the most popular choices in robotics. It is very flexible but is less robust than matching signalized markers, and can be computationally intensive. The two original ICP implementations that are still widely used today are the point-to-point ICP by Besl and McKay [33] (conventionally solved using Horn's method [34]) and the point-to-plane ICP by Chen and Medioni [35] (formulated as a least-squares minimization problem). To improve the algorithm, over the years many variations of ICP have been proposed and tested. For more details, the reader can refer to taxonomy of ICP in [36]. As no ICP algorithm is superior in all situations, the designer is responsible for selecting and/or modifying the ICP algorithm to fit their needs.

Many different VO/SLAM algorithms based on the above concepts exist and a review of all of them is beyond the scope of this paper. Instead, a few popular state-of-the-art methods relevant to this project are explained with their pros and cons highlighted.

Typical VO operates like Parallel Tracking and Mapping (PTAM) [37]. A pair of 2D images is used to intersect points in the scene with the scale being arbitrary or introduced based on a priori information. Then consecutive images that are captured use these intersected 3D points to perform a resection to solve for their egomotion. Once this transformation is known, they are once again used in the intersection process to create more points in the scene. The end result is usually a sparse map of the environment that carries little value for most mapping operations compared to dense point clouds from LiDAR or 3D cameras. Furthermore, three drawbacks arise with PTAM: 1) the "catch 22 problem" where more reconstructed 3D points give a better camera tracking solution, but cause the state vector to grow rapidly; 2) the same 3D point needs to be observable in at least three images before it can be used for tracking the camera's motion; 3) an initial metric scale is difficult to obtain. The last two statements are less of a problem because through fusion with an IMU the scale can be estimated as quickly as 15 seconds [38]. Or if depth data are available it can be solved instantaneously from a single exposure and camera tracking can begin immediately following the first camera exposure [39].

Henry et al. [40] had a single PrimeSense RGB-D camera carried by a human in a forward-facing configuration for indoor mapping. They explained that the depth data ignores valuable cues in the RGB images and RGB images alone do not perform well in dark and sparsely textured areas. Therefore they presented 1-step and 2-step matching methods that exploit both pieces of information. In the 1-step case, a joint optimization was performed that minimized the point-to-point distances for the detected 3D keypoints and point-to-plane distances for the depth maps using ICP. To improve the speed of the algorithm, they suggested splitting this into two steps: running the keypoint matching first and only performing ICP after if the match was poor (e.g. insufficient keypoints were matched). This resulted in only minor compromise in terms of accuracy. In both cases, they performed RANSAC on the 3D keypoints for the initial alignment step. Compared to their previous work [41], they have replaced SIFT with the Features from Accelerated Segment Test (FAST) detector, minimized the reprojection errors of the matched 3D keypoints instead of Euclidean distances, and did a global optimization using sparse bundle adjustment (SBA) instead of the Tree-based netwORk Optimizer (TORO) [30]. In their indoor mapping experiment, they travelled over 71 m in an office space and their ICP-only solution showed 15 cm error, while their proposed RGB-D ICP method showed 10-11 cm error. Their proposed RGB-D ICP scheme eliminated the need to weight the colour space relative to the Euclidean space [42], but created the new problem of having to weight 3D keypoints against the point-to-plane ICP. In this paper they give the two different point matches the same weight in their joint-optimization. Moreover, they only mention the possibility of weighting the points in ICP based on distances, normal angle, etc. to improve the matching but the same weight was assigned to all points. The authors acknowledge and stress the importance of minimizing reprojection errors, but their point-to-plane ICP did not adopt this scheme, only the 3D keypoints used this cost function. They also mention that because they tightly-coupled depth and colour information, whenever the depth sensor is out of range they do not have 3D keypoints for RANSAC matching even though they were detectable in 2D. This can be rather common with their single forward facing sensor configuration. For example, when looking down a hallway, the majority of the center pixels will likely be out of range.

The Kinect is capable of delivering 9.2 million points per second, a rate that far exceeds any terrestrial laser scanners. A popular method for handling this vast amount of data in real-time is the KinectFusion solution [43]. They parallelized the point-to-plane ICP algorithm to be executed on the graphics processing unit (GPU) for speed and fused the dense point clouds using the Truncated Signed Distance Function (TSDF) [44]. This novel implementation was considered for this project but at the time its measurement volume was restricted by the memory of the GPU. Even with a lower resolution for the TSDF, mapping over hundreds of metres in distances was not possible. This method can map in the dark because only the depth images were considered and all the semantic RGB data were ignored. Although it was designed for handheld sequences, slow and steady egomotion was assumed, as sudden jerks and insufficient overlap can cause the ICP to fail. In addition, a lack of geometry in the scene (a well-known problem with ICP) makes it difficult for the KinectFusion method to converge to the global minimum.

More recently [45] has lifted this limitation of measurement volume by downloading the old data onto the central processing unit (CPU) and using a moving TSDF to only map the most recent and immediate point clouds. In their later paper they even took advantage of the RGB information [46], and added loop-closure [47]. Keller et al. [48] improved upon KinectFusion by replacing the volumetric representation by a point-based representation of the scene. This eliminated many computational overheads and conversions between different scene representations, making it more scalable and faster. Each point in the depth map is projectively associated with the scene and reliable points are merged using a weighted average. Besides being able to map larger areas, the biggest improvement is the ability to handle a certain degree of dynamics in the scene. Even with these significant improvements, being a single Kinect system, its field of view (FOV) is limited to less than $60^{o}$ and their ICP algorithms will fail when the scene lacks features (e.g. only a single homogenous wall is in view). Also, the timing error between the colour and depth information was usually neglected [49].

Unlike the above papers, another stream of research used Kalman filtering to solve the SLAM problem. Li et al. [50] demonstrated that the performance and convergence of SLAM is affected by landmark initialization uncertainty and linearization error. The former can be addressed by using the depth information provided by the Kinect to better initialize the 3D landmarks. The latter can be improved by replacing the extended Kalman filter (EKF) with the IEKF, which re-linearizes the measurement model by iterating an approximate maximum a posteriori estimate around the updated states rather than relying on the dynamics model. They showed using real data from their keypoint based SLAM that in some situations where EKF will fail, the IEKF would not. They also proposed a way to add and remove landmarks from the state vector to avoid memory overload. The IEKF can help reduce the effect of non-linearities in the measurement model; however just like the EKF they are still based on a measurement model where a single observation is expressed as a function of the unknown states.

Aghili et al. [51] realized the importance of initial alignment for the ICP and attempts to keep track of the pose when good measurement updates were unavailable. They solved this by fusing the ICP pose in a KF under a closed-loop configuration. In this case, the measurements were loosely-coupled and the EKF only provided an initial guess for the ICP method. With a similar motivation, Hervier et al. [52] fused the Kinect with a three axes gyroscope for improved mapping and localisation accuracy. They used the ICP algorithm in the Visualization Toolkit (VTK) to first solve for the pose, and then fuse it in the KF in a loosely-coupled manner. In their current system, they mentioned it is difficult to evaluate movements parallel to a flat wall.

All these SLAM solutions will fail in cases when a single homogenous plane is imaged. Often when the ICP method is used with a KF, they are loosely-coupled rather than tightly-coupled. Most of the work either ignores the RGB information or if included they require the depth and RGB information to be available simultaneously and assumed no synchronization errors.

In Grisetti et al. [30] they distinguished SLAM into the filtering approach and smoothing approach. Filtering approaches are incremental by nature and often use KF (also referred to as online SLAM), whereas smoothing approaches estimate the complete trajectory from all the measurements (a.k.a the full SLAM problem). This paper focuses on the forward filtering part of SLAM and proposes new ways to solve monocular VO and ICP in a tightly-coupled KF.

## 2. The Scannect Mobile Mapping System

The Scannect is an indoor mapping system that is unique in its *System Design*, *Choice of Sensors*, and *Methods for Localisation and Mapping*. Each of these aspects of the Scannect is explained in the following sections.

*2.1. Proposed System Design*

The proposed system attempts to harvest the accuracy and stability of stop-and-go systems while maintaining a degree of speed from full-kinematic systems. High quality 3D scans are captured when the robot has "stopped", and lower quality data is captured while it is "going" for mapping and keeping track of the relative change in pose. For this reason, the system design can be seen as a hybrid between static and kinematic mapping, termed continuous stop-and-go mode in this paper. According to a recent review by [12], systems operating in continuous stop-and-go mode have not yet been released. If higher quality data are desired the robot can make more frequent stops, hence approaching the stop-and-go solution's quality or if speed and time is of the essence, it can make zero stops (approaching the full-kinematic solution). This trade-off between speed and accuracy does not affect other components of the Scannect, and can therefore be easily adapted to the project at hand.

2.1.1. Designed Robot Behaviour

The robot's behaviour during the mission is described in the following stages with the desirable trait of the system highlighted:

- Stage 1: The robot enters an unfamiliar environment without prior knowledge about the map or its current position.
    - The absolute position may never be known, but an absolute orientation based on the Earth's gravity and magnetic fields is determinable.
    - The system should "look" in every direction to its maximum range possible before moving around to establish a map at this arbitrary origin.

- Stage 2: Due to occlusions and limited perceptible range, the robot needs to move and explore the area concurrently.
    - When exploring new areas the map should be expanded while maintaining the localisation solution based on the initial map.
- Stage 3: Occasionally when more detail is desired or the pose estimation is uncertain, the system can stop and look around again.
    - As the system is for autonomous robots with no assumptions about existing localisation infrastructure (e.g. LED position systems), every movement based on the dead-reckoning principle will increase the position uncertainty. Typically it is more accurate to create a map using long-range static remote sensing techniques because they are typically less than the dead-reckoning errors.
    - Looking forward and backward over long ranges from the same position can possibly introduce loop-closure.
    - One of the Kinects and the IMU are rigidly mounted together and force centred on the mobile platform. Through a robotic arm, quadcopter or by other means the Kinect and IMU can be dismounted from the platform during “stop” mode for mapping, making it more flexible/portable for occlusion filling. Afterwards it can return and dock at the same position and orientation and normal operations is resumed.

*2.2. Proposed Choice of Sensors*

The starting point of the Scannect is the FARO Focus$^{3D}$ S terrestrial laser scanning (TLS) instrument. This is motivated by the accuracy, stability, speed, range, and field of view of this modern static 3D TLS instrument, (e.g. millimeter-level range accuracy and 360$^{o}$/310$^{o}$ horizontal/vertical FOV, respectively). This sensor is very suitable for indoor mapping applications, especially with the recent upgrades (i.e. digital compass and tilt sensors), and reductions in cost, size, and weight. Its success for indoor mapping can be seen in the papers published using this system for as-built surveys and cultural heritage documentation [53] [54].

The Microsoft Kinect is a trinocular vision system capable of capturing dense RGB-D information at up to 30 Hz. The RGB information comes from an onboard VGA resolution camera while the depth information is determined using the coded-light principle: a pattern is emitted by the projector and simultaneously imaged by the infrared (IR) camera to perform photogrammetric intersection. More details about the Kinect can be found in many articles, e.g. [55] [56].

In large open spaces (e.g. a gymnasium, shopping malls, and airports) TLS can be more efficient and effective at mapping than 3D cameras like the Kinect. However, TLS falls short when many occlusions exist in the scene so many static scans are needed for complete coverage, which is very time-consuming. When setup on a mobile platform (e.g. trolley) the setup-move-setup time is reduced [12]. A selective FOV in theory can shorten the scan time when compared to a full 360$^{o}$ scan, however because of the rapid scan speed of modern scanners, the process of capturing a preview scan followed by the manual FOV selection and doing a dense scan can actually result in longer overall acquisition time. In cases where small occlusions needed to be filled, the Kinect is perhaps more efficient. Unlike 2D profile scanners which essentially have no overlap in kinematic mode, the Kinect provides a 3D point cloud within its FOV instantaneously at 30 Hz. Consequently, a high-degree of overlap exists between consecutive frames, which facilitate continuous point cloud stitching for higher quality mapping while maintaining the robot's pose. Because the Kinect, as well as other 3D cameras currently on the market, all have a relatively small FOV (the largest being $90^{o}$), a simple and cost-effective solution is to add extra Kinects [57].

The Scannect was equipped with two sideward facing Kinects (Figure 1); this architecture was chosen to maximize the overall FOV for indoor SLAM and to prevent possible interference between the two Kinects. If desired, additional Kinects are possible and the interference issue can be mitigated by introducing small vibrations to each Kinect [58]. Although a forward facing and skyward facing Kinect design has also been considered, it was not implemented because 1) a forward facing Kinect usually does not have sufficient range to reach the end of a long corridor to provide any useful information; 2) the considered situations in this paper assume a flat horizontal floor for the building, which is often the case, so a non-holonomic constraint was used for constraining the height of the robot instead since a sideward facing Kinect is more valuable for mapping.

Nonetheless, all the sensors discussed thus far require a direct line-of-sight to conjugate features in the overlapping regions. If the overlap between two visual datasets is small or lacking in features/texture, optical-based matching techniques may be inaccurate or even fail in some cases.

A suitable aiding device is an inertial measurement unit, which operates well regardless of its environment. A MEMS-based IMU from Xsens, the MTi, was adopted for the task at hand. When the robot enters a dark corridor with walls too far for the Kinect, the integrated rotation rate will keep track of the orientation, and the rotated acceleration after double integration will keep track of the position of the platform.

Overall, the Focus$^{3D}$ S is envisioned for mapping large open areas in static mode. When the robot is moving the Kinects are used for localisation and filling in the shadowed areas until the system stops to perform Zero velocity UPdaTes (ZUPT) and capture another laser scan. The IMU continuously assists the Kinect matching and bridges the gaps when vision-based localisation fails due to lack of details in the scene. To the authors' best knowledge this is the first paper using more than one Kinect for an inside-out type system. Popular multi-Kinect systems are designed for body scanning or motion capture with an outside-in design [57]. Furthermore, the Kinect is usually used as the sole mapping sensor, instead of playing a more assistive role to a more accurate laser scanner, as in this paper.

**Figure 1.** Sensor configuration of the Scannect.

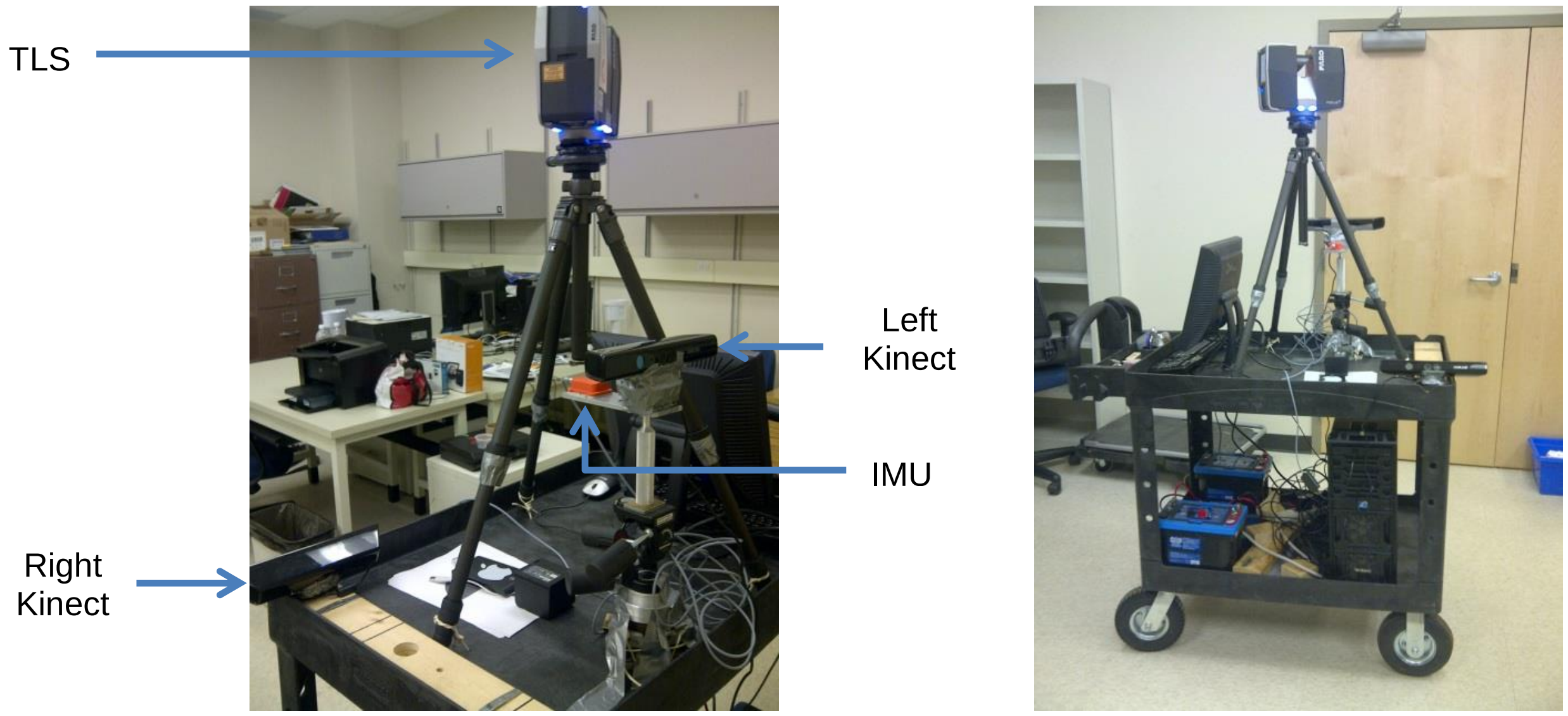

2.2.1. System Calibration

Systematic errors are prevalent in all man-made instruments, for instance due to manufacturing flaws. Before integrating the four proposed sensors for SLAM, they were first individually calibrated for their intrinsic parameters. Afterwards, the relative rotation (boresight) and relative translation (leverarm) between all the sensors were determined through an extrinsic calibration process. Only the intrinsic calibrations of the optical sensors are considered in this paper. The calibration procedure for MEMS-based IMUs has reached a level of maturity that is widely accepted [59]. The commonly adopted methods solve for biases, scale factor errors, and axes non-othogonalities for the triad of accelerometers and gyroscopes, and misalignments between them. For the accelerometer calibration, the sensor can be placed in various orientations and capturing a sequence of static data. These measurements are related to the local gravity vector, which is used as the reference signal. For the gyroscopes, a rotation rate table can be used to establish the reference signal. Since such a calibration requires specialized equipment that is not easily accessible, the manufacturer's calibration was relied upon. Furthermore, MEMS IMUs are known to have a poor bias stability, so the residual systematic errors need to be modelled stochastically in every operation.

*2.2.1.1. Microsoft Kinect RGB-D Camera Calibration*

The Kinect is originally intended for gaming applications, where metric measurements are not critical. In order to integrate the Kinect with a laser scanner for mapping, the sensor needs to be individually calibrated to ensure the residual systematic errors from the manufacturer's calibration are minimized. Distortions in the Kinect's depth map have previously been reported and various calibration approaches have been proposed [55] [60]. As the Kinect is a relatively new sensor, a commonly accepted calibration procedure has not yet been established, which is in contrast to MEMS IMU calibration. For convenience, a user-self calibration approach that requires only a planar checkerboard pattern for performing a total system calibration was selected.

The Kinect consists of three optical sensors, all of which can be modelled using the pin-hole camera model. The well-established bundle adjustment with self-calibration method was modified in [61] to solve for all the intrinsic and extrinsic parameters of all the optical sensors in the Kinect simultaneously. This method was adopted in this paper and was responsible for independently calibrating the two Kinects installed onboard the Scannect. The results from the calibration are published in [61].

*2.2.1.2. FARO Focus$^{3D}$ S 120 3D Terrestrial Laser Scanner Calibration*

The Focus$^{3D}$ S is the lowest-cost sensor in its category of laser scanners. Compared to more expensive scanners within the same class, it has been found to exhibit more significant systematic distortions, in particular angular errors. The multi-station self-calibration approach is a popular and effective means for reducing the systematic errors internal to TLS instruments without the need of specialized tools [62] [63]. Analogous to camera calibration, by observing the same targets from different stations a least-squares adjustment can be performed for solving the biases in its distance and bearing measurements and various axes misalignments, eccentricities, and wobbling. Recently [64] showed that signalized targets can be replaced by planar features to achieve similar calibration results with reduced manual labour. The Focus$^{3D}$ S on the Scannect was calibrated using this plane-based self-calibration method, and the calibration results can be found in [65].

*2.2.1.3. Boresight and Leverarm Calibration Between the Kinect and Laser Scanner*

Both the Kinect and laser scanner are visual sensors, therefore target fields such as a planar checkerboard pattern is visible to both. These common targets can be independently extracted in their corresponding point clouds and then related through a 3D rigid-body transformation [66]. This approached was adopted to simultaneously solve for their boresight and leverarm parameters. Since the Kinects point away from each other, they were related through the intermediate laser scanner.

*2.2.1.4. Boresight and Leverarm Calibration Between the Kinect and IMU*

The problem of solving for the 3D rigid body transformation parameters between an IMU and camera is often encountered in multi-sensor mapping systems [10]. Some common methods include simple surveying techniques via a total station, solving it as a state estimation problem in an EKF, or formulating it as an offline global optimization problem. The first method is most suitable for higher-end IMUs and metric cameras with clear markers on the exterior casing, for instance in airborne mapping systems. The locations of the Kinect's perspective centres are unknown and CAD models are not published by Microsoft. Therefore, the other two algebraic calibration methods are perhaps more suitable.

Lobo and Dias [67] proposed a popular two-step calibration procedure that first estimates the boresight parameters and then uses a turntable that spins about the IMU's origin to solve for the leverarm parameters. Not only does this require specialized tools and careful placement of the system on the turntable, correlations between the rotational and translational parameters are ignored in this method. Hol et al. [68] formulated the calibration as a gray-box model where the navigation states were solved in the EKF using their current best estimate of the unknown parameters (boresight, leverarm, gyro bias, accelerometer bias, and gravity vector). A post-filtering adjustment was then used to improve their estimate of the unknown parameters simultaneously by minimizing the differences between the predicted and measured target positions in image space. This online filtering and offline adjustment is then repeated until convergence. This type of calibration is preferred in this project because it is more user-friendly and can account for the correlations between the boresight and leverarm parameters. Based on the approach in [68], the measurement model shown in Equation 1 was used for calibration in this project. Two subtle differences exist in the implementation presented herein: first, the slowly time-varying biases are solved in the IEKF like [69]; second, the gravity vector in the navigation frame has been replaced by the rotation angles between the navigation frame and world frame of the checkerboard. The rotation matrix notation was used here and in other equations for convenience in representation; in the actual adjustment, quaternions were used to give the system full rotational freedom without encountering gimbal lock.

$$p_i^{RGB}(t) = \mu_i^{RGB}(t) R_{IR}^{RGB} \left\{ R_s^{IR} \left\{ R_n^s(t) \left\{ R_W^n O_i^W - T^n(t) \right\} - \Delta b^s \right\} - \Delta b^{IR} \right\} \tag{1}$$

where $p_i^{RGB}(t)$ is the observation vector for point *i* expressed in the RGB image space at time *t*; $\mu_i^{RGB}(t)$ is the scale factor for point *i* at time *t* (which can be eliminated by dividing the image measurements model by the principal distance model [19]); $R_{IR}^{RGB}$ is the relative rotation between the IR and RGB cameras; $R_s^{IR}$ is the relative rotation between the IMU sensor and IR camera; $R_n^s(t)$ is the orientation of the IMU sensor relative to the navigation frame at time *t*; $R_W^n$ is the relative rotation between the world frame defined by the checkerboard and the navigation frame; $O_i^W$ is the 3D coordinates of point *i* in the world frame; $T^n(t)$ is the position of the IR camera in the navigation frame at time *t*; $\Delta b^s$ is the leverarm between the IMU sensor and IR camera; $\Delta b^{IR}$ is the relative translation between the IR and RGB cameras.

### *2.3. Proposed Methods for Localisation and Mapping*

Figure 2 shows the interaction between the sensors onboard the Scannect. The different colours separate the various components of the Scannect (e.g. cyan highlights the egomotion estimation using RGB images and purple highlights the 3D image processing). The IMU observations after correction (Equation 2) were integrated to provide changes in orientation and position (Equation 3), which were useful for initializing the 2D/3D matching and keep track of the states between measurement updates. The accelerometer and gyroscope biases were solved in the KF as a first order Gauss-Markov process (Equation 4), whose corresponding spectral densities are given in Equation 5. The Gauss-Markov parameters were determined from autocorrelation analysis [70]. The TLS point clouds were matched using the conventional point-to-plane ICP [35] with the first point cloud oriented relative to magnetic north in the local level frame. The proposed RGB-D localisation and mapping method is based on a novel ICP implementation suitable for triangulation-based 3D cameras such as the Kinect. The new cloud was always registered with the global scene to partially address the loop-closure problem. The localisation was separated into depth-only and RGB-only for two reasons: 1) the depth and RGB streams are not perfectly synchronized; and 2) the depth measurements have a limited range -- by processing the RGB data separately, localisation of the robot is still possible when depth data are unavailable. Both depth-only and RGB-only processing used a tightly-coupled implicit iterative extended Kalman filter to estimate the states.

**Figure 2.** The overall workflow of performing SLAM using the Scannect.

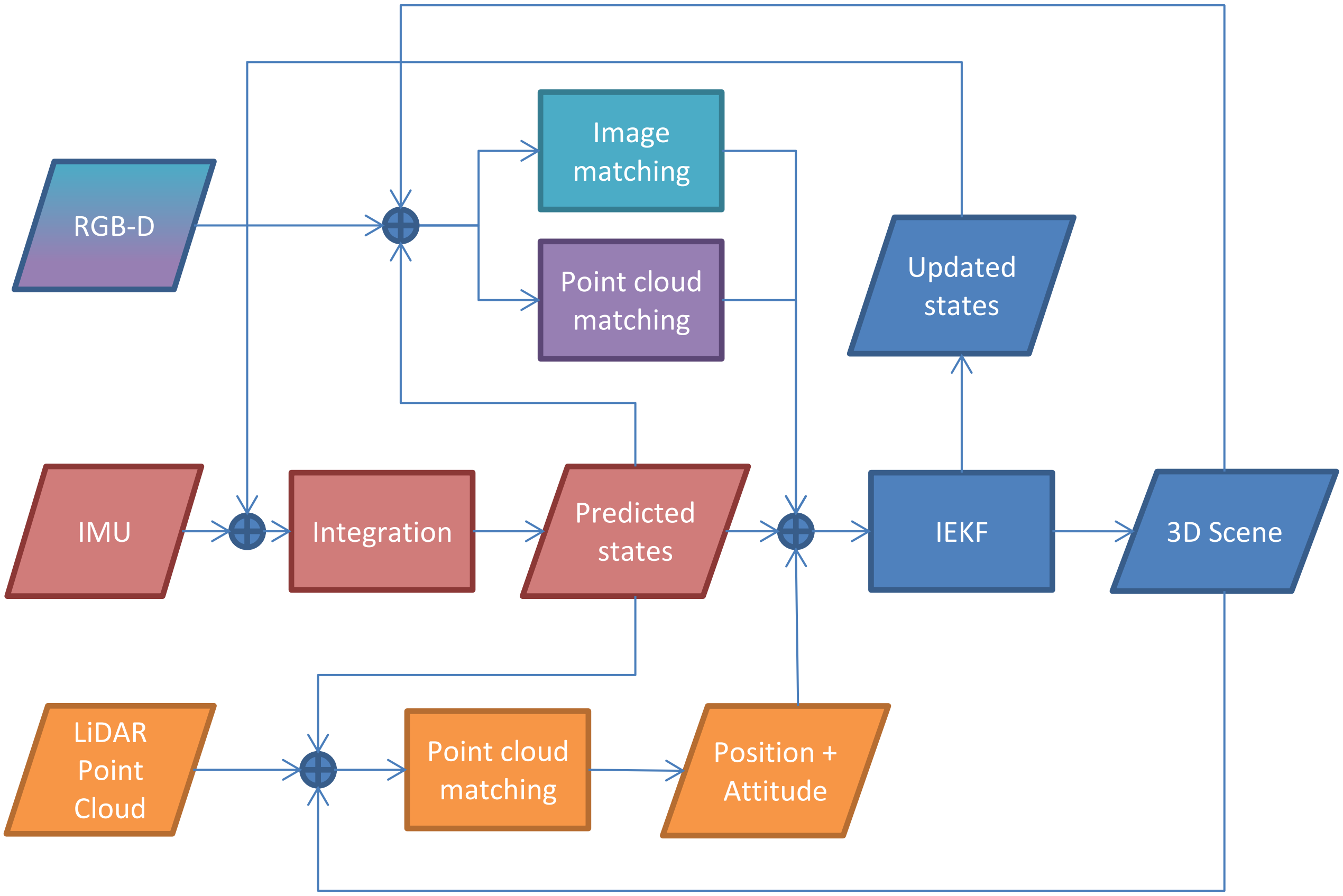

$$\begin{aligned} \ddot{x}^n(t) &= R_s^n(t)\left[f^s(t) - \delta^a(t)\right] - 2\omega_n^{e\to i} \times \dot{x}^n(t) + g^n \\ \omega_s^{s\to n}(t) &= \omega_s^{s\to i}(t) - R_n^s(t)\,\omega_n^{e\to i} - \delta^\omega(t) \end{aligned} \tag{2}$$

$$\begin{aligned} x^n(t+1) &= x^n(t) + \dot{x}^n(t)\Delta t + \ddot{x}^n(t)\,{}^{\Delta t^2}\!/_2 \\ \dot{x}^n(t+1) &= \dot{x}^n(t) + \ddot{x}^n(t)\Delta t \\ q^{s\to n}(t+1) &= q^{s\to n}(t) \otimes \exp\left[\omega_s^{s\to n}(t)\,{}^{\Delta t}\!/_2\right] \end{aligned} \tag{3}$$

$$\begin{aligned} \delta^a(t+1) &= e^{-\Delta t/\tau^a}\,\delta^a(t) + \eta^a(t) \\ \delta^\omega(t+1) &= e^{-\Delta t/\tau^\omega}\,\delta^\omega(t) + \eta^\omega(t) \end{aligned} \tag{4}$$

$$\psi^{a}(t)=\sigma_{a}^{2}\left[1-e^{-2\Delta t/\tau^{a}}\right]$$
$$\psi^{\omega}(t)=\sigma_{\omega}^{2}\left[1-e^{-2\Delta t/\tau^{\omega}}\right] \tag{5}$$

where $x^n, \dot{x}^n, and\ \ddot{x}^n$ are the position, velocity, and acceleration of the IMU sensor in navigation frame, respectively; $f^s$(t) and $\omega_s^{s\to i}(t)$ are the accelerometer and gyroscope measurements at time $t$, respectively; $\varepsilon^a(t)$ and $\varepsilon^\omega(t)$ are their corresponding measurement noises; $\omega_n^{e\to i}$ is the rotation rate of the earth as seen in the navigation frame; $g^n$ is the local gravity in the navigation frame; $q^{s\to n}$ is the rotation from IMU sensor frame to navigation frame expressed using quaternions; $\delta^a(t)$ and $\delta^\omega(t)$ are the slowly time-varying accelerometer and gyroscope biases, respectively, which are modelled as a first order Gauss-Markov process; $\eta^a(t)$ and $\eta^\omega(t)$ are the corresponding Gauss-Markov process driving noises, whose spectral densities are $\psi^a(t)$ and $\psi^\omega(t)$; $\sigma_a$ and $\sigma_\omega$ are the Gauss-Markov process temporal standard deviations; $\tau^a$ and $\tau^\omega$ are the correlation time for the accelerometers and gyroscopes; and $\Delta t$ is the time interval between the current and previous measurement.

2.3.1. Tightly-Coupled Implicit Iterative Extended Kalman Filter

The Kalman filter has been a popular choice for sensor fusion and navigation for decades. The Kalman filter assumes a linear relationship between the measurements and the states. Non-linearities can be approximated by using a first order Taylor series expansion. If the non-linearities are severe, higher order terms of the Taylor series can be included, and/or an iterative approach can be taken. All the filtering in this paper was done following the latter approach where an IEKF is used for updating the states [71]. A total of 16 states were used by the system (namely 3 for position, 3 for velocity, 4 for attitude in the quaternion, 3 for gyro bias, and 3 for accelerometer bias) and the size of this vector remains constant throughout the entire large-scale mapping project. This is desired and done to reduce the likelihood of divergence with a large state vector, and the reduction of system performance over time as more states are added [71]. In addition, many geometrical formulas used in vision actually violate the parametric model assumed by the KF (i.e. a single observation is a function of the unknown states, as shown in Equation 6); instead they are implicitly defined (i.e. the observations and unknown states are inseparable in the measurement update model, as shown in Equation 7). The extension of the conventional IEKF for implicit math models are given in [72], and was used in all the matching processes described in this paper. For outlier removal, the normalized residuals were used instead of testing the innovations [73]. This implicit IEKF formulation permits the Kinect measurements to update the state vector in a tightly-coupled manner, which allows for the system to be updated even if the matching is underdetermined (e.g. less than the minimum number of matches is found, or their distribution is collinear).

$$y = h(X) \tag{6}$$

$$h(X, Y) = 0 \tag{7}$$

where $y$ represents a single observation; $Y$ represents multiple observations; $X$ represents the unknown states; and $h$ denotes a functional model relating the observation(s) and unknowns.

### 2.3.2. Dense 3D Point Cloud Matching for the Kinect

Computer vision techniques for egomotion estimation and 3D reconstruction are very suitable for robotics SLAM applications because of their balance between speed, accuracy, and reliability. Often in freeform scan-matching, the ICP method is used. The ICP matching is usually first solved independently to reconstruct the scene, and then the relative translation and rotation between the point clouds are fused using a Kalman filter to update the pose. When the point clouds being matched are rich in geometric features, this method can provide high quality pose estimates. Typical indoor urban environments present additional challenges such as a single flat wall and sparse/poorly-distributed point clouds due to sensor saturation (the Kinect has difficulties producing a point cloud under direct sunlight, even through windows) and scene's reflectivity (e.g. mirrors and glass). For the former case, point-to-point ICP can converge to an infinite number of solutions based on the initial alignment, and the point-to-plane ICP will have a singular matrix that would not invert due to lack of geometric constraints. If the wall exhibits a lot of texture, RGB information can be combined with the standard geometry-only ICP in a joint optimization to obtain a solution [42], however if the wall is homogenous (e.g. white walls without textures are common) this approach is not applicable. Furthermore, based on the OpenNI programmer's guide [74] the depth and colour streams of the Kinect are not perfectly synchronized. In some extreme cases, data captured using Kinect Studio from the Microsoft Kinect SDK even revealed significant lag between the two streams. Figure 3 shows the temporal availability of colour and depth information while a Kinect was logging both simultaneously. The red and blue lines indicate when a RGB image and depth map was captured, respectively; lag up to 0.26 seconds can be observed.

**Figure 3.** Recorded colour and depth data using Kinect Studio with the red and blue lines showing the time when an image is acquired.

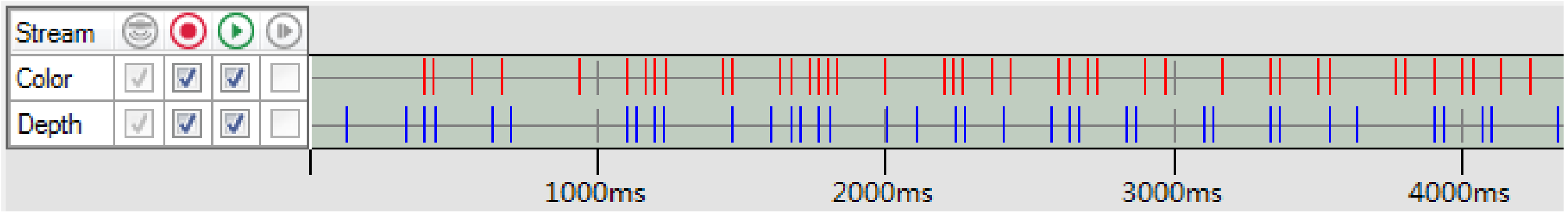

A general overview of the depth matching methodology is summarized graphically in Figure 4. A calibrated point cloud augmented with RGB and normals was generated in near real-time using OpenMP, Point Cloud Library (PCL), and [61]. Stable keypoints were then extracted from this new point cloud and matched to a previous keyframe. The estimated pose after outlier removal (using the uniqueness constraint, Signature of Histograms of OrienTations COLOR (SHOTColor) descriptor, IMU predicted image point locations, and RANSAC) was used as the initial alignment for the tightly-coupled ICP. During scan-matching, the input point cloud was iteratively matched with the reconstructed scene (i.e. the map) while incorporating the a priori information from the IMU. After a laser scan, when mobile mode was resumed, this approach becomes comparable to a localisation only solution in a known environment until reaching occluded areas. The output of this method includes the 16 updated states and an extended map of the environment. The state parameters were estimated in the KF, while the map was obtained from averaging new points at the current pose with the scene using a voxel grid. The adopted model-to-scene scan-matching approach estimates the map by allowing the scene to be updated through a moving average instead of augmenting the state vector with the map, which is computationally demanding for the KF since every Kinect exposure contains up to 307 200 points.

The central concept of the proposed depth processing pipeline is the ICP algorithm. Numerous variations of ICP can be found in literature, each with their own advantages and disadvantages. Rusinkiewicz and Levoy [36] summarized five key components for tuning the ICP algorithm: the sampling method, searching, cost function, outlier rejection, and weighting. Following this guideline, a new point-to-plane ICP solved in a tightly-coupled IEKF is proposed to account for the solo homogenous flat wall situation described above. Through the additional information provided by the dynamics model, not only does the initial alignment become trivial, it makes the single plane case solvable. The following sections explain the proposed ICP algorithm in relation to the five key components, initial alignment, and the loop closure problem.

**Figure 4.** SLAM using the Kinect's depth information.

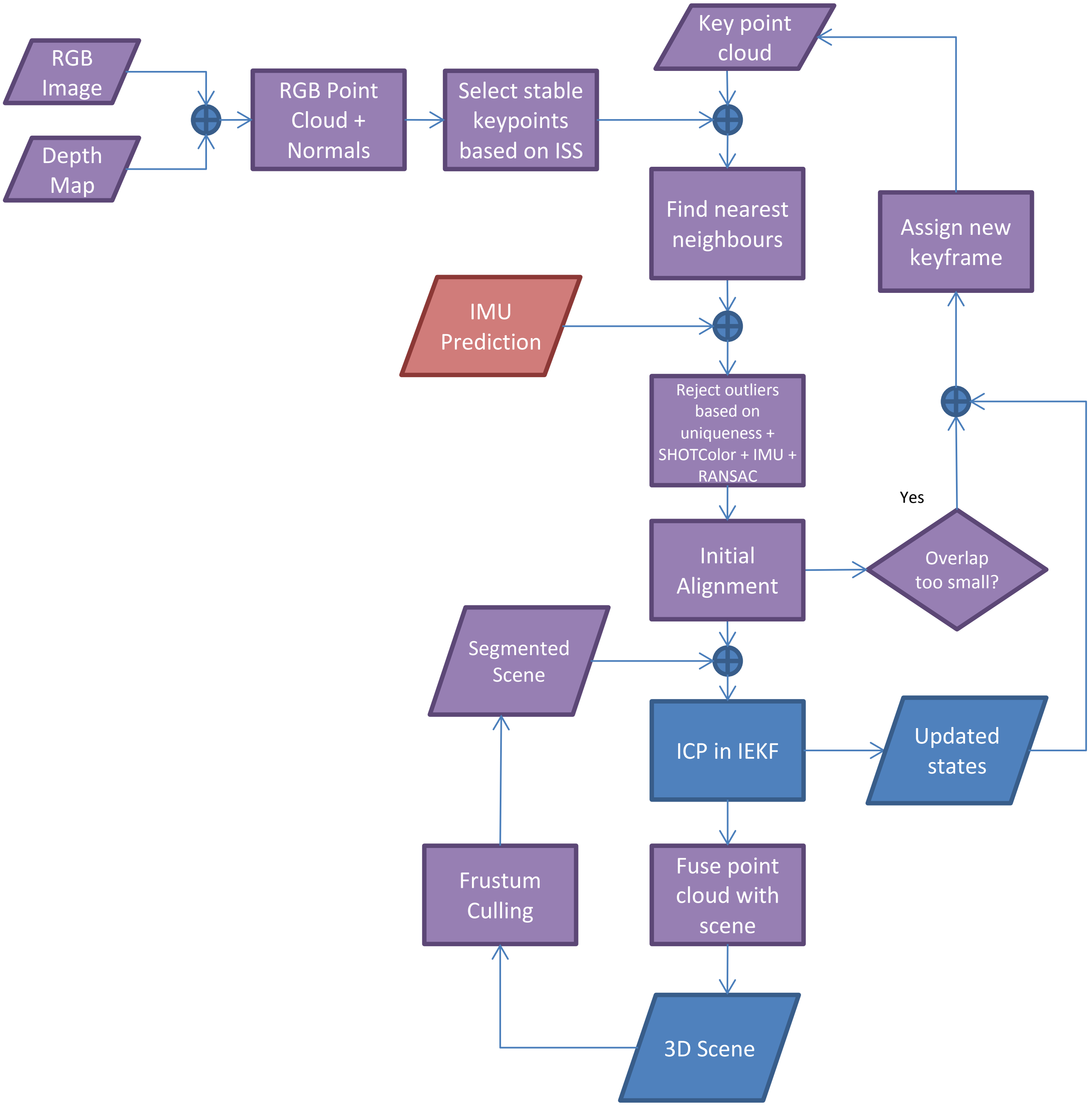

*2.3.2.1. Sampling*

A single Kinect depth map is 640 by 480 pixels, which can result in a point cloud with a maximum size of 307 200 points. This is a lot of data, especially considering the Kinect simultaneously captures RGB images at 30 Hz. Since the corresponding points between the IR camera and projector are determined using a 9 x 9 kernel [55], depth values of adjacent pixels are highly correlated and do not carry as much information as pixels that are further away. Therefore, to reduce the computation time, the point clouds were downsampled. For this, the normal downsampling technique suggested by [36] was used, because it showed better performance than uniform and random downsampling.

*2.3.2.2. Searching*

The popular Fast Library for Approximate Nearest Neighbors (FLANN) [75], which is a fast KD-tree, was used for speeding up the query process, which is one of the most computationally intensive steps of the ICP algorithm. Instead of just finding a single nearest neighbour in the target cloud for every point in the source cloud, the k-nearest points in the target cloud based on the Euclidean distance to the query point were selected, and the point with the minimum orthogonal distance was chosen as the corresponding point. This was then repeated with every point in the target cloud as query points to find the correspondences. Only correspondences that pass this uniqueness constraint were considered as matches.

Instead of building a KD-tree for a growing scene and querying a large amount of points, two steps were taken to reduce the memory and the processing load. First, every new ICP-registered point cloud was fused with the scene model using a voxel grid representation, where points within the same voxel were reduced to their centroid. The second step involves a fast segmentation of the scene to extract only regions that were overlapping with the new incoming point cloud using frustum culling [76] based on the FOV of the Kinect and IMU-predicted pose. This essentially sets a threshold on the maximum number of points to be queried, thus prevented certain iterations from running significantly longer than others.

*2.3.2.3. Cost Function*

The point-to-plane cost function minimizes the orthogonal distances instead of the Euclidean distances as in the point-to-point ICP; this is known to perform better in terms of accuracy and convergence speed for most cases [36]. More importantly, it is preferred in this application because it would not "anchor" the along-track translation of the robot when it is travelling down a corridor. The point-to-plane ICP will only minimize the 1D orthogonal distance, therefore for a side-looking Kinect it will not prevent it from sliding along the wall as desired. When solving it in a KF framework, information from the dynamics model (i.e. IMU) can help push the solution forward in the along-track direction while the ICP corrects the across-track position and the two orientations that are not parallel to the normal of that wall. Furthermore, the proposed tightly-coupled point-to-plane ICP will not be singular even if only one plane is within view.

As for all camera-based models, minimization of the reprojection errors yields higher quality results than simply minimizing the Euclidean distances because it better describes the physical measurement principle. Following the derivations from [61], for a rigidly attached stereo camera, its measurement model can be written as Equation 8 (for the master camera) and 9 (for the slave camera), where the projector is treated as a reverse camera. The inherent scale ambiguity for both the IR camera ($\mu_{i,k}^{IR}$) and the projector ($\mu_{i,k}^{PRO}$) are solved through spatial intersection with the point-to-plane constraint (realized by substituting Equations 8 and 9 into Equation 10, respectively). The depth camera measurement update model implemented in the proposed KF framework is then obtained by equating Equations 8 and 9. It is worth noting that the functional relationship between the observations and unknown states are implicitly defined. The math model presented here minimizes the reprojection errors of the Kinect's IR camera and projector pair to reduce the orthogonal distance of every point in the source point cloud relative to the target point cloud.

$$O_i^n = R_s^n(t)\left\{R_{IR}^s R_k^{IR}\left\{\mu_{i,k}^{IR}(t) p_{i,k}^{IR}(t) - \Delta b^k\right\} + \Delta b^s\right\} + T^n(t) \tag{8}$$

$$O_i^n = R_s^n(t)\left\{R_{IR}^s R_k^{IR}\left\{\left[R_{PRO}^{IR}\right]_k \mu_{i,k}^{PRO}(t) p_{i,k}^{PRO}(t) + \Delta c^k - \Delta b^k\right\} + \Delta b^s\right\} + T^n(t) \tag{9}$$

$$\begin{bmatrix} a_i & b_i & c_i \end{bmatrix} \bullet \begin{bmatrix} O_i^n \end{bmatrix} - d_i = 0 \tag{10}$$

where, $R_k^{IR}$ is the relative rotation from the IR camera of Kinect $k$ to the IR camera of the reference Kinect; $\mu_{i,k}^{IR}(t)$ and $\mu_{i,k}^{PRO}(t)$ are the scale factors for point $i$ observed by the IR camera and projector of Kinect $k$ at time $t$, respectively; $p_{i,k}^{IR}(t)$ and $p_{i,k}^{PRO}(t)$ are the image measurement vectors of point $i$ observed by the IR camera and projector of Kinect $k$ at time $t$, respectively; $[R_{PRO}^{IR}]_k$ is the relative rotation between the IR camera and projector of Kinect k; $\Delta b^k$ is the relative translation between the IR camera of Kinect $k$ and the reference Kinect; $\Delta c^k$ is the relative translation between the IR camera and projector of Kinect $k$; $a_i$, $b_i$, $c_i$, and $d_i$ are the best-fit plane parameters of point $i$ in the navigation frame.

*2.3.2.4. Outlier Rejection*

The correspondence matching between two point clouds can only be seen as approximate because exact point matches cannot be assumed in dense point clouds captured by the Kinect or laser scanner. This causes the ICP algorithm to be highly sensitive to outliers in the approximated correspondences. To minimize the possibility of outliers, several precautions were taken, resulting in the following three outlier removal steps.

<u>2.3.2.4.1 Rejection When Estimating Correspondences</u>

One of the benefits of including an IMU is that it can provide good initial alignment for the ICP. Using this prediction and a fixed radius around each query point, correspondences established during the searching step above that were outside of each sphere can be removed. The remaining correspondences were further compared using their normal vector determined in their local neighbourhood using orthogonal regression. Only points with differences in spatial angle less than a threshold were retained.

<u>2.3.2.4.2 Rejection Using RANSAC</u>

When more than 3 non-collinear points were matched between the source and target clouds, RANSAC based on the point-to-point ICP from Besl and Mckay [33] can be used for finding the inliers efficiently. This approach is similar to the RANSAC approach explained in [40].

<u>2.3.2.4.3 Rejection in the Kalman Filtering</u>

In case outliers still exist in the correspondences, standard outlier detection method based on the normalized residuals was adopted [73].

*2.3.2.5. Weighting*

Assigning weights that are inversely proportional to the depth measurements when registering point clouds coming from the Kinect has shown to have significant improvements in terms of accuracy [49]. For the Scannect this was achieved based on the pin-hole camera model and collinearity equations with assumed additive Gaussian noise for the pixel measurements, which have been shown to be capable of describing the distance measurement uncertainty of the Kinect over its full range [27]. This weighting scheme simplifies the observation variance-covariance matrix to a diagonal matrix while properly describing uncertainties in the point cloud due to the baseline-to-range ratio and parallax angle over the Kinect's full frame. In contrast to [49], which would assign the same weight to all the points when registering coplanar points between two point clouds while the Kinect is oriented orthogonal to the wall, the adopted method will assign slightly different weights based on the variation of the parallax angles in these coplanar points. The end result is similar to the empirical weighting scheme adopted in [49], but with the underlying physical measurement principles of the Kinect expressed mathematically.

*2.3.2.6. Initial Alignment*

Generally, consecutive Kinect depth maps are captured merely fractions of a second after another. This plus transporting the Kinect slowly and smoothly during the SLAM process results in the initial transformation of the ICP to be highly predictable by a suitable motion model, and sometimes can even be assumed to be the same as the previous pose. Nevertheless, the IMU predictions are usually sufficient to get the ICP to converge to the correct solution because of the relatively short time update. However, because the ICP tends to converge to a local minimum as oppose to the global minimum, whenever possible the initial alignment is derived via other means to keep the ICP updates as independent from the IMU predictions as possible. This reduces the chance of a poor IMU prediction causing the ICP to converge to a local minimum and then using this distorted scene for camera localisation.

This alternative method for deriving the initial alignment is based on 3D keypoint matching [40]. Similar to the conventional 2D keypoint matching technique [77], 3D keypoint matching begins with keypoint detection, every keypoint is then associated with a description, and points with the most similar description are matched. Most of the papers directly solve for the egomotion in EKF using this keypoint method, indicating its fidelity. However, in this paper they are only treated as initial approximations for ICP because a reliable implementation typically requires texture information, which as mentioned, RGB images are not triggered simultaneously as the depth images in the Kinect.

The Implicit Shape Signature (ISS) [78] was used for extracting keypoints with a unique and stable underlying surface. Many existing 3D keypoint detectors are just extensions of 2D image edge/corner detectors (e.g. Harris). While edges/corners are good for 2D images, in 3D their depth measurements are usually less reliable due to for example the mixed pixels effect. Therefore, the ISS, which was originally designed for 3D point clouds was favoured. Also, it was chosen because [79] showed it has better overall performance than other 3D keypoint detectors that are publically available in PCL. To have a strong description for the detected keypoints, SHOTColor (a.k.a. Color Signature of Histograms of OrienTations (CSHOT)) in PCL was used [80]. This is one of the few descriptors that incorporates both shape and color information into their description, making its signature the most unique (i.e. highest dimension) in PCL. RANSAC was then used to select a group of inliers and estimate the initial alignment between the two clouds to be used in ICP. Since RANSAC is an expensive process, the IMU predictions were used to eliminate obvious false positives by setting a radius threshold around each keypoint.

#### *2.3.2.7. Loop-Closure*

The proposed method presented here focussed only on the filtering solution, also known as the SLAM frontend. For a full SLAM solution, the global relaxation and optimization is crucial. To partially address the issue of loop closure in this paper, the ICP is carried out in a model-to-scene fashion like KinectFusion [43]. This way, every new cloud coming from the sensors was matched to the global scene. This approach can be even more effective when performing on laser scanner data because of their wide FOV and long-ranges.

### 2.3.3. RGB Visual Odometry

As mentioned, RGB data was processed without consideration of the depth information. This reduces errors in egomotion estimation introduced through time synchronization errors [49] and more importantly keeps track of the pose even when depth data is unavailable and/or unreliable (e.g. imaging glass) [39]. The basic assumption is that the scene is static and any changes in the appearance of the RGB image must be caused by the camera's movement. For this task of identifying changes in consecutive images, the popular SIFT keypoints are detected along with descriptions computed by SURF. A blob detector is preferred over edge/corner detectors because such geometric information is already encapsulated in the depth data processing pipeline [81]. Correspondences were established by searching for points in the two images with the most similar description. These potential matches were then filtered based on uniqueness constraint, the IMU predicted keypoint positions in the new image, 5-point RANSAC in OpenCV, and later via outlier rejection at the measurement update stage. When the pixel differences between conjugate image points are smaller than a threshold ZUPT is performed, otherwise the camera measurement update is executed. The proposed VO workflow is summarized in Figure 5.

Monocular camera trajectory estimation in general can be solved as a SLAM problem or using VO [32]. The former achieves better global consistency at the expense of higher complexity and being more computationally demanding. While the latter focuses on local consistency and has shown great success for example in the Mars Rover exploration. VO is preferred in this project because the mapping and localisation are mainly achieved using ICP, while the RGB data is just an aid to improve localisation when depth is not available or when traversing over highly texturized areas of the map. The proposed VO algorithm differs from PTAM in the sense that it implicitly solves for the 3D points from a pair of images. This simultaneous photogrammetric intersection and resection is mathematically realized through solving the set of linear equations in Equation 11 and obtaining an implicit functional model. This was the main motivation to choose the VO route because it is possible to formulate a 2D-to-2D matching VO so as to not increase the size of the state vector while incorporating all the camera geometry information, correlation between the pose and reconstructed points, and minimizing the reprojection errors.

Its advantages over PTAM include egomotion estimation by matching across 2 images instead of 3 images, no need for a recovery routine in case camera tracking fails, and it is more scalable because of the fixed state vector dimension. A noticeable disadvantage arises from the fact that reconstructed points are not memorized; in other words they are immediately forgotten in the KF and later images cannot resect from continuously improving 3D object points, but only from selected keyframes.

**Figure 5.** Visual odometry using monocular vision.

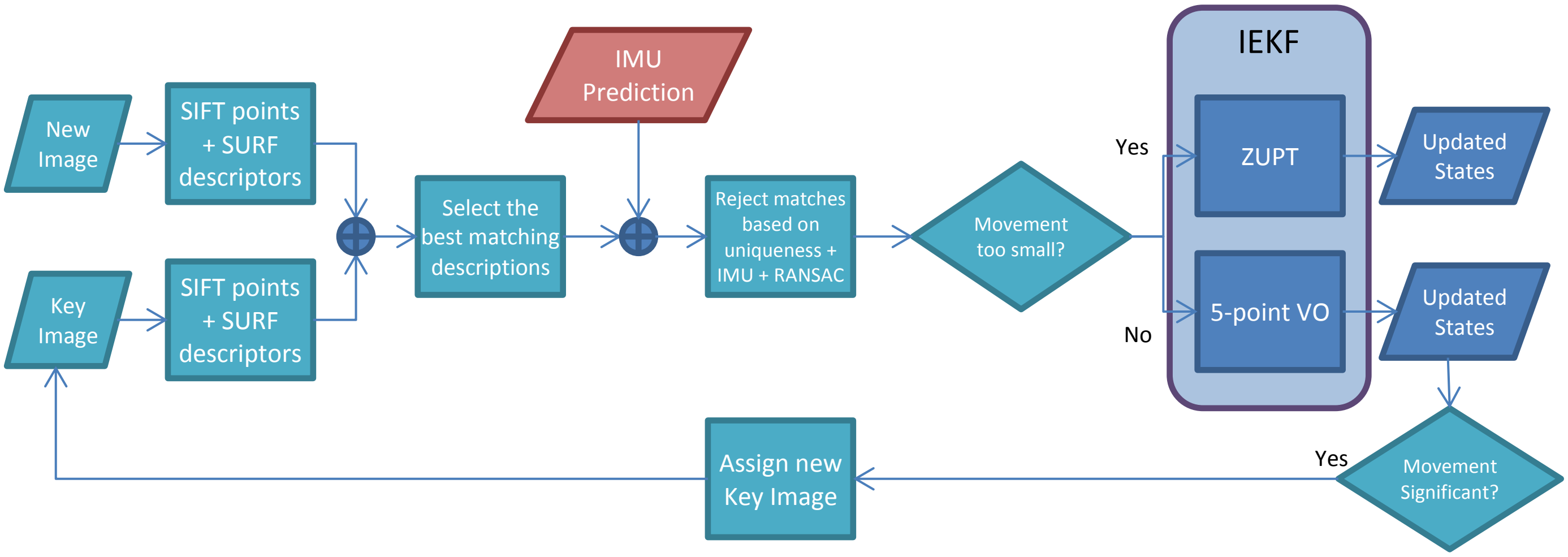

$$R_s^n(t+1)\left\{R_{IR}^s R_k^{IR}\left\{\left[R_{RGB}^{IR}\right]_k \mu_i^{RGB}(t+1)p_i^{RGB}(t+1)+\Delta d^k-\Delta b^k\right\}+\Delta b^s\right\}+T^n(t+1)$$
$$-\left\{R_s^n(t)\left\{R_{IR}^s R_k^{IR}\left\{\left[R_{RGB}^{IR}\right]_k \mu_i^{RGB}(t)p_i^{RGB}(t)+\Delta d^k-\Delta b^k\right\}+\Delta b^s\right\}+T^n(t)\right\}=0 \quad (11)$$

where $[R_{RGB}^{IR}]_k$ is the relative rotation between the RGB and IR camera of Kinect k; $\Delta d^k$ is the relative translation between the RGB and IR camera of Kinect *k*.

## 3. Results and Discussion

### *3.1. Boresight and Leverarm Calibration Between the Kinect and Laser Scanner*

Once all the optical systems were independently calibrated their measurements need to be related in 3D space before they can be tightly-coupled and fused in a KF. A standard checkerboard pattern was placed in 17 different static positions and orientations in front of one of the Kinects, where point clouds with the RGB texture overlaid were captured. At every setup, a laser scan of the checkerboard was also captured. The corners were extracted in 3D using a least-square cross-fitting [82]. The necessary boresight and leverarm parameters were then computed using the 3D rigid-body transformation equations solved in a least-squares adjustment. The same procedure was then repeated between the other Kinect and the Focus$^{3D}$ S. Although the two sets of boresight and leverarm parameters were defined with respect to the Focus$^{3D}$ S, for the convention adopted, they were redefined to use the left Kinect (mounted with the IMU) as the reference coordinate system. To assess the fit between the point clouds, 10 planes at various orientations and positions were extracted from both Kinects and compared with the Focus$^{3D}$ S as the reference. The RMSE computed using the orthogonal distances between the Kinect point clouds and the plane defined by the Focus$^{3D}$ S is 4.5 mm and 6.2 mm for the left and right Kinects, respectively. These estimates were higher than anticipated based on the expected accuracy from [61] and [65] but can probably be attributed to the instability of the camera mounts relative to the trolley in this prototype.

*3.2. Boresight and Leverarm Calibration Between the Kinect and IMU*

As all the optical sensors were already calibrated both internally and externally relative to each other, only the boresight and leveram parameters between the IMU and one of the optical sensors, in this case the left Kinect was necessary to complete the system calibration. A checkerboard pattern was placed horizontally on a levelled surface. The rigidly-attached Kinect and IMU were then moved in front of it to ensure sufficient excitations about all three axes. To remove any range restrictions, only a video sequence from the RGB camera was captured at 10

Hz along with the IMU data at 100 Hz. A photogrammetric resection was performed in a tightly-coupled IEKF to update the IMU predictions online. After filtering, a global adjustment was used to estimate the boresight and leverarm parameters simultaneously offline. The errors between the IMU predicted corner locations and the measured locations in the RGB images before and after calibration is given in Table 1.

**Table 1.** Errors between the predicted and measured target positions in image space before and after calibration

| | Before Calibration | After Calibration | Improvements |
|---|---|---|---|
| RMSE x | 29.1 pix | 0.8 pix | 97% |
| RMSE y | 23.1 pix | 1.1 pix | 95% |

*3.3. Point-to-Plane ICP for the Kinect*

To compare the proposed ICP algorithm to the widely accepted point-to-plane ICP from Chen and Medioni, the Kinect was mounted on a translation stage while imaging a complex scene between 0.9 m and 2.9 m away (Figure 6). A known translation of 5 cm was then introduced in the depth direction of the Kinect with a standard deviation of 0.05 mm and then a second point cloud was captured. The two point clouds were then registered using the two ICP algorithms. The RMSE of the fit between the two point clouds computed using the point-to-plane distances metric was 7 mm using either algorithm. Also, based on a qualitative analysis of the registered point clouds, no apparent differences between the two methods were identifiable. However, there is a slight difference between the recovered movements of the Kinect. The method by Chen and Medioni [35] underestimated the translation by 1.1 mm and the proposed method underestimated the translation by 0.1 mm. This small improvement comes from the weights assigned to correspondences (based on the triangulation geometry) and a cost function which minimizes the reprojection error of the IR camera and projector pair. Differences of a millimetre may seem insignificant at first, but in the absence of additional information the errors can accumulate rather quickly in a Kinect-only dead-reckoning solution, especially when the Kinect is capturing with a frequency as high as 30 Hz.

**Figure 6.** Calibrated RGB-D data from Kinect of the scene used for testing ICP

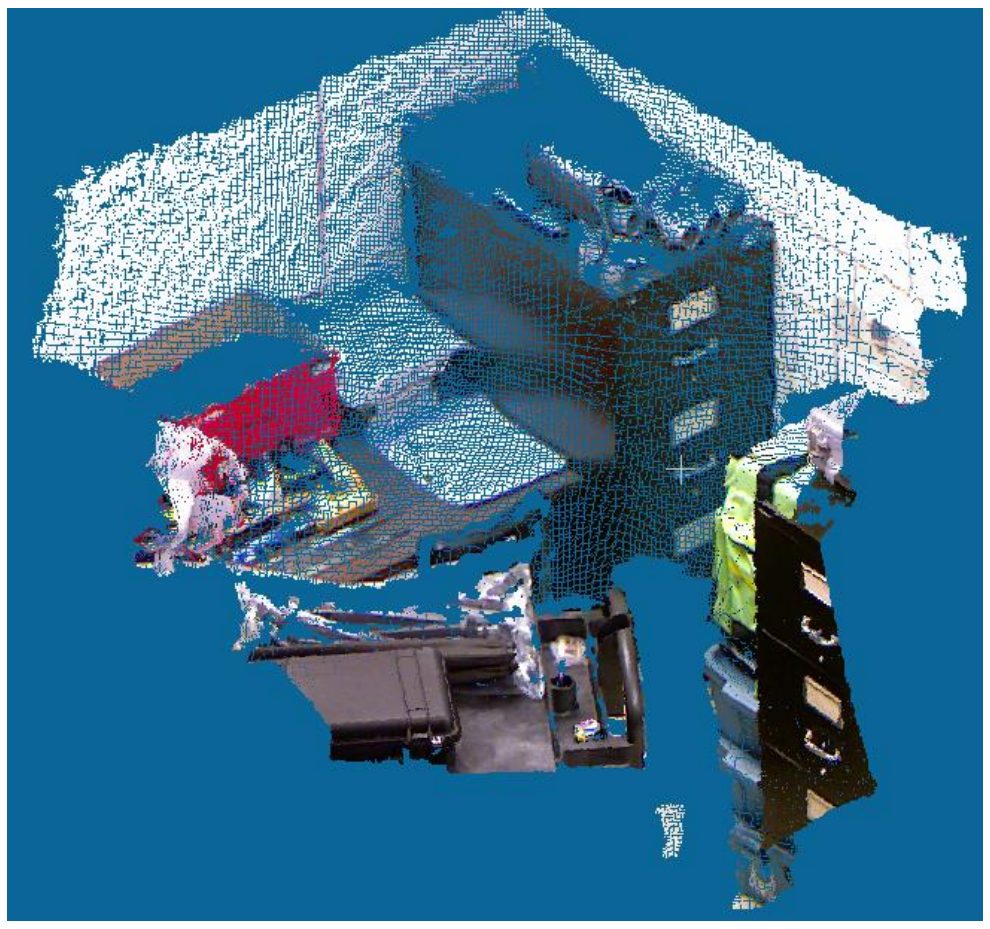

To test the suitability of the devised ICP algorithm for mapping applications when assisted by an Xsens MTi-30 IMU, a sequence of handheld data was captured. This illustrates the mapping abilities when this module is removed from the main trolley platform for exploring inaccessible spaces like stairs before returning to its forced centred position. The Kinect and the MTi-30 were rigidly mounted together with the Kinect logging at 10 Hz and the MTi-30 logging at 100 Hz. A person carried the Kinect and IMU by hand, ascended a flight of stairs, turned around and then descended the same flight of stairs. A total of 315 RGB-D datasets were captured and processed. The final 3D point cloud of the staircase is shown in Figure 7 where the colors are assigned based on the local normals. Considering that the distances travelled was short, the model-to-scene registration scheme was capable of partially handle the loop-closure problem and no apparent misfits can be identified near the bottom of the staircase. For a quantitative accuracy assessment, 20 horizontal and vertical planes were extracted and the overall RMSE from this check plane analysis was 7.0 mm.

**Figure 7.** Stairs reconstructed using the Kinect and IMU in a human carried walking sequence.

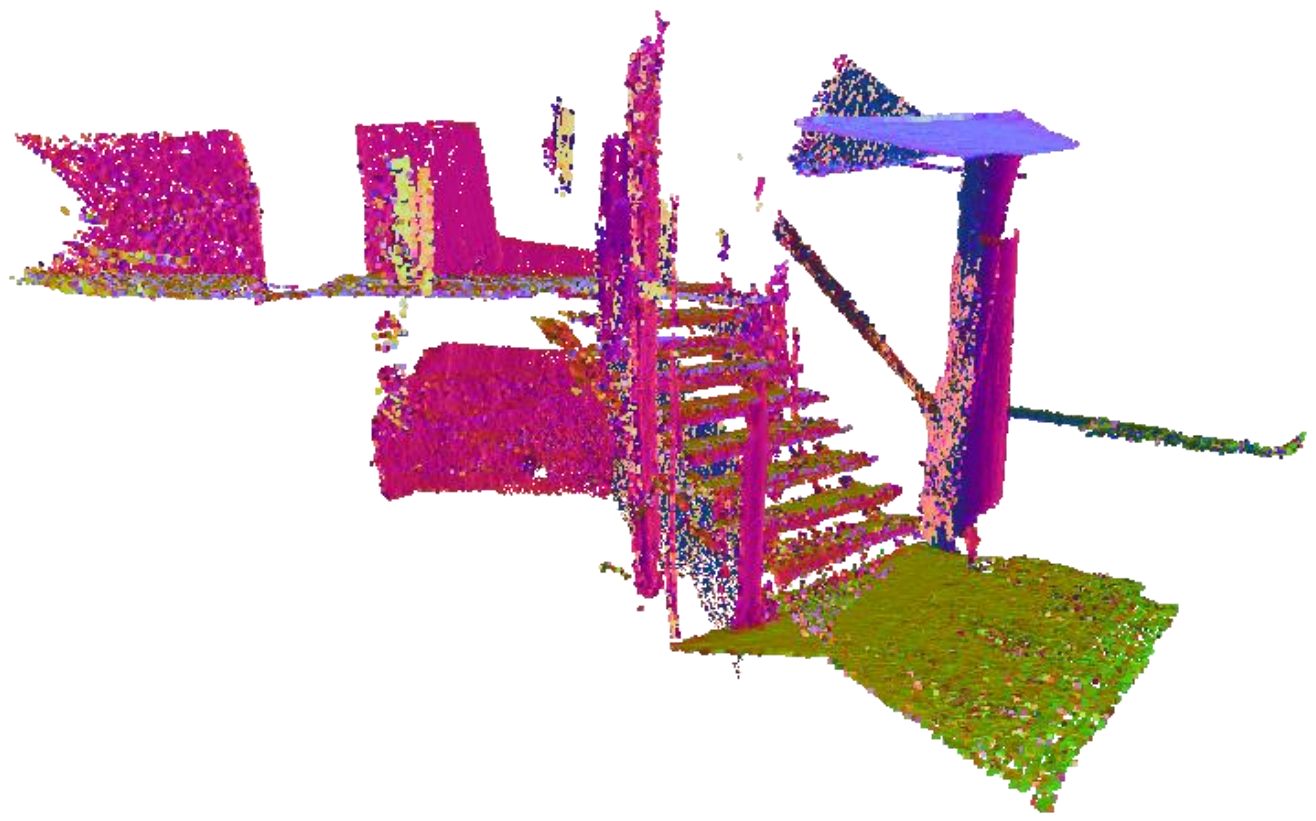

*3.4. 2D-to-2D Visual Odometry for Monocular Vision*

A checkerboard pattern with known dimensions was imaged with the Kinect's RGB camera. The Kinect was handheld and waved in front of the checkerboard for approximately 80 seconds. The reference solution was generated using photogrammetric resection solved in a least-squares adjustment frame-by-frame. This represents the best case scenario in camera navigation where known control points are always visible. VO was performed by solving for the camera's pose using the mathematical model presented in Equation 11 with the first image held as the reference keyframe the entire time. The camera orientations converted from quaternion to Euler angles are shown in Figure 8. It can be observed that the VO solution follows the reference solution quite closely over certain sections of the plot (e.g. between 350 and 362 seconds). But in other sections a noticeable difference in their estimation can be observed, with errors reaching as high as 80o. The main cause for such a discrepancy is the intersection geometry. When the pair of images were captured near proximity of each other, the parallax angle is small, this results in a less accurate scene reconstruction, which translates into poor egomotion estimation. In fact, the proposed math model has a singularity when the light rays in the pair of images are all parallel with each other. However, this can easily be identified by checking the disparity of conjugate points. The case when the disparities are all below a threshold actually indicates the camera has not moved significantly, suggesting the possibility of the camera being static, therefore ZUPT was performed. For a MEMS-based IMU, integration of the gyroscope readings alone over a period of approximately 80 seconds would have introduced a noticeable rotation drift. But as witnessed in Figure 8, the fusion of VO with IMU prevented the estimated orientation from drifting. At the same time the IMU smoothed out the vision-based orientation estimates and provided a better orientation estimate when the uncertainty in VO is high, resulting in a final solution that follows the reference solution quite closely over the full trial. This suggests that VO and gyroscope combined can provide a better orientation estimate than either method alone.

**Figure 8.** Comparison between various ways of estimating monocular camera rotations. α, β, and γ represents the rotation about the x, y, and z axis, respectively. Res represents the photogrammetric resection solution.

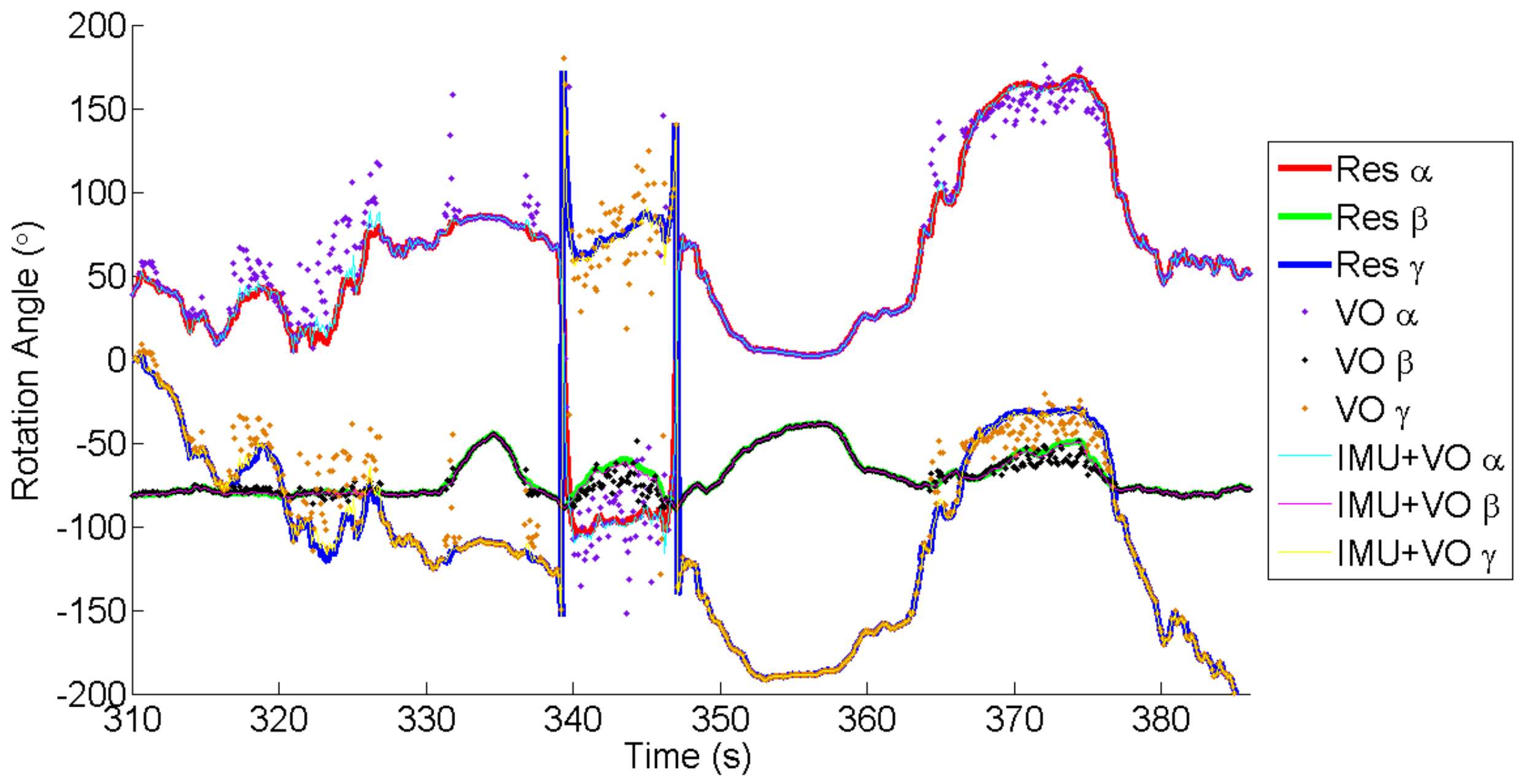

*3.5. Scannect Testing at the University of Calgary*

The full Scannect system was tested by mapping a floor in the Math Science building located at the University of Calgary, Canada (Figure 9). The hallways are mainly composed of doors (some are recessed), posters, bulletin boards, and white homogenous flat walls. The Scannect travelled 120 m in continuous mode and only stopped at the four corners to capture a 7 minute scan with the Focus$^{3D}$ S. While in continuous mode, both Kinects were capturing data at 10 Hz while the IMU was logging at 200 Hz. Because the floor was relatively flat, a height constraint was applied. Also, ZUPTS were applied every time sufficiently small movements were detected by the RGB VO and when the Focus$^{3D}$ S was scanning. The total time it took to map the floor in the field was 35 minutes, significantly less than a typical laser scanner only project over such a large area. The map generated by the Kinect with RGB texture information added for visual appearance is shown in Figure 10. Although only the points measured by the Kinects are shown, these results were generated with the assistance of the Focus$^{3D}$ S. At the beginning a 360$^{o}$ scan was first captured by the scanner and added to the scene. Then the Kinects used this as the scene in the dense 3D matching step. Meanwhile it filled in some occlusions and expands the map until it arrived at the next corner where another laser scan was captured and matched to the scene through the point-to-plane ICP with the IMU predicted pose as the initial alignment. This process was repeated until the floor had been explored (completely mapped).

Throughout the floor, 30 checkpoints were also captured by the Kinects. A top view of the Kinects' map along with the horizontal distribution of these signalized targets is shown in Figure 11. Due to the limited vertical FOV of the Kinects, all the checkpoints were approximately at the same height. The RMSE in the X, Y, and Z direction was 10.0 cm, 10.8 cm, and 9.0 cm, respectively. These error estimates only indicates the forward filtering solution. After backward smoothing or SBA, the results are expected to improve [40]. Note that the laser scanner only solution shown in Figure 12 (with the green stars indicating the static scan locations) missed a lot of the doorways (in particular the top corridor between targets A26 and A18) but the general shape of the floor is prevalent. This general shape from the laser scans aided the Kinects' localisation and provided some degree of loop-closure. From Figure 11 it can be perceived that the details such as doorways that were missed by the laser scanner were well localized in the same coordinate system by the Kinect. Some misalignments in the scanner data can be observed (e.g. the top corridor), this is likely a result of errors in the initial alignment provided by the Kinect and IMU filter and the accumulated mapping errors.

**Figure 9.** Image of one of the hallways in the Math Science building at the University of Calgary.

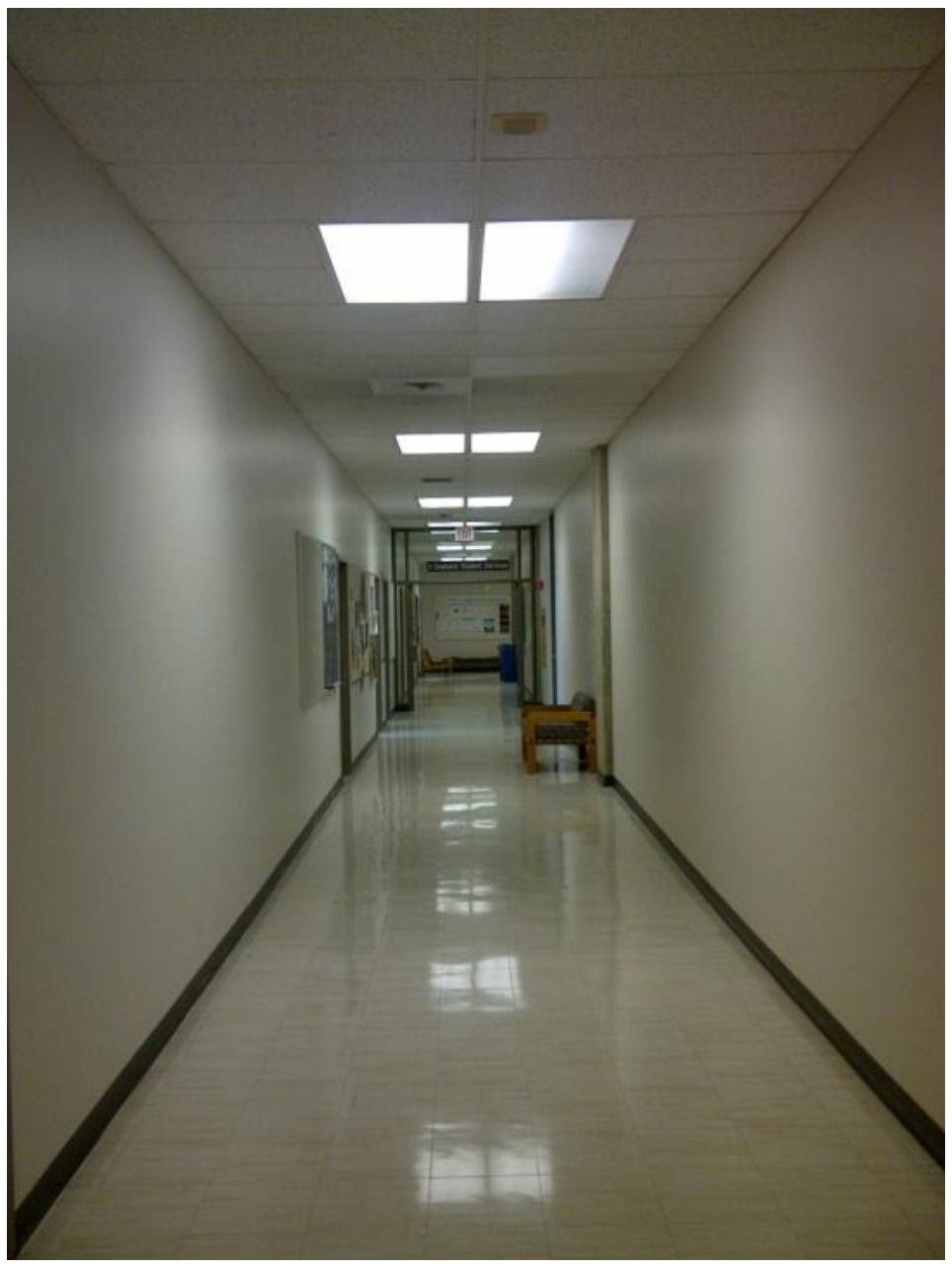

**Figure 10.** Oblique view of the Kinects point clouds from the 3D reconstructed scene.

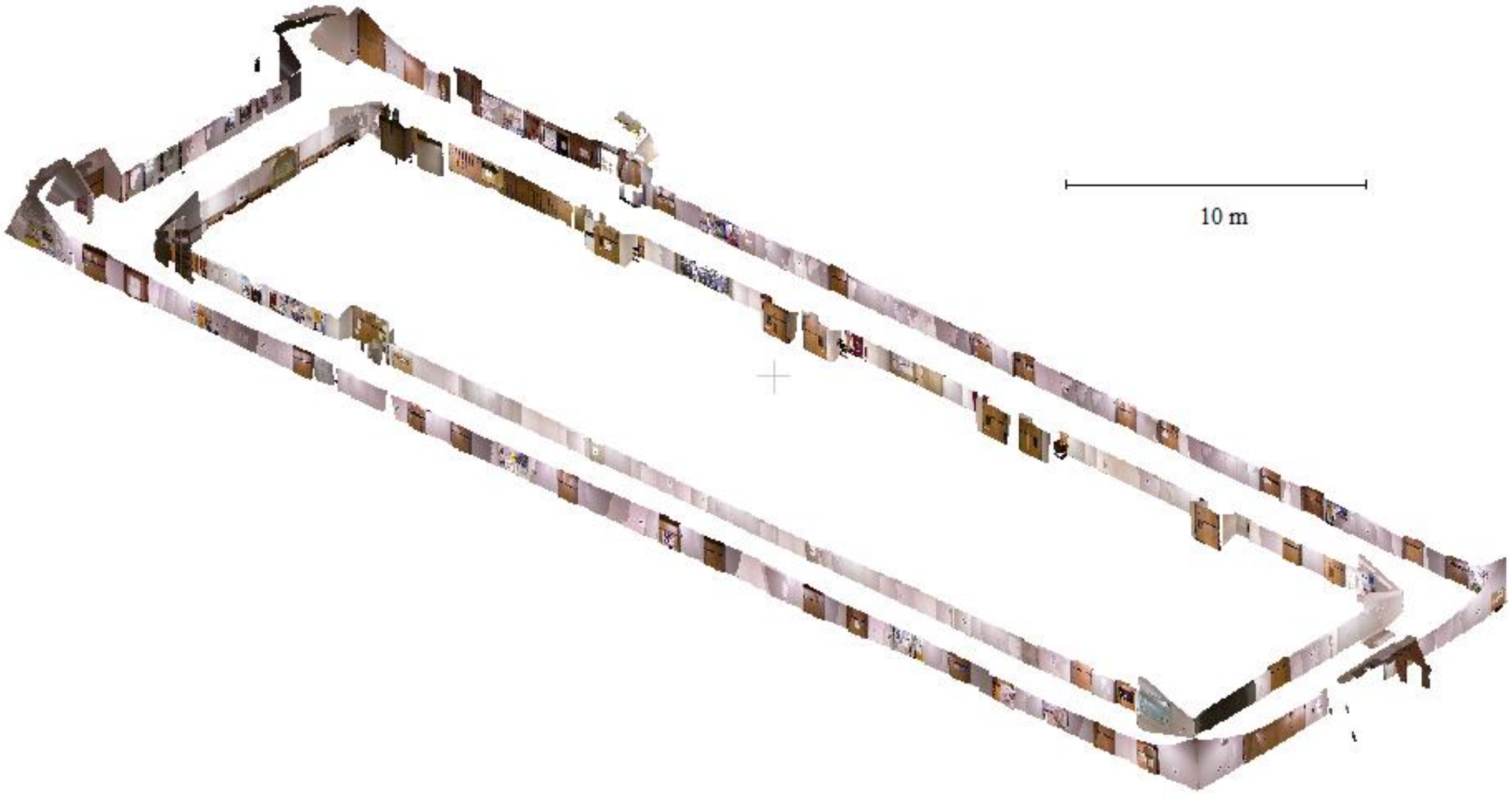

**Figure 11.** Top view of the *Kinects* point clouds from the 3D reconstructed scene. The magenta labels indicate the location of the check points.

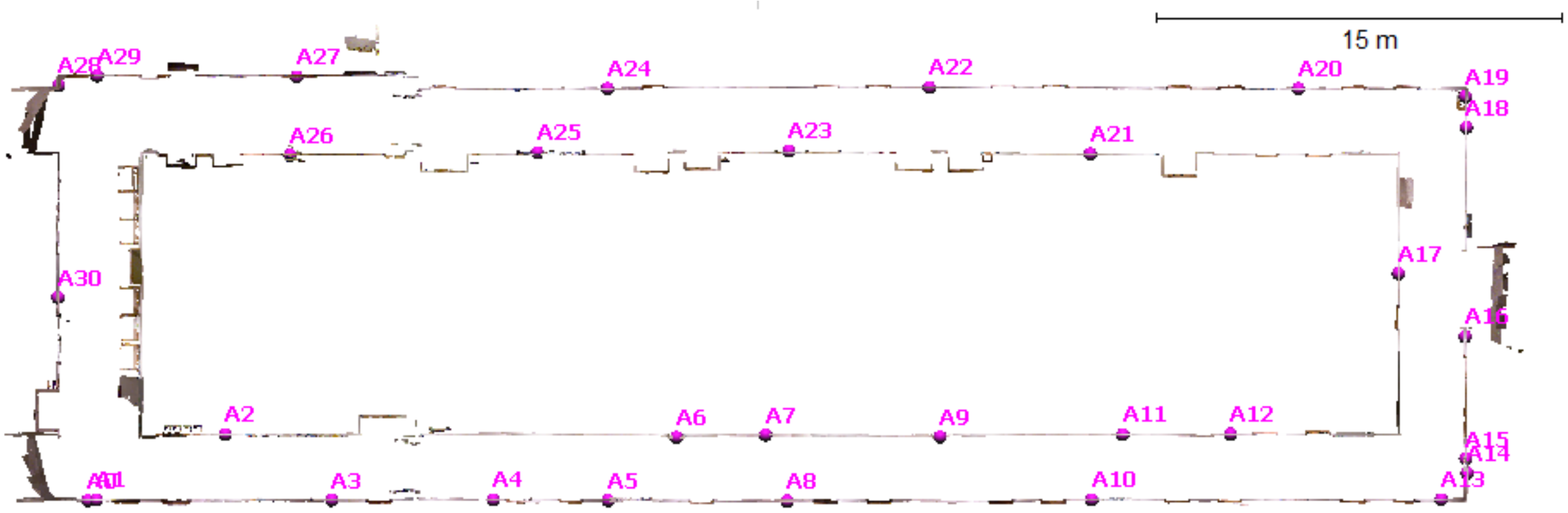

**Figure 12.** Top view of the *laser scanner* point clouds from the 3D reconstructed scene with the ceiling and floor removed. The green stars indicate the scan stations.

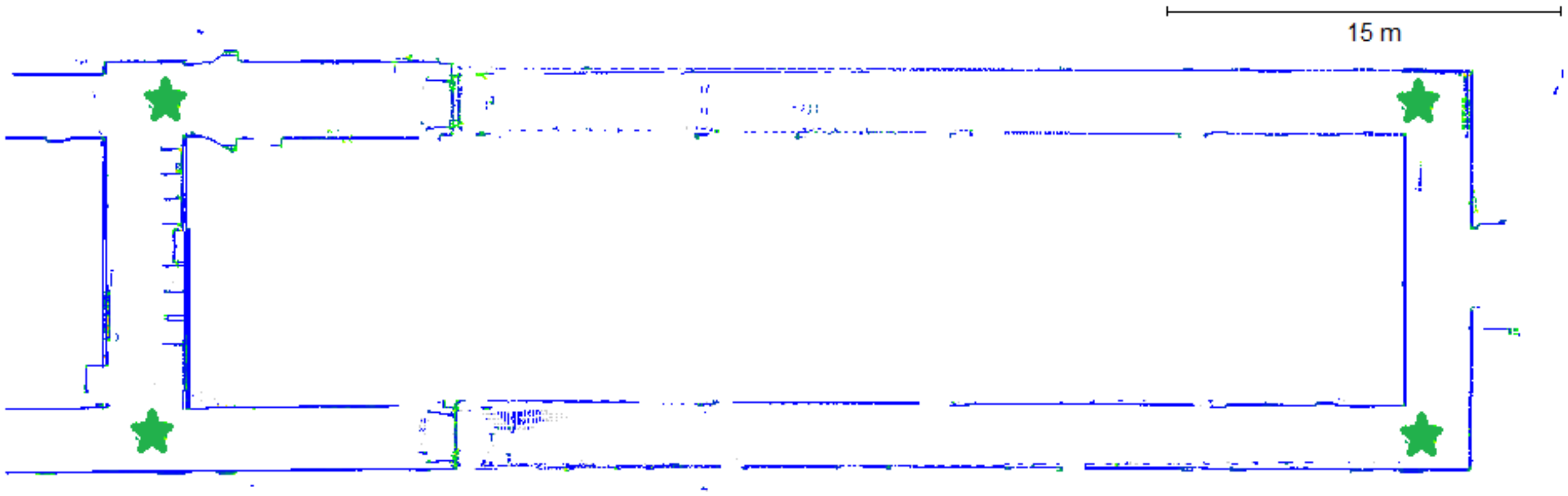

## 4. Conclusions and Future Work

SLAM is a theoretically matured subject with many practical implementation challenges. A few of these challenges were addressed in this paper: the implicit iterative extended Kalman filter was responsible for fusing measurement updates from the Kinect's RGB and depth stream separately to account for their timing errors and non-linearity of implicit photogrammetry equations; using only the RGB images, a 5-point visual odometry was performed in a tightly-coupled manner while eliminating the need to add or remove states in the filter; a new tightly-coupled ICP method for stereo-vision systems was used for processing the depth maps. The presented Scannect system is the first multi-Kinect mobile mapping system and the first to adopt the continuous stop-and-go design. This system can map an entire building floor efficiently with a Mean Radial Spherical Error of 17 cm. It was capable of navigating in areas with as little features as a single white wall.

In the future, more Kinects will be added to further increase the field of view of the Scannect. For instance, to make the Scannect more flexible and account for sloped floors, a skyward facing Kinect can be added without any modifications to the algorithms and processing pipeline presented in this paper. Newer 3D cameras with a wider field of view (e.g. PMD CamBoard nano) or higher accuracy (e.g. Kinect 2) will be considered. Adding other autonomous non-vision-based sensors such as wheel odometers and magnetometers will be beneficial for localisation. Other non-holonomic constraints such as levelling updates can also be included if applicable. A more accurate scene representation such as surfels will be adopted [83]. To better handle loop-closures, methods like sparse bundle adjustment will be implemented to improve the global consistency. This was shown to be more accurate than TORO (a graph-based method) when using the Kinect [40]. For real-time online processing and visualisation, general-purpose computing on graphics processing units (GPGPU) will be investigated to expedite the filtering process.

**Acknowledgments**

This research is funded by the Natural Science and Engineering Research Council (NSERC) of Canada, Alberta Innovates, the Canada Foundation for Innovation, and the Killam Trust.

**References**

1. T. Gadeke, J. Schmid, M. Zahnlecker, W. Stork and K. Muller-Glaser, "Smartphone pedestrian navigation by foot-IMU sensor fusion," in 2nd International Conference on Ubiquitous Positioning, Indoor Navigation, and Location Based Service, pp. 1-8, Oct. 3-4, Helsinki, Finland, 2012.

2. H. Durrant-Whyte and T. Bailey, "Simultaneous localization and mapping: part I," IEEE Robotics & Automation Magazine 13 (2), pp. 99-110, 2006.

3. T. Bailey and H. Durrant-Whyte, "Simultaneous localisation and mapping (SLAM): Part II state of the art," IEEE Robotics and Automation Magazine 13(3), pp. 108-117, 2006.

4. A. Huang, A. Bachrach, P. Henry, M. Krainin, D. Maturana, D. Fox and N. Roy, "Visual odometry and mapping for autonomous flight using an RGB-D camera," in 15th International Symposium on Robotics Research, August 28 - September 1, Flagstaff, USA, 2011.

5. B. He, K. Yang, S. Zhao and Y. Wang, "Underwater simultaneous localization and mapping based on EKF and point features," in International Conference on Mechatronics and Automation, pp. 4845 - 4850, August 9-12, Changchun, China, 2009.

6. B. Kuo, H. Chang, Y. Chen and S. Huang, "A light-and-fast SLAM algorithm for robots in indoor environments using line segment map," in Journal of Robotics, Article ID 257852, 12 pages, doi:10.1155/2011/257852, 2011.

7. P. Newman, D. Cole and K. Ho, "Outdoor SLAM using visual appearance and laser ranging," in IEEE International Conference on Robotics and Automation, pp. 1180-1187, May 15-19, Orlando, USA, 2006.

8. A. Georgiev and P. Allen, "Localization methods for a mobile robot in urban environments," IEEE Transactions on Robotics 20(5), pp. 851-864, 2004.

9. R. Kümmerle, B. Steder, C. Dornhege, A. Kleiner, G. Grisetti and W. Burgard, "Large scale graph-based SLAM using aerial images as prior information," Journal of Autonomous Robots 30(1), pp. 25-39, 2011.

10. G. Vosselman and H.-G. Maas, Airborne and Terrestrial Laser Scanning, Scotland, UK: Whittles Publishing, 2010.

11. A. Nüchter, "Parallelization of scan matching for robotic 3D mapping," in 3rd European Conference on Mobile Robots, September 19-21, Freiburg, Germany, 2007.

12. Y. Lin, J. Hyyppä and A. Kukko, "Stop-and-go mode: sensor manipulation as essential as sensor development in terrestrial laser scanning," Sensors 13(7), pp. 8140-8154, 2013.

13. B. Ferris, D. Fox and N. Lawrence, "WiFi-SLAM using Gaussian process latent variable models," in 20th International Joint Conference on Artificial Intelligence, pp. 2480-2485, January 6-12, Hyderabad, India, 2007.

14. B. Des Bouvrie, "Improving rgbd indoor mapping with imu data," Master's Thesis, Delft University of Technology, 2011.

15. H. Strasdat, J. Montiel and A. Davison, "Real-time monocular SLAM: why filter?," in IEEE International Conference on Robotics and Automation, pp. 2657-2664, May 3-7, Anchorage, Alaska, 2010.

16. M. Tomono, "Robust 3D SLAM with a stereo camera based on an edge-point ICP algorithm," in IEEE International Conference on Robotics and Automation, pp. 4306-4311, May 12-17, Kobe, Japan, 2009.

17. M. Pfennigbauer, P. Rieger, N. Studnicka and A. Ullrich, "Detection of concealed objects with a mobile laser scanning system," in SPIE Vol. 7323, Laser Radar Technology and Applications XIV, 732308, May 2, Orlando, USA, 2009.

18. S. May, S. Fuchs, D. Droeschel, D. Holz and A. Nüchter, "Robust 3D-mapping with time-of-flight cameras," in IEEE/RSJ International Conference on Intelligent Robots and Systems, pp. 1673-1678, October 10-15, St. Louis, USA, 2009.

19. T. Luhmann, S. Robson, S. Kyle and I. Harley, Close range photogrammetry: principles, techniques and applications, Caithness, UK: Whittles Publishing, 2006.

20. R. Mishra and Y. Zhang, "A review of optical imagery and airborne LiDAR data registration methods," The Open Remote Sensing Journal Vol. 5, pp. 54-63, 2012.

21. T. Kurz, S. Buckley, D. Schneider, A. Sima and J. Howell, "Ground-based hyperspectral and lidar scanning: a complementary method for geoscience research," in Proceedings of International Association for Mathematical Geosciences, September 5-9, Salzburg, Austria, 2011.

22. J. Han, L. Shao, D. Xu and J. Shotton, "Enhanced computer vision with Microsoft Kinect sensor: a review," IEEE Transactions on Cybernetics 43 (5), pp. 1318-1334, 2013.

23. S. Zug, F. Penzlin, A. Dietrich, T. Nguyen and S. Albert, "Are laser scanners replaceable by Kinect sensors in robotic applications?," IEEE International Symposium on Robotic and Sensors Environments (ROSE), Nov. 16-18, Magdeburg, Germany, pp. 144-149, 2012.

24. S. Meister, P. Kohli, S. Izadi, M. Hämmerle, C. Rother and D. Kondermann, "When can we use KinectFusion for ground truth acquisition?," Proceedings of IEEE/RSJ International Conference on Intelligent Robots and Systems, Workshop on Color-Depth Camera Fusion in Robotics, 2012.

25. J. Smisek, M. Jancosek and T. Pajdla, "3D with kinect," in Consumer Depth Cameras for Computer Vision, Barcelona, Spain, November 12, 2011.

26. T. Stoyanov, A. Louloudi, H. Andreasson and A. Lilienthal, "Comparative evaluation of range sensor accuracy for indoor environments," in Proceedings of European Conference on Mobile Robots, pp. 19-24, September 7-9, Örebro, Sweden, 2011.

27. J. Chow, K. Ang, D. Lichti and W. Teskey, "Performance analysis of a low-cost triangulation-based 3D camera: Microsoft Kinect system," in International Society of Photogrammetry and Remote Sensing Congress, Melbourne, Australia, 2012.

28. V. Renaudin, O. Yalak, P. Tomé and B. Merminod, "Indoor navigation of emergency agents," European Journal of Navigation vol. 5, pp. 36-45, 2007.

29. C. Ghilani and P. Wolf, Elementary Surveying: An Introduction to Geomatics 12 Edition, Prentice Hall, 2008.

30. G. Grisetti, C. Stachniss and W. Burgard, "Improving grid-based SLAM with with Rao-Blackwellized particle filters by adaptive proposals and selective resampling," in IEEE International Conference on Robotics and Automation, pp. 2432-2437, April 18-22, Barcelona, Spain, 2005.

31. F. Gustafsson, Statistical sensor fusion 2nd edition, Studentlitteratur AB, 2010.

32. D. Scaramuzza and F. Fraundorfer, "Visual odometry part I: the first 30 years and fundamentals," IEEE Robotics and Automation Magazine 18(4), pp. 80-92, 2011.

33. P. Besl and N. McKay, "A method for registration of 3D shapes," IEEE Transactions on Pattern Analysis and Machine Intelligence 14(2), pp. 239-256, 1992.

34. B. Horn, "Closed-form solution of absolute orientation using unit quaternions," J. Opt. Soc. Amer. A, 4(4), pp. 629-642, 1987.

35. Y. Chen and G. Medioni, "Object modelling by registration of multiple range images," Image and Vision Computing 10(3), pp. 145-155, 1992.

36. S. Rusinkiewicz and M. Levoy, "Efficient variants of the ICP algorithm," in 3rd International Conference on 3-D Digital Imagining and Modeling, pp. 145-152, May 28 - June 1, Quebec City, Canada, 2001.

37. G. Klein and D. Murray, "Parallel tracking and mapping for small AR workspaces," in 6th IEEE and ACM International Symposium on Mixed and Augmented Reality, pp. 225-234, November 13-16, Nara, Japan, 2007.

38. G. Nützi, S. Weiss, D. Scaramuzza and R. Siegwart, "Fusion of IMU and vision for absolute scale estimation in monocular SLAM," Journal of Intelligent and Robotic Systems Vol. 61, pp. 287-299, 2011.

39. S. Scherer, D. Dube and A. Zell, "Using depth in visual simultanesous localisation and mapping," in IEEE International Conference on Robotics and Automation, May 14-18, St. Paul, USA, 2012.

40. P. Henry, M. Krainin, E. Herbst, X. Ren and F. D, "RGB-D mapping: using Kinect-style depth cameras for dense 3D modeling of indoor environments," International Journal of Robotics Research 31(5), pp. 647-663, 2012.

41. P. Henry, M. Krainin, E. Herbst, X. Ren and D. Fox, "RGB-D mapping: using depth cameras for dense 3d modeling of indoor environments," in RGB-D: Advanced Reasoning with Depth Cameras in Conjunction with Robotics Science and Systems, June 27, Zaragoza, Spain, 2010.

42. A. Johnson and S. Kang, "Registration and integration of textured 3D data," Image and Vision Computing Vol. 17, pp. 135-147, 1999.

43. S. Izadi, D. Kim, O. Hilliges, D. Molyneaux, R. Newcombe, P. Kohli, J. Shotton, S. Hodges, D. Freeman, A. Davison and A. Fitzgibbon, "KinectFusion: real-time 3D reconstruction and interaction using a moving depth camera," in 24th ACM Symposium on User Interface Software and Technology, pp. 559-568, Santa Barbara, USA, 2011.

44. B. Curless and M. Levoy, "A volumetric method for building complex odels from range images," in 23rd Annual Conference on Computer Graphics and Interactive Techniques - SIGGRAPH, pp. 303-312, August 4-9, New York, USA, 1996.

45. T. Whelan, J. McDonald, M. Kaess, M. Fallon, H. Johannsson and J. Leonard, "Kintinuous: spatially extended KinectFusion," in 3rd RSS Workshop on RGB-D: Advanced Reasoning with Depth Cameras, July 9-10, Sydney, Australia, 2012.

46. T. Whelan, H. Johannsson, M. Kaess, J. Leonard and J. McDonald, "Robust real-time visual odometry for dense RGB-D mapping," in IEEE International Conference on Robotics and Automation, pp. 5724-5731, May 6-10, Karlsruhe, Germany, 2013.

47. T. Whelan, M. Kaess, J. Leonard and J. McDonald, "Deformation-based loop closure for large scale dense RGB-D SLAM," in IEEE/RSJ International Conference on Intelligent Robots and Systems, November 3-8, Tokyo, Japan, 2013.

48. M. Keller, D. Lefloch, M. Lambers, S. Izadi, T. Weyrich and A. Kolb, "Real-time 3D reconstruction in dynamic scenes using point-based fusion," in International Conference on 3D Vision, pp. 1-8, June 29 - July 1, Seattle, USA, 2013.

49. K. Khoshelham, D. Dos Santos and G. Vosselman, "Generation and weighting of 3D point correspondences for improved registration of RGB-D data," in ISPRS Annals of the Photogrammetry and Remote Sensing and Spatial Information Sciences, Volume II-5/W2, November 11-13, Antalya, Turkey, 2013.

50. L. Li, E. Cheng and I. Burnett, "An iterated extended Kalman filter for 3D mapping via Kinect camera," in IEEE International Conference on Acoustics, Speech and Signal Processing, pp. 1773-1777, May 26-31, Vancouver, Canada, 2013.

51. F. Aghili, M. Kuryllo, G. Okouneva and D. McTavish, "Robust pose estimation of moving objects using laser camera data for autonomous rendezvous and docking," International Society of Photogrammetry and Remote Sensing Archives Volume XXXVIII-3/W8, pp. 253-258, 2009.

52. T. Hervier, S. Bonnabel and F. Goulette, "Accurate 3D maps from depth images and motion sensors via nonlinear Kalman filtering," in IEEE/RSJ International Conference on Intelligent Robots and Systems, pp. 5291-5297, October 7-12, Vilamoura, Portugal, 2012.

53. A. Burens, P. Grussenmeyer, S. Guillemin, L. Carozza, F. Lévêque and V. Mathé, "Methodological developments in 3D scanning and modelling of archaeological French heritage site: the bronze age painted cave of Les Fraux, Dordogne (France)," International Archives of the Photogrammetry, Remote Sensing and Spatial Information Sciences, Volume XL-5/W2, pp. pp. 131-135, 2013.

54. F. Chiabrando and A. Spanò, "Point clouds generation using TLS and dense-matching techniques. A test on approachable accuracies of different tools.," ISPRS Annals of the Photogrammetry, Remote Sensing and Spatial Information Sciences, Volume II-5/W1, pp. 67-72, 2013.

55. K. Khoshelham and S. Oude Elberink, "Accuracy and resolution of kinect depth data for indoor mapping applications," Sensors, vol. 12, pp. 1437-1454, 2012.

56. C. Dal Mutto, P. Zanuttigh and G. Cortelazzo, Time-of-Flight Cameras and Microsoft Kinect, Springer, 2013.

57. B. Kainz, S. Hauswiesner, G. Reitmayr, M. Steinberger, R. Grasset, L. Gruber, E. Veas, D. Kalkofen, H. Seichter and D. Schmalstieg, "OmniKinect: real-time dense volumetric data acquisition and applications," in Proceedings of 18th ACM Symposium on Virtual Reality Software and Technology, pp. 25-32, December 10-12, Toronto, Canada, 2012.

58. A. Butler, S. Izadi, O. Hilliges, D. Molyneaux, S. Hodges and D. Kim, "Shake'n'sense: reducing interference for overlapping structured light depth cameras," in Proceedings of the SIGCHI Conference on Human Factors in Computing Systems, pp. 1933-1936, May 5-10, Austin, USA, 2012.

59. P. Aggarwal, Z. Syed, A. Noureldin and N. El-Sheimy, MEMS-based integrated navigation, Norwood, USA: Artech House, 2010.

60. D. Herrera, J. Kannala and J. Heikkilä, "Joint depth and color camera calibration with distortion correction," IEEE Transactions on Pattern Analysis and Machine Intelligence 34(10), pp. 2058-2064, 2012.

61. J. Chow and D. Lichti, " Photogrammetric bundle adjustment with self-calibration of the PrimeSense 3D camera technology: Microsoft Kinect," IEEE Access 1(1), pp. 465-474, 2013.

62. D. Lichti, "Modelling, calibration and analysis of an AM-CW terrestrial laser scanner," ISPRS Journal of Photogrammetry and Remote Sensing 61 (5), pp. 307-324, 2007.

63. Y. Reshetyuk, "A unified approach to self-calibration of terrestrial laser scanners," ISPRS Journal of Photogrammetry and Remote Sensing 65 (5) , pp. 445-456, 2010.

64. J. Chow, D. Lichti, C. Glennie and P. Hartzell, "Improvements to and comparison of static terrestrial LiDAR self-calibration methods," Sensors 13(6), pp. 7224-7249, 2013.

65. J. Chow, D. Lichti and W. Teskey, "Accuracy assessment of the Faro Focus$^{3D}$ and Leica HDS6100 panoramic type terrestrial laser scanner through point-based and plane-based user self-calibration," in FIG Working Week 2012: Knowing to manage the territory, protect the environment, evaluate the cultural heritage, May 6-10, Rome, Italy, 2012.

66. K. Al-Manasir and C. Fraser, "Registration of terrestrial laser scanner data using imagery," The Photogrammetric Record 21(115), pp. 255-268, 2006.

67. J. Lobo and J. Dias, "Relative pose calibration vetween visual and inertial sensors," The International Journal of Robotics Research 26(6), pp. 561-575, 2007.

68. J. Hol, T. Schön and F. Gustafsson, "Modeling and calibration of inertial and vision sensors," The International Journal of Robotics Research 29(2-3), pp. 231-244, 2009.

69. F. Mirzaei and S. Roumeliotis, "A Kalman filter-based algorithm for IMU-camera calibration: observability analysis and performance evaluation," IEEE Transaction on Robotics 24(5), pp. 1143-1156, 2008.

70. D. Titterton and J. Weston, Strapdown inertial navigation technology, Stevenage, UK: IEE Radar, Sonar, Navigation and Avionics Series. Peter Peregrinus Ltd., 1997.

71. A. Gelb, Applied Optimal Estimation, The M.I.T. Press, 1974.

72. R. Steffen and C. Beder, "Recursive estimation with implicit constraints," in Proceedings of the 29th DAGM Conference on Pattern Recognition, pp. 194-203, September 12-14, Heidelberg, Germany, 2007.

73. R. Steffen, "A robust iterative Kalman filter based on implicit measurement equations," Photogrammetrie - Fernerkundung - Geoinformation 2013(4), pp. 323-332, 2013.

74. OpenNI, "OpenNI Programmer's Guide," OpenNI 2.0 API, [Online]. Available: http://www.openni.org/openni-programmers-guide/. [Accessed 16 December 2013].

75. M. Muja and D. Lowe, "Fast approximate nearest neighbors with automatic algorithm configuration," VISAAP (1), pp. 331-340, 2009.

76. J. Luck, C. Little and W. Hoff, "Registration of range data using a hybrid simulated annealing and iterative closest point algorithm," in IEEE International Conference on Robotics and Automation, pp. 3739-3733, April 24-28, San Francisco, USA, 2000.

77. G. Bradski and A. Kaehler, Learning OpenCV, Sabastopol, USA: O'Reilly Media, 2008.

78. Y. Zhong, "Intrinsic shape signatures: A shape descriptor for 3D object recognition," in IEEE 12th International Conference on Computer Vision , pp. 689-696, September 29 - October 2, Kyoto, Japan, 2009.

79. S. Filipe and L. Alexandre, "A comparative Evaluation of 3D keypoint detectors," in Proceeding of 9th Conference on Telecommunications, pp. 145-148, May 8-10, Castelo Branco, Portugal, 2013.

80. F. Tombari, S. Salti and L. Di Stefano, "A combined texture-shape descriptor for enhanced 3D feature matching," in 18th IEEE International Conference on Image Processing, pp. 809-812, September 11-14, Brussels, Belgium, 2011.

81. F. Fraundorfer and D. Scaramuzza, "Visual odometry part II: matching, robustness, optimization, and applications," IEEE Robotics and Automation Magazine 19(2), pp. 78-90, 2012.

82. J. Chow, A. Ebeling and W. Teskey, "Point-based and plane-based deformation monitoring of indoor environments using terrestrial laser scanners," Journal of Applied Geodesy 6(3-4), pp. 193-202, 2012.

83. H. Pfister, M. Zwicker, J. van Baar and M. Gross, "Surfels: surface elements as rendering primitives," in Proceedings of the 27th Annual Conference on Computer Graphics and Interactive Techniques, pp. 335-342, July 23-28, New Orleans, USA, 2000.

## 5.2 Contributions of Authors

The first author prepared the manuscript; acquired the datasets using the laser scanner, Kinects, and IMU; developed software for manipulating, processing, and integrating data from the sensors; formulated the modified ICP algorithm and 5-point VO algorithm; deciding on the choice of sensors; coming up with the hybrid system design; and testing the Scannect and analyzing its results. The second author provided valuable input to the manuscript. The third, fourth, and fifth authors assisted in the IMU processing and had valuable insights for the project.

# Chapter Six: Conclusions and Recommendations for Future Work

Over the years, technological advancements in LiDAR have made this instrument the preferred tool in 3D reconstruction surveys. Its ability to rapidly acquire a 3D point cloud in almost every direction with long ranges has secured its place as the leading technology for capturing dense geometric information. One of the bottlenecks in modern TLS is the registration workflow. Point cloud registration, the process of estimating the instrument's exterior orientation parameters, is fundamental for "stitching" local scans together to build a global 3D model with complete coverage. For terrestrial applications, this is typically achieved via indirect georeferencing (i.e. using control/tie points), which can be time-consuming in the field, requires large degrees of overlap, and an experienced user to design the geomatics network. In recent years, direct georeferencing (i.e. using additional onboard sensors) has gained a lot of interest in the surveying and mapping community because of its efficiency and simplification of the field operations. Specifically, under open sky conditions, positioning and navigation sensors such as GNSS and INS are commonly used to aid the registration process. However, this relies on the availability of accurate satellite positioning solutions, resulting in difficulties in urban canyons, underground, and indoors. Systems that can acquire 3D geospatial information inside buildings in an efficient, robust, and accurate manner for surveying applications is still lacking. Methods that can automatically and effectively position and orient a vast amount of 3D spatial images in a common coordinate system relative to each other is an active area of research and development. This thesis focused on designing an indoor direct georeferencing solution suitable for a 3D terrestrial laser scanner by integrating it with a low-cost MEMS inertial measurement unit, 2D cameras, and state-of-the-art 3D cameras. Development of such a system was further motivated by the huge demand for accurate indoor as-built surveys over large areas in a timely manner, which has applications in cultural heritage documentation, urban planning, structural health monitoring, forensics, motion capture, virtual reality, and location-based services.

Usually in terrestrial photogrammetry projects there exists a trade-off between selecting a slower but more accurate tripod-mounted static optical system and a faster but less accurate mobile kinematic system. The developed trolley-based mobile mapping system called the Scannect comprised both static and kinematic systems and is the first indoor adaptive hybrid stop-and-go/full-kinematic system. Its integrated architecture is unique and the equipped instruments have different failure modes, resulting in a more robust system. The balance between speed and accuracy can be continuously tuned external to the filtering algorithms; therefore it can be conveniently modified to suit the task. An important component of sensor integration is calibration, as residual systematic errors can exist even after the manufacturer's intricate calibration. For each sensor installed, significant residual systematic errors were compensated in the software prior to integration.

For terrestrial laser scanners, the existing multi-station point-based and plane-based self-calibration methods were studied and compared for their applicability in reducing systematic distortions. It was shown that with appropriate network designs, the more efficient plane-based calibration can deliver results comparable to the better explored point-based approach. The low-cost FARO Focus$^{3D}$ S laser scanner used was found to exhibit significant angular systematic errors (i.e. up to 2 arcmin) due to axes misalignments. After self-calibrations, on average the range, horizontal direction, and vertical angle measurement precisions of the Focus$^{3D}$ S were improved by 2.6%, 29.6%, and 1.1%, respectively. Improvements up to 60% in the horizontal direction had also been observed.

For the camera systems, the widely adopted gaming peripheral from Microsoft called the Kinect was utilized. Although this system was originally intended for recreation it has been demonstrated that a calibrated system has potential in many engineering applications. To further improve the Kinect sensor's accuracy, a novel, accurate, rigorous, and user-friendly total-system bundle adjustment calibration approach was proposed and demonstrated. The resulting point clouds had a metric scale and were smoother. Compared to the uncalibrated textured point clouds from the Kinect, post self-calibration the measurement accuracy was improved by 53%, resulting in RGB-D data with errors less than 4 mm.

Existing Kalman filter-based mapping systems tend to solve cloud-to-cloud matching and measurement update as a two-step process. In this research the point clouds from the Kinect were fused in a tightly-coupled Kalman filter framework using a new matching algorithm along with inertial measurements. This provides estimation of relative exterior orientation parameters between terrestrial laser scanner stations, and also removes the shadowed regions in the laser scanner point clouds. Such a system can function in long corridors with textureless walls, which are frequently encountered in indoor urban environments. When texture information becomes available and/or point clouds become unavailable, a new 5-point visual odometry algorithm was used to improve the state estimation in the Kalman filter automatically. This decoupling of the depth and colour information permits individual and joint processing while compensating for their possible time synchronization errors. Unlike most visual odometry algorithms that are solved either independent of the Kalman filter or by augmenting the state vector with many landmark positions, the proposed method in this research was tightly-coupled in the Kalman filter and solved for the triangulated 3D points implicitly. The Scannect can deliver 3D point clouds with an accuracy of 10 cm in each of the principal directions in office environments over a travelled distance of 120 m when only performing forward filtering. When detached from the Focus$^{3D}$ S and trolley, the Kinect and IMU alone can reconstruct small volumes (e.g. a flight of stairs) with a misclosure of 7 mm (from a handheld sequence).

This prototype of the Scannect can improve productivity for 3D reconstruction projects of indoor urban environments when compared to conventional static terrestrial laser scanning done in surveying. This is accomplished by using a mobile platform. By introducing Kinects and IMUs as additional sensors to aid the TLS registration process, the labour-intensive procedure of establishing control points was eliminated and the requirement for overlap between scans was reduced. Point clouds captured by both the Focus$^{3D}$ laser scanner and Kinect were automatically registered and fused together to efficiently build a 3D map of a large area with many occlusions. Although the results from Chapter 5 indicated that the accuracy of the system was at the decimeter level and was insufficient for most surveying applications, it had nevertheless demonstrated its potential to expedite the conventional workflow and may be sufficient for lower accuracy consumer applications. Future work will focus on improving the Scannect's accuracy, robustness, and speed. To achieve these goals, the following recommendations can be made:

- The 3D camera world is rapidly developing; newer sensors such as the Microsoft Kinect 2 can be considered. The same applies to MEMS IMUs and laser scanners – more accurate and newer sensors can be considered.
- Additional navaids, such as odometers and magnetometers, can be integrated.
- Perform backward smoothing and/or global optimization to ensure loop-closure, and thus improve the overall consistency.
- Use more rigid mounts to increase the Scannect's stability.
- Achieve real-time performance by implementing the Kinect's registration algorithms on the GPU.
- Replace the SIFT operator with faster and/or more accurate operators.
- Conduct more thorough testing in various indoor environments, especially in scenes with moving objects and people. For robust SLAM, techniques are necessary to isolate the dynamics in the scene and reconstruct only the static portions.
- Replace the trolley platform with a self-controlled robot for automatic exploration and mapping.

## REFERENCES

[1] World Health Organizatoin, "Urban population growth," Global Health Observatory (GHO), [Online]. Available: http://www.who.int/gho/urban_health/situation_trends/urban_population_growth_text/en/. [Accessed 26 January 2014].

[2] Henry Davis Consulting, "The Earliest Known Map," [Online]. Available: http://www.henry-davis.com/MAPS/Ancient%20Web%20Pages/100mono.html. [Accessed 26 January 2014].

[3] A. Georgiev and P. Allen, "Localization methods for a mobile robot in urban environments," *IEEE Transactions on Robotics 20(5),* pp. 851-864, 2004.

[4] P. Allen, I. Stamos, A. Gueorguiev, E. Gold and P. Blaer, "AVENUE: Automated site modeling in urban environments," Quebec City, Canada, 2001.

[5] Applanix, "Trimble," 2011. [Online]. Available: http://www.trimble.com/Indoor-Mobile-Mapping-Solution/pdf/TIMMSOverview.pdf. [Accessed 4 August 2011].

[6] S. El-Hakim, P. Boulanger, F. Blais and J. Beraldin, "A system for indoor 3D mapping and virtual environments," in *Proceedings of SPIE Videometrics V, Vol. 3174,* San Diego, CA, 1997.

[7] S. El-Hakim, C. Brenner and G. Roth, "A multi-sensor approach to creating accurate virtual environments," *International Journal of Photogrammetry and Remote Sensing,* pp. 379-391, 1998.

[8] P. Allen, I. Stamos, A. Troccoli, B. Smith, M. Leordeanu and Y. Hsu, "3D modeling of

historic sites using range and image data," *Proceedings of Internation Conference on Robotics and Automation*, pp. 11-16, 2003.

[9] Z. Yuan, "Plane-based 3D mapping for structured indoor environment," *PhD Thesis, Politecnico di Torino, 112 pages*, 2013.

[10] C. Wen, L. Qin, Q. Zhu, C. Wang and J. Li, "Three-dimensional indoor mobile mapping with fusion of two-dimensional laser scanner and RGB-D camera data," *IEEE Geoscience and Remote Sensing Letters 11 (4)*, pp. 843-847, 2014.

[11] M. Harris, "How new indoor navigation systems will protect emergency responders," IEEE Spectrum, [Online]. Available: http://spectrum.ieee.org/static/how-new-indoor-navigation-systems-will-protect-emergency-responders. [Accessed 26 January 2014].

[12] Google Maps, "What is indoor Google Maps," Google, [Online]. Available: http://maps.google.com/help/maps/indoormaps/. [Accessed 26 January 2014].

[13] S. Soudarissanane and R. Lindenbergh, "Optimizing terrestrial laser scanning measurement set-up," in *ISPRS Laser Scanning Workshop* , Calgary, Canada, 2011.

[14] X. Zhao, C. Goodall, Z. Syed, B. Wright and N. El-Sheimy, "Wi-Fi assisted multi-sensor personal navigation system for indoor environments," in *Proceedings of the 2010 International Technical Meeting of The Institute of Navigation*, San Diego, CA, January 2010, 2010.

[15] D. Vissière, A. Martin and N. Petit, "Using distributed magnetometers to increase IMU-based velocity estimation into perturbed area," in *46th IEEE Conference on Decision and Control*, New Orleans, LA. December 12-14, 2007.

[16] J. Hol, F. Dijkstra, H. Luinge and T. Schon, "Tightly coupled UWB/IMU pose esitmation," in *In IEEE International Conference on Ultra-Wideband,* 2009.

[17] F. Gielsdorf, A. Rietdorf and L. Gründig, "A concept for the calibration of terrestrial laser scanners," in *FIG Working Week*, Athens, Greece, May 22-27, 2004.

[18] T. Rabbani, S. Dijkman, F. van den Heuvel and G. Vosselman, "An integrated approach for modelling and global registration of point clouds," *ISPRS Journal of Photogrammetry and Remote Sensing 61(6),* pp. 355-370, 2007.

[19] P. Besl and N. McKay, "A method for registration of 3D shapes," *IEEE Transactions on Pattern Analysis and Machine Intelligence 14(2),* pp. 239-256, 1992.

[20] Y. Chen and G. Medioni, "Object modelling by registration of multiple range images," *Image and Vision Computing 10(3),* pp. 145-155, 1992.

[21] K. Al-Manasir and C. Fraser, "Registration of terrestrial laser scanner data using imagery," *The Photogrammetric Record 21(115),* pp. 255-268, 2006.

[22] E. Renaudin, A. Habib and A. Kersting, "Featured-based registration on terrestrial laser scans with minimum overlap using photogrammetric data," *ETRI Journal 33(4),* pp. 517-527, 2011.

[23] J. Talaya, R. Alamús, E. Bosch, A. Serra, W. Kornus and A. Baron, "Integration of a terrestrial laser scanner with GPS/IMU orientation sensors," *International Archives of Photogrammetry, Remote Sensing and Spatial Sciences Vol. 35,* 2004.

[24] T. Asai, M. Kanbara and N. Yokoya, "3D modeling of outdoor environments by integrating omnidirectional range and color images," in *Proceedings of the Fifth International Conference on 3-D Digital Imaging and Modelling,* Ottawa, Canada, 2005.

[25] A. Habib, A. Kersting and K. Bang, "Comparative Analysis of Different Approaches for the incorporation of position and orientation information in integrated sensor orientation procedures," *International Society for Photogrammetry and Remote Sensing Archives Vol. XXXVIII - Part 1,* 2010.

[26] V. Renaudin, O. Yalak, P. Tomé and B. Merminod, "Indoor navigation of emergency agents," *European Journal of Navigation vol. 5,* pp. 36-45, 2007.

[27] H. Kim, D. Kim, S. Yang, Y. Son and S. Han, "An indoor visible light communication positioning system using a RF carrier allocation technique," *Journal of Lightwave Technology 31(1),* pp. 134-144, 2013.

[28] Y. Gu, A. Lo and I. Niemegeers, "A survey of indoor positioning systems for wireless personal networks," *IEEE Communications Surveys & Tutorials 11(1),* pp. 13-32, 2009.

[29] Riegl Mobile Scanning, "RIEGL VMX-450," RIEGL Laser Measurement Systems GmbH, [Online]. Available: http://www.riegl.com/nc/products/mobile-scanning/produktdetail/product/scannersystem/10/. [Accessed 26 January 2014].

[30] Topcon, "IP-S2 Compact + Integrated High Density Mobile Mapping System," Topcon Positioning Systems, Inc., [Online]. Available: http://www.topconpositioning.com/products/mobile-mapping/ip-s2-compact. [Accessed 26 January 2014].

[31] Google Street View, "Behind the Scenes of Google Street View," [Online]. Available: http://www.google.ca/maps/about/behind-the-scenes/streetview/. [Accessed 26 January 2014].

[32] R. Staiger, "Terrestrial laser scanning - Technology, systems and applications," in *In: Proceedings of 2nd FIG Regional Conference*, Marrakech, Morocco, 2-5 December. http://www.fig.net/pub/morocco/index.htm (accessed 02.02.09), 2003.

[33] Trimble, "Trimble GX," Trimble Navigation Limited, [Online]. Available: http://www.trimble.com/trimblegx.shtml. [Accessed 16 March 2010].

[34] Zoller + Fröhlich, "Z+F IMAGER 5010C," Zoller + Fröhlich GmbH, [Online]. Available: http://www.zf-laser.com/Z-F-IMAGER-5010C-3D.135.0.html?&L=1. [Accessed 26 January 2014].

[35] Konica Minolta, "3-D Hardware and Support," Konica Minolta Sensing Singapore Pte Ltd, [Online]. Available: http://sensing.konicaminolta.asia/search-by-services/3-d-hardware-and-support/. [Accessed 27 January 2014].

[36] Optech, "ILRIS-3D Intelligent Laser Ranging and Imaging System," Optech Incorporated: Industrial & 3D Imaging, [Online]. Available: http://www.optech.ca/i3dprodline-ilris3d.htm. [Accessed 27 January 2014].

[37] D. Lichti, "ENGO699.15 Optical Imaging Metrology Course Notes," *Geomatics Engineering, University of Calgary,* 2009.

[38] Riegl Terrestrial Scanning, "RIEGL VZ-1000," RIEGL Laser Measurement Systems GmbH, [Online]. Available: http://www.riegl.com/nc/products/terrestrial-

scanning/produktdetail/product/scanner/27/. [Accessed 27 January 2014].

[39] Trimble, "Trimble TX8," Trimble Navigation Limited, [Online]. Available: http://www.trimble.com/3d-laser-scanning/tx8.aspx?dtID=overview&. [Accessed 27 January 2014].

[40] D. Lichti, M. Stewart, M. Tsakiri and A. Snow, "Calibration and testing of a terrestrial laser scanner," *The International Archives of the Photogrammetry, Remote Sensing and Spatial Information Sciences 33 (part B5/2)*, pp. 485-492, 2000.

[41] D. Lichti, S. Gordon, M. Stewart, J. Franke and M. Tsakiri, "Comparison of digital photogrammetry and laser scanning," in *CIPA WG 6 International Workshop on Scanning Cultural Heritage Recording*, Corfu, Greece: 1-2 September 2002, 2002.

[42] W. Böhler, M. Bordas Vicent and A. Marbs, "Investigating laser scanner accuracy," *The International Archives of the Photogrammetry, Remote Sensing and Spatial Information Sciences 34 (Part5/C15)* , pp. 696-701, 2003.

[43] T. Kersten, H. Sternberg, K. Mechelke and C. Acevedo Pardo, "Terrestrial laser scanning system Mensi GS100/GS200 - Accuracy tests, experiences and projects at the Hamburg University of Applied Sciences," *ISPRS, Vol. XXXIV, Part 5/W16, www.tu-dresden.de/fghfipf/photo/PanoramicPhotogrammetryWorkshop2004/Proceedings.htm,* 2004.

[44] T. Kersten, H. Sternberg, K. Mechelke and C. Acevedo pardo, "Investigations into the Accuracy Behaviour of the Terrestrial Laser Scanning System Mensi GS100," *Optical 3-D Measurement Techniques VII, Gruen & Kahmen (Eds.), Vol. I,* pp. 122-131, 2005.

[45] D. Lichti, J. Chow and H. Lahamy, "Parameter de-correlation and model-identification in hybrid-style terrestrial laser scanner self-calibration," *ISPRS Journal of Photogrammetry and Remote Sensing, 66 (3)*, pp. 317-326, 2011.

[46] D. Lichti, " Modelling, calibration and analysis of an AM-CW terrestrial laser scanner," *ISPRS Journal of Photogrammetry and Remote Sensing 61 (5)*, pp. 307-324, 2007.

[47] Leica Geosystems, "Leica ScanStation P20," Leica Geosystems, [Online]. Available: http://www.leica-geosystems.com/en/Leica-ScanStation-P20_101869.htm. [Accessed 27 January 2014].

[48] Faro, "FARO Focus3D S," FARO, [Online]. Available: http://www.faro.com/en-us/products/3d-surveying/faro-focus3d/overview. [Accessed 27 January 2014].

[49] Panasonic, "D-IMager," Panasonic Electric Works Corporation of America, [Online]. Available: http://pewa.panasonic.com/components/built-in-sensors/3d-image-sensors/d-imager/. [Accessed 27 January 2014].

[50] Mesa, "SwissRanger™ SR4500," MESA Imaging, [Online]. Available: http://www.mesa-imaging.ch/swissranger4500.php. [Accessed 27 January 2014].

[51] SoftKinetic, "DepthSense 325 for Hand & Finger Tracking," SoftKinetic, [Online]. Available: www.softkinetic.com/products/depthsensecameras.aspx. [Accessed 27 January 2014].

[52] PMDTechnologies, "PMD Reference Design pico," PMDTechnologies gmbh, [Online]. Available: http://www.pmdtec.com/products_services/reference_design_pico_pico_s.php. [Accessed 27 January 2014].

[53] Microsoft Xbox One, "Xbox One: Innovation," [Online]. Available: http://www.xbox.com/en-CA/xbox-one/innovation. [Accessed 27 January 2014].

[54] Microsoft Xbox 360, "Kinect," [Online]. Available: http://www.xbox.com/en-CA/Kinect. [Accessed 26 January 2014].

[55] C. Dal Mutto, P. Zanuttigh and G. Cortelazzo, Time-of-Flight Cameras and Microsoft Kinect, Springer, 2013.

[56] J. Boehm and T. Pattinson, "Accuracy of exterior orientation for a range camera," *The International Archives of the Photogrammery, Remote Sensing and Spatial Information Sciences 38 (Part 5),* pp. 103-108, 2010.

[57] J. Biswas and M. Veloso, "Depth camera based localization and navigation for indoor mobile robots," in *RGB-D: Advanced Reasoning with Depth Cameras*, Washington, Seattle, 2011.

[58] C. Herrera, J. Kannala and J. Heikkilä, "Accurate and practical calibration of a depth and color camera pair," in *14th International Conference on Computer Analysis of Images and Patterns*, Seville, Spain, 2011.

[59] F. Menna, F. Remondino, R. Battisti and E. Nocerino, "Geometric investigation of a gaming active device," in *Videometrics, Range Imaging, and Applications XI*, Munich, Germany, 2011.

[60] K. Khoshelham, "Accuracy analysis of kinect depth data," in *ISPRS Laser Scanning Workshop*, Calgary, Canada, 2011.

[61] S. Godha, "Performance evaluation of low cost MEMS-based IMU integrated with GPS for land vehicle navigation application," *MSC Thesis, Department of Geomatics Engineering, University of Calgary, Canada, UCGE Report No. 20239*, 2006.

[62] Xsens, "MTi," [Online]. Available: http://www.xsens.com/en/general/mti. [Accessed 26 January 2014].

[63] E. Shin and N. El-Sheimy, "A New Calibration Method for Strapdown Inertial Navigation Systems," *Zeitschrift fűr Vermessungswesen Journal 127 (1),* pp. 1-10, 2002.

[64] T. Bailey and H. Durrant-Whyte, "Simultaneous localisation and mapping (SLAM): Part II state of the art," *IEEE Robotics and Automation Magazine 13(3),* pp. 108-117, 2006.

[65] P. Smith and R. Cheeseman, "On the representation and estimation of spatial uncertainty," *International Journal of Robotics Research,* pp. 85-94, 1986.

[66] K. Chen and W. Tsai, "Vision-based autonomous vehicle guidance for indoor security patrolling by a SIFT-based vehicle-localization technique," *IEEE Transaction Vehicular Technology 59(7),* pp. 3261-3271, 2010.

[67] T. Okuma, K. Sakaue, H. Takemura and N. Yokoya, "Real-time camera parameter estimation from images for a mixed reality system," in *Proc. 15th International Conference on Pattern Recognition, vol. 4,* Barcelona, Spain, 2004.

[68] C. Wu and W. Tsai, "Location estimation for indoor autonomous vehicle navigation by omni-directional vision using circular landmarks on ceilings," *Robotics and Autonomous Systems 57(5),* pp. 546-555, 2009.

[69] N. Ayache and O. Faugeras, "Building, registrating, and fusing noisy visual maps," *International Journal of Robotics Research, 7(6),* pp. 45-65, 1988.

[70] F. Lu and E. Milios, "Globally consistent range scan alignment for environment mapping," *Autonomous Robots, vol. 4,* pp. 333-349, 1997.

[71] J. Gutmann and K. Konolige, "Incremental mapping of large cyclic environments," in *International Symposium on Computational Intelligence in Robotics and Automation*, Monterey, CA, November, 1999.

[72] S. Thrun, "A probabilistic on-line mapping algorithm for teams of mobile robots," *International Journal of Robotics Research 20(5),* pp. 335-363, 2001.

[73] N. Engelhard, F. Endres, J. Hess, J. Sturm and W. Burgard, "Real-time 3D visual SLAM with a hand-held RGB-D camera," in *RGB-D Workshop on 3D Perception in Robotics*, Västerås, Sweden, 2011.

[74] J. Cunha, E. Pedrosa, C. Cruz, A. Neves and N. Lau, "Using a depth camera for indoor robot localization and navigation," in *RGB-D: Advanced Reasoning with Depth Cameras*, Washington, Seattle, 2011.

[75] P. Piniés, T. Lupton, S. Sukkarieh and J. Tardos, "Inertial aiding of inverse depth SLAM using a monocular camera," in *Robotics and Automation, 2007 IEEE International Conference*, Rome, Italy, 2007.

[76] A. Gelb, Applied Optimal Estimation, The M.I.T. Press, 1974.

[77] R. Smith, M. Self and P. Cheeseman, "Estimating uncertain spatial relationships in robotics," *Autonomous Robot Vehnicles, Springer*, 1990.

[78] P. Aggarwal, "Hybrid extended particle filter (HEPF) for INS/GPS integrated system," *ION GNSS 2008. Institute of Navigation, Savannah, GA*, pp. 1600-1609, 2008.

## APPENDIX A: Study of the Microsoft Kinect's Behaviour and Suitability for 3D Reconstruction Applications in Geomatics.

### Article: Performance Analysis of a Low-Cost Triangulation-Based 3D Camera: Microsoft Kinect System

J.C.K. Chow*, K.D. Ang, D.D. Lichti, and W.F. Teskey

Department of Geomatics Engineering, University of Calgary, 2500 University Dr NW, Calgary, Alberta, T2N 1N4, Canada (jckchow, kdang, ddlichti, and wteskey)@ucalgary.ca

**ABSTRACT:**

Recent technological advancements have made active imaging sensors popular for 3D modelling and motion tracking. The 3D coordinates of signalised targets are traditionally estimated by matching conjugate points in overlapping images. Current 3D cameras can acquire point clouds at video frame rates from a single exposure station. In the area of 3D cameras, Microsoft and PrimeSense have collaborated and developed an active 3D camera based on the triangulation principle, known as the Kinect system. This off-the-shelf system costs less than $150 USD and has drawn a lot of attention from the robotics, computer vision, and photogrammetry disciplines. In this paper, the prospect of using the Kinect system for precise engineering applications was evaluated. The geometric quality of the Kinect system as a function of the scene (i.e. variation of depth, ambient light conditions, incidence angle, and object reflectivity) and the sensor (i.e.

* Corresponding author.

warm-up time and distance averaging) were analysed quantitatively. This system's potential in human body measurements was tested against a laser scanner and 3D range camera. A new calibration model for simultaneously determining the exterior orientation parameters, interior orientation parameters, boresight angles, leverarm, and object space features parameters was developed and the effectiveness of this calibration approach was explored.

## 1. INTRODUCTION

Passive photogrammetric systems have the disadvantage of high computation expense and require multiple cameras when measuring a dynamic scene. A single camera can be used when reconstructing a static scene, but more than one exposure station is needed. Terrestrial laser scanners are an attractive alternative for 3D measurements of static objects. These instruments can yield geospatial measurements at millimetre-level accuracy, at over a million points per second. Most importantly, they can determine 3D positions on textureless surfaces with no topography from a single scan. However, scanners cannot be used for measuring moving objects due to the scan time delay. With the aforementioned limitations, the development of 3D range cameras began (Lange & Seitz, 2001). Time-of-flight (TOF) 3D cameras measure the range between the sensor and the object space at every pixel location in real time with almost no dependency on background illumination and surface texture. Pulse-based range cameras can measure longer ranges but the distance accuracy is dependent on the clock accuracy (e.g. DragonEye). Although stronger laser pulses can be used for distance measurement, the frame rate of the sensor is limited due to the difficulty in generating short pulses with fast rise and fall time (Shan & Toth, 2008). This low data acquisition rate and requirement for expensive clocks can theoretically be mitigated by using AM-CW light (e.g. PMD, DepthSense, Fotonic,

SwissRanger, and D-Imager). However, in reality a high integration time is required to reduce noise, which limits the frame rate for real-time applications (Kahlmann et al., 2006). One of the most important errors for phase-based 3D cameras stems from the internal scattering effect, where distance observations made to the background objects are biased by the strong signals returned by the foreground objects. This scene-dependent error can be difficult to model and is one of the main limiting factors for using this category of cameras in many applications (Mure-Dubois & Hugli., 2007). In addition, the 3D range cameras currently on the market are still relatively expensive and have low pixel resolution.

On the other hand, the Kinect is a structured light (or coded light) system where the depth is measured based on the triangulation principle. The Kinect consists of three optical sensors: an infrared (IR) camera, IR projector, and RGB camera. The projector emits a pseudo-random pattern of speckles in the IR light spectrum. This known pattern is imaged by the IR camera and compared to the reference pattern at known distances stored in its memory. Through a 9x9 spatial correlation window, a disparity value is computed at every pixel location which is proportional to a change in depth. The Kinect emits speckles of 3 different sizes to accommodate for objects appearing at different depths and according to the manufacturer it produces a measurable range of 1.2m – 3.5m. More details about the inner mechanics of the Kinect can be found in (Freedman et al., 2010; and Konolige & Mihelich, 2010).

Before commencing user self-calibration of the Kinect, the off-the-shelf sensor error behaviour is tested and the results are summarized in Section 2. The performances of the Kinect for 3D object space reconstruction (more specifically for measuring the human body) is evaluated in Section 3. This system is not free of distortions; noticeable data artefacts and systematic errors in the Kinect have been reported in literature (Menna et al., 2011). The proposed mathematical model for performing self-calibration of the Kinect is presented in Section 4, followed by some empirical results illustrating the effectiveness of the new calibration method in Section 5. Please note, in the Preliminary Tests (Section 2) the Kinect data is acquired using either the Microsoft Kinect SDK Beta 1 (sub-headings marked with a 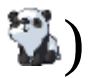) or software named Brekel Kinect (Brekelmans, 2012) (sub-headings marked with a 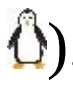). Thereafter, all Kinect data in subsequent sections is acquired using the Microsoft Kinect SDK Beta 1 (Microsoft, 2012) unless otherwise stated. Note, depending on the drivers and libraries used for data capture, different raw outputs are streamed by the Kinect.

## 2. PRELIMINARY TEST RESULTS

An out-of-the-box Kinect system was tested for its distance measurement capabilities. A series of tests were carried out in an indoor environment and the results are presented below.

### 2.1 Warm-up Test

The distance measurement quality of 3D range cameras (Chiabrando et al., 2009) and laser scanners (Glennie & Lichti, 2011) have been shown to be affected by the warm-up time. In a 23.2°C, 883.9mb, and 36.8% humidity room, a white Spectralon target located approximately 1.1m from the Kinect was observed every 5 minutes over a period of 2 hours. The flat target was nominally orthogonal to the Kinect and a plane was fitted to the point cloud (Figure 1). During the first hour of warming up, the estimated normal distance to the best-fit plane changed by 1 cm. Based on Figure 2, it is advisable to warm-up the Kinect for at least 60 minutes prior to data capture. For all the results shown in this paper, the Kinect was turned on at least 90 minutes prior to data capture.

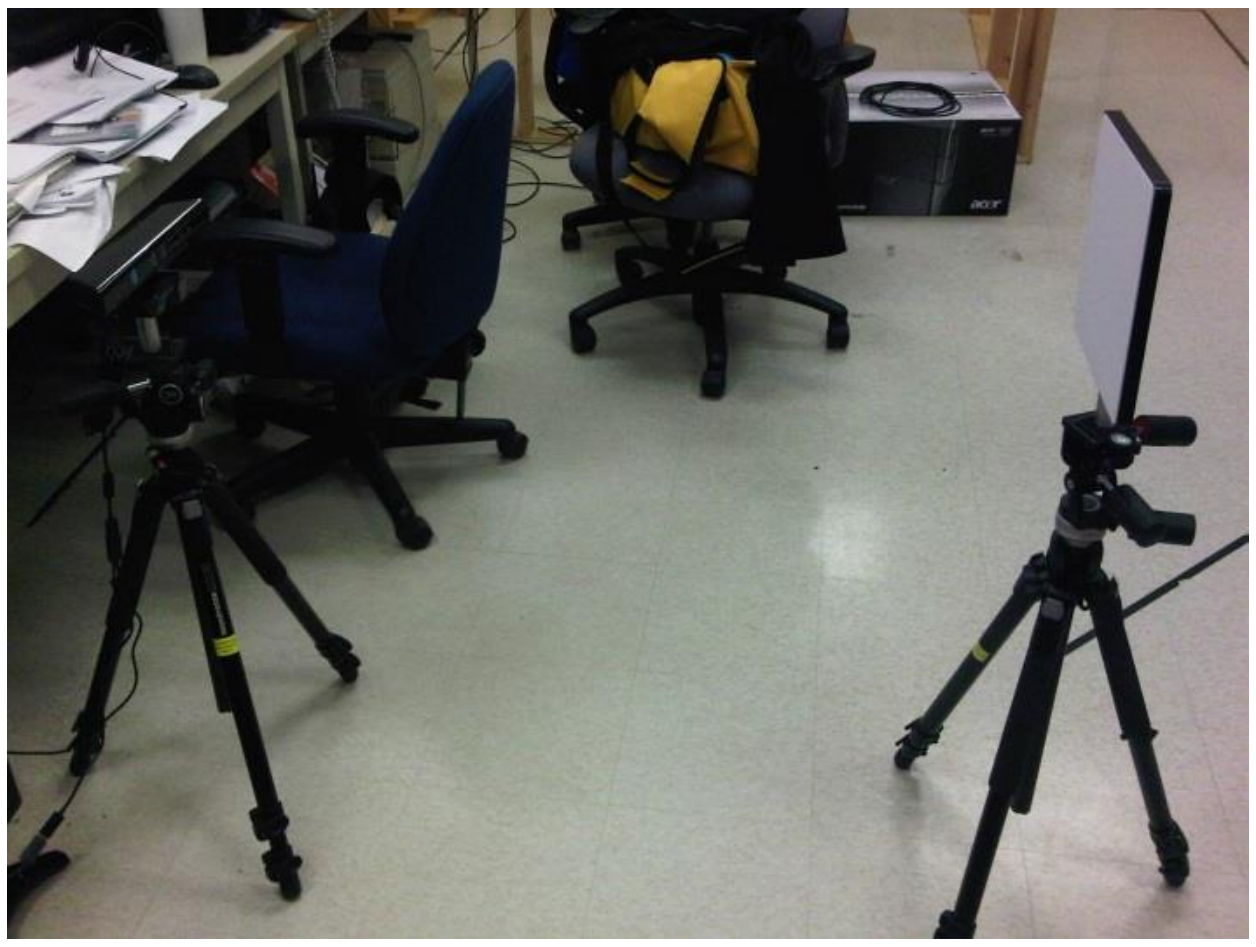

*Figure 1: Data capture of a white planar Spectralon target using the Microsoft Kinect*

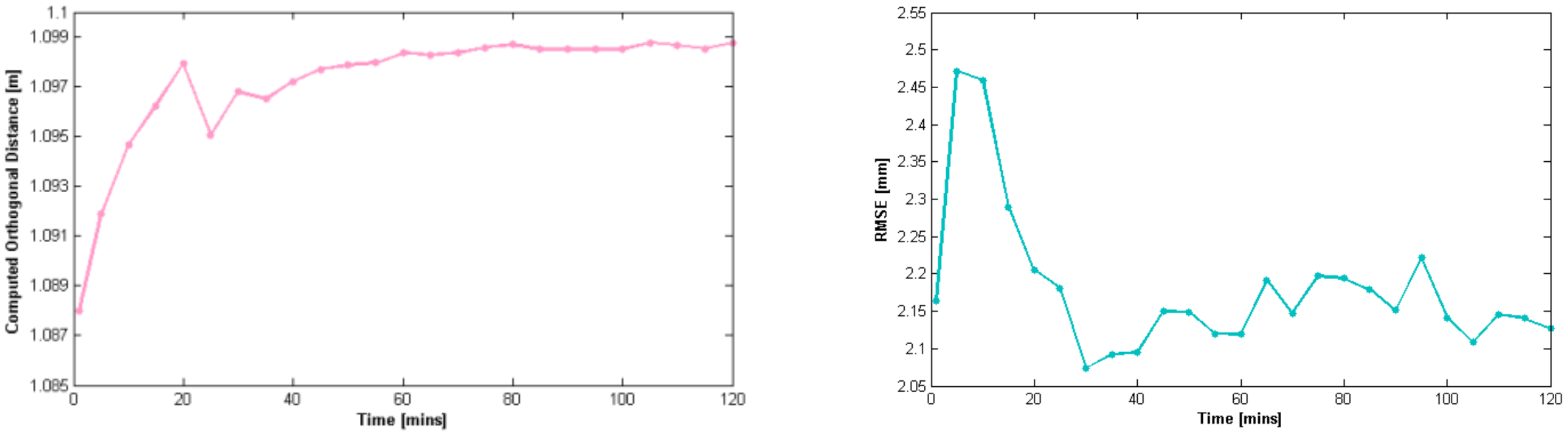

*Figure 2: (a) Orthogonal distance to best-fit plane as a function of warm-up time. (b) Depth measurement noise as a function of warm-up time.*

## 2.2 Ambient Light Test

A flat wall was imaged by the Kinect with the room's fluorescent lights turned on and off. This test was repeated with and without a strong light from a desk lamp illuminating the wall. The measured point cloud of the wall with and without white light illumination is shown in Figure 3. The depth measurements appear to be fairly robust against changes in the environment's lighting condition as demonstrated by the similarity between the computed least-squares best-fit plane parameters.

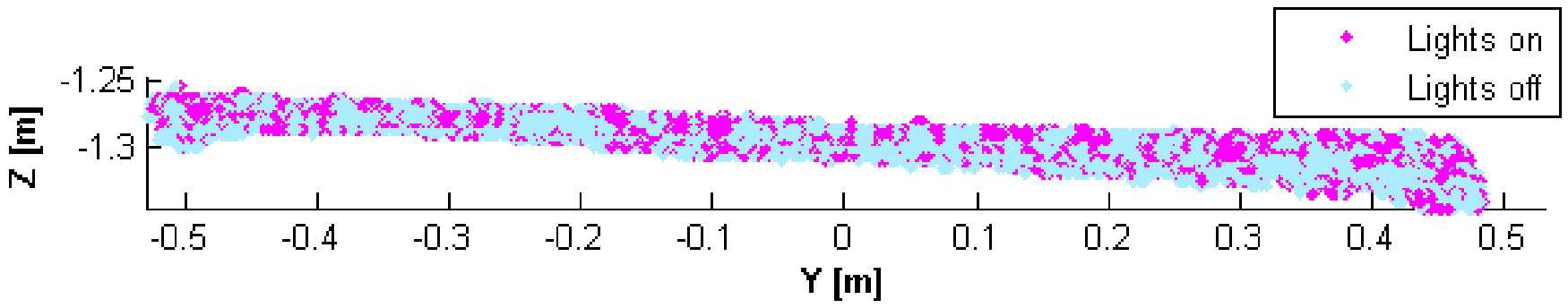

*Figure 3: Top view of the point cloud of a wall with the background lights turned on and off.*

## 2.3 Incidence Angle Test

In Soudarissanane et al. (2011) it has been reported that the noise of TOF distance measurements made with a laser escalates with increasing incidence angle. In a similar setup as shown in Figure 1, the Spectralon target was rotated about the vertical axis. The direction of the plane's normal was changed from nominally parallel to the optical axis of the camera to nearly orthogonal. The RMSE of the plane fitting as a function of the incidence angle is reported in Figure 4. Despite the fact that points are projected onto the scene and at large incidence angle they are elongated, through the correlation window process the effect of the incidence angle on the depth measurement precision appears to be small.

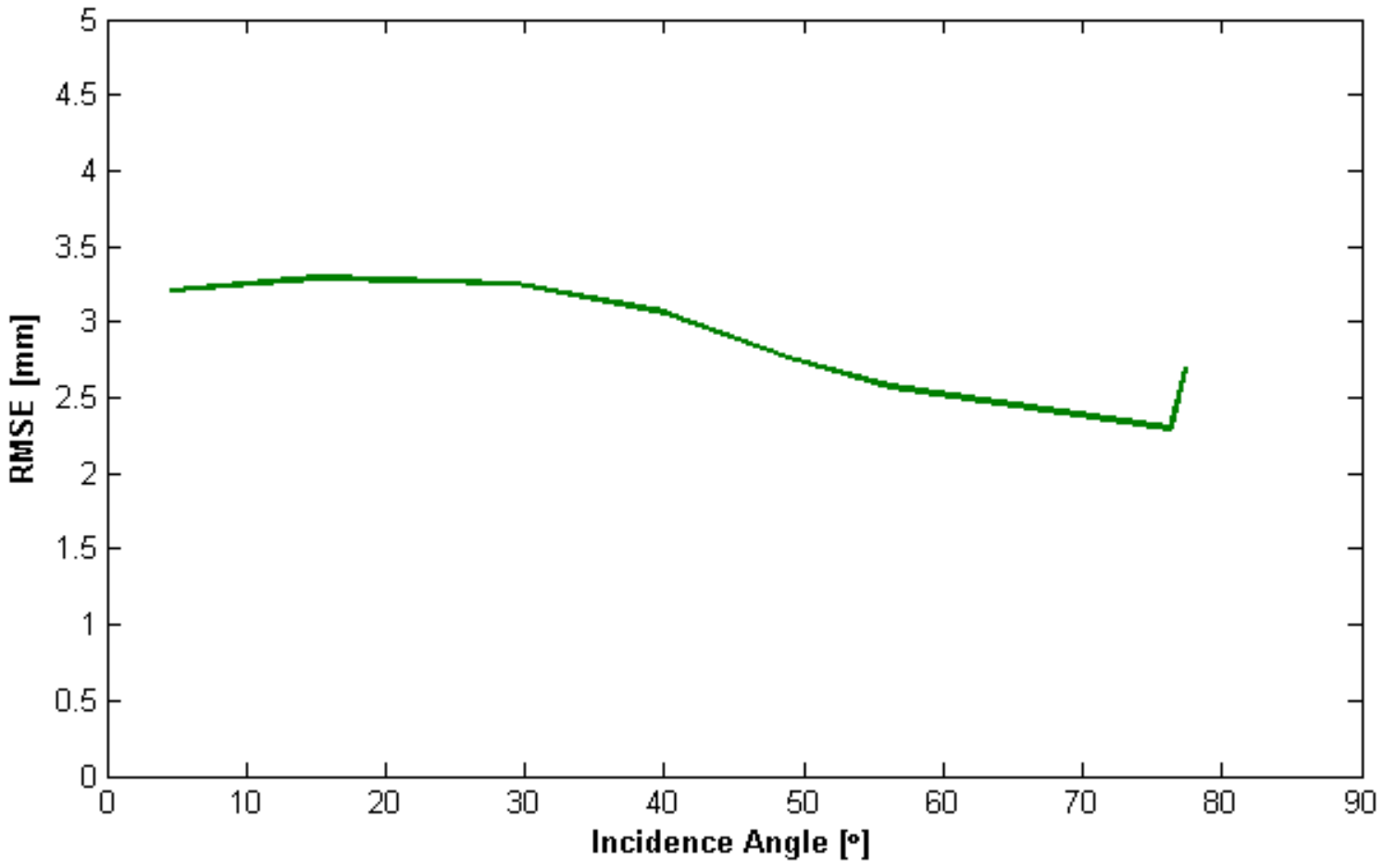

*Figure 4: RMSE of plane fitting to planar Spectralon target as a function of the incidence angle.*

### 2.4 Radiometric Influence

TOF range measurements by laser scanners have shown dependency on the surface colour (Hanke et al, 2006), i.e. a black surface reflects less energy than a white surface, which results in a range bias. To determine if the Kinect's depth sensor can measure ranges independent of the surface texture, a white metallic plate with a black circle printed in the centre was imaged with the Kinect located at 1 m distance up to 3 m at increments of 0.5 m. The RGB image of the target and the reconstructed point cloud using the depth sensor of the Kinect is shown in Figure 5. The colours in Figure 5b depict the range to the target, and no significant discrepancies to the range measurements can be observed between the black and white regions. The RMSE of the plane fitting at various distances is comparable to the results from imaging a white wall (Figure 10).

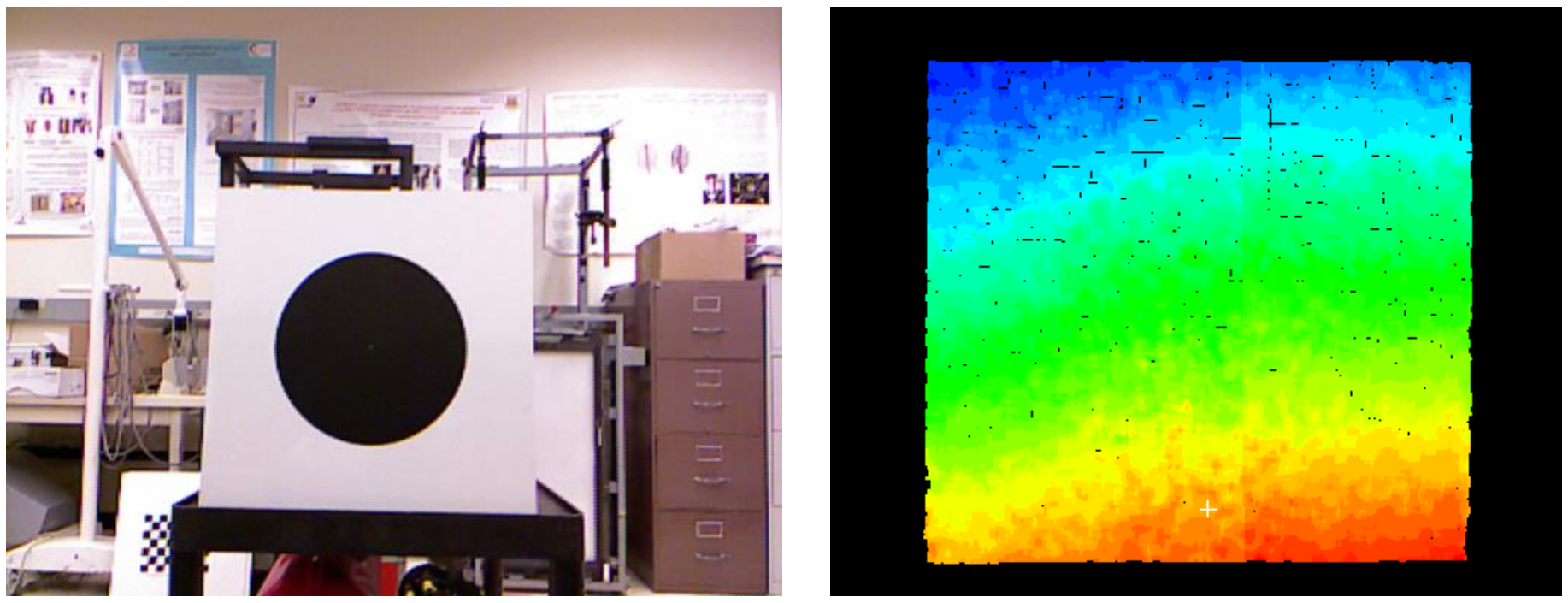

*Figure 5: (a) RGB image of a black and white target. (b) Point cloud of the black and white image captured using the Kinect.*

### 2.5 Distance Averaging

To reduce the random noise of depth measurements in TOF laser scanners and range cameras, it is common to take multiple distance observations for each point and use the average (Karel et al., 2010). To learn whether or not distance averaging is necessary, a planar wall located at approximately 1.5 m and 3 m from the Kinect was measured and the RMSE from the plane-fitting is plotted as a function of the number of depth images averaged. From Figure 6 it can be deduced that distance averaging at close-range is probably not necessary as the RMSE is only improved by approximately 0.1 mm even after averaging 100 images. However, at longer ranges such as 3 m, averaging 10 or more depth images improves the RMSE by a millimetre. For longer range applications that do not require 30 frames per second, distance averaging seems to be a viable solution for reducing the random errors in the depth measurements.

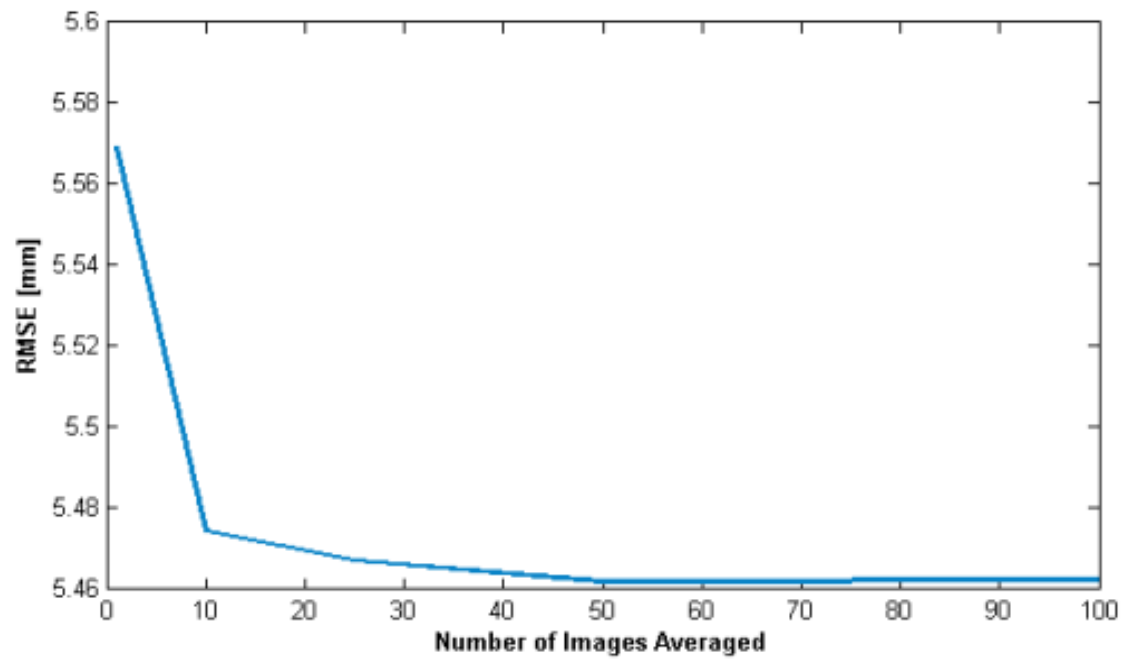

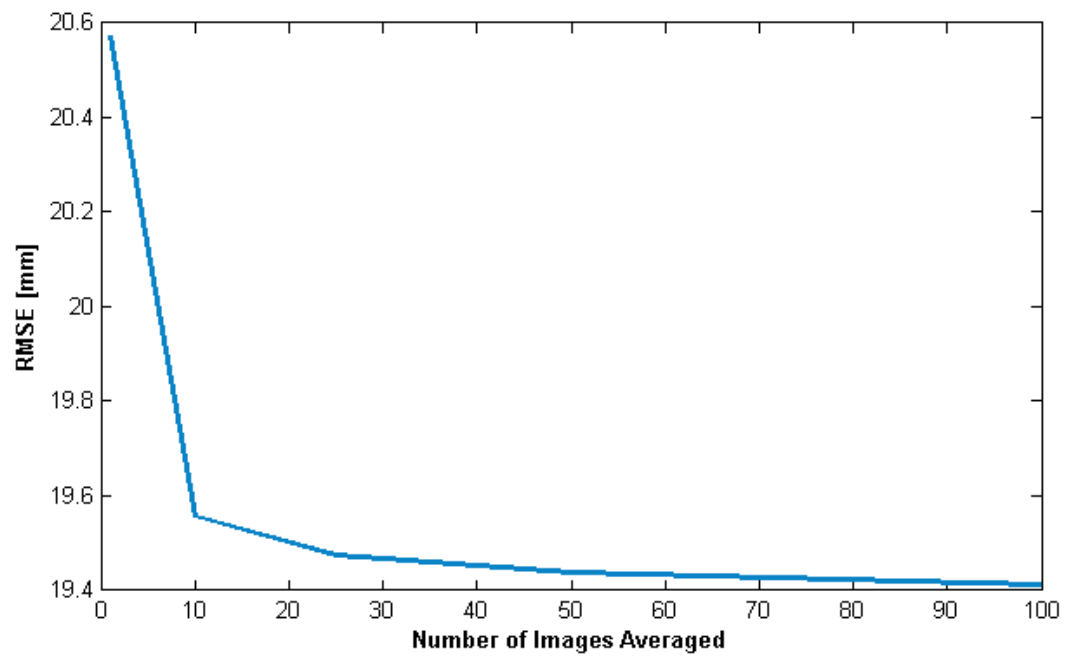

*Figure 6: RMSE of plane-fitting as a function of pixel-by-pixel distance averaging at (a) 1.5 m and (b) 3 m*

## 3. APPLICATIONS

To better understand the potential of using this low-cost gaming device for engineering-type applications, an experiment was conducted to explore the achievable accuracy in a common photogrammetric, biomedical, and Geomatics type problem. It is important to note that the Kinect results in this section are performed before any self-calibration.

### 3.1 3D Human Body Reconstruction

Precise 3D human body measurements are necessary for biometrics, animation, medical science, apparel design, and much more (Loker et al., 2005; Leyvand et al., 2011; and Weise et al., 2011). In this experiment the mannequin shown in Figure 7a was reconstructed using the Leica ScanStation 2 terrestrial laser scanner (Figure 7b), SwissRanger SR3000 TOF range camera (Figure 7c), and the Kinect (Figure 7d). Multiple point clouds were captured around the mannequin and registered using the iterative closest point (ICP) algorithm (Chen & Medioni, 1992) in Leica Cyclone to create a 3D model. By visual inspection, the laser scanner results are the most detailed, features such as the nose and eye sockets are easily identifiable. The results from the SR3000 are greatly degraded because of the internal scattering distortion (Jamtsho &

Lichti, 2010). Distance measurements made to objects farther away from the camera are biased by signals returned from closer objects, which cause internal multipath reflection between the lens and CCD/CMOS sensor. Unlike systematic error such as lens distortions, internal scattering changes from scene to scene and can be a challenge to model mathematically. Even after calibrating the IOPs and range errors of the camera (Lichti & Kim, 2011), significant distortions of the mannequin can still be visually identified due to existence of foreground objects. In comparison, the Kinect delivered a more visually elegant model (Figure 7d). Smisek et al. (2011) quantitatively compared the Kinect with the newer SR4000 and also reported the Kinect as more accurate. One of the main reasons is because triangulation-based 3D cameras make direction measurements instead of TOF measurements so it does not experience any scattering. This has been confirmed empirically by the authors. The RMSE of the ICP registration for the Kinect point cloud alone was 3 mm. When compared to the model reconstructed by the ScanStation 2, a RMSE of 11 mm was computed using ICP after registration. In Figure 7e the cyan depicts the model from laser scanning and the pink is from the Kinect. Some systematic deviations between the two models can be observed and are expected because the Kinect has not yet been calibrated. However, the Kinect showed promising results even before calibration in this experiment and should be applicable to a wide range of 3D reconstruction tasks.

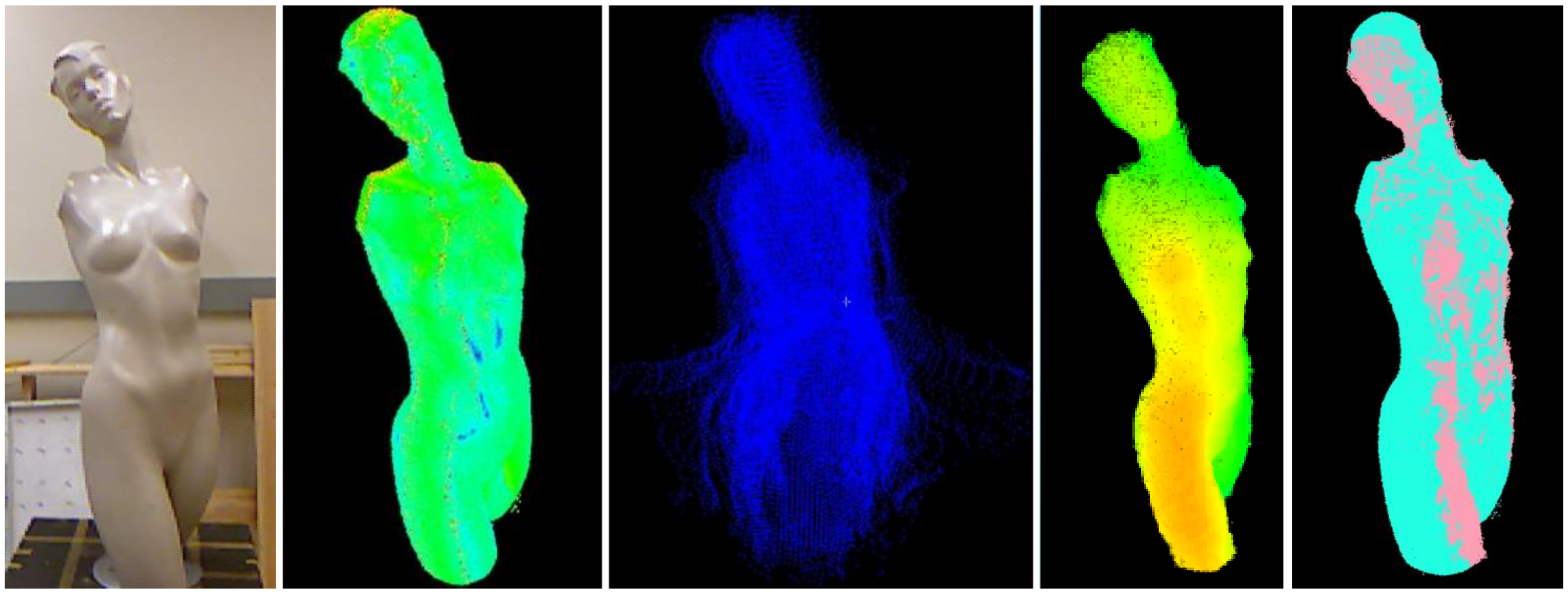

*Figure 7: (a) RGB of the mannequin. Point cloud of the mannequin acquired using (b) ScanStation 2, (c) SR3000, and (d) Kinect. (e) ICP registered model from ScanStation 2 (cyan) and Kinect (pink).*

## 4. MATHEMATICAL MODEL

The Kinect is composed of a RGB camera, IR camera, projector, multi-array microphone, and MEMs accelerometer. In this paper, only the optical sensors are considered. Computer vision based calibrations for the Kinect are widely available (Burrus, 2012). Herrera et al. (2011) used checkerboard targets and a single plane to calibrate the Kinect. The method presented in this paper has a similar concept but instead of a fast computer vision approach, an accuracy-driven photogrammetric approach is taken. The IR camera and projector together form the depth sensor which gives distance information for every pixel in the IR camera. The calibration for the RGB camera can follow the conventional collinearity model shown in Equation 1. The EOPs, IOPs, and object space coordinates of the signalised targets can be determined simultaneously using least-squares adjustment. To model the systematic errors in both cameras and projector, Brown's distortion model (Brown, 1971) has been adopted (Equation 2).

$$x_{ij} = x_p - c\frac{m_{11}(X_i - X_{oj}) + m_{12}(Y_i - Y_{oj}) + m_{13}(Z_i - Z_{oj})}{m_{31}(X_i - X_{oj}) + m_{32}(Y_i - Y_{oj}) + m_{33}(Z_i - Z_{oj})} + \Delta x$$
$$y_{ij} = y_p - c\frac{m_{21}(X_i - X_{oj}) + m_{22}(Y_i - Y_{oj}) + m_{23}(Z_i - Z_{oj})}{m_{31}(X_i - X_{oj}) + m_{32}(Y_i - Y_{oj}) + m_{33}(Z_i - Z_{oj})} + \Delta y \quad (1)$$

where $x_{ij}$ & $y_{ij}$ are the image coordinate observations of point i in image j
$x_p$ & $y_p$ are the principal point offsets in the x and y directions
c is the principal point distance
$X_i$, $Y_i$, & $Z_i$ are the object space coordinates of point i
$X_{oj}$, $Y_{oj}$, & $Z_{oj}$ are the position of image j in object space
$m_{11}\ldots m_{33}$ are the elements of the rotation matrix (R) describing the orientation of image j
$\Delta x$ & $\Delta y$ are the additional calibration parameters

$$\Delta x = x'_{ij}\left(k_1 r_{ij}^2 + k_2 r_{ij}^4 + k_3 r_{ij}^6\right) + p_1\left(r_{ij}^2 + 2{x'_{ij}}^2\right) + 2p_2 x'_{ij}\, y'_{ij} + a_1 x'_{ij} + a_2 y'_{ij}$$
$$\Delta y = y'_{ij}\left(k_1 r_{ij}^2 + k_2 r_{ij}^4 + k_3 r_{ij}^6\right) + p_2\left(r_{ij}^2 + 2{y'_{ij}}^2\right) + 2p_1 x'_{ij}\, y'_{ij} \quad (2)$$

where $k_1$, $k_2$, & $k_3$ describes the radial lens distortion
$p_1$ & $p_2$ describes the decentring lens distortion
$a_1$, $a_2$ describes the affinity and shear
$r_{ij}$ is the radial distance of point i in image j referenced to the principal point
$x'_{ij}$ & $y'_{ij}$ are the image coordinates of point i in image j after correcting for the principal point offset

Co-registration of the RGB image and depth image is necessary to colourize the point cloud. This involves solving for the translational and rotational offsets (a.k.a. leverarm and boresight parameters) of the two cameras as well as their IOPs. Since the RGB camera and IR camera are rigidly mounted on the same platform, a boresight and leverarm constraint can be applied to strengthen the bundle adjustment (Equation 3).

$$\begin{bmatrix} \bar{x}_i - \Delta x \\ \bar{y}_i - \Delta y \\ -c \end{bmatrix}_{RGB} - \frac{1}{\mu_i^{RGB}} R_{IR}^{RGB} R_{Map}^{IR} \left( \begin{bmatrix} X_i \\ Y_i \\ Z_i \end{bmatrix}_{RGB}^{Map} - \begin{bmatrix} X_o \\ Y_o \\ Z_o \end{bmatrix}_{IR}^{Map} - R_{IR}^{Map} \begin{bmatrix} b_x \\ b_y \\ b_z \end{bmatrix}_{RGB}^{IR} \right) = 0 \qquad (3)$$

where $b_x$, $b_y$, & $b_z$ are the leverarm parameters
$\mu_i$ is the unique scale factor for point i
R is the 3D rotation matrix

The depth sensor can also be modelled using the collinearity equations. However, the raw output from the depth sensor when using the Microsoft SDK is $Z_i$ at every pixel location of the IR camera. This is fundamentally different from most Kinect error models published to date, which use open source drivers (e.g. libfreenect and OpenNI) that stream disparity values as 11 bit integers. Unlike in Menna et al. (2011) and Khoshelham & Oude Elberink (2012) whose observations are disparity values and the depth calibration is performed by solving the slope and bias of a linear mapping function that relates depth to disparity, an alternative calibration method suitable for the Microsoft SDK is proposed in this paper.

Since the Microsoft SDK does not give access to the IR image, traditional point-based calibration where signalized targets are observed is not applicable. Instead, a plane-based calibration following the collinearity equation is adopted for calibrating the depth sensor. Since the Z coordinate of every point is provided, the corresponding X and Y coordinates can be computed in the IR camera's coordinate system based on the relationship shown in Equation 4. For the IR camera, image observations for each point are taken as the centre of every pixel. The IR camera uses the Aptina MT9M001 monochrome CMOS sensor, which has a nominal focal length of 6 mm and pixel size of 5.2 μm. The effective array size after 2x2 binning and cropping is 640 by 480 pixels with a pixel size of 10.4 μm.

$$X_i^{IR} = \frac{Z_i^{IR}}{c_{IR}} x_i^{IR} \qquad\qquad Y_i^{IR} = \frac{Z_i^{IR}}{c_{IR}} y_i^{IR} \tag{4}$$

Although the projector cannot "see" the image, it still obeys the collinearity condition and can be modelled as a reverse camera. It is assumed for every point $[X_i, Y_i, Z_i]_{IR}$ there is a corresponding $[x_{ij}, y_{ij}]_{IR}$ and $[x_{ij}, y_{ij}]_{Pro}$. The image coordinate observations can be determined by back-projecting the object space coordinates $[X_i, Y_i, Z_i]_{Pro}$ (which is equivalent to $[X_i, Y_i, Z_i]_{IR}$) into the projector's image plane using Equation 1. Very limited documentation about the projector is publically available so it is assumed that the projector has the same properties as the MT9M001 sensor. It is important to note that this assumption should not have major impacts on the effectiveness of the calibration. With the boresight and leverarm expressed relative to the IR camera, the functional model for calibrating the depth sensor is given in Equation 5.

$$\mu_i^{Pro} R_{IR}^{Map} R_{Pro}^{IR} \begin{bmatrix} \bar{x}_i - \Delta x \\ \bar{y}_i - \Delta y \\ -c \end{bmatrix}_{Pro} + \begin{bmatrix} X_o \\ Y_o \\ Z_o \end{bmatrix}_{IR}^{Map} + R_{IR}^{Map} \begin{bmatrix} b_x \\ b_y \\ b_z \end{bmatrix}_{Pro}^{IR} - \left( \mu_i^{IR} R_{IR}^{Map} \begin{bmatrix} \bar{x}_i - \Delta x \\ \bar{y}_i - \Delta y \\ -c \end{bmatrix}_{IR} + \begin{bmatrix} X_o \\ Y_o \\ Z_o \end{bmatrix}_{IR}^{Map} \right) = 0 \tag{5}$$

Instead of solving for the scale factor $\mu_{ij}$ for every point, it is expressed as a function of the plane parameters ($a_k$, $b_k$, $c_k$, and $d_k$), EOPs of the IR camera, IOPs of the applicable optical sensors, and the boresight and leverarm offsets for the RGB camera and projector. The functional model for the unique scale factor can be determined by solving for $\mu_{ij}$ in Equation 6, where $X_i$, $Y_i$, $Z_i$ is defined by Equation 3.

$$[a_k \quad b_k \quad c_k]\begin{bmatrix} X_i \\ Y_i \\ Z_i \end{bmatrix} - d_k = 0 \tag{6}$$

The proposed functional model minimizes the discrepancy of conjugate light rays at each tie point location while constraining every point reconstructed by the RGB camera and depth sensor to lie on the best-fit plane. Unlike most existing Kinect calibrations where the depth is calibrated independently of the bundle adjustment process, the proposed method solves for the EOPs, IOPs, boresights, leverarms, object space coordinates of targets measured by the RGB camera, and the plane parameters simultaneously in a combined least-squares adjustment model. The proposed method also takes into consideration the fact that the output depth/disparity values are a function of lens distortion. This mathematical model is kept as general as possible and should be applicable for other triangulation-based 3D cameras (e.g. Asus Xtion) and camera-projector systems.

# 5. EXPERIMENTAL RESULTS AND ANALYSES

## 5.1 Kinect Self-calibration

Two experiments were conducted to calibrate the Kinect. In the first calibration a single texturized plane was used for the calibration. In the second calibration, three roughly orthogonal planes were utilized. In both calibrations the datum was defined using inner constraints on the object space target coordinates and Baarda's data snooping was adopted. In the first experiment a checkerboard pattern was projected onto a flat wall to provide some targets for calibrating the RGB camera (Figure 8a). If only the depth sensor needs to be calibrated, homogenous flat surfaces will suffice. Multiple convergent images were captured from different positions and orientations (Figure 8b) while ensuring the target field covers the majority of the image format (Fraser, 2012).

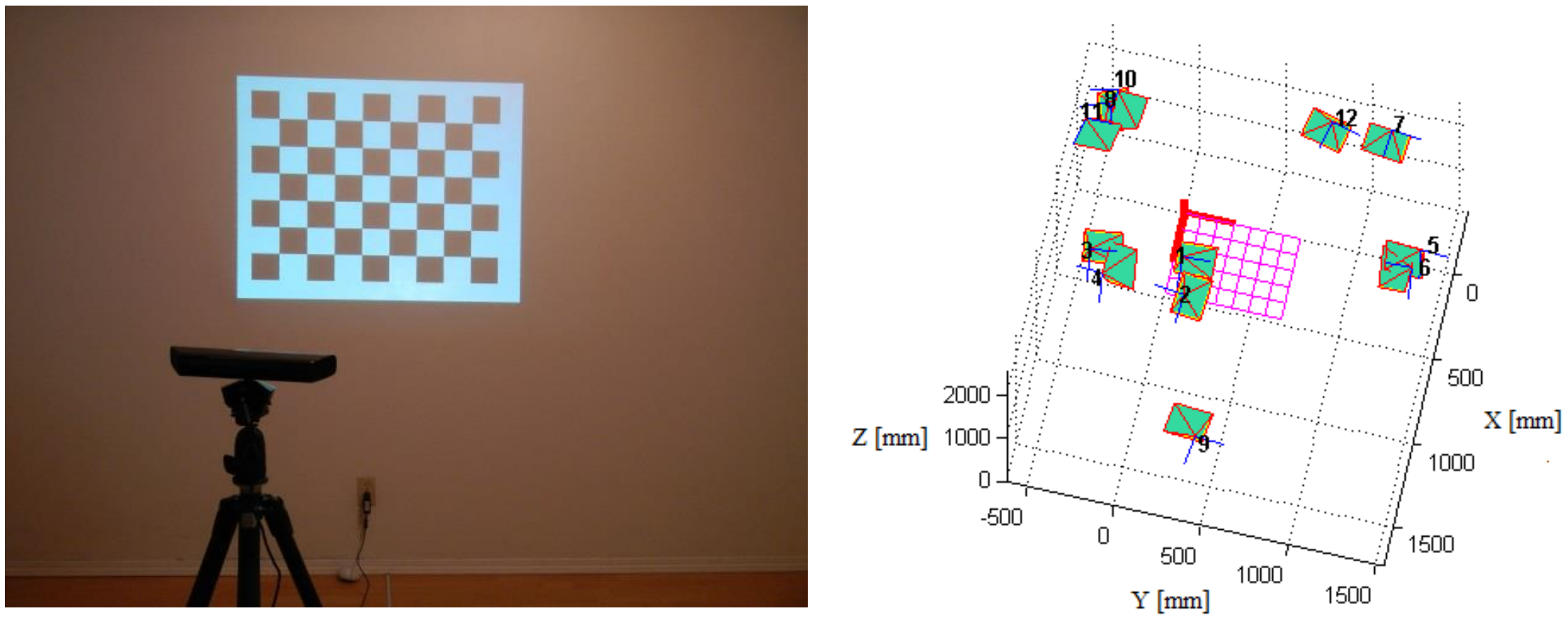

*Figure 8: (a) Experimental design for calibrating the Kinect. (b) Network configuration of the calibration.*

The corner points of the checkerboard pattern were extracted using the computer vision camera calibration toolbox from Caltech (Bouguet, 2010). Fifty depth images were averaged at each exposure station and points from the wall were semi-automatically extracted. Some statistics about this calibration are summarized in Table 1. It was initially assumed that relative to the IR camera, the projector has no rotational offsets and is located 7.5 cm away in the positive x-direction of the IR sensor, while the RGB camera is 2.5 cm away and has zero rotational offsets. The leverarm and boresight parameters determined in this adjustment are given in Table 2. To quantify the effectiveness of the calibration for the depth sensor, the misclosure of light rays (computed using Equation 5) at every object point used in the calibration was computed. From Table 3, it can be observed that after self-calibration the quality of fit between conjugate light rays are significantly improved.

| Table 1: Single plane least-squares calibration statistics | |
|---|---|
| # of exposure stations | 12 |
| # of unknowns | 232 |
| # of signalized targets | 48 |
| # of observations | 3456 |
| Average Redundancy | 0.94 |

| Table 2: Computed leverarm and boresight in the single plane calibration | | | | |
|---|---|---|---|---|
| | IR-Projector | | IR-RGB | |
| | Value [$^o$ & m] | σ [“ & mm] | Value [$^o$ & m] | σ [“ & mm] |
| $\Delta\omega$ | -0.001 | 16 | -0.865 | 542 |
| $\Delta\phi$ | 0.018 | 34 | 0.071 | 560 |
| $\Delta\kappa$ | 0.003 | 19 | -0.959 | 1033 |
| $\Delta b_x$ | 0.074 | 0.6 | 0.031 | 4.4 |
| $\Delta b_y$ | 0.000 | 0.1 | 0.028 | 4.8 |
| $\Delta b_z$ | 0.000 | 0.6 | 0.022 | 7.0 |

| Table 3: Quality of fit for conjugate light rays in the depth sensor before and after the single plane calibration | | | |
|---|---|---|---|
| | Before Calibration [mm] | After Calibration [mm] | % Improvement |
| $RMSE_X$ | 0.56 | 0.18 | 68 |
| $RMSE_Y$ | 0.88 | 0.32 | 64 |
| $RMSE_Z$ | 0.02 | 0.01 | 64 |

From the single plane calibration, the estimation of the rotational and translational offsets between the IR camera and RGB camera was quite poor. One of the biggest sources of uncertainty arises from the weak determination of the IR camera’s EOPs since only one plane was deployed. In the second experiment, three differently oriented planes were used to calibrate the Kinect (Figure 9). A brief summary of the least-squares adjustment, the recovered boresights and leverarms, and the RMSE of the misclosure vectors are documented in Tables 4, 5, and 6, respectively. By adding two additional planes, even with fewer images, the precision of the 3D rigid body transformation parameters relating the IR and RGB camera is significantly improved.

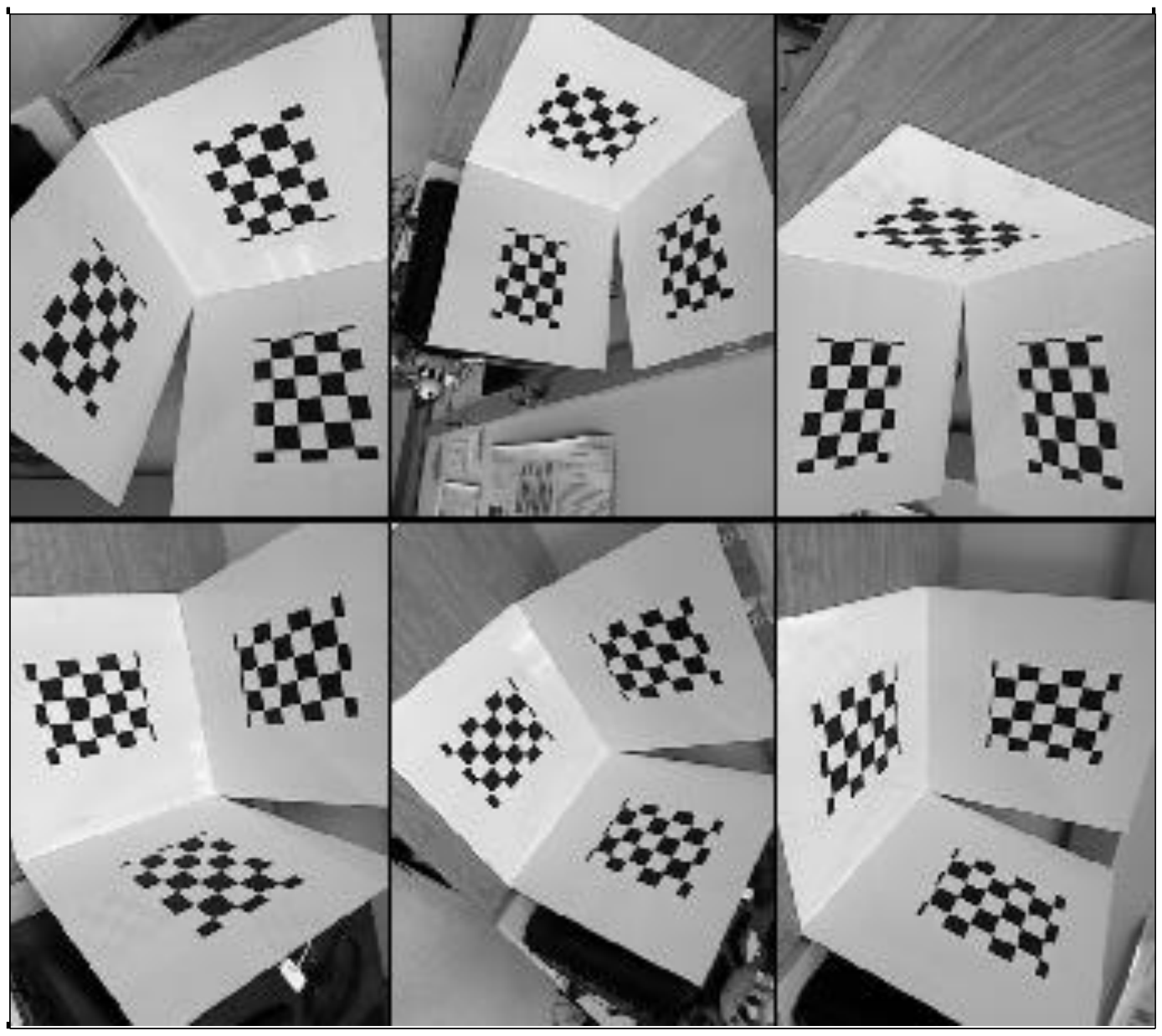

*Figure 9: Images used for the Kinect multi-plane calibration*

| Table 4: Multi-plane least-squares calibration statistics | |
|---|---|
| # of exposure stations | 6 |
| # of unknowns | 275 |
| # of signalized targets | 72 |
| # of observations | 2368 |
| Average Redundancy | 0.89 |

| Table 5: Computed leverarm and boresight in the multi-plane calibration | | | | |
|---|---|---|---|---|
| | IR-Projector | | IR-RGB | |
| | Value [° & m] | σ [" & mm] | Value [° & m] | σ [" & mm] |
| Δω | -0.002 | 29 | -0.775 | 362 |
| Δϕ | -0.006 | 38 | -0.119 | 312 |
| Δκ | -0.003 | 14 | -0.359 | 207 |
| $\Delta b_x$ | 0.080 | 0.3 | 0.017 | 1.3 |
| $\Delta b_y$ | 0.000 | 0.1 | -0.008 | 1.4 |
| $\Delta b_z$ | 0.000 | 0.1 | 0.007 | 2.7 |

| Table 6: Quality of fit for conjugate light rays in the depth sensor before and after the multi-plane calibration | | | |
|---|---|---|---|
| | Before Calibration [mm] | After Calibration [mm] | % Improvement |
| $RMSE_X$ | 0.49 | 0.37 | 23 |
| $RMSE_Y$ | 0.42 | 0.24 | 41 |
| $RMSE_Z$ | 0.62 | 0.29 | 53 |

Due to high correlations in both calibrations, only the IOPs of the RGB camera can be recovered at this point (Table 7). It is worth mentioning that the recovered $y_p$ parameter from the first and second experiment corresponds to a shift of 31 and 32 pixels, respectively. This is similar to the empirical values reported by Khoshelham & Oude Elberink (2012) and the theoretical value of 32 pixels due to image cropping. But very different $x_p$ and $y_p$ values were reported in Menna et al. (2011).

Future work will attempt to estimate the IOPs of the IR camera either through stronger network configuration or by gaining access to the IR images using open source software such as RGBDemo (Burrus, 2012). As shown in Smisek et al. (2011), under certain illumination conditions the IR camera can observe signalised targets that are visible to the RGB camera. In that case, the conventional bundle adjustment with self-calibration can be applied to calibrate the IR camera. This should also improve the boresight and leverarm estimation between the IR camera and RGB camera.

| Table 7: Recovered IOPs for the RGB camera in both calibrations | | | | |
|---|---|---|---|---|
| | Single Plane | | Multi-Plane | |
| | Value | σ | Value | σ |
| $x_p$ [mm] | 0.06 | 0.005 | 0.02 | 0.002 |
| $y_p$ [mm] | -0.18 | 0.005 | -0.17 | 0.002 |
| c [mm] | 2.97 | 0.008 | 2.95 | 0.004 |
| $k_1$ [$mm^{-2}$] | 1.71e-2 | 4.47e-4 | 1.62e-2 | 4.04e-4 |
| $k_2$ [$mm^{-4}$] | -3.03e-3 | 1.14e-4 | -3.15e-3 | 1.03e-4 |

## 5.2 Precision versus depth

To verify the proposed sensor error model for calibrating the depth sensor, an independent experiment was made. In a 20' by 40' by 20' racquetball court under stable and controlled atmospheric conditions (20.0°C, 884.8mb and 48.8% humidity) a flat wall was observed with an uncalibrated Kinect situated at 1 m distance up to 10 m at increments of 0.5 m using the Brekel Kinect software. The RMSE from the least-squares plane fitting is shown in Figure 10 as the purple curve. Based on the proposed mathematical model in this paper, the same experiment was simulated with 100 well distributed gridded points observed on each plane and no systematic errors. It was assumed the IR camera and projector has a nominal baseline of 7.5 cm. Using Variance Component Estimations in the multi-plane self-calibration an image measurement precision of 0.8 μm (1/13 pixel) was determined for the IR camera. Using these values, the RMSE of the fitted planes from the simulated data were computed and they follow closely the experimental depth error behaviour (orange curve in Figure 10). This depth error behaviour is similar to those reported in Khoshelham (2011).

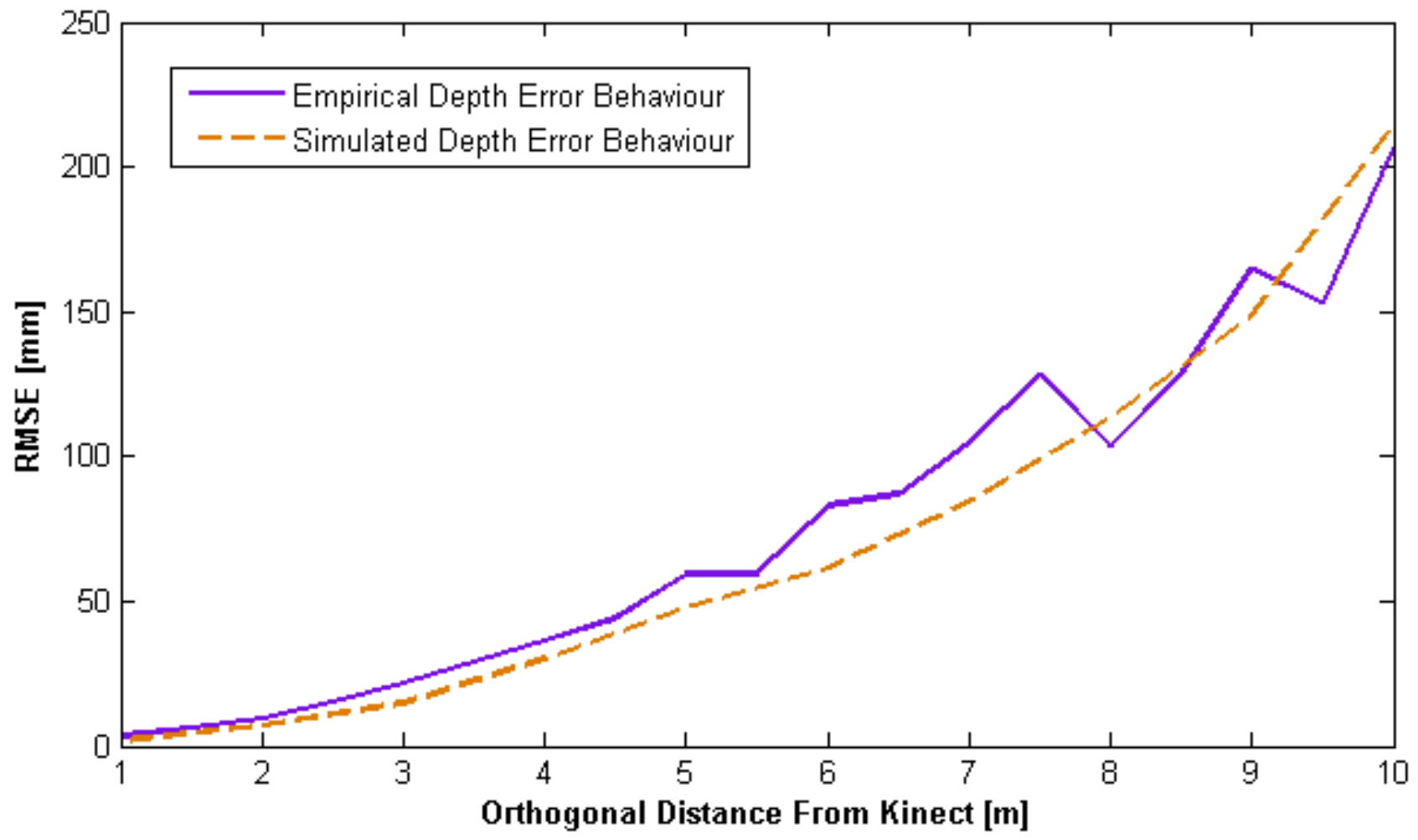

*Figure 10: Empirical and theoretical error behaviour for the depth measurements of the Kinect.*

## 6. CONCLUSION & FUTURE WORK

The Kinect system has demonstrated potential to be used for engineering applications. At long-range it may not be as accurate as a terrestrial laser scanner, photogrammetric system, or structured light system, but for the cost and portability it is delivering fairly high geometric accuracy at close-range. The Kinect was used to capture point clouds in a series of tests. Since this is a triangulation-based system, it does not exhibit data distortions that would otherwise be identifiable in TOF systems, of which the most pronounced range error is the internal scattering experienced by most TOF 3D range cameras. In one of the 3D reconstruction experiments, the uncalibrated Kinect system was capable of modelling a mannequin much more accurately than a calibrated SR3000. Before this gaming device should be used for any metric type applications the systematic errors should be properly modelled. A new calibration routine is presented in this paper which models the IR camera, RGB camera, and projector using the collinearity equation with boresight, leverarm, and point-on-plane constraints. Although the IOPs of the IR camera cannot be recovered at this point, by updating the boresight and leverarm parameters the volume of uncertainty for each point determined by the depth sensor was improved significantly. Future calibrations will try to incorporate IR images of some target field to improve the IOPs, boresight, and leverarm observability.

## ACKNOWLEDGEMENTS

The authors would like to sincerely thank Natural Sciences and Engineering Research Council of Canada, the Canada Foundation for Innovation, Alberta Innovates, Informatics Circle of Research Excellence, Terramatic Technologies Inc., and SarPoint Engineering Ltd. for funding this research. The authors are also grateful for Tammy Smith and Justin Waghray for assisting with data capture and processing the SR3000 and ScanStation 2 point clouds.

## REFERENCES

Bouguet, J. (2010, July 9). *Camera Calibration Toolbox for Matlab*. Retrieved March 30, 2012, from Computational Vision at CALTECH: http://www.vision.caltech.edu/bouguetj/calib_doc/index.html

Brekelmans, J. (2012). *Brekel Kinect*. Retrieved March 30, 2012, from http://www.brekel.com/?page_id=155

Brown, D. (1971). Close-range camera calibration. *Photogrammetric Engineering, 37 (8)*, 855-866.

Burrus, N. (2012). *RGBDemo*. Retrieved March 30, 2012, from Manctl: http://labs.manctl.com/rgbdemo/

Chen, Y., & Medioni, G. (1992). Object modelling by registration of multiple range images. *Image and Vision Computing*, 145-155.

Chiabrando, F., Chiabrando, R., Piatti, D., & Rinaudo, F. (2009). Sensors for 3D imaging: metric evaluation and calibration of a CCD/CMOS time-of-flight camera. *Sensors (9)*, 10080-10096.

Fraser, C. (2012). Automatic camera calibration in close-range photogrammetry. *ASPRS 2012 Annual Conference.* Sacramento, USA.

Freedman, B., Shpunt, A., Machline, M., & Arieli, Y. (2010). *Patent No. US2010/0018123 A1*. United States of America.

Glennie, C., & Lichti, D. (2011). Temporal stability of the Velodyne HDL-64E S2 scanner for high accuracy scanning applications. *Remote Sensing 3(3)*, 539-553.

Hanke, K., Grussenmeyer, P., Grimm-Pitzinger, A., & Weinold, T. (2006). First experiences with the Trimble GX scanner. *International Archives of Photogrammetry, Remote Sensing and Spatial Information Sciences, Vol. XXXVI, Part 5* (pp. 1682-1759). Dresden, Germany, Sept. 25-27: ISPRS Comm. V Symposium.

Herrera, C., Kannala, J., & Heikkilä, J. (2011). Accurate and practical calibration of a depth and color camera pair. *14th International Conference on Computer Analysis of Images and Patterns*, (pp. 437-445). Seville, Spain.

Jamtsho, S., & Lichti, D. (2010). Modeling scattering distortion of 3D range camera. *The International Archives of Photogrammetry, Remote Sensing, and Spatial Information Sciences, XXXVIII (5)*, (pp. 299-304).

Kahlmann, T., Remondino, H., & Ingensand, H. (2006). Calibration for increased accuracy of the range imaging camera SwissRangeeTM. *The International Archives of Photogrammetry, Remote Sensing and Spatial Information Sciences, 35 (5)*, 136-141.

Karel, W., Ghuffar, S., & Pfeifer, N. (2010). Quantifying the distortion of distance observations caused by scattering in time-of-flight range cameras. *The International Archives of Photogrammetry, Remote Sensing, and Spatial Information Sciences, XXXVIII (5)*, (pp. 316-321). Newcastle upon Tyne, UK.

Khoshelham, K. (2011). Accuracy analysis of kinect depth data. *ISPRS Laser Scanning Workshop.* Calgary, Canada.

Khoshelham, K., & Oude Elberink, S. (2012). Accuracy and resolution of kinect depth data for indoor mapping applications. *Sensors, vol. 12*, 1437-1454.

Konolige, K., & Mihelich, P. (2010, December 9). *Kinect Calibration: Technical*. Retrieved March 30, 2012, from Robot Operating System: http://www.ros.org/wiki/kinect_calibration/technical

Lange, R., & Seitz, P. (2001). Solid-state time-of-flight range camera. *IEEE Journal of Quantum Electronics, 37 (3)*, 390-397.

Leyvand, T., Meekhof, C., Yi-Chen, W., Jian, S., & Baining, G. (2011). Kinect identity: technology and experience. *Computer, vol. 44*, 94-96.

Lichti, D., & Kim, C. (2011). A comparison of three geometric self-calibration methods for range cameras. *Remote Sensing, vol 3 (5)*, 1014-1028.

Loker, S., Ashdown, S., & Schoenfelder, K. (2005). Size specific analysis of body scan data to improve apparel fit. *Textile and Apparel Technology Management, vol 4 (3)*, 103-120.

Menna, F., Remondino, F., Battisti, R., & Nocerino, E. (2011). Geometric investigation of a gaming active device. *Videometrics, Range Imaging, and Applications XI.* Munich, Germany: SPIE Optical Metrology.

Microsoft. (2012). *Kinect for Windows*. Retrieved March 30, 2012, from Microsoft: http://www.microsoft.com/en-us/kinectforwindows/

Mure-Dubois, J., & Hugli, H. (2007). Optimized scattering compensation for time-of-flight camera. *Proc. SPIE: Two- and Three-Dimensional Methods for Inspection and Metrology V, 6762, 67620H,* 1-10.

Shan, J., & Toth, C. (2008). *Topographic Laser Ranging and Scanning: Principles and Processing.* Boca Raton, Florida: CRC Press.

Smisek, J., Jancosek, M., & Pajdla, T. (2011). 3D with kinect. *Consumer Depth Cameras for Computer Vision*, (pp. 1154-1160). Barcelona, Spain, November 12.

Soudarissanane, S., Lindenbergh, R., Menenti, M., & Teunissen, P. (2011). Scanning geometry: Influencing factor on the quality of terrestrial laser scanning points. *ISPRS Journal of Photogrammetry and Remote Sensing 66(4),* 389–399.

Weise, T., Bouaziz, S., Li, H., & Pauly, M. (2011). Real time performance-based facial animation. *ACM Transactions on Graphics, Proceedings of the 38th ACM SIGGRAPH Conference, vol 30 (4).* Los Angeles, CA.

**Contributions of Authors**

The first author wrote the paper, which was later edited by the second and third authors. The first author was responsible for capturing the Kinect data, using them for calibration, and analyzing the results. The second author assisted in the preliminary testing and wrote the first version of the Kinect acquisition software. The third author helped analyze the results and made valuable recommendations. The fourth author provided guidance throughout the project.

## APPENDIX B: TLS Self-Calibration Results of the FARO Focus$^{3D}$ S and its Comparison to the Higher Quality Leica HDS6100

### Article: Accuracy Assessment of the FARO Focus$^{3D}$ and Leica HDS6100 Panoramic-Type Terrestrial Laser Scanners through Point-Based and Plane-Based User Self-Calibration

Jacky C.K. CHOW, Derek D. LICHTI, and William F. TESKEY, Canada

### SUMMARY

Recent developments of terrestrial laser scanners have made these accurate and efficient survey instruments more affordable and portable. In this paper, the geometric accuracy of two terrestrial laser scanners (i.e. the FARO Focus$^{3D}$ and the Leica HDS6100) is being evaluated. Both of these mid-range instruments are phase-based and have a panoramic architecture. The Focus$^{3D}$ is currently one of the most compact and affordable 3D terrestrial laser scanners on the market. The geometric quality of this latest generation laser scanner will be compared to the HDS6100. The quality assessments performed in this paper are centred on the self-calibration method for terrestrial laser scanners. This method can remove systematic defects in the instrument without hardware modifications or specialized equipment. Through the observation of either signalised targets or planar-features, the residual systematic errors in the laser scanner can be modelled mathematically. From previous studies, these unmodelled systematic errors can drastically deteriorate the point cloud quality and the self-calibration approach has been proven to be an effective tool for eliminating systematic effects caused by flaws of individual components and misalignment between components. Through redundant observations, the distance and angular measurement precision can be estimated in a least-squares adjustment and used as a quantitative measure for comparing the systems. Both laser scanners were tested and calibrated multiple times at the University of Calgary. Based on the experimental results presented in this paper, it was discovered that data coming from the more affordable Focus$^{3D}$ are contaminated with significant systematic errors. Even after self-calibration, the measurement random noise is still higher than the HDS6100. However, at close-range the contribution of the higher random noise to the positioning solution is small and does not have a significant detrimental impact on the mapped scene.

# Accuracy Assessment of the FARO Focus$^{3D}$ and Leica HDS6100 Panoramic-Type Terrestrial Laser Scanners through Point-Based and Plane-Based User Self-Calibration

Jacky C.K. CHOW, Derek D. LICHTI, and William F. TESKEY, Canada

## 1. INTRODUCTION

Terrestrial laser scanning (TLS) instruments have already established their status as a reliable, accurate, and high-speed active 3D mapping solution (Vosselman & Mass, 2010). By measuring spatial distances at uniform increments of arc in two orthogonal directions, 3D coordinates on any laser-suitable surface can be measured without specially-designed targets. In the past, companies have mainly focused on improving the positioning accuracy, acquisition speed, measurable range, and functionalities (e.g. built-in memory storage, Wi-Fi connectivity, built-in batteries, and on-board touch-screen controller) of the scanner. Recently, the cost of 3D terrestrial laser scanners has begun to reduce. Companies such as FARO and Leica are releasing new scanners (i.e. Focus$^{3D}$ and C5, respectively) that are opening new markets to the active 3D imaging industry. Consumers are beginning to find shorter range applications with lower accuracy requirements for these economical laser scanners. In this paper, the geometric accuracy of the FARO Focus$^{3D}$ will be analyzed and compared to a higher accuracy laser scanner from Leica (i.e. the HDS6100). The range, horizontal direction, and elevation angle measurement precision of both instruments were estimated repeatedly using variance component estimations in a point-based user self-calibration routine (Lichti, 2007). Residual systematic errors in both scanners were also recovered and modelled empirically by observing a large quantity of signalised targets or planar-features. Self-calibration is useful because despite the manufacturer's laboratory calibration, systematic errors are still identifiable in various scanners (Kersten et al., 2008) and through self-calibration the geometric accuracy of the scanner can be improved; in some cases improvement from the millimetre level to the sub-millimetre level is possible (Chow et al., 2011b). To assist the point-cloud registration, most modern TLS instruments are equipped with additional sensors for defining the exterior orientation parameters (EOPs) of the instrument, for example built-in electronic compass, dual-axis compensator, and electronic barometer. These additional observations are valuable for reducing correlations between EOPs and the other parameters in the self-calibration adjustment (Lichti, 2010).

## 2. MATHEMATICAL MODEL

Software calibration of optical sensors is a well-established concept in photogrammetry and computer vision. For analog and digital cameras, bundle adjustment based on the collinearity equations is the preferred self-calibration method for most cases (Brown, 1971). Regarding terrestrial laser scanner self-calibration, registration using 3D rigid body transformation with the observations expressed in the spherical coordinate system has proven to be an effective approach (Lichti, 2007; Reshetyuk, 2010). In this paper, both the Focus$^{3D}$ and HDS6100 were calibrated using this method but with different geometric primitives. In the point-based self-calibration (Lichti et al., 2007; Reshetyuk, 2009; Schneider, 2009), a large quantity of signalised targets distributed evenly on the walls, ceiling, and floor of a room were observed. These targets, acquired by the scanner occupying different positions and orientations, were then used for registering the point clouds. Additional parameters (APs) were appended to the observation equations to simultaneously correct for biases, axes misalignments, wobbling, etc. As an alternative to the point-based self-calibration approach, well defined geometric features such as lines, planes, cylinders, spheres, and tori can be used to replace signalised targets for registering point clouds and calibrating the scanner. In this paper, besides point-based calibration, planar features were also considered for self-calibration (Bae & Lichti, 2007; Chow et al., 2011a).

### 2.1 Point-based user self-calibration of terrestrial laser scanners

The point-based self-calibration method is based on the 3D rigid body transformation given in Equation 1. The scanner's range and angles observation with APs appended are shown in Equation 2. The calibration models for range, horizontal direction, and elevation angle observations are given in Equations 3, 4, and 5, respectively. The EOPs of each scanner setup, 3D object space coordinates of every target, and APs are estimated simultaneously in a parametric least-squares adjustment. The stochastic model for the observations is also estimated in the adjustment using variance component estimation. It is assumed that the observations are uncorrelated and the standard deviation of the angular observations is independent of the scanning geometry. The range observations on the other hand are known to vary according to the secant function of the incidence angle (Soudarissanane et al., 2011). The elongation of the laser footprint at large incidence angles causes distance measurements to be integrated over a larger surface area, which results in a lower range measurement precision.

$$\begin{bmatrix} x_{ij} \\ y_{ij} \\ z_{ij} \end{bmatrix} = M_j \left( \begin{bmatrix} X_i \\ Y_i \\ Z_i \end{bmatrix} - \begin{bmatrix} X_{oj} \\ Y_{oj} \\ Z_{oj} \end{bmatrix} \right) \tag{1}$$

$$M_j = R_3(\kappa_j) R_2(\phi_j) R_1(\omega_j)$$

where $X_i$, $Y_i$, and $Z_i$ are the object space coordinates of point i.
$x_{ij}$, $y_{ij}$, and $z_{ij}$ are the Cartesian coordinates of point i in scanner space j.
$X_{oj}$, $Y_{oj}$, and $Z_{oj}$ are the position of scanner j in object space.
$\omega_j$, $\varphi_j$, and $\kappa_j$ are the primary, secondary, and tertiary rotation angles that describes the orientation of scanner j in object space.
$R_1$, $R_2$, and $R_3$ are the rotation matrices about the primary, secondary, and tertiary axis, respectively.

$$\rho_{ij} = \sqrt{x_{ij}^2 + y_{ij}^2 + z_{ij}^2} + \Delta\rho$$
$$\theta_{ij} = \tan^{-1}\left(\frac{y_{ij}}{x_{ij}}\right) + \Delta\theta \qquad (2)$$
$$\alpha_{ij} = \tan^{-1}\left(\frac{z_{ij}}{\sqrt{x_{ij}^2 + y_{ij}^2}}\right) + \Delta\alpha$$

where $\rho_{ij}$, $\theta_{ij}$, and $\alpha_{ij}$ are the range, horizontal circle, and vertical circle reading, respectively to point i in scanner space j
$\Delta\rho$, $\Delta\theta$, and $\Delta\alpha$ are the additional systematic correction parameters for range, horizontal direction, and vertical direction, respectively

$$\Delta\rho = A_0 + A_1\rho_{ij} + A_2\sin(\alpha_{ij}) + A_3\sin\left(\frac{4\pi}{U_1}\rho_{ij}\right) + A_4\cos\left(\frac{4\pi}{U_1}\rho_{ij}\right) + ET_\rho \qquad (3)$$

where $A_0$ describes the rangefinder offset
$A_1$ describes the scale factor error
$A_2$ describes the laser axis vertical offset
$A_3$ and $A_4$ describe the cyclic errors
$ET_\rho$ means other empirical range error terms

$$\Delta\theta = B_1\theta + B_2\sin(\theta) + B_3\cos(\theta) + B_4\sin(2\theta) + B_5\cos(2\theta) + B_6\sec(\alpha) + B_7\tan(\alpha) + B_8\rho^{-1} + B_9\sin(\alpha) + B_{10}\cos(\alpha) + ET_\theta \quad (4)$$

where $B_1$ describes the scale factor error
$B_2$ and $B_3$ describe the horizontal circle eccentricity
$B_4$ and $B_5$ describe the non-orthogonality of encoder and vertical axis
$B_6$ describes the collimation axis error
$B_7$ describes the trunnion axis error
$B_8$ describes horizontal eccentricity of collimation axis
$B_9$ and $B_{10}$ describe the trunnion axis wobble
$ET_\theta$ means other empirical horizontal direction error terms

$$\Delta\alpha = C_0 + C_1\alpha + C_2\sin(\alpha) + C_3\sin(2\alpha) + C_4\cos(2\alpha) + C_5\rho^{-1} + C_6\sin(2\theta) + C_7\cos(2\theta) + C_8\sin(4\theta) + ET_\alpha \quad (5)$$

where $C_0$ describes the vertical circle index error
$C_1$ describes the scale factor error
$C_2$ describes the vertical circle eccentricity
$C_3$ and $C_4$ describe the non-orthogonality of encoder and trunnion axis
$C_5$ describes the vertical eccentricity of collimation axis
$C_6$, $C_7$, and $C_8$ describe the vertical axis wobble
$ET_\alpha$ means the other empirical elevation angle error terms

The centre of each signalised target can be determined using least-squares geometric form fitting or template matching. Typically a laser scanner cannot aim directly at the centre of a target like a total station, therefore an ample amount of observations are made on the surface of the target and the centroid is then calculated precisely by exploiting information/properties about the signalised target such as size, shape, and reflectivity. In cases where the laser scanner was leveled and/or the heading of the scanner was measured, additional observations/constraints can be included in the least-squares adjustment. For example, with the Focus$^{3D}$, when all three attitude angles are measured internally and applied to the point cloud, the following observations can be added to the adjustment (Equation 6).

$$\begin{aligned} \omega &= \omega_{obs} \pm \sigma_{\omega} \\ \phi &= \phi_{obs} \pm \sigma_{\phi} \\ \kappa &= \kappa_{obs} \pm \sigma_{\kappa} \end{aligned} \tag{6}$$

## 2.2 Plane-based user self-calibration of terrestrial laser scanners

Instead of a functional model that minimizes the discrepancies of tie points in the X, Y, and Z direction as explained in Section 2.1, a functional model that constrains every point to lie on the best-fit plane is applied instead in the plane-based self-calibration. Although other geometric features can be utilized, planes are beneficial because they are abundant in urban environments, which make in-situ self-calibration more feasible. Following the same rigid body transformation model and spherical parameterization of the observations in Equations 1 and 2, Equation 7 can be adopted to constrain every point to lie on a plane. Instead of estimating the 3D object space coordinates of every target, the plane-based calibration estimates the four plane parameters while minimizing the sum of squares of the residuals. In a combined least-square adjustment, the scanner's EOPs, plane parameters, and APs are all estimated simultaneously.

$$\begin{pmatrix} a_k & b_k & c_k \end{pmatrix} \left\{ M_j \begin{pmatrix} (\rho_{ij} - \Delta\rho)\cos(\alpha_{ij} - \Delta\alpha)\cos(\theta_{ij} - \Delta\theta) \\ (\rho_{ij} - \Delta\rho)\cos(\alpha_{ij} - \Delta\alpha)\sin(\theta_{ij} - \Delta\theta) \\ (\rho_{ij} - \Delta\rho)\sin(\alpha_{ij} - \Delta\alpha) \end{pmatrix} + \begin{pmatrix} X_{o_j} \\ Y_{o_j} \\ Z_{o_j} \end{pmatrix} \right\} - d_k = 0 \tag{7}$$

where $a_k$, $b_k$, $c_k$, $d_k$ are the direction normal and the orthogonal distance to plane k

## 3. EXPERIMENT

The Leica HDS6100 was calibrated seven times in the past three years using the point-based self-calibration approach and four times using the plane-based self-calibration method. Self-calibration was carried out for the newer Focus$^{3D}$ six times using the point-based method and once using the plane-based method this year. The calibrations were all performed at the University of Calgary in one of two rooms. The smaller room (Figure 1) has dimensions 5 m by 5 m by 3 m and the larger room (Figure 2) is 14 m by 11 m by 3 m. In both situations, either a redundant number of targets were observed or a redundant number of planes were observed. For the point-based calibration either circular or checkerboard type paper targets were deployed and their centroids were determined using least-squares geometric form-fitting as explained in Chow et al. (2010) and Chow et al. (2011b), respectively. In the plane-based self-calibration, besides natural planes in the environment (i.e. walls) additional metal plates with a glossy white finish were introduced to strengthen the network geometry. The number of scans, number of targets/planes, number of observations, number of unknowns, and the average redundancy for the HDS6100 and Focus$^{3D}$ calibrations are summarized in Tables 1 and 2, respectively.

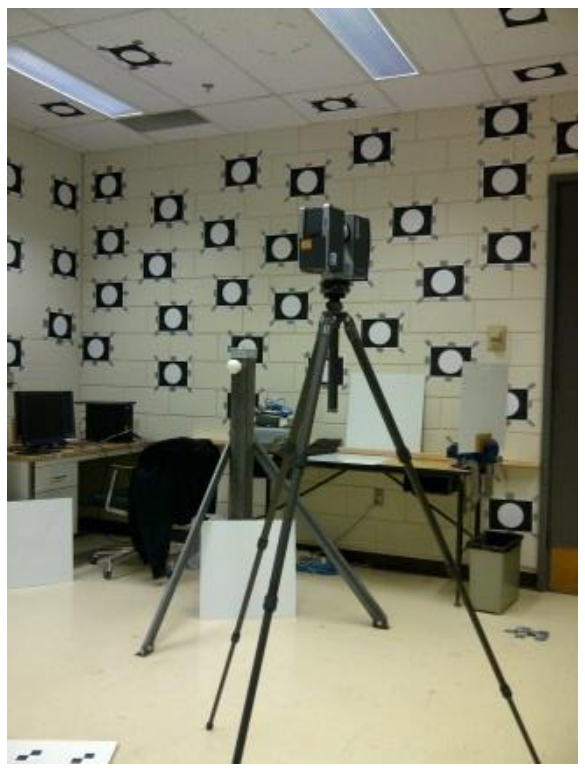

*Figure 1: Small 5 x 5 x 3m calibration room*

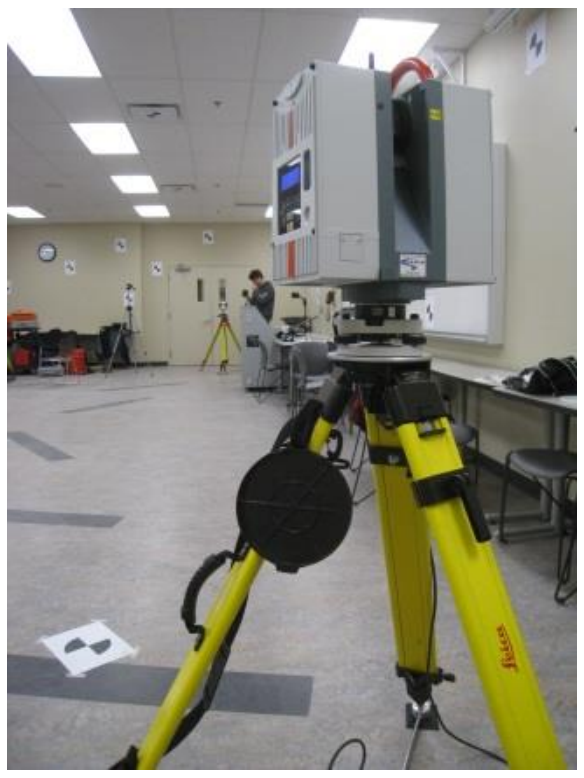

*Figure 2: 14 x 11 x 3m Large calibration room*

| Table 1: Summary of some statistics from the HDS6100 self-calibrations | | | | | | | |
|---|---|---|---|---|---|---|---|
| **Dataset** | **Room** | **Point/plane based** | **# of scans** | **# of targets/planes** | **# of obs.** | **# of unk.** | **Avg. Red.** |
| 1 | Small | Point | 6 | 264 | 3495 | 833 | 0.76 |
| 2 | Large | Point | 6 | 63 | 759 | 231 | 0.70 |
| 3 | Large | Point | 6 | 63 | 762 | 230 | 0.70 |
| 4 | Large | Point | 4 | 102 | 963 | 335 | 0.66 |
| 5 | Large | Point | 4 | 104 | 837 | 339 | 0.60 |
| 6 | Small | Point | 4 | 181 | 2069 | 571 | 0.73 |
| 7 | Large | Point | 6 | 300 | 3591 | 936 | 0.74 |
| 8 | Large | Plane | 6 | 9 | 43137 | 75 | 0.33 |
| 9 | Large | Plane | 6 | 9 | 40653 | 74 | 0.33 |
| 10 | Large | Plane | 4 | 70 | 33900 | 308 | 0.33 |
| 11 | Large | Plane | 4 | 60 | 28194 | 268 | 0.33 |

| Table 2: Summary of some statistics from the Focus$^{3D}$ self-calibrations | | | | | | | |
|---|---|---|---|---|---|---|---|
| **Dataset** | **Room** | **Point/plane based** | **# of scans** | **# of targets/planes** | **# of obs.** | **# of unk.** | **Avg. Red.** |
| 1 | Small | Point | 4 | 206 | 1833 | 646 | 0.65 |
| 2 | Small | Point | 4 | 183 | 2001 | 573 | 0.72 |
| 3 | Small | Point | 4 | 176 | 2040 | 557 | 0.73 |
| 4 | Small | Point | 4 | 166 | 1821 | 526 | 0.71 |
| 5 | Large | Point | 7 | 300 | 3786 | 942 | 0.75 |
| 6 | Large | Point | 7 | 300 | 3429 | 942 | 0.73 |
| 7 | Small | Plane | 4 | 52 | 31200 | 235 | 0.33 |

## 4. RESULTS & ANALYSES

Non-random trends due to systematic defects can usually be visually identified in the residuals when plotted versus the scanner's raw observations. For instance, significant trunnion axis error and collimation axis error were observed in calibration dataset 2 for the Focus$^{3D}$ (Figure 3).

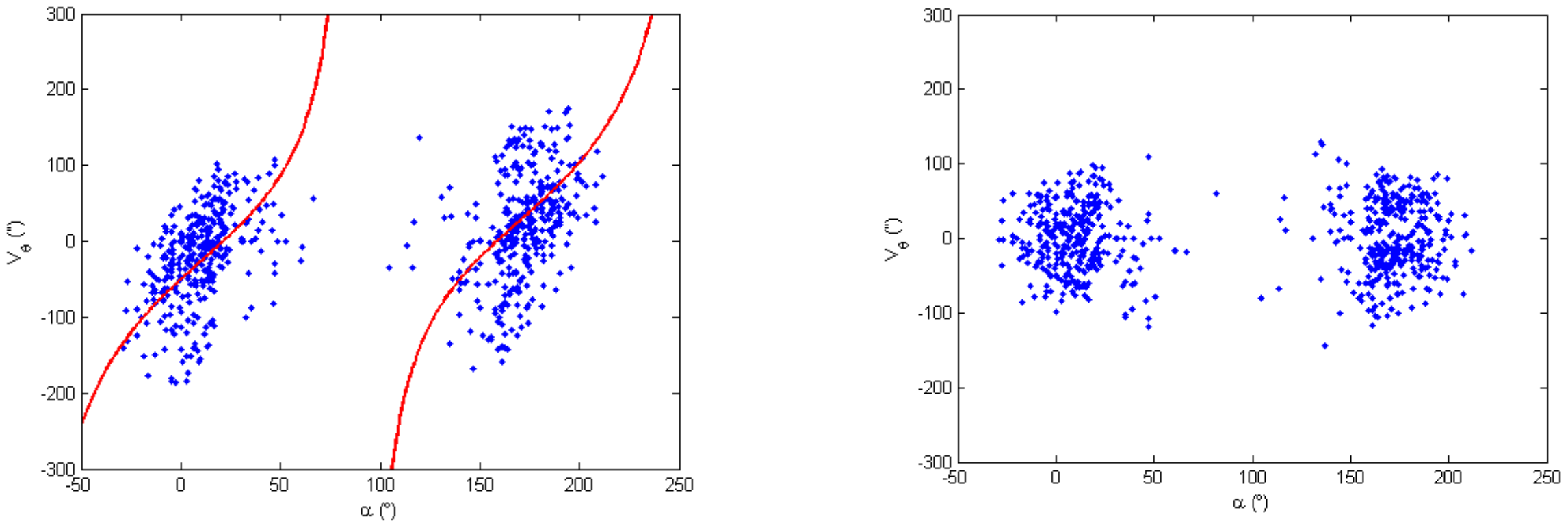

*(a)* *(b)*

*Figure 3: Residuals of the horizontal circle reading as function of the elevation angle (a)* ***before*** *self-calibration and (b)* ***after*** *self-calibration.*

The estimated raw observation precision of both scanners before and after applying the point-based self-calibration is given in Figure 4. The average $\sigma_\rho$, $\sigma_\theta$, and $\sigma_\alpha$ before and after point-based self-calibration for the HDS6100 and Focus$^{3D}$ are shown in Table 3. It is evident that in general the calibration routine can help improve the observation precision of the scanner and the standard deviation of the HDS6100's observations are in general half of the Focus$^{3D}$. The estimated range precision of the HDS6100 is comparable to the independent accuracy assessment carried out in Nuttens et al. (2010). It is worthwhile mentioning that the HDS6100 was sent back to the manufacturer for repairs after dataset 1. It is evident from Figure 4 that after the manufacturer's precise laboratory calibration, the noise level of the instrument was reduced, but it was still improved further using the self-calibration method. As the scanner experiences wear and tear, the HDS6100's elevation angle measurement precision declined over time and reached the same level of precision as the Focus$^{3D}$.

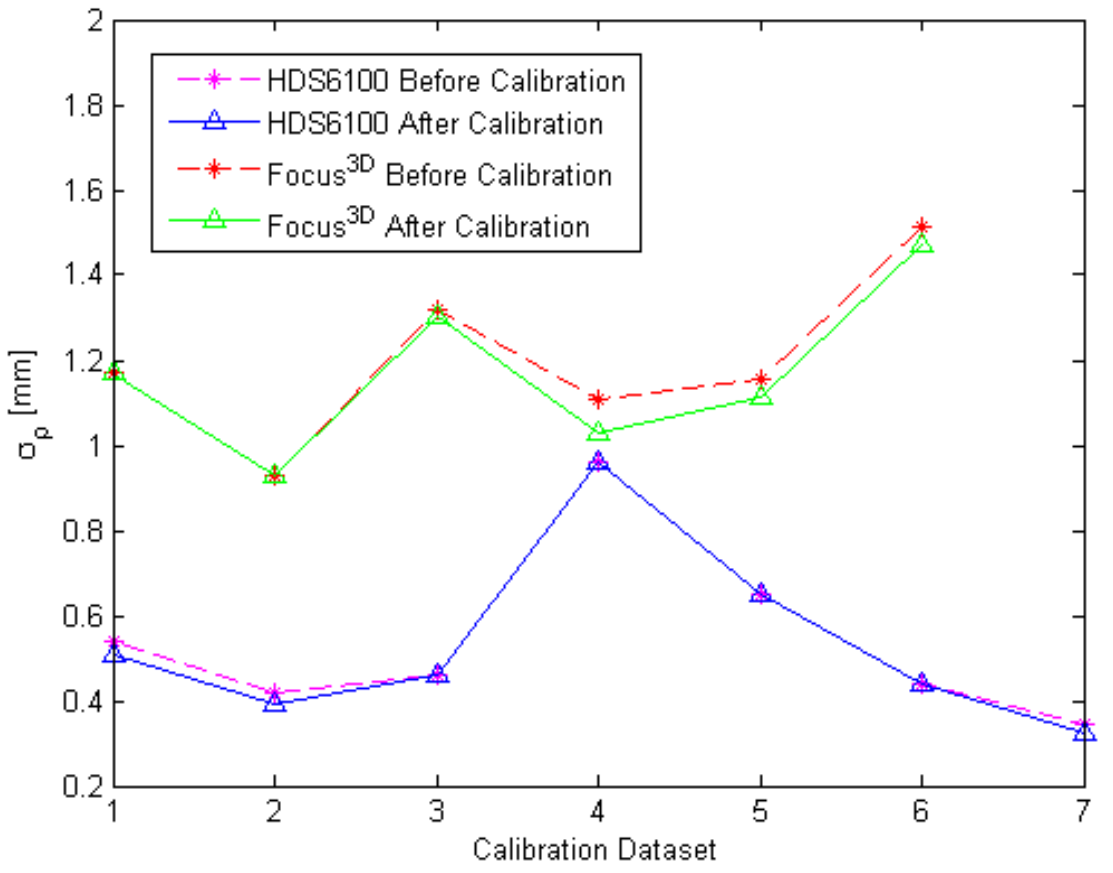

*(a)*

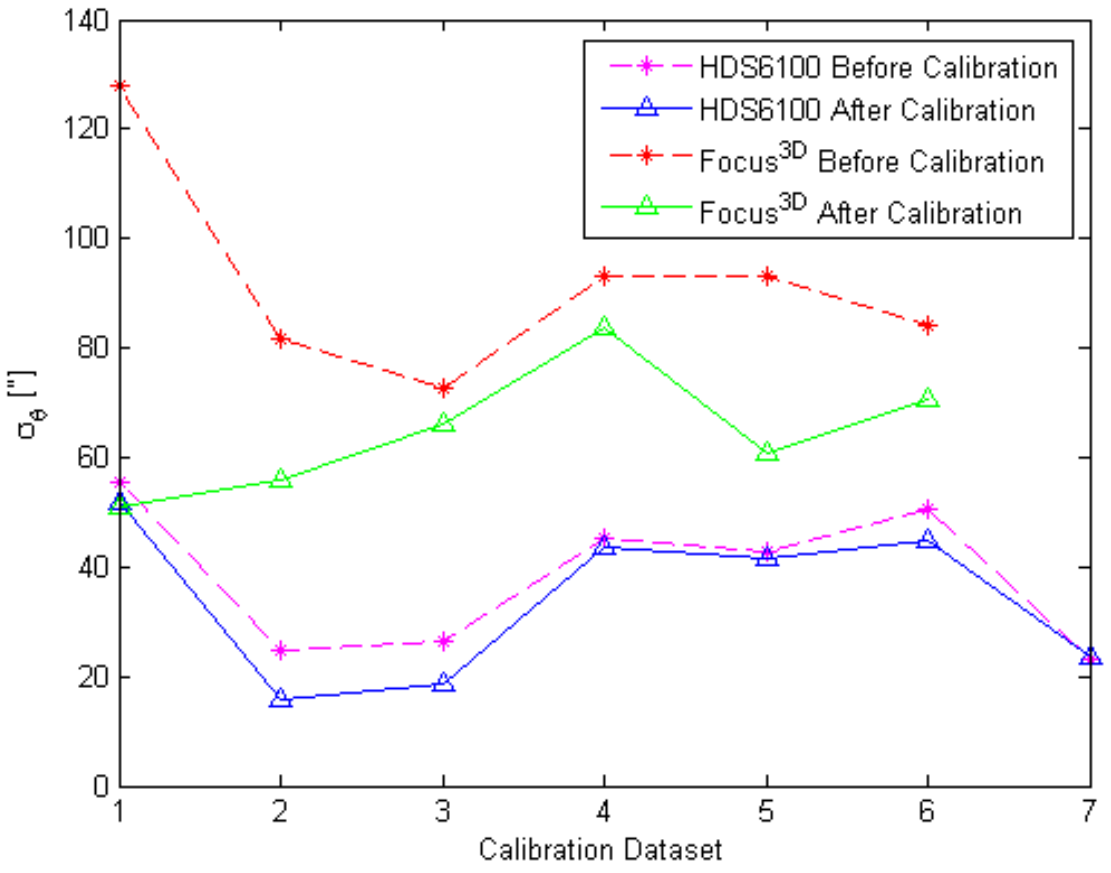

*(b)*

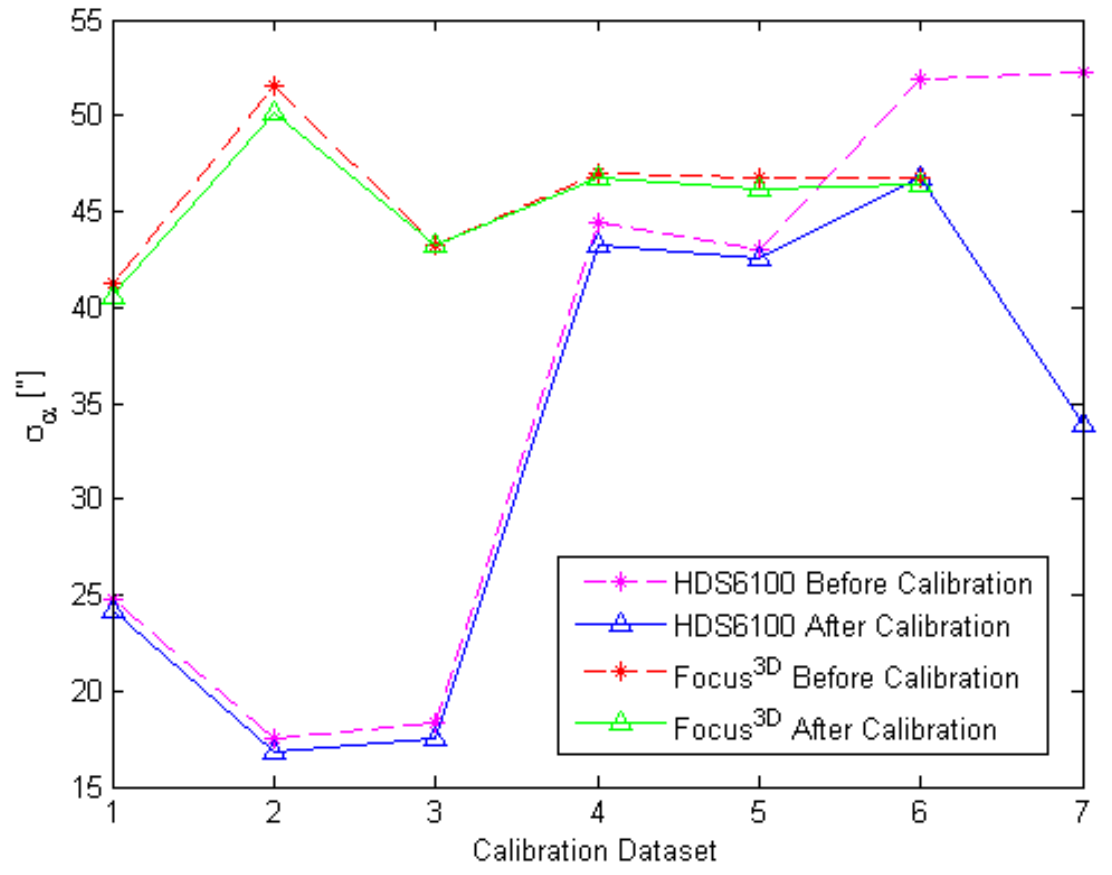

*(c)*

*Figure 4: (a) Estimated* ***range*** *observation precision for the HDS6100 and Focus$^{3D}$ before and after point-based self-calibration. (b) Estimated* ***horizontal circle*** *observation precision for the HDS6100 and Focus$^{3D}$ before and after point-based self-calibration. (c) Estimated* ***elevation angle*** *observation precision for the HDS6100 and Focus$^{3D}$ before and after point-based self-calibration.*

*Table 3: Average observation standard deviation before and after point-based self-calibration*

| | HDS6100 | | | Focus$^{3D}$ | | |
|---|---|---|---|---|---|---|
| | Before | After | Improv. | Before | After | Improv. |
| $\sigma_\rho$ [mm] | 0.55 | 0.53 | 2.2% | 1.20 | 1.17 | 2.6% |
| $\sigma_\theta$ [“] | 38.4 | 34.2 | 11.0% | 92.0 | 64.8 | 29.6% |
| $\sigma_\alpha$ [“] | 36.0 | 32.1 | 10.8% | 46.1 | 45.6 | 1.1% |

The recovered systematic errors that are either statistically significant and/or observable in the range, horizontal direction, and elevation angle residual plots are shown in Tables 4 and 5 for the HDS6100 and the Focus$^{3D}$, respectively. Even though residual systematic errors can be observed in most modern TLS instruments they might be small for some scanners. On the contrary, it can be argued that self-calibration is more essential for older or low-cost scanners because they appear to exhibit more significant residual systematic errors. The trunnion axis error and collimation axis error in the Focus$^{3D}$ are the most significant, and after calibration improvements up to 60% in the horizontal direction measurement precision can be observed. If not modelled properly, a 100 arcsec trunnion axis error and 100 arcsec collimation axis error can result in a horizontal error of -5 mm and 7 mm, respectively at a 45$^{o}$ elevation angle and 10 m distance from the scanner. The reason behind the fluctuation of the scanner's APs is unknown; it might be due to the scanner's instability and/or temperature changes internal to the instrument (Glennie & Lichti, 2011). But as explained in Habib & Morgan (2005) and Lichti (2008), there are some shortfalls when directly comparing APs for checking temporal stability.

*Table 4: Recovered systematic errors for the HDS6100 through point-based and plane-based self-calibration*

| Dataset | $A_0$ | $A_2$ | $B_2$ | $B_3$ | $B_4$ | $B_5$ | $B_6$ | $B_7$ | $C_0$ | $C_6$ | $C_7$ | $C_8$ |
|---|---|---|---|---|---|---|---|---|---|---|---|---|
| 1 | -1.08 ± 0.09 | | | | | | 11.0 ± 1.8 | -18.0 ± 3.7 | -50.5 ± 8.3 | | | |
| 2 | -0.83 ± 0.16 | | 24.0 ± 3.8 | | -10.7 ± 1.6 | | | -22.7 ± 3.9 | -34.0 ± 9.9 | | | 6.2 ± 1.6 |
| 3 | -1.1 ± 0.20 | | 26.2 ± 0.53 | | -9.6 ± 2.2 | | | -29.0 ± 5.4 | -27.2 ± 13.1 | | | |
| 4 | -0.57 ± 0.25 | | | | -11.6 ± 3.6 | -12.3 ± 3.6 | | | | -9.0 ± 3.6 | -7.5 ± 3.6 | |
| 5 | | 1.76 ± 0.47 | | | -9.1 ± 3.8 | -14.1 ± 3.6 | | | | | | |
| 6 | | | | | | -17.4 ± 4.7 | -14.6 ± 4.2 | -49.5 ± 10.4 | | | | |
| 7 | -0.42 0.05 | | | | | | -2.0 0.9 | -40.7 1.7 | -117.2 4.0 | | | |
| 8 | -2.12 ± 0.19 | | | | | 11.4 ± 3.4 | | -28.9 ± 15.9 | | | | |
| 9 | -1.99 ± 0.20 | | | -16.3 ± 3.1 | | | | | | | | |
| 10 | -0.94 ± 0.08 | | | | | | -6.6 ± 1.6 | -26.5 ± 10.6 | 8.9 ± 2.4 | | | |
| 11 | 0.52 ± 0.12 | | | | | | | -50.9 ± 10.8 | 10.3 ± 2.5 | | | |

*Table 5: Recovered systematic errors for the Focus$^{3D}$ through point-based and plane-based self-calibration*

| Dataset | $A_0$ | $B_2$ | $B_3$ | $B_4$ | $B_5$ | $B_6$ | $B_7$ | $C_0$ | $C_3$ | $C_7$ |
|---|---|---|---|---|---|---|---|---|---|---|
| 1 | | | | 13.6 ± 3.1 | 10.9 ± 3.2 | 87.4 ± 4.5 | -203.8 ± 8.8 | | | |
| 2 | 0.54 ± 0.23 | -27.5 ± 9.2 | -51.8 ± 3.6 | | | 50.3 ± 2.9 | -138.3 ± 6.8 | | 24.2 ± 4.4 | |
| 3 | 1.12 ± 0.23 | | | | | 49.6 ± 4.8 | -32.6 ± 9.8 | | | 12.3 ± 2.8 |
| 4 | 2.12 ± 0.31 | | -54.5 ± 5.2 | | | | 44.2 ± 9.8 | | | |
| 5 | 0.48 ± 0.18 | | | | | 58.1 ± 2.2 | -49.6 ± 3.6 | -37.3 ± 7.5 | | |
| 6 | 0.96 ± 0.23 | | | | | 50.2 ± 2.7 | -59.1 ± 4.1 | -38.9 ± 8.1 | | |
| 7 | 2.02 ± 0.42 | | | | | 102.0 ± 6.5 | -113.3 ± 17.7 | 128.7 ± 13.0 | | |

Although the estimated standard deviation of the observations is a valid approach to compare the two instruments, in most cases the user is only concerned about the object space accuracy. To assess the difference in the reconstructed object space by the two scanners, a check point analysis was performed. The HDS6100 and Focus$^{3D}$ were calibrated in the same room on the same day and the determined object space coordinates from both adjustments were compared. In the small room, 168 targets from dataset 6 of the HDS6100 and dataset 3 of the Focus$^{3D}$ were compared. In the large room, 200 targets from dataset 7 of the HDS6100 and dataset 5 of the Focus$^{3D}$ were compared. The object space coordinates determined by both scanners were transformed into a common coordinate system using a 7-parameter 3D similarity transformation. The computed RMSE of the targets before and after self-calibration in the X, Y, and Z directions are shown in Table 6. This check point analysis indicates high compatibility between the two scanners. In the small room, the overall differences between the target positions are less than a millimeter in all directions. At larger distances, the effect of the angular systematic errors is more pronounced, and after self-calibration the compatibility between the point clouds acquired by the HDS6100 and Focus$^{3D}$ was improved.

*Table 6: Differences between the signalised target positions determined by the HDS6100 and Focus$^{3D}$*

| | Before Calibration [mm] | | | After Calibration [mm] | | |
|---|---|---|---|---|---|---|
| Room | $RMSE_X$ | $RMSE_Y$ | $RMSE_Z$ | $RMSE_X$ | $RMSE_Y$ | $RMSE_Z$ |
| Small | 0.7 | 0.8 | 0.5 | 0.7 | 0.7 | 0.5 |
| Large | 0.6 | 0.8 | 2.2 | 0.5 | 0.8 | 1.4 |

Systematic artifacts can be observed in the Focus$^{3D}$ point cloud as shown in Figures 5 and 6a. Near the zenith, a hole in the data can be observed. At low elevations, a mismatch between the point clouds measured on the floor captured in face 1 and face 2 is apparent. After calibration in dataset 7, the 5 mm horizontal displacement between the data captured in front and behind the sensor is eliminated as shown in Figure 6. Note that when capturing a 360$^{o}$ scan, data from face 1 and face 2 overlapped, this is probably because the instrument scanned beyond 360°. For the analysis shown in Figure 6, the overlapping points are removed. The hole on the ceiling situated 1.7 m above the scanner has a maximum diameter of approximately 1.3 cm before calibration and 1.2 cm after calibration. Future work will attempt to improve the self-calibration method and eliminate this systematic defect near zenith.

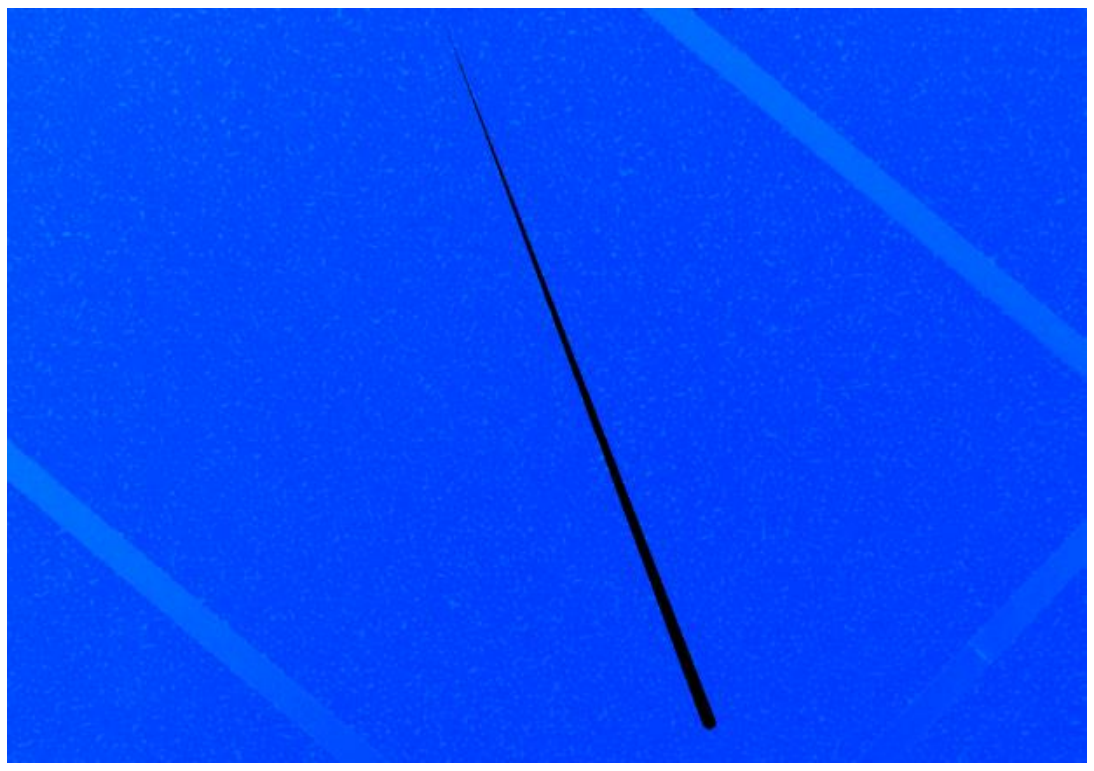

*Figure 5: A hole in the Focus$^{3D}$ point cloud near zenith*

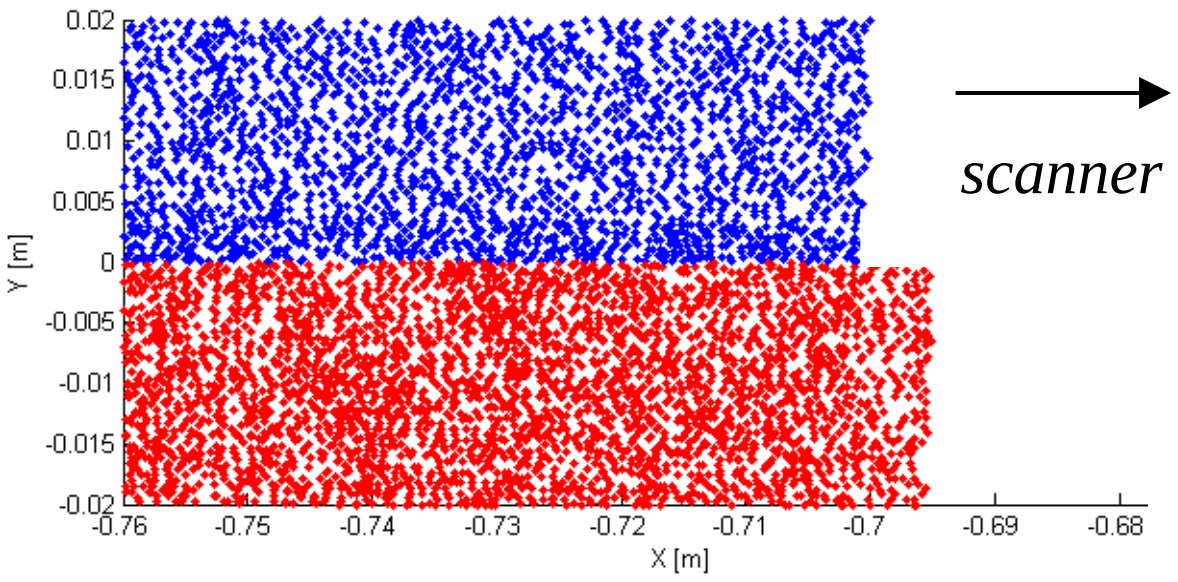

(a)

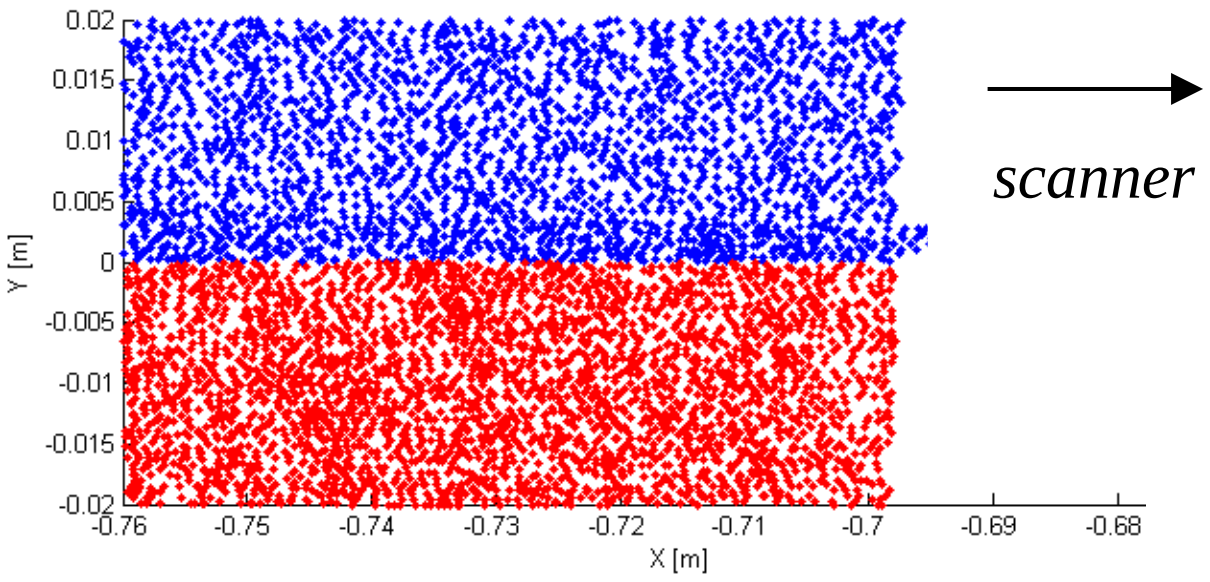

(b)

*Figure 6: Top view of the displacement between face 1 scan data (blue) and face 2 scan data (red) at low elevation angles near the tripod (a)* ***before*** *calibration and (b)* ***after*** *calibration in scanner space*

## 5. CONCLUSION

The geometric quality of the point clouds acquired by two different scanners (i.e. a Leica HDS6100 and a FARO Focus$^{3D}$) was compared quantitatively in this paper. Independent point-based and plane-based self-calibrations were carried out for both sensors to study their systematic errors and raw observation precision. The object space reconstructed by both scanners was also compared directly to evaluate the geometric performance of the more affordable Focus$^{3D}$. From the empirical data, it was discovered that the point clouds generated by both scanners were contaminated by unmodelled systematic errors. The Focus$^{3D}$ showed significant systematic errors in the horizontal circle measurements in all cases. The HDS6100's observation quality appears to be deteriorating slightly, perhaps due to wear and tear, especially in the elevation angle measurements. Even after the residual systematic errors have been modelled using the self-calibration method, the random noise of the Focus$^{3D}$'s observations remains approximately twice the magnitude of that of the HDS6100. However, in close-range, it has been demonstrated in this paper that the discrepancy in the reconstructed object space is small and negligible for most applications. At longer ranges, self-calibration has advantages in improving the geometric accuracy of the reconstructed environment. Future work will attempt to improve the self-calibration technique to remove the systematic artifacts near the zenith.

## ACKNOWLEDGEMENTS

Research funding provided by the Natural Sciences and Engineering Research Council of Canada, the Canada Foundation for Innovation, Alberta Innovates, Informatics Circle of Research Excellence, Terramatic Technologies Inc., and SarPoint Engineering Ltd. is gratefully acknowledged.  The authors are sincerely appreciative for the help they have received when setting up the experiments and collecting the data; thank you Axel Ebeling, Ivan Detchev, Claudius Schmitt, John Tang, and Kathleen Ang.

## REFERENCES

Bae, K., & Lichti, D. (2007). On-site self-calibration using planar features for terrestrial laser scanners. *The international Archives of the Photogrammery, Remote Sensing and Spatial Information Sciences 36 (Part 3/W52)*, 14-19.

Brown, D. (1971). Close-range camera calibration. *Photogrammetric Engineering, 37 (8)*, 855-866.

Chow, J., Ebeling, A., & Teskey, W. (2010). Low Cost Artificial Planar Target Measurement Techniques for Terrestrial Laser Scanning. *FIG Congress 2010: Facing the Challenges - Building the Capacity*. Sydney, Australia, April 11-16.

Chow, J., Lichti, D., & Glennie, C. (2011a). Point-based versus plane-based self-calibration of static terrestrial laser scanners. *ISPRS Laser Scanning Workshop 2011*. Calgary, Canada.

Chow, J., Teskey, W., & Lovse, J. (2011b). In-situ self-calibration of terrestrial laser scanners and deformation analysis using both signalized targets and intersection of planes for indoor applications. *Joint International Symposium on Deformation Monitoring*. Hong Kong, China.

Glennie, C., & Lichti, D. (2011). Temporal stability of the Velodyne HDL-64E S2 scanner for high accuracy scanning applications. *Remote Sensing 3(3)*, 539-553.

Habib, A., & Morgan, M. (2005). Stability analysis and geometric calibration of off-the-shelf digital cameras. *Photogrammetric Engineering and Remote Sensing 71(6)*, 773-741.

Kersten, T., Mechelke, K., Lindstaedt, M., & Sternberg, H. (2008). Geometric Accuracy Investigations of the Latest Terrestrial Laser Scanning Systems. *FIG Working Week: Integrating Generations.* Stockholm, Sweden 14-19 June 2008.

Lichti, D. (2007). Modelling, calibration and analysis of an AM-CW terrestrial laser scanner. *ISPRS Journal of Photogrammetry and Remote Sensing 61 (5)*, 307-324.

Lichti, D. (2008). A method to test differences between additional parameter sets with a case study in terrestrial laser scanner self-calibration stability analysis. *ISPRS Journal of Photogrammetry and Remote Sensing, 63 (2)*, 169-180.

Lichti, D. (2010). Terrestrial laser scanner self-calibration: correlation sources and their mitigation. *ISPRS Journal of Photogrammetry and Remote Sensing 65 (1)* , 93-102.

Lichti, D., Brustle, S., & Franke, J. (2007). Self-calibration and analysis of the Surphaser 25HS 3D scanner. *Strategic Integration of Surveying Services, FIG Working Week 2007.* Hong Kong, China: May 13-17, 2007.

Nuttens, T., De Wulf, A., Bral, L., De Wit, B., Carlier, L., De Ryck, M., et al. (2010). High resolution terrestrial laser scanning for tunnel deformation measurements. *FIG Congress 2010: Facing the Challenges - Building the Capacity.* Sydney, Austrailia, April 11-16.

Reshetyuk, Y. (2009). Self-calibration and direct georeferencing in terrestrial laser scanning. *Doctoral Thesis. Department of Transport and Economics, Division of Geodesy, Royal Institute of Technology (KTH), Stockholm, Sweden, January.*

Reshetyuk, Y. (2010). A unified approach to self-calibration of terrestrial laser scanners. *ISPRS Journal of Photogrammetry and Remote Sensing 65 (5)* , 445-456.

Schneider, D. (2009). Calibration of a Riegl LMS-Z420i based on a multi-station adjustment and a geometric model with additional parameters. *The International Archives of the Photogrammetry, Remote Sensing and Spatial Information Sciences 38 (Part 3/W8)*, 177-182.

Soudarissanane, S., Lindenbergh, R., Menenti, M., & Teunissen, P. (2011). Scanning geometry: Influencing factor on the quality of terrestrial laser scanning points. *ISPRS Journal of Photogrammetry and Remote Sensing 66(4)*, 389–399.

Vosselman, G., & Maas, H.-G. (2010). *Airborne and Terrestrial Laser Scanning.* Scotland, UK: Whittles Publishing.

**Contributions of Authors**

The first author wrote the article; and acquired, processed, and analyzed the data. The second author helped revise the article and made valuable suggestions for improving the paper. The third author provided valuable advice for the project.